\documentclass[accepted]{uai2025} 
                        
\usepackage[american]{babel}

\usepackage{natbib} 
\usepackage{mathtools} 
\usepackage{booktabs} 
\usepackage{tikz} 

\usepackage{balance} 
\usepackage{hyperref}
\usepackage{url}

\usepackage[utf8]{inputenc} 
\usepackage[T1]{fontenc}    
\usepackage{hyperref}       
\usepackage{url}            
\usepackage{booktabs}       
\usepackage{amsfonts}       
\usepackage{nicefrac}       
\usepackage{microtype}      
\usepackage{xcolor}         

\usepackage{amsmath,amsfonts}
\usepackage{graphicx}
\usepackage{textcomp}
\usepackage{amsmath}
\DeclareMathOperator*{\argmax}{argmax}
\usepackage{tabularx}
\usepackage{float}
\usepackage{graphicx}
\usepackage{caption}
\usepackage{subcaption}
\usepackage{algpseudocode}
\usepackage{algorithm}
\usepackage{cuted}
\usepackage{array}
\usepackage{multirow}

\algnewcommand\algorithmicforeach{\textbf{for each}}
\algdef{S}[FOR]{ForEach}[1]{\algorithmicforeach\ #1\ \algorithmicdo}

\title{Group-Agent Reinforcement Learning with Heterogeneous Agents}

\author{Kaiyue Wu}
\author{Xiao-Jun Zeng}
\author{Tingting Mu}
\affil{%
    Computer Science Dept.\\
    The University of Manchester\\
    Manchester M13 9PL, UK
}
  
\begin{document}
\maketitle

\begin{abstract}
Group-agent reinforcement learning (GARL) is a newly arising learning scenario, where multiple reinforcement learning agents study together in a group, sharing knowledge in an asynchronous fashion. The goal is to improve the learning performance of each individual agent. 
Under a more general heterogeneous setting  where different agents learn using different algorithms, we advance GARL by designing novel and effective group-learning mechanisms.  
They guide the agents on whether and how to learn from 
action choices from  the others, and allow the agents to  adopt available policy and value function models sent by another  agent if they perform better.
We have conducted extensive experiments on a total of 43 different Atari 2600 games to demonstrate the superior performance of the proposed method. 
After the group learning, among the 129 agents examined, 96\% are able to achieve a learning speed-up, and 72\% are able to learn $\geq$100 times faster. 
Also, around 41\% of those agents have achieved a higher accumulated reward score by learning in  $\leq$5\% of the time steps required by a single agent when learning on its own.
\end{abstract}

\section{Introduction}
\label{intro}

There has been an increasing interest in reinforcement learning involving multiple agents, emphasizing and strengthening different aspects of learning. 
One mainstream of such research is multi-agent reinforcement learning (MARL) \citep{christianos2020shared, sunehag2018value, mahajan2019maven, wang2020roma, rashid2018qmix, lowe2017multi}. It emphasizes cooperation and/or competition between multiple agents, in addition to individual success of each agent, in a shared environment with all the agents acting synchronously. 
The behaviour of cooperation or competition plays an important role, given that the outcome of each agent's behaviour is affected by the behaviour of the other agents. 
MARL has been shown to improve the learning for playing  the game of Go \citep{silver2016mastering, silver2017mastering, silver2018general, schrittwieser2020mastering} and robotic soccer, with the former being a competitive scenario and the later being a mixed task of cooperation and competition.
Another mainstream is ensemble reinforcement learning (ERL) which aims at achieving better performance by combining the learned policies of multiple independent  agents of either the same type \citep{fausser2011ensemble, duell2013ensembles, fausser2015neural, fausser2015selective, osband2016deep, saphal2021seerl, lee2021sunrise, chen2017ucb, anschel2017averaged, smit2021pebl}  or different types \citep{yang2020deep, wiering2008ensemble, chen2018ensemble, nemeth2022split}, referred to as base agents.
There is no communication between the base agents, while their decisions are combined only after completing the whole learning process. 
ERL does not intend to improve each individual base agent, but aims at a robust ensemble, of which an example application is stock trading \citep{yang2020deep, nemeth2022split}. 

More recently, the concept of group-agent reinforcement learning (GARL) has been proposed by \cite{garl}. 
It takes the inspiration from social learning where people learn by observing and imitating the behaviour of the others \citep{mcleod2011albert}. 
In addition to learning by the classical trial-and-error exploration, the agents   share knowledge with each other during the learning process.
The goal is to  enable  the improvement of each individual agent, i.e., their learning speed and quality, through their received knowledge.
Different from MARL, each GARL agent acts separately in its own environment, and learns asynchronously without any intention to cooperate or compete. 
This asynchronous group-learning paradigm is suitable for  applications involving a group of agents with its members seeking to share and improve knowledge whenever  available, e.g. video game playing, autonomous driving, network routing, etc.

Given the different learning objectives and assumptions imposed  by MARL, ERL and GARL, it is not possible to transfer effortlessly the technical success between them. 
We will review main advances in MARL and ERL, and discuss  in comparison to GARL in more detail in Section \ref{related}.  Significant research efforts have been put separately on improving each of these learning paradigms, addressing specific learning requirements. 
In this paper, we focus on   advancing the GARL scenario, which is a 
new learning paradigm still under development. 
Currently, GARL \citep{garl} has successfully improved the  stability and scalability of  agent training for a classic control task of CartPole-v0 on OpenAI Gym \citep{1606.01540}. 
However, it works under a simple homogeneous setting,  where the agents are restricted  to act in the same environment and learn by the same  algorithm. 
When being required to support agents that use different learning algorithms,  such method becomes inapplicable, limited by its core design of communicating knowledge by  sharing gradients.

In this paper, we focus on developing more general GARL learning mechanisms to support heterogeneous agents  underpinned by different algorithms, referred to  as \emph{heterogeneous group-agent reinforcement learning} (HGARL). 
We specify  the knowledge to share between agents as their policy/value model parameters and their accumulated reward score.
Each agent selects the best one from a set of actions suggested by not only their own policy model  but also the received policy models from the other agents, for which three action selection rules are suggested, including  probability addition (PA), probability multiplication (PM) and reward-value-likelihood combination (Combo).
After this,  the agent applies the selected action and continues its learning from the trajectory resulted from this selected action.
In addition, we allow each agent to replace their own policy model with a better received model guided by a set of model adoption rules.
Extensive experiments are conducted to assess the proposed  approach, through learning to play a total of  43 different Atari 2600 games \citep{bellemare2013arcade}. 
We simulate a group of  three agents each supported by a different Actor-Critic deep reinforcement learning algorithm, including A2C \citep{mnih2016asynchronous}, PPO \citep{schulman2017proximal} and ACER \citep{wangsample}.  
The effectiveness of HGARL is evidenced by performance improvement of  these three types of agents, comparing  learning in  a group as opposed to learning  individually. 
Among the total 129 agents examined, 96\% are able to achieve a learning speed-up, and 72\% learn over 100 times faster. 
Also, around 41\% of the agents have achieved a higher reward score within less than 5\% of the time steps required by a single agent when learning on its own.
Overall, the contribution of this paper is three-fold: 
\begin{itemize}
\item Advancing GARL to a more general heterogeneous setting, supporting different agents underpinned by different learning algorithms. 
\item Designing novel and effective learning mechanisms to improve each agent by using better action choices suggested by  the others and adopting better policy/value function models sent by another.
\item Demonstrating significant performance improvement of learning speed-up and improved reward score.
\end{itemize}

\section{Related Work and Discussion}
\label{related}

\subsection{Multi-Agent Reinforcement Learning} 
We review research work on MARL with respect to two main groups of learning setups, where one assumes a global view of all the agents without communication, while the other allows the agents to communicate to gain sufficient views. 

Example work from the first non-communication group include \citep{sunehag2018value} that decomposes the team value function into a sum of agent-wise value functions, \citep{rashid2018qmix} that relaxes the strong additive assumption of \citep{sunehag2018value},  \citep{mahajan2019maven} that improves \citep{sunehag2018value} as it severely limits the complexity of the team value function, and \citep{wang2020roma} that applies the role concept to MARL to allow agents with similar roles to share similar behaviours. 
Another representative work from this group is \citep{christianos2020shared}, which shares experiences between multiple agents and incorporates all those into the loss function.
All the agents are  synchronised in a common environment, thus all of them are at the same learning stage at any time, and their decisions can always help each other. 
However, we have observed through experiments that such a  synchronised  sharing of experiences does not suit the  GARL scenario.
This is  because our agents are normally in different learning stages and sharing the activities of the agents that are less mature to a senior agent will lead to potential learning failure. 
 
For the second  group of communicative approaches,   all the agents are assumed to synchronise with each other when taking actions in a common environment \citep{lowe2017multi, sukhbaatar2016learning, peng2017multiagent, wang2022fcmnet, qu2019value}.
For instance, \cite{lowe2017multi} consider explicit communication between agents but does not specify any particular structure on the communication method. 
\cite{qu2019value} explore a fully decentralised setting for the agents and locates each of them at a node of a communication network. 
\cite{sukhbaatar2016learning} maintain a centralised communication network and allows multiple continuous communication cycles at each time step to decide the actions of all the agents. 
\cite{peng2017multiagent} further deal with heterogeneous agents with bi-directional communication channel using recurrent neural networks. 
\cite{wang2022fcmnet} focus on the problem of learning to communicate, introducing a new framework for the agents to learn a communication protocol.

In general, it is not  suitable to directly adapt an MARL approach for GARL that requires asynchronous communication.
It is non-trivial to add any communication component to the first group of MARL approaches. 
And for  the  second group of MARL approaches, their communication components require synchronisation that not only will limit the group agents too much but also is non-realistic for real-world GARL applications.

\subsection{Ensemble Reinforcement Learning} 

Ensemble learning is popular for its capacity of combining machine learning models to make a better prediction.
It is well explored  under the supervised learning paradigm, with well-known methods including bagging \citep{breiman1996bagging}, AdaBoost \citep{freund1999short} and random forest \citep{ho1995random}. 
More recently, its effectiveness has been also demonstrated in reinforcement learning, through combining multiple policies learned by one common reinforcement learning algorithm \citep{fausser2011ensemble, duell2013ensembles, fausser2015neural, fausser2015selective, osband2016deep, saphal2021seerl, lee2021sunrise, chen2017ucb, anschel2017averaged, smit2021pebl}, or combining multiple algorithms \citep{yang2020deep, wiering2008ensemble, chen2018ensemble, nemeth2022split}. 

To combine policies, the commonly used combiners include majority voting and averaging, e.g. for combining base value functions that are either parameterised or approximated by neural networks \citep{fausser2011ensemble, fausser2015neural, fausser2015selective}, for combining policies in discrete action space \citep{duell2013ensembles}, and for developing an ensemble version of  DQN algorithm \citep{mnih2013playing} by averaging the Q-value functions \citep{anschel2017averaged}.
There are also approaches developed to increase diversity between base policies, e.g.,  bootstrapping by random initialisation of policy networks \citep{osband2016deep}, exploration strategy based on confidence bounds \citep{chen2017ucb},  and hybrid method for combining the previous two \citep{lee2021sunrise},  as well as the directed policy perturbation strategy \citep{saphal2021seerl} and the method for dealing with value overestimation \citep{smit2021pebl}. 
To combine algorithms, in addition to the standard approaches like majority voting, rank voting, Boltzmann multiplication and Boltzmann addition \citep{wiering2008ensemble}, \cite{chen2018ensemble} have developed  a two-step ensemble strategy to combine DQN algorithms \citep{mnih2013playing}.
More specialized ensemble strategies have also been developed for stock trading \citep{yang2020deep, nemeth2022split}. 

A main difference between GARL and ERL is that GARL aims at improving the performance of every individual agent in the group, while ERL aims at improving the final overall performance regardless of  the individual performance. 
In an ERL problem each individual algorithm updates its model independently without affecting each other.
The priority of ERL is not to improve the individual learning of each base agent.
Differently,  each agent in a group-learning system aims at self-improvement by using knowledge from the other agents in the group.
Despite the difference, the ensemble rules developed for ERL can inspire the design of how to combine knowledge in GARL.

\begin{figure}
\includegraphics[width=\linewidth]{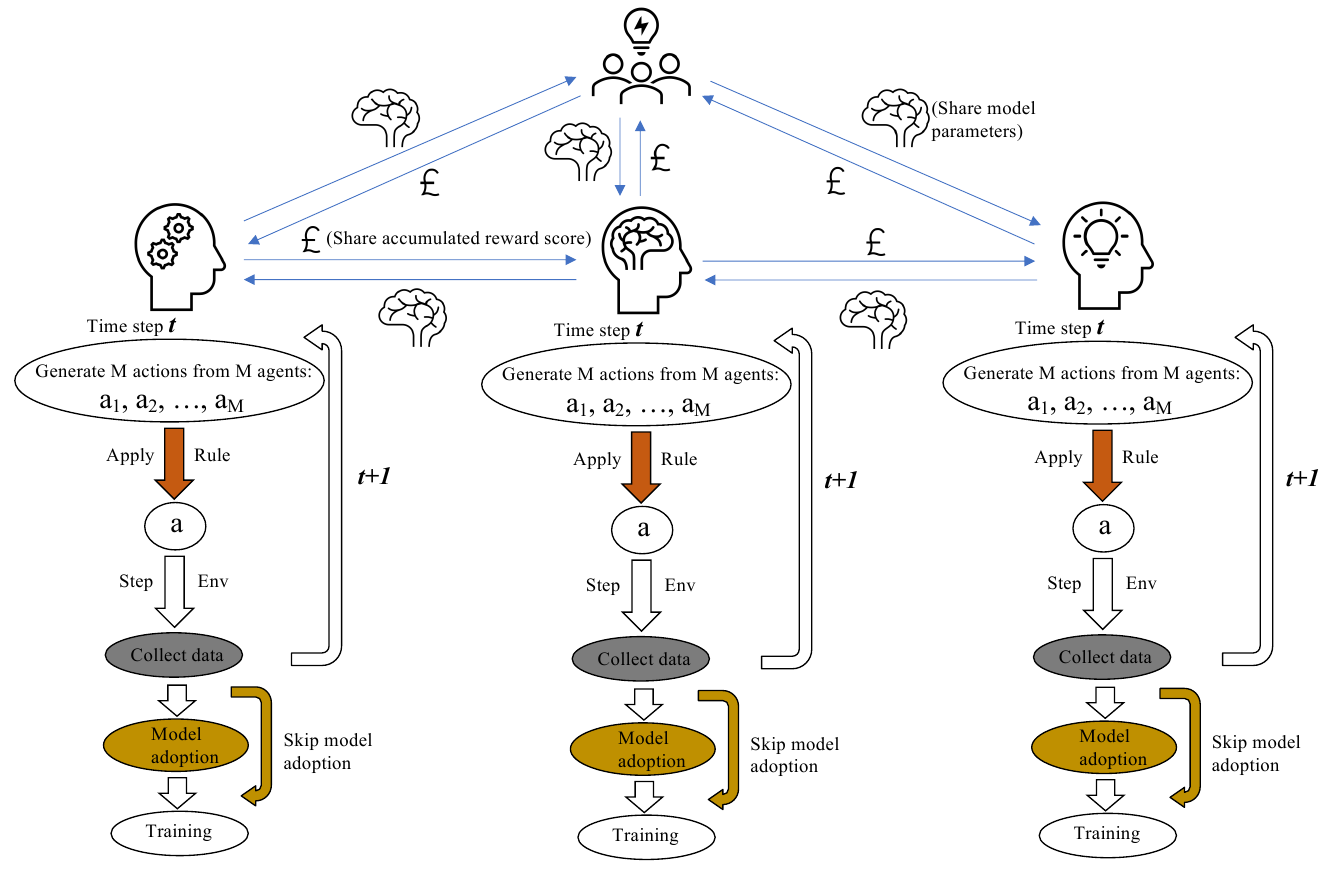}
\caption{Flowchart: The agents of different types share knowledge with each other during their learning processes. The knowledge are of two types, one is the policy and value function model parameters and the other is the accumulated reward score that they've achieved so far. At each time step, an agent will get a set of suggested actions to take from itself and all its peers in the learning group, then select a best action according to one of our action selection rules. After performing the selected action in its environment, the agent will collect trajectory data resulted from this action and train itself with the data. There is one more step of model adoption which will only happen when the used action selection rule is Combo.}
\label{flowchart}
\end{figure}

\section{Proposed HGARL}

We study effective ways to achieve HGARL,  where a group of agents are trained by different learning algorithms  from scratch in separate environments  and the agents are allowed to communicate  every few updates. 
The extension from GARL to HGARL is non-trivial, as different learning algorithms update their models in different ways, therefore the gradient sharing scheme used by GARL becomes inapplicable.  
More sophisticated design on deciding what  knowledge to share  and how to  use the shared knowledge is required.
Group learning  becomes helpful when the agents are asked to solve similar tasks. 
Also the  group can be informed to adopt a fixed neural network architecture for building their policy models as a useful prior. 
So we  assume  the same action and state spaces for all the agents in a group, and the   same architecture for all policy models. 
We define the shared knowledge between agents as the policy and value model parameters  and  the accumulated reward score per episode, achieved by an agent at a certain time step.
For the simulated agents as explained below,  we let A2C and PPO adopt state value models, while ACER adopts action value model.

\subsection{Agent Simulation} \label{agent_sim}

We simulate different agents using different Actor-Critic reinforcement learning algorithms, selecting A2C \citep{mnih2016asynchronous}, PPO \citep{schulman2017proximal} and ACER \citep{wangsample} without loss of generality. 

\textbf{Advantage Actor-Critic} (A2C) approximates a policy function $\pi_{\theta}$ parameterized by $\theta$ and a state-value function $V(s)$, then calculates an advantage value $A(s_t,a_t) = Q(s_t,a_t) - V(s_t)$ where $Q(s_t,a_t) = r + \gamma V(s_{t+1})$ 
and $Q(s_{t+1},a_{t+1}) =  r$ for the terminal state $s_{t+1}$  and $r$ as the immediate reward. 
A policy update is given as
\begin{equation}
\nabla_{\theta} \log \pi_{\theta}(a_t|s_t)A(s_t,a_t) = \nabla_{\theta} \log \pi_{\theta}(a_t|s_t)(Q(s_t,a_t) - V(s_t)).
\end{equation}

\textbf{Proximal Policy Optimization} (PPO) introduces a clipped surrogate objective function which discourages large policy change at each update and improves training stability, as below
\begin{equation}
L^{\textmd{CLIP}} (\theta) = \hat{E_t}\left [\min\left(r_t(\theta)\hat{A_t}, \textmd{clip}(r_t(\theta), 1-\epsilon, 1+\epsilon)\hat{A_t} \right) \right],
\end{equation}
where $\hat{E_t}[\dots]$ is the empirical average over a finite batch of experiences, $\hat{A_t}$ is an estimator of the advantage function at time step $t$, also $r_t(\theta) =\frac{\pi_{\theta} (a_t|s_t)}{\pi_{\theta_{\textmd{old}}} (a_t|s_t)}$, and the term $\textmd{clip}(r_t(\theta), 1-\epsilon, 1+\epsilon)$ restricts $r_t (\theta)$ within the range $[1-\epsilon, 1+\epsilon]$ with $\epsilon$ as a small constant.

\textbf{Actor-Critic with Experience Replay} (ACER) introduces several innovations into an off-policy Actor-Critic reinforcement learning algorithm. It adopts the Retrace algorithm \citep{munos2016safe} in the estimation of the Q-function of the target policy.  
A Retrace estimator is expressed as
\begin{align}
Q^{\textmd{ret}}(s_t, a_t) = \;&  R(s_t, a_t)  + \gamma V(s_{t+1}) \\
\nonumber
& + \gamma \overline{\rho}_{t+1} \left[Q^{\textmd{ret}}(s_{t+1}, a_{t+1}) - Q(s_{t+1}, a_{t+1})\right]
\end{align}
where $\overline{\rho}_t = \min\{c, \rho_t\}$, $\rho_t = \frac{\pi(a_t | s_t)}{\mu(a_t | s_t)}$  with $\pi$ as the target policy while $\mu$ as the behaviour policy, $Q$ is the current estimate of the action value, and $V$ is the current estimate of the state value. 
$Q^{\textmd{ret}}$ is used in  gradient calculation, as $(Q^{\textmd{ret}}(s_t, a_t) - Q(s_t, a_t)) \nabla Q(s_t, a_t)$.  
ACER  adopts the importance weighted policy gradient that is approximated using marginal value functions over the limiting distribution of the learning process \citep{degris2012off}.
It applies a truncation technique to the importance weights for the purposes of bounding the variance of the gradient estimate and correcting bias. 

The proposed HGARL method applies to an arbitrary number of agents in a group. 
In our empirical study, we choose the above  algorithms  to simulate a group of three agents possessing different learning strengths and varying effectiveness over different learning tasks. 
Such a variation helps highlight the agent difference.  
We  do not intend to include more agents  in the experiments, in order to  be more focused on  studying how and how much they  affect each other in a group. 

\subsection{Action Selection and Adoption} \label{action_sel}

Denote the agent group by $\mathcal{M}$ and the action set  by $\mathcal{A}$. 
At time step $t$, an agent receives a set of policies from the other agents in the group, which together with its own policy form the policy set $\mathcal{P}_t = \{\pi_m(a_t | s_t) \}_{m \in \mathcal{M}}$. 
We  design selection rules  to  obtain the best action from $\mathcal{P}_t$.
 We start  from  the following two straightforward selection rules.

\textbf{Probability Addition (PA) Rule.}  It computes an accumulated policy by adding up all the candidate policies in $\mathcal{P}_t$, i.e.
\begin{equation}
\pi(a_t | s_t) = \sum_{m \in \mathcal{M}} \pi_m(a_t | s_t).
\end{equation}
The best action is then given by 
\begin{equation}
\label{eq:sel}
a^{\textmd{best}}_t  = \argmax_{a \in \mathcal{A}} \pi(a | s_t).
\end{equation}

\textbf{Probability Multiplication (PM) Rule.}  Alternatively, an accumulated policy can be computed by multiplying all the candidate policies in $\mathcal{P}_t$, i.e. 
\begin{equation}
\pi(a_t | s_t) = \prod_{m \in \mathcal{M}} \pi_m(a_t | s_t).
\end{equation}
Similar to PA,  the best action is obtained by Eq. (\ref{eq:sel}).

Both rules above consider equally policies provided by all the agents.  
However, policies generated during  a very early stage of training  heavily rely on random exploration. 
Thus, it is  useful to filter   out  low-quality polices and focus on more reliable ones supported by evidence like the accumulated award score and prediction confidence.  
Motivated by this, we  propose a more sophisticated action selection rule as below.

\textbf{Reward-Value-Likelihood Combination  (Combo) Rule.}  We focus  on the $i$-th agent in $\mathcal{M}$ at a time step $t$, and let $a_{i, t}$ denote its predicted action.   
The Combo rule  considers multiple factors including the accumulated reward scores per episode, state/action values, and action prediction confidence when selecting the best action. 
It contains the following steps:
\begin{enumerate}
\item   Identify all the agents from the group that possess  higher accumulated reward scores ($AR$)  than the target agent $i$, and include those to the set  $ \mathcal{M}_{i,t}= \left\{m |AR_{m,t}> AR_{i,t}, \forall m \in \mathcal{M}\right\}$.
\item  For each agent $m$ in the identified set $m \in\mathcal{M}_{i,t}$, apply its action to obtain the next state denoted by $s_{t+1, m}$ and predict its value $V(s_{t+1, m})$. Then we calculate an accumulated value $V_t^{acc}(m)$ by adding up all the predicted $V(s_{t+1, m})$ so far 
within the current batch of experiences. Select the agent $m^*$ with the highest value, i.e. 
\begin{equation}
\label{state_value}
m^* = \argmax_{m \in \mathcal{M}_{i,t}} V_t^{acc}(m).
\end{equation}
\item  Examine the action probability predicted by the selected agent $m^* $, and return the best action 
\begin{equation}
\label{action_thresholding}
a^{\textmd{best}}_t =  
\left\{
\begin{array}{ll}
a_{m^*, t},  &  \textmd{if}   -\log\pi_{m^*} (a_{m^*, t} | s_t) <  \phi, \\
a_{i, t},  &     \textmd{otherwise},   
\end{array}
\right.
\end{equation}
where $\phi>0$ is a user-specified threshold (set as a hyper-parameter).
\end{enumerate}
After selecting the action $a^{\textmd{best}}_t$ for the target agent,  a new state is obtained by applying $a^{\textmd{best}}_t$, and the learning continues from this new state. 
When $a^{\textmd{best}}_t$ is different from the original action predicted by this agent, a different trajectory of training data is resulted in.

\textbf{Discussion on Combo Rule.}
We consider  three selection factors in  Combo Rule, including the accumulated reward score per episode, state/action values and action probabilities. 
The accumulated reward score, calculated by adding up all the rewards that an agent gets from an episode, is ground truth, serving as the real evidence for assessing the quality of an agent. 
On the contrary, both state values and action probabilities are  estimated   by an agent,   indicating how confident the agent is currently  feeling about its action choice.  
Such  estimations can be unreliable in early training, as the agent has not learned well yet. 
Although the accumulated reward score is reliable,  it is insufficient to represent 
the whole state space, as the agent that achieves the best overall performance (accumulated reward per episode) may have only explored partially the state space, with  limited knowledge on some unexplored parts. 
Therefore, it is important to consider all the three factors.

In Eq. (\ref{action_thresholding}), we examine the action confidence by thresholding the negative log likelihood, rather than the  probability predicted directly.
This is because  a negative log function maps a probability value between 0 and 1 to a much wider range of values spreading over the entire positive axis, and this reduces the threshold sensitivity. 
When the threshold $\phi$ is small,  it  favours very confident actions suggested by other agents,  but this may exclude some good action choices.
Choice of $\phi$ can differ over problems. 
The probabilities will start with the average probability. For example, if a game has 6 actions in its action space, the initial probability for each action is $\frac{1}{6}$.
In some cases, the probability of the best-belief action may go up very high in the middle of the training but decrease towards the end of  the training. 
A final learned model of one environment  mostly outputs a high  probability for its best actions,  but it may output lower probabilities for the best actions  in  a different environment. 
In our experiments, we attempt to locate a threshold choice that is generally good for all the problems that we have tested on.  
The process for the selection of $\phi$ is described in Appendix A.

In Step (2), the state value is calculated after applying an action, referred to as the next-state value. 
In order to mitigate individual errors,  we consider the accumulated next-state value obtained in the current batch of experience in Eq. (\ref{state_value}). 
It indicates how good the action is under the current circumstance  and also how frequently an agent takes a good action.    
Therefore, as in Eq. (\ref{state_value}),  the agent with the highest accumulated next-state value should be the most trustworthy one in the group to the best of our knowledge.
Overall,  we consider an action the current best when it is predicted with a high probability by an agent producing better accumulated reward score and the highest accumulated next-state value.

\begin{algorithm}[t]
\caption{HGARL training of the $i$-th agent in the group $\mathcal{M}$} \label{alg1}
\begin{algorithmic}[1]
\Require Initialise policy and value neural network models for all agents in the group $\mathcal{M}$ at the $i$-th agent
\ForEach{$episode$}
\State Send my models to all the other agents in the group
\State Receive models from other agents in the group
\ForEach{$time step \ t \in T$}  \Comment{training in batches}
\ForEach{$agent \in \mathcal{M}$}
\State Get its action choice
\State Append the action to an action set $\mathcal{A}_t$
\EndFor
\State Compare the actions in $\mathcal{A}_t$ with a rule of either Combo, PA or PM  \Comment{apply a rule to select the best action}
\State Apply the selected action $a_t \in \mathcal{A}_t$ for all agents in $\mathcal{M}$
\State Collect trajectory data $(S_t, a_t, S_{t+1}, R_t)$, action probabilities, state/action values and other relevant info for all agents in $\mathcal{M}$
\State Append the above data to the corresponding training data set $D^T_m (m \in \mathcal{M})$
\EndFor
\If{(the most of the selected actions $a_t$ are from another agent $k \in \mathcal{M}$ but $k \neq i$) $AND$ ($N_i^k  \geq \frac{N}{2}$) $AND$ (agent $k$ currently produces the highest accumulated reward score per episode)}
\State Replace the $i-th$ agent's models with the $k-th$ agent's models  \Comment{model adoption}
\EndIf
\State Train the $i-th$ agent with dataset $D_k^T$ (or $D_i^T$ if model adoption did not happen)  \Comment{training with the correct trajectory data}
\EndFor
\end{algorithmic}
\end{algorithm}

\subsection{Model Adoption}

We include an additional step of model adoption  to our HGARL design, when using Combo rule for action selection. 
As explained earlier,  the policy and value model parameters  are shared between agents. 
Specifically, we enable each agent to send its entire model (parameters) to other agents every few updates, meanwhile each agent  to receive  models  from the other agents in the group. 
At every single time step, each agent predicts actions for the current state by using all the models it has, and selects the best action using the Combo rule as explained in Section \ref{action_sel}. 
For  a target agent $i$, after applying this selection rule for a number of $N$ time steps  where $N$ is the batch size, we  record the number of times among these $N$ steps when it selects an action predicted by  the policy of  agent $k \in \mathcal{M}$, denoted by $N_i^k$. 
Identify whether there exists an agent $k$ in the group that satisfies  the following three conditions simultaneously:
\begin{enumerate}
\item The usage of agent $k$'s policy by the target agent $i$ ranks 1st.
\item  The usage frequency satisfies $N_i^k  \geq \frac{N}{2}$.
\item Agent $k$ produces the highest accumulated reward score per episode.
\end{enumerate}
If agent $k$ exists, the target agent $i$ replaces its own policy and value function models with those of agent $k$'s, and continues the training by using the trajectories, predicted action probabilities, state/action values and other relevant data of the new current models that are received from agent $k$.
With model adoption, we intend to make a bigger move in terms of improving policy and value function models, in addition to the gradual improvement through adopting better actions from others one by one. 
Note that we skip model adoption when using PA and PM selection rules, since they use actions according to joint decisions of all the agents instead of decisions of any particular agent.
%
%
Finally, we provide in Algorithm \ref{alg1} the pseudo code of the proposed HGARL and in Figure \ref{flowchart} the flowchart of the whole process.%

 \section{Experiments and Results}
\label{Eval}

 \begin{figure}
         \begin{subfigure}[]{0.49\linewidth}
                \includegraphics[width=0.333\linewidth]{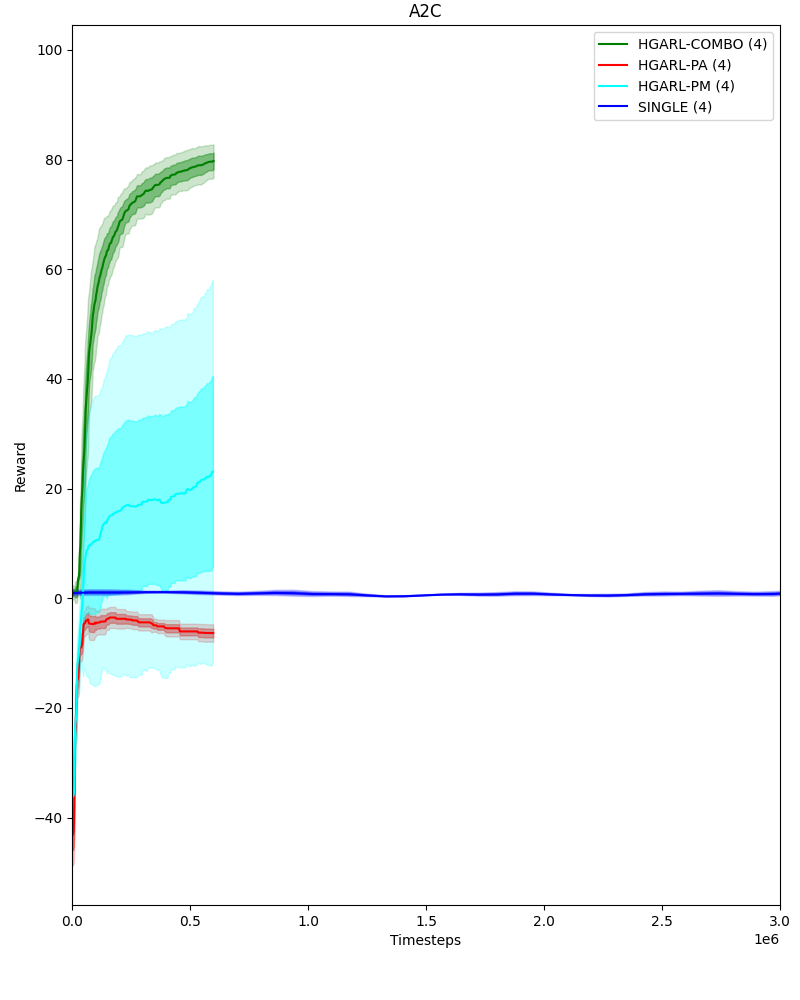}\hfill
                \includegraphics[width=0.333\linewidth]{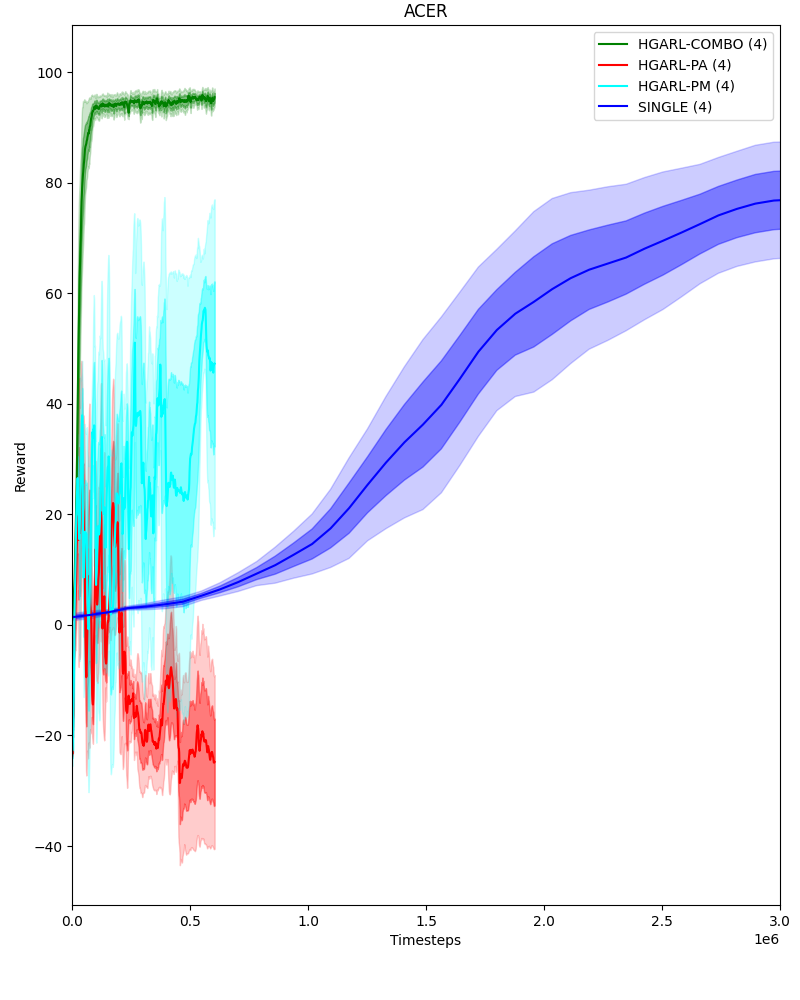}\hfill
                \includegraphics[width=0.333\linewidth]{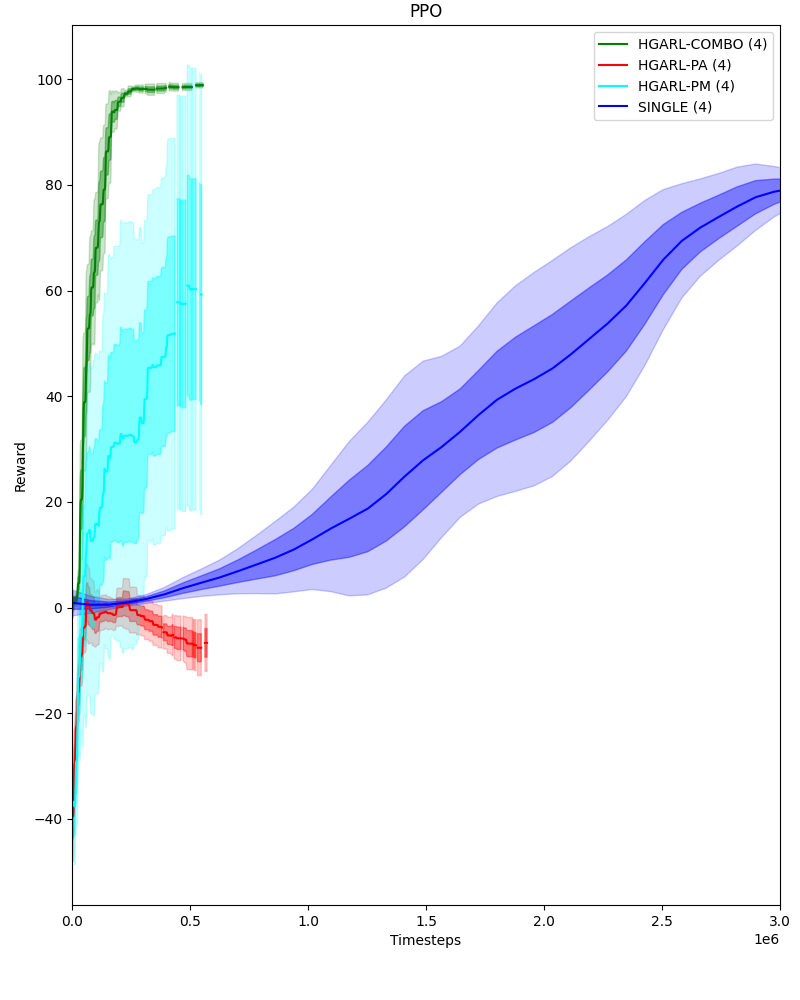}
                \caption{Boxing}
         \end{subfigure}
         \begin{subfigure}[]{0.49\linewidth}
                \includegraphics[width=0.333\linewidth]{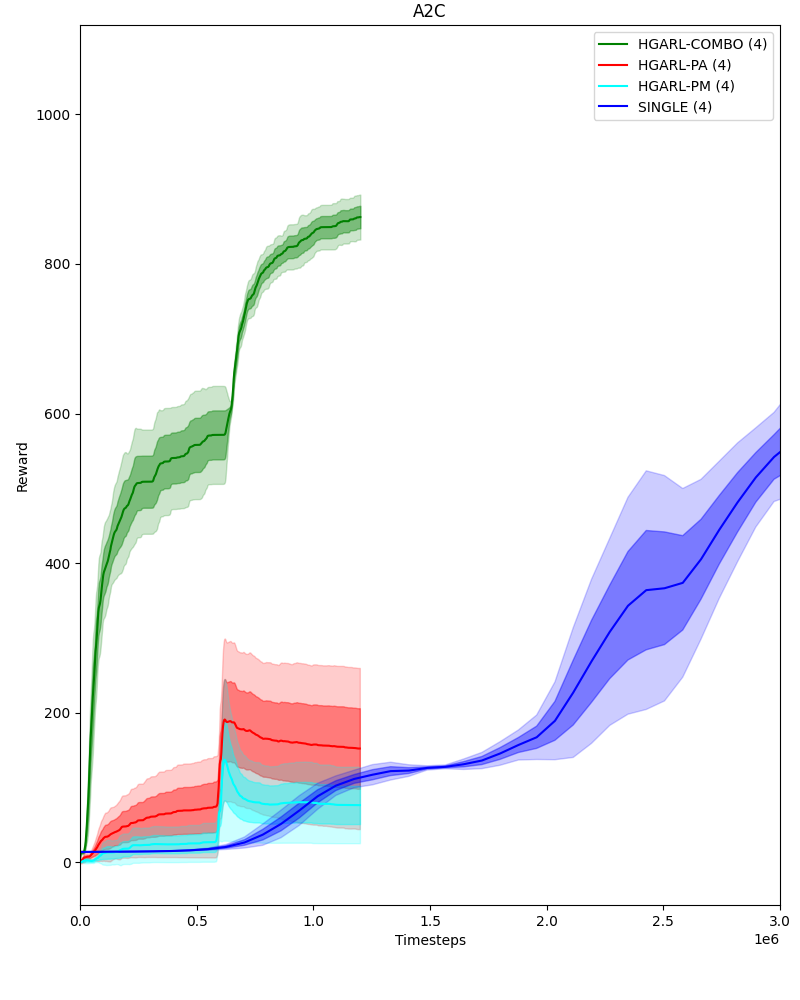}\hfill
                \includegraphics[width=0.333\linewidth]{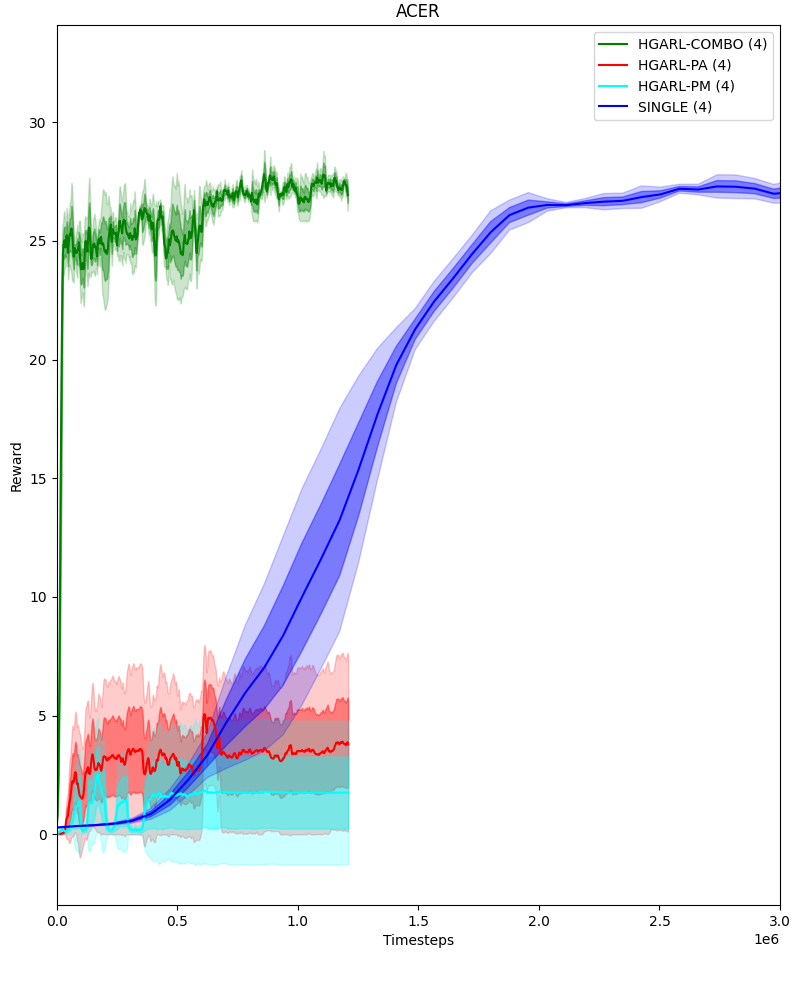}\hfill
                \includegraphics[width=0.333\linewidth]{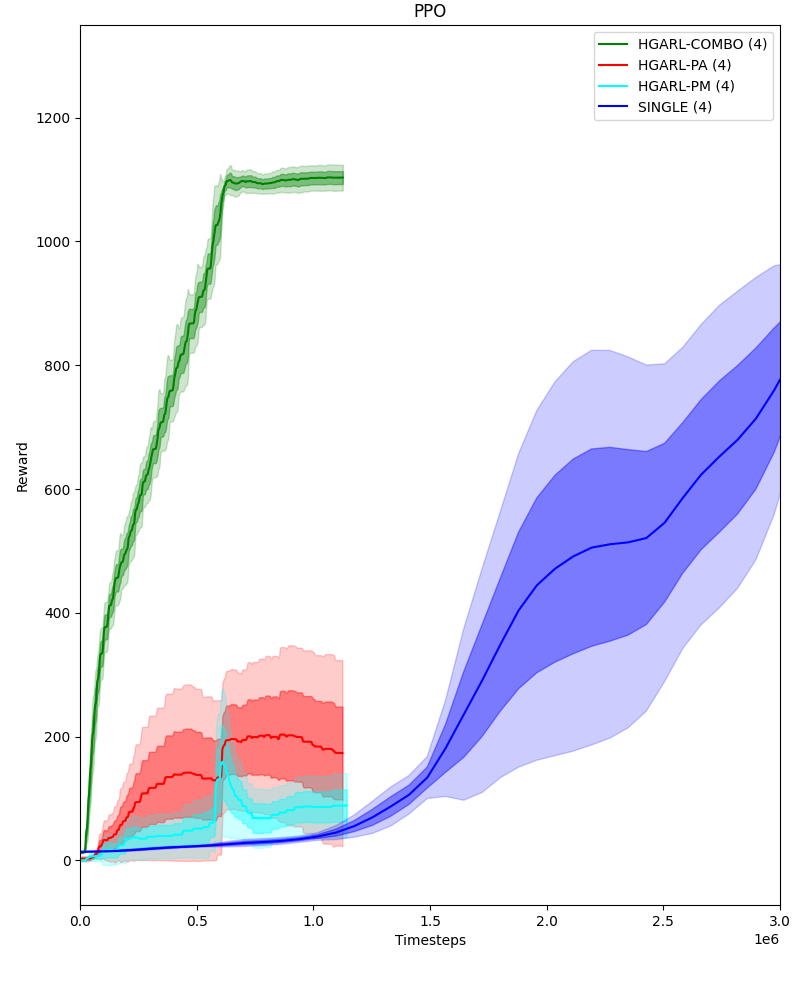}
                \caption{BankHeist}
         \end{subfigure}
         \begin{subfigure}[]{0.49\linewidth}
                \includegraphics[width=0.333\linewidth]{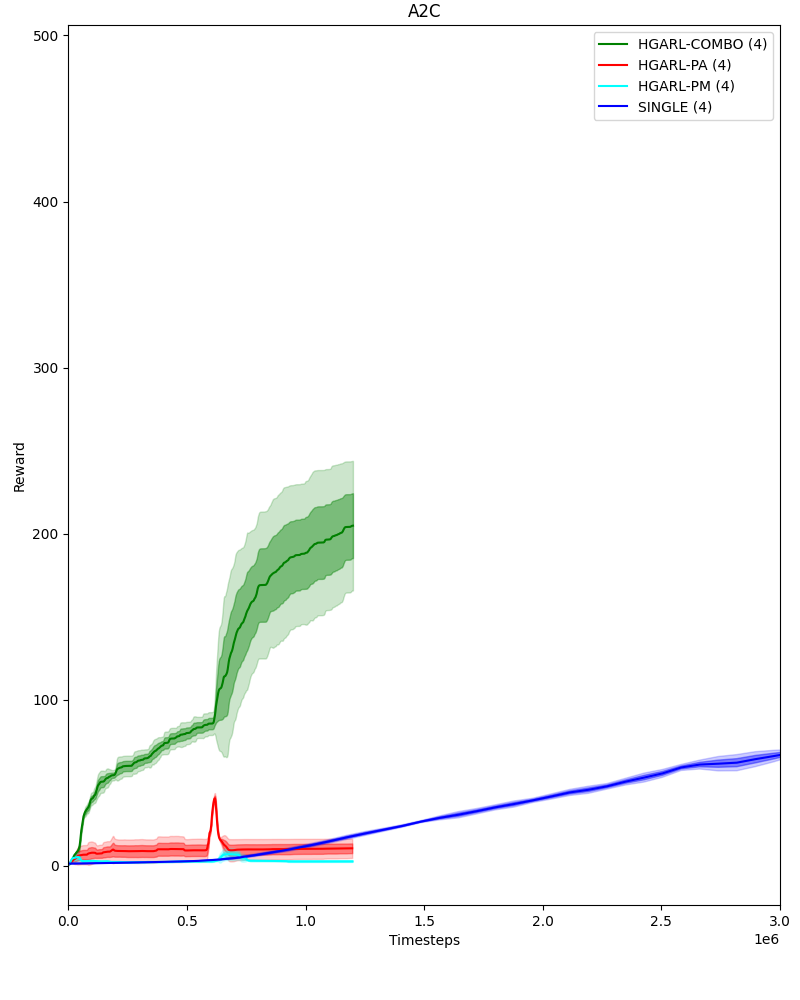}\hfill
                \includegraphics[width=0.333\linewidth]{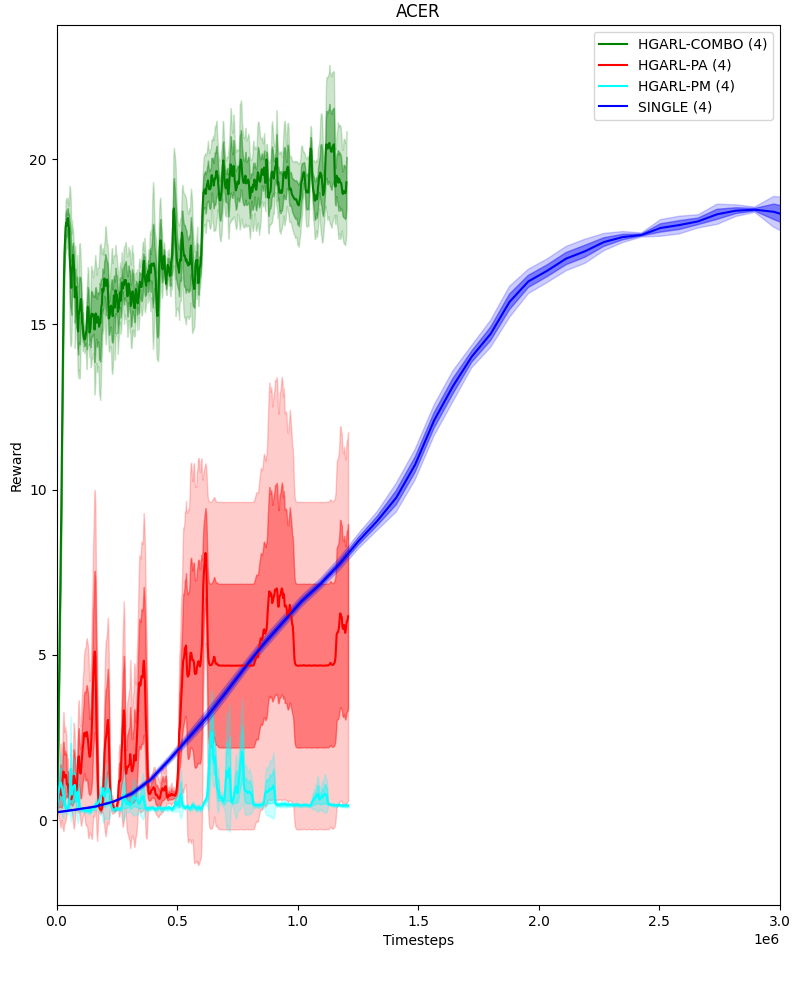}\hfill
                \includegraphics[width=0.333\linewidth]{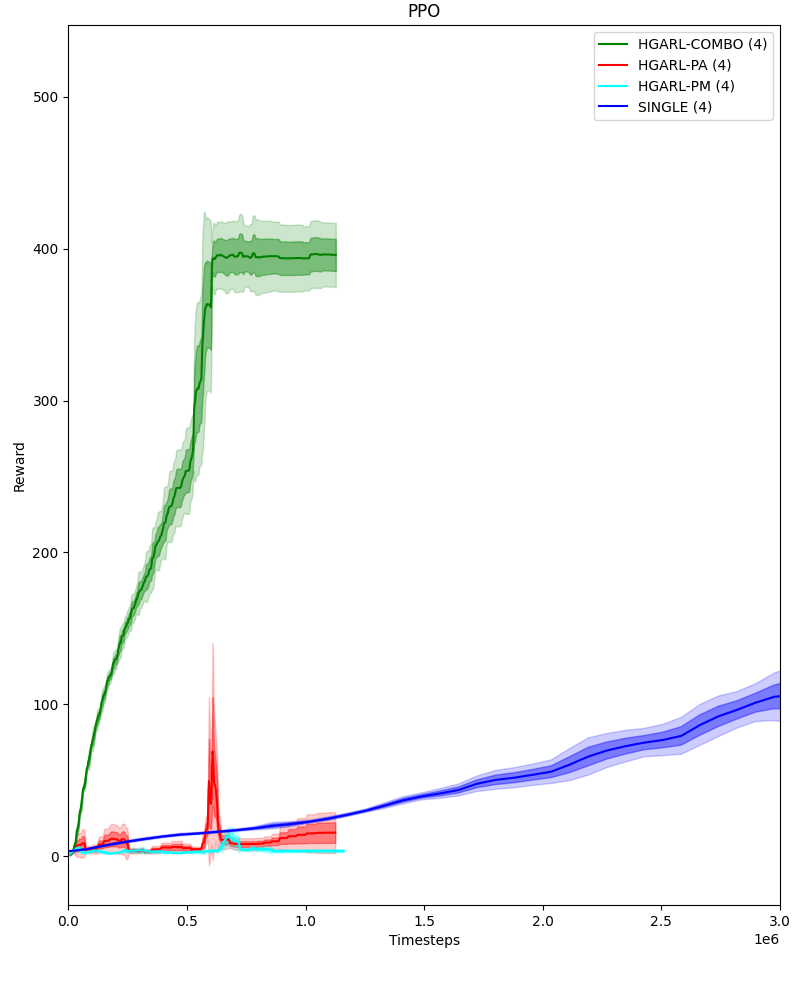}
                \caption{Breakout}
         \end{subfigure}
         \begin{subfigure}[]{0.49\linewidth}
                \includegraphics[width=0.333\linewidth]{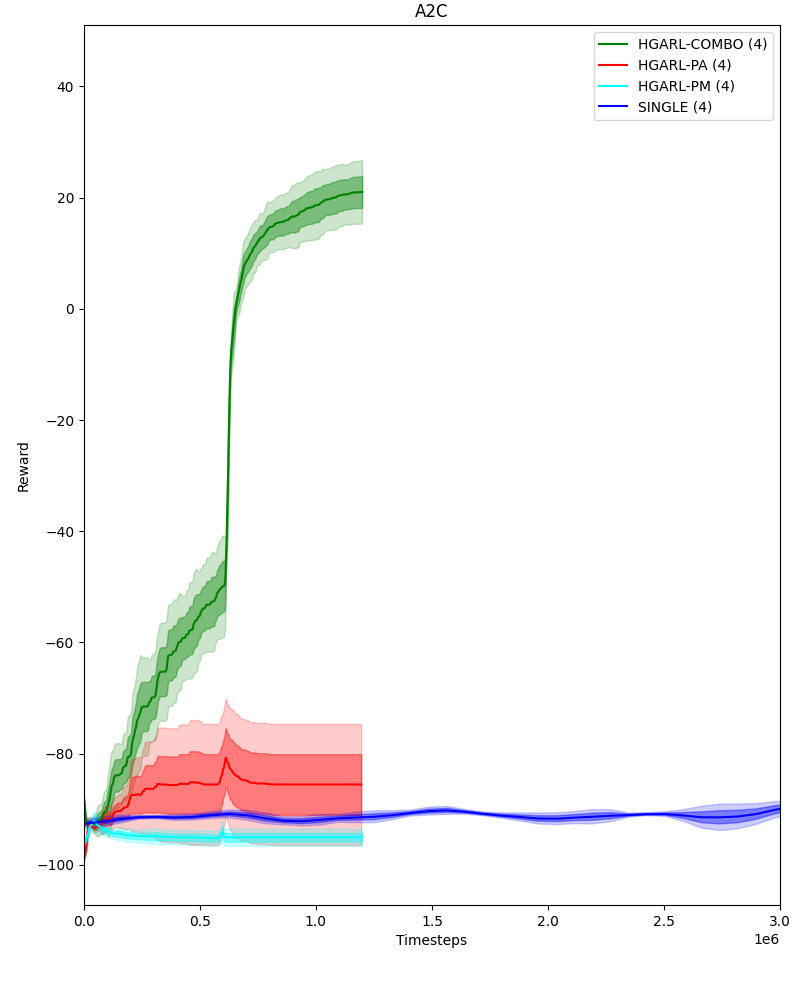}\hfill
                \includegraphics[width=0.333\linewidth]{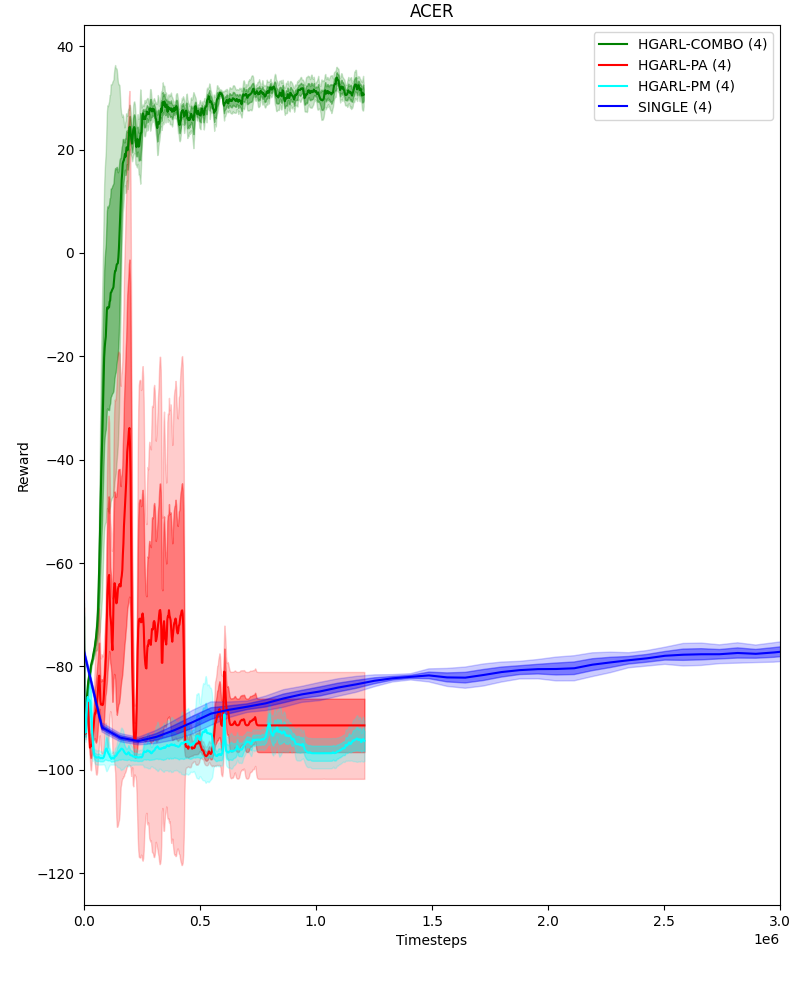}\hfill
                \includegraphics[width=0.333\linewidth]{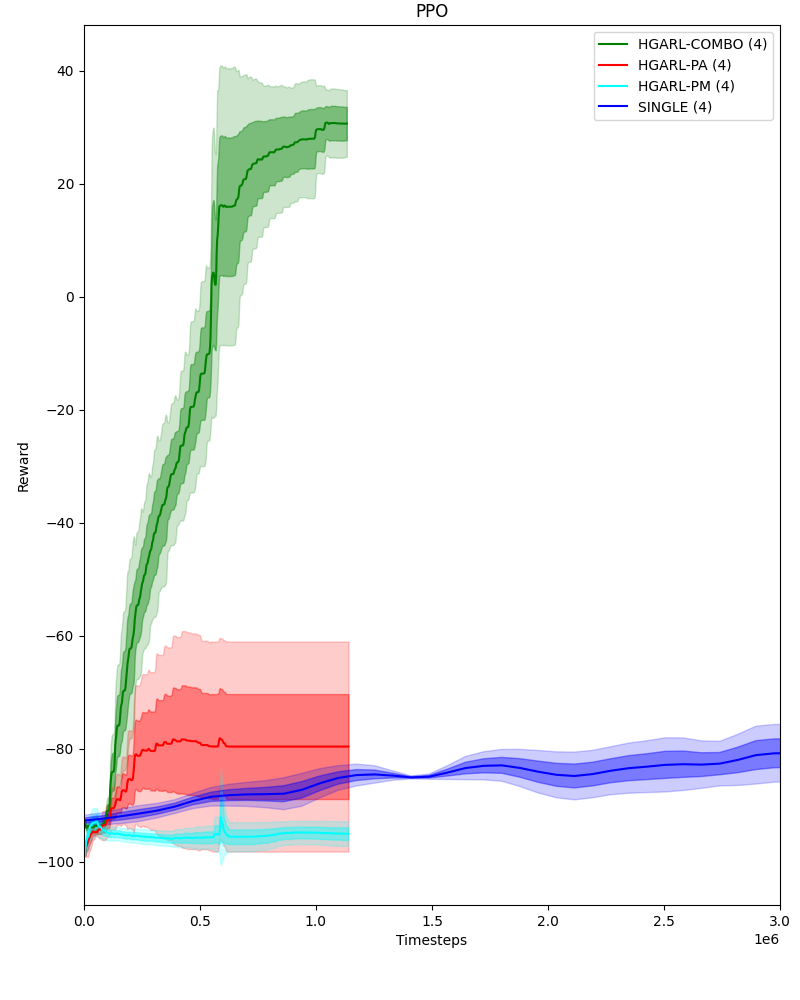}
                \caption{FishingDerby}
         \end{subfigure}

         \caption{Performance comparison for different agents and games in the first 3e6 time steps.}
        \label{Atari0}
\end{figure}

\subsection{Experiment Setting} 

We test intensively the proposed HGARL on a total of 43 different Atari 2600 games \citep{bellemare2013arcade}, and repeat the experiment  4 times for each game with 4 different seeds to obtain a more robust view on the learning performance. 
As explained in Section \ref{agent_sim}, we simulate a group of  3 agents  supported by the learning algorithms of A2C, PPO and ACER, respectively.  
The goal is to test whether and how much HGARL improves the learning performance of each agent by letting them learn in a group,  compared to learn on their own.
We experimented with all the three selection rules, i.e., PA, PM and Combo, for HGARL.
The experiments consume,  for each run,  64 CPU cores with a total memory of 384GB for at least 7 days, which accounts for  more than 10M core hours in total.

We used the A2C, PPO and ACER implementations by \cite{baselines} released under MIT license. 
The ACER implementation learns an action value function, while the A2C and PPO implementations learn a state value function, where convolutional neural networks consisting of three convolutional layers with ReLU activation  followed by one linear layer  are used to model the policy and value functions.
As a result, in the model adoption step, A2C and PPO agents can adopt both the policy  and value  networks (essentially their parameters)  from each other, but ACER agent  adopts only the policy network from A2C or PPO agent and vice versa. 
To implement the  Combo rule, A2C and PPO  predict  the state values used in  Eq. (\ref{state_value}) by their value networks,  while for  the ACER agent,  this value is computed from the predicted action probabilities by $V(s) = \sum_a \pi(a|s) \cdot Q(s,a)$, where $Q(s,a)$ is the action value, $V(s)$ is the state value and $\pi(a|s)$ is the policy. 
%
The batch size is set as   5 for A2C,    2048 for PPO and 20 for ACER,  which is in general a sensitive parameter affecting the learning performance. These settings are kept as the original settings in \citep{baselines} in order to get best tuned performance.
The threshold $\phi$ for the negative log likelihood is set as  $80\%$ of the initial negative log likelihood, which is fine-tuned.  

\begin{table}
  \centering
  \begin{tabular}{l|l|l|l|l}
    \toprule
    \multirow{2}{*}{Agent} &
    \multicolumn{3}{c}{Number of Games (Percentage)} \\
    \cmidrule(r){2-4}
     & $N_1$ & $N_2$ & $N_3$ \\
    \midrule
    ACER & 43/43(100\%) & 42/43(97.7\%) & 41/43(95.4\%) \\
    \hline
    A2C & 34/43(79.1\%) & 34/43(79.1\%) & 39/43(90.7\%) \\
    \hline
    PPO & 36/43(83.7\%) & 37/43(86.1\%) & 36/43(83.7\%) \\
    \bottomrule
  \end{tabular}
  \caption{Statistics: Percentages of the improved games in terms of accumulated reward score at time step $N_1$ = 2e5, $N_2$ = 5e5 and $N_3$ = 7e5. The total number of different games is 43 (Atari 2600 games). Note that for each agent type and test case (game), we report the performance of the rule that gives the best performance among the three rules. This is because the three rules are supposed to show performance strengths for different test cases, and we're happy as long as there is at least one rule that works well for each case. According to the table, we've achieved improvement of over 95\% of the tests for ACER agent at the three values of $N$ and over 83\% for PPO agent. For A2C agent, the initial improvement is a bit lower at 79\% but later rises to 90.7\%.}
   \label{tableN}
\end{table}

\begin{table}
  \centering
  \begin{tabular}{l|l|l|l|l}
    \toprule
    \multirow{2}{*}{Speed-up} &
    \multicolumn{3}{c}{Number of Games} \\
    \cmidrule(r){2-4}
     & A2C & ACER & PPO \\
    \midrule
    0-1x & 0 & 5 & 0 \\
    \hline
    1x-100x & 7 & 8 & 16  \\
    \hline
    100x-1000x & 12 & 1 & 12\\
    \hline
    >1000x & 6 & 0 & 9\\
    \hline
    inf & 18 & 29 & 6 \\
    \bottomrule
  \end{tabular}
  \caption{Statistics: Speed-up = $\frac{T}{T_G}$ where $T_G$ represents the number of time steps that the group agents took to achieve the highest performance and $T$ represents the number of time steps that the corresponding single agents took to reach the same performance. We classified Speed-up into 5 groups and counted the original statistics (presented in Table \ref{table2} in Appendix B) that belong to each group. The $inf$ value means that the single agents were never able to reach the same performance. Note that for each agent type and test case (game), we report the performance of the rule that gives the best performance among the three rules. This is because the three rules are supposed to show performance strengths for different test cases, and we're happy as long as there is at least one rule that works well for each case. According to the table, we've achieved learning speed-up for 96.12\% of the tests and improved reward for 41.09\% of the tests.}
   \label{table1}
\end{table}

\subsection{Results and  Analysis} 

In each plot, we  report the learning curve for one particular agent and one particular game, under three group learning setups corresponding to the three selection rules, and  under the independent setup of learning on its own.
We distinguish the four setups by  HGARL-PA, HGARL-PM, HGARL-COMBO and SINGLE in the plot legend.
Each learning curve has its x-axis as the time step which is the number of steps that a reinforcement learning agent has taken in its environment, while its y-axis as the accumulated reward score per episode which grows as the agent gains more knowledge to perform the given task better.
The four curves are distinguished  by colours in a plot.
Each curve is associated with two kinds of shade, where the  lighter  one shows the standard deviation of the accumulated reward,  while the darker one shows  the normalised standard deviation divided by the square root of the number of seeds.
 Our experiments study in total 3 agents for 43 games, resulting in a total of $3\times 43 =129$ plots.
 We display all the 129 plots through Fig. \ref{Atari1}, Fig. \ref{Atari2}, Fig. \ref{Atari3} and Fig. \ref{Atari4} which are placed in Appendix B because of page limit.

\subsubsection{On Learning Speed-up.} 

Limited by the computing resource, we stopped all the HGARL group training much earlier than the independent  training of a single agent. 
In all the plots, we compare HGARL learning curves in much fewer time steps to  the independent learning curves that run a lot more time steps.
It is worth highlighting that, even in the very early stage, the agents under group learning have already showed much better performance than the single agents. 
Agents in a group learn much faster. 
By continuing their learning with more time steps, we expect them to  improve further
%
To illustrate the accelerated learning speed, we compare in Fig. \ref{Atari0} the learning curves of the three agents for 4 games  by observing their first $N=3e6$ training steps. 
 %
 It can be seen that the group learning agents reached very good performance with the Combo rule  in a lot fewer time steps than the single-agent learning.

We further examine the  learning speed-up achieved by HGARL from two aspects, of which one is on the performance increase achieved within a fixed number of time steps, the other is on the reduction of time steps required to reach the best performance.
For performance increase, we record the value $\textmd{AR}_N$ 
as the accumulated reward score that a single agent and the same agent but learning in a group achieve, within $N$ times steps.
For time reduction, we define another quantity $r= \frac{T}{T_G}$, where $T_G$ represents the number of time steps that an agent in the group takes to achieve the highest performance, while $T$ represents the number of time steps that the same single agent takes to achieve the same performance. 
We have recorded the values of $r$ in Table \ref{table2} and $\textmd{AR}_N$ with three different values of $N$ ($N_1$ = 2e5, $N_2$ = 5e5, $N_3$ = 7e5) for each agent, game and rule in Table \ref{table4} for ACER agents, Table \ref{table3} for A2C agents and Table \ref{table5} for PPO agents, which are placed in Appendix B because of page limit.
For ease of reading, we present statistics of the three tables of $\textmd{AR}_N$ in Table \ref{tableN}. 
For each agent type, we count the number of different test cases where the agent gained better performance by group learning than learning as a single agent at the three different values of time steps $N$. We calculated percentages based on the fact that the total number of different test cases is 43 (Atari 2600 games). For ACER agent we get the best improvements of over 95\% at the three $N$s. For PPO agent we reached over 83\% improvement. For A2C agent the percentages are a bit lower in the beginning at around 79\%, but later improve to 90\% at $N_3$.
%
We then calculated statistics based on Table \ref{table2} and get Table \ref{table1} presenting the learning speed-up statistics. There are only 5 games where the ACER agents produced similar learning performance between group agents and single agents. For all the other cases (96.12\%), our group learning techniques achieved impressive speed-up. For 41\% of the tests we reached improved final reward score with less than 5\% of the time steps required for an agent in single-agent learning to reach same performance.

\begin{figure}
         \begin{subfigure}[b]{\linewidth}
                \includegraphics[width=0.48\linewidth]{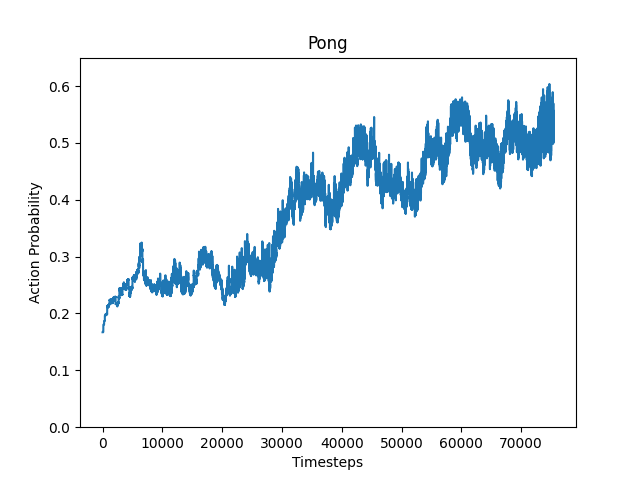}
                \includegraphics[width=0.48\linewidth]{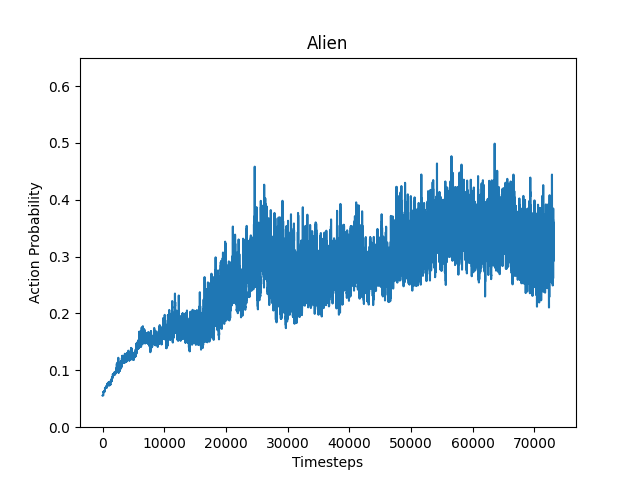}
        \end{subfigure}
         \caption{Action Probabilities: The probability that each possible action gets from agents' decisions. A higher probability means the agents think the corresponding action is the better one in the current situation. And an agent will usually choose an action with the highest probability under the current state.}
        \label{Probs}
\end{figure}  

\subsubsection{On Action Selection Rules.}  

Assisted by the action selection, the better-performing agents can help the other agents in a group by expediting their learning processes.
With proper guidance from better-performing peers, the time spent on random exploration in the initial stage of learning is greatly reduced.
%
%
For the games in Fig. \ref{Atari1}, all the three agents in the group system show superb performance under the Combo rule, which is fast and stable. For the games in Fig. \ref{Atari2}, the ACER agent benefits a lot from the other two agents in the group system under the Combo rule as well. For the games in Fig. \ref{Atari3}, the PA or PM rule shows outstanding performance for the three agents in group learning, a lot better than single-agent learning. For the rest games in Fig. \ref{Atari4}, the three rules produced similar performance which are still much better than single agents. 
In the scenarios where the Combo rule works best, the performance under the Combo rule is also very stable with no performance drops, while the PA and PM rules often introduce obvious performance fluctuations. Even for the scenarios where PA and PM rules work better, they sometimes also introduce big performance fluctuations. 
Looking deeper into the plots in Fig. \ref{Atari1}, the green curves represent the performance of the group learning agents under the Combo rule, while the blue curves represent the single agent performance. They show that the group agents under Combo rule reached better performance than single agents much faster -- the green curves go up higher than blue curves a lot earlier. 

We try to explain the different performance of the PA, PM and Combo rules through analysing the developing trend of the action probabilities of two representative games -- Pong and Alien, in Fig. \ref{Probs}. Pong is a typical case where the Combo rule shows superb performance. All the three agents of A2C, ACER and PPO in the group learn a lot faster with very stable performance compared to when they work as single agents. Alien is a typical case where the PA and PM rules produced very good learning performance but the Combo rule did not do very well. 

The figures in Fig. \ref{Probs} show the developing trend of the probability of the chosen actions at each time step during the learning process. The probabilities are average over 4 different seeds. 
It's obvious that the action probability of the Alien game fluctuates much more as it develops compared to the Pong game. Since the Combo rule has three barriers (the three steps) to filter out any possible bad action choices to the best of its knowledge, it turns out to be too conservative in terms of utilising the knowledge from other agents under the Alien game environment -- their action probabilities drop below the threshold and the rule determines them as bad action choices, which are in fact good enough. Instead the PA and PM rules provided impressive performance for Alien because they are much more aggressive when utilising other agents' knowledge -- they just directly use the probabilities without any filtering process. For Pong, because the fluctuations of the action probabilities are very minor, the Combo rule did not filter too much and made enough use of better actions from the other agents in the group, resulting in great performance, better than the PA and PM rules. 

The performance fluctuations and drops are because the intuitive PM and PA rules do not have a mechanism to avoid any bad influence from any agent that's in a bad learning status for any state. They always combine all the decisions of all agents in the group. The Combo rule has three barriers to filter out any possible bad action choices to the best of its knowledge, which prove to be very effective and generate stable performance. 



\section{Conclusion}

GARL is a new concept which identifies an unexplored area in reinforcement learning. 
This work extends the method in \citep{garl} which is restricted in a scenario where the agents are of the same type and in same environments to a much more advanced approach which works in the more general learning scenario with different types of agents in same environments. The evaluations show that we significantly improved the learning speed and quality of the agents in a GARL system compared to single-agent learning. 

%

\bibliography{iclr2025_conference}

\newpage

\onecolumn

\title{Appendix}
\maketitle

\appendix
\section{Selection of hyper parameter $\phi$}

Since we transform the original action probability into a negative log likelihood, the larger the probability, the smaller the negative log likelihood. Because a larger probability means that the agent is very confident in its action decisions, naturally we would prefer a larger probability, hence we would prefer lower likelihood. Therefore, we start with a reasonably low likelihood threshold $\phi_1$. However, as we mentioned in the paper, for quite some time we don’t need the agent to be too confident and a lower probability can be also acceptable. We don’t want to filter out these reasonable decisions that are with lower probabilities. To do this, we then try a reasonably large likelihood threshold $\phi_2$, and gradually narrow the interval [$\phi_1$, $\phi_2$] till locating a best likelihood threshold $\phi$ in the middle.

\section{Original statistics and figures}
 \begin{table*}[htbp]
\centering
\begin{tabular}{c|c|c|c|c|c|c|c|c|c|c|c|c}
 \hline
 & \multicolumn{4}{c}{$N_1$} & \multicolumn{4}{c}{$N_2$} & \multicolumn{4}{c}{$N_3$}    \\
 \cmidrule(lr){2-5} \cmidrule(lr){6-9} \cmidrule(lr){10-13} 
Game  & Single & PA & PM & Combo & Single & PA & PM & Combo & Single & PA & PM & Combo  \\ [0.5ex] 
 \midrule
 \hline
 Alien & 10.8 & 19.1 & 19.8 & 7.2 & 17.3 & 13.3 & 28.1 & 11.9 & 20.1 & 23.2 & 29.3 & 7.6 \\
 \hline
 Assault & 4.9 & 6.8 & 4.9 & 32.5 & 10.3 & 6.8 & 5.1 & 33.6 & 13.0 & 7.9 & 5.5 & 39.1 \\
 \hline
 Asterix & 2.2 & 24.3 & 14.0 & 28.0 & 5.9 & 27.6 & 14.1 & 31.9 & 7.7 & 33.7 & 25.9 & 33.5 \\
 \hline
 Asteroids & 5.3 & 6.1 & 5.7 & 5.8 & 5.2 & 6.0 & 6.3 & 6.1 & 5.0 & 6.1 & 6.6 & 5.7 \\
 \hline
 Atlantis & 5.3 & 283.2 & 1719.7 & 1180.9 & 7.0 & 69.7 & 873.1 & 1679.6 & 7.9 & 41.1 & 1045.1 & 1518.3 \\
 \hline
 BankHeist & 0.4 & 3.3 & 1.1 & 24.5 & 2.3 & 2.9 & 1.6 & 25.0 & 4.7 & 3.4 & 1.8 & 27.1 \\
 \hline
 BattleZone & 0.7 & 1.9 & 2.0 & 2.2 & 0.8 & 1.9 & 1.4 & 2.0 & 1.0 & 2.7 & 2.2 & 2.0  \\
 \hline
 BeamRider & 3.2 & 10.7 & 10.0 & 8.9 & 3.6 & 14.1 & 10.8 & 5.6 & 4.3 & 8.7 & 11.5 & 5.8 \\
 \hline
 Bowling & 7.5 & 6.6 & 7.4 & 9.8 & 7.5 & 10.0 & 5.0 & 9.6 & 7.7 & 7.5 & 7.5 & 9.3 \\
 \hline
 Boxing & 3.0 & -1.4 & 23.7 & 94.1 & 5.3 & -23.0 & 30.1 & 95.2 & 7.8 & -23.0 & 30.0 & 95.2 \\
 \hline
 Breakout & 0.6 & 2.3 & 0.8 & 16.3 & 2.5 & 1.2 & 0.4 & 16.9 & 3.9 & 4.7 & 1.0 & 19.1 \\
 \hline
 Centipede & 28.1 & 31.6 & 32.5 & 29.2 & 27.1 & 34.0 & 31.2 & 32.3 & 27.1 & 34.1 & 31.7 & 33.4 \\
 \hline
 CrazyClimber & 71.6 & 21.5 & 20.7 & 93.5 & 107.1 & 7.6 & 14.7 & 73.1 & 107.9 & 5.2 & 23.5 & 95.6 \\
 \hline
 DemonAttack & 5.4 & 11.5 & 7.5 & 17.6 & 12.1 & 7.8 & 6.2 & 16.9 & 13.6 & 8.4 & 10.9 & 17.1 \\
 \hline
 DoubleDunk & -18.0 & -6.8 & -19.6 & -16.5 & -17.6 & -7.0 & -7.3 & -15.8 & -17.3 & -7.0 & -7.3 & -15.8 \\
 \hline
 Enduro & 0.0 & 350.0 & 260.8 & 0.0 & 0.0 & 248.3 & 529.9 & 0.0 & 0.0 & 248.3 & 529.9 & 0.0 \\
 \hline
 FishingDerby & -94.5 & -42.9 & -97.2 & 23.5 & -89.1 & -95.1 & -94.3 & 26.9 & -87.8 & -91.4 & -94.6 & 30.7 \\
 \hline
 Freeway & 5.9 & 21.5 & 30.8 & 22.6 & 0.0003 & 23.8 & 28.1 & 23.2 & 0.002 & 23.8 & 28.1 & 23.2 \\
 \hline
 Frostbite & 5.3 & 16.5 & 21.0 & 4.2 & 6.3 & 15.4 & 11.2 & 5.1 & 6.5 & 15.4 & 11.2 & 5.1 \\
 \hline
 Gopher & 9.2 & 5.9 & 22.4 & 177.5 & 12.1 & 17.6 & 2.1 & 184.2 & 12.1 & 17.6 & 2.1 & 184.2 \\
 \hline
 Gravitar & 0.3 & 0.06 & 0.05 & 0.3 & 0.3 & 0.4 & 0.5 & 0.3 & 0.3 & 0.6 & 0.7 & 0.3 \\
 \hline
 IceHockey & -8.9 & -14.1 & -7.0 & -7.3 & -9.2 & -9.1 & -7.7 & -8.0 & -8.6 & -7.9 & -6.2 & -7.0 \\
 \hline
 Jamesbond & 0.1 & 0.6 & 1.3 & 1.4 & 0.1 & 1.2 & 1.6 & 1.3 & 0.1 & 1.3 & 1.7 & 1.4 \\
 \hline
 Kangaroo & 0.1 & 3.0 & 2.5 & 2.7 & 0.1 & 0.3 & 2.0 & 14.5 & 0.1 & 1.0 & 2.4 & 15.6 \\
 \hline
 Krull & 89.1 & 195.2 & 195.1 & 208.2 & 137.6 & 255.6 & 180.3 & 214.1 & 159.8 & 161.8 & 201.2 & 262.4 \\
 \hline
 KungFuMaster & 8.8 & 28.2 & 36.9 & 43.5 & 23.2 & 11.9 & 34.7 & 45.4 & 26.2 & 11.9 & 34.7 & 45.4 \\
 \hline
 MsPacman & 17.3 & 31.3 & 42.3 & 19.5 & 24.1 & 27.7 & 23.1 & 28.9 & 27.6 & 27.4 & 29.9 & 36.9 \\
 \hline
 NameThisGame & 25.6 & 68.8 & 64.2 & 138.0 & 43.0 & 80.8 & 68.9 & 133.1 & 53.6 & 91.3 & 49.3 & 154.9 \\
 \hline
 Pitfall & -1.7 & -1.2 & -3.3 & -1.0 & -1.0 & -0.8 & -2.6 & -0.8 & -0.9 & -0.8 & -2.6 & -0.8 \\
 \hline
 Pong & -20.2 & 18.2 & -1.4 & 20.1 & -18.5 & -2.2 & -8.9 & 20.5 & -15.4 & 2.0 & 9.0 & 20.7 \\
 \hline
 PrivateEye & -0.4 & 0.2 & 0.7 & 0.7 & -0.3 & 0.5 & 1.0 & 0.1 & -0.2 & 0.8 & 1.0 & 0.8 \\
 \hline
 Qbert & 3.2 & 2.3 & 1.5 & 16.7 & 5.4 & 6.1 & 1.5 & 19.0 & 6.8 & 6.6 & 1.5 & 15.7  \\
 \hline
 Riverraid & 7.5 & 10.5 & 12.1 & 29.6 & 9.1 & 11.9 & 17.1 & 35.9 & 9.5 & 11.9 & 17.1 & 35.9 \\
 \hline
 RoadRunner & 1.6 & 4.1 & 21.6 & 41.2 & 4.7 & 1.4 & 15.7 & 28.4 & 11.2 & 3.0 & 33.1 & 31.1 \\
 \hline
 Robotank & 0.6 & 2.5 & 2.0 & 1.4 & 0.6 & 2.8 & 2.1 & 1.1 & 0.5 & 2.8 & 2.1 & 1.1 \\
 \hline
 Seaquest & 6.2 & 16.6 & 17.7 & 20.9 & 8.4 & 7.5 & 19.8 & 17.3 & 9.1 & 9.5 & 37.3 & 16.5 \\
 \hline
 SpaceInvaders & 3.9 & 21.1 & 17.1 & 22.8 & 5.8 & 15.0 & 10.2 & 30.0 & 7.7 & 20.4 & 15.2 & 38.5 \\
 \hline
 StarGunner & 1.0 & 1.4 & 5.0 & 15.2 & 1.7 & 1.5 & 0.4 & 17.6 & 2.0 & 1.5 & 0.4 & 17.6 \\
 \hline
 TimePilot & 2.5 & 4.0 & 1.2 & 3.5 & 2.5 & 2.1 & 1.3 & 3.3 & 2.5 & 2.0 & 1.7 & 4.0 \\
 \hline
 UpNDown & 8.2 & 10.9 & 17.1 & 36.2 & 20.8 & 8.5 & 12.4 & 29.9 & 27.3 & 10.5 & 25.2 & 16.1 \\
 \hline
 VideoPinball & 43.8 & 74.6 & 44.5 & 154.4 & 43.5 & 210.2 & 75.1 & 174.8 & 42.6 & 186.3 & 53.2 & 302.8 \\
 \hline
 WizardOfWor & 1.8 & 3.5 & 4.4 & 2.2 & 1.9 & 2.1 & 3.9 & 2.6 & 1.9 & 3.2 & 4.4 & 2.6 \\
 \hline
 Zaxxon & 0.009 & 0.004 & 0.1 & 1.1 & 0.01 & 0.1 & 0.6 & 2.1 & 0.01 & 0.1 & 0.6 & 2.1 \\ [1ex]  
 \hline
\end{tabular}
\caption[htbp]{The accumulated reward scores for the ACER agent at time steps $N_1$ = 2e5, $N_2$ = 5e5, $N_3$ = 7e5.}
\label{table4}
\end{table*}

\begin{table*}[htbp]
\centering
\begin{tabular}{c|c|c|c|c|c|c|c|c|c|c|c|c}
 \hline
 & \multicolumn{4}{c}{$N_1$} & \multicolumn{4}{c}{$N_2$} & \multicolumn{4}{c}{$N_3$}    \\
 \cmidrule(lr){2-5} \cmidrule(lr){6-9} \cmidrule(lr){10-13} 
 Game  & Single & PA & PM & Combo & Single & PA & PM & Combo & Single & PA & PM & Combo  \\ [0.5ex] 
 \midrule
 \hline
 Alien & 279.3 & 858.5 & 828.1 & 243.9 & 296.7 & 862.7 & 848.4 & 269.3 & 300.0 & 859.6 & 1173.0 & 273.7 \\
 \hline
 Assault & 313.4 & 410.9 & 405.1 & 262.7 & 396.8 & 460.5 & 405.1 & 269.9 & 475.7 & 638.1 & 448.8 & 271.2 \\
 \hline
 Asterix & 239.3 & 1.4e3 & 2.1e3 & 249.7 & 251.0 & 2.0e3 & 2.3e3 & 252.4 & 283.8 & 1.2e4 & 6.8e3 & 251.7 \\
 \hline
 Asteroids & 1.1e3 & 853.0 & 854.3 & 859.2 & 1.3e3 & 1.0e3 & 1.1e3 & 884.0 & 1.3e3 & 1.1e3 & 1.6e3 & 1.3e3 \\
 \hline
 Atlantis & 1.7e4 & 3.9e4 & 1.9e4 & 1.7e4 & 1.9e4 & 3.6e4 & 3.2e4 & 1.6e4 & 2.0e4 & 4.6e4 & 4.6e4 & 1.8e4 \\
 \hline
 BankHeist & 14.0 & 48.1 & 17.9 & 476.8 & 17.3 & 70.3 & 25.0 & 558.0 & 26.2 & 177.7 & 84.8 & 726.7 \\
 \hline
 BattleZone & 4.2e3 & 3.1e3 & 3.9e3 & 4.1e3 & 4.3e3 & 4.6e3 & 4.5e3 & 4.3e3 & 4.2e3 & 1.6e3 & 1.1e4 & 4.3e3  \\
 \hline
 BeamRider & 398.9 & 943.6 & 881.8 & 380.6 & 403.9 & 1103.5 & 946.8 & 399.4 & 413.9 & 1.5e3 & 1.3e3 & 476.0 \\
 \hline
 Bowling & 23.9 & 21.7 & 10.6 & 23.7 & 24.1 & 21.5 & 13.9 & 24.1 & 23.9 & 22.6 & 30.0 & 25.7 \\
 \hline
 Boxing & 1.1 & -3.8 & 15.9 & 68.6 & 1.0 & -6.1 & 19.8 & 78.5 & 0.8 & -6.1 & 19.8 & 78.5 \\
 \hline
 Breakout & 2.0 & 9.1 & 2.4 & 57.1 & 3.0 & 9.3 & 2.7 & 80.0 & 4.8 & 9.6 & 6.9 & 136.6 \\
 \hline
 Centipede & 3.4e3 & 3.8e3 & 4.3e3 & 2.8e3 & 2.9e3 & 4.1e3 & 4.5e3 & 3.2e3 & 2.8e3 & 5.0e3 & 4.8e3 & 4.1e3 \\
 \hline
 CrazyC & 3.0e4 & 2.1e4 & 1.9e4 & 1.0e4 & 5.4e4 & 2.2e4 & 2.0e4 & 1.3e4 & 6.1e4 & 2.2e4 & 2.0e4 & 2.0e4 \\
 \hline
 DemonA & 173.0 & 1.2e3 & 1.0e3 & 164.1 & 316.1 & 1278.6 & 1.0e3 & 161.6 & 483.3 & 1531.3 & 1204.5 & 163.4 \\
 \hline
 DoubleD & -18.4 & -22.7 & -23.1 & -17.7 & -18.2 & -22.9 & -20.5 & -17.3 & -18.3 & -22.9 & -20.5 & -17.3 \\
 \hline
 Enduro & 0.0 & 75.6 & 87.6 & 0.0 & 0.0 & 93.3 & 128.0 & 0.0 & 0.0 & 93.3 & 128.0 & 0.0 \\
 \hline
 FishingD & -91.5 & -87.7 & -94.7 & -80.2 & -91.1 & -85.3 & -95.1 & -54.8 & -91.1 & -84.9 & -95.1 & 8.6 \\
 \hline
 Freeway & 0.002 & 24.9 & 27.0 & 0.0 & 0.5e-3 & 24.4 & 27.8 & 0.0 & 0.003 & 24.4 & 27.8 & 0.0 \\
 \hline
 Frostbite & 132.2 & 360.0 & 232.3 & 94.1 & 195.6 & 377.6 & 277.4 & 139.7 & 200.9 & 377.6 & 277.4 & 139.7 \\
 \hline
 Gopher & 423.8 & 401.2 & 467.5 & 407.0 & 543.1 & 486.1 & 503.7 & 416.3 & 612.0 & 486.1 & 503.7 & 416.3 \\
 \hline
 Gravitar & 216.3 & 43.9 & 48.5 & 205.6 & 209.8 & 83.6 & 92.4 & 215.8 & 212.5 & 260.9 & 385.7 & 215.6 \\
 \hline
 IceHockey & -10.0 & -10.4 & -6.8 & -9.4 & -10.4 & -10.4 & -7.3 & -9.2 & -10.4 & -8.9 & -4.9 & -8.8 \\
 \hline
 Jamesbond & 34.0 & 85.7 & 111.4 & 33.5 & 31.4 & 98.3 & 150.2 & 35.9 & 29.1 & 306.9 & 393.2 & 35.8 \\
 \hline
 Kangaroo & 49.4 & 204.8 & 541.0 & 43.2 & 46.4 & 478.3 & 668.7 & 42.5 & 46.0 & 656.6 & 1288.6 & 43.5 \\
 \hline
 Krull & 2.5e3 & 3.4e3 & 4.8e3 & 2.3e3 & 2.7e3 & 4.1e3 & 5.6e3 & 2.3e3 & 2.9e3 & 4.5e3 & 5.7e3 & 2.4e3 \\
 \hline
 KungFuM & 2.2e3 & 1.4e3 & 818.1 & 490.4 & 6.1e3 & 1.9e3 & 2.4e3 & 555.2 & 7836.8 & 1.9e3 & 2.4e3 & 555.2 \\
 \hline
 MsPacman & 564.1 & 975.2 & 1129.7 & 447.9 & 628.9 & 1.0e3 & 1.2e3 & 491.7 & 670.2 & 1.0e3 & 1.4e3 & 502.5 \\
 \hline
 NameThisG & 2.3e3 & 3e3 & 2.6e3 & 2.1e3 & 2.3e3 & 3.2e3 & 2.8e3 & 2.2e3 & 2.3e3 & 4.1e3 & 2.0e3 & 2.2e3 \\
 \hline
 Pitfall & -205.8 & -89.4 & -114.9 & -183.1 & -193.4 & -84.8 & -97.7 & -178.6 & -183.5 & -84.8 & -97.7 & -178.6 \\
 \hline
 Pong & -20.3 & -15.2 & -10.0 & 1.0 & -20.2 & -11.3 & -10.2 & 4.6 & -20.1 & -7.8 & -0.1 & 15.1 \\
 \hline
 PrivateE & 21.2 & 28.1 & -167.4 & 14.6 & 47.7 & 29.3 & -136.8 & 30.5 & 43.3 & 89.9 & 90.2 & 36.6 \\
 \hline
 Qbert & 248.1 & 422.9 & 269.7 & 203.5 & 309.3 & 304.9 & 171.2 & 249.9 & 326.8 & 606.0 & 224.6 & 280.4  \\
 \hline
 Riverraid & 1.7e3 & 699.5 & 1.4e3& 1.4e3& 2.1e3 & 1.0e3 & 1.9e3 & 1.4e3 & 2.1e3 & 1.0e3 & 1.9e3 & 1.4e3 \\
 \hline
 RoadR & 79.5 & 230.0 & 4.0e3 & 6.7e3 & 640.7 & 156.8 & 4.3e3 & 9.9e3 & 971.2 & 1.8e3 & 1.8e4 & 1.1e4 \\
 \hline
 Robotank & 2.3 & 5.3 & 7.3 & 2.3 & 2.1 & 5.8 & 7.3 & 2.2 & 2.1 & 5.8 & 7.3 & 2.2 \\
 \hline
 Seaquest & 330.0 & 364.8 & 521.9 & 111.7 & 491.5 & 380.6 & 619.4 & 135.1 & 531.0 & 407.1 & 2.2e3 & 196.4 \\
 \hline
 SpaceInv & 227.0 & 496.9 & 411.2 & 155.0 & 234.7 & 552.0 & 451.1 & 158.9 & 220.5 & 782.3 & 565.9 & 168.4 \\
 \hline
 StarGun & 636.9 & 1.3e3 & 1.4e3 & 674.1 & 790.2 & 1.3e3 & 1.7e3 & 667.1 & 932.3 & 1254.9 & 1697.3 & 667.1 \\
 \hline
 TimePilot & 3.5e3 & 2.4e3 & 1.6e3 & 3.4e3 & 3.4e3 & 2.7e3 & 1.6e3 & 3.4e3 & 3.5e3 & 3.0e3 & 2.1e3 & 3.6e3 \\
 \hline
 UpNDown & 870.5 & 1.2e3& 1.8e3 & 570.0 & 1.3e3 & 1.2e3 & 1.8e3 & 771.8 & 1.5e3 & 1.5e3 & 5.5e3 & 996.4 \\
 \hline
 VideoPin & 2.2e4 & 1.1e4 & 1.4e4 & 2.0e4 & 2.4e4 & 1.4e4 & 1.4e4 & 2.0e4 & 2.3e4 & 2.8e4 & 3.0e4 & 2.1e4 \\
 \hline
 WizardOfW & 777.6 & 660.5 & 821.4 & 684.6 & 726.8 & 822.5 & 1079.5 & 721.7 & 750.1 & 988.9 & 1.8e3 & 819.5 \\
 \hline
 Zaxxon & 19.0 & 8.6 & 12.3 & 19.4 & 20.2 & 6.7 & 142.8 & 20.3 & 26.1 & 6.7 & 142.8 & 20.3 \\ [1ex]  
 \hline
\end{tabular}
\caption[htbp]{The accumulated reward scores for the A2C agent at time steps $N_1$ = 2e5, $N_2$ = 5e5, $N_3$ = 7e5.}
\label{table3}
\end{table*}

\begin{table*}[htbp]
\centering
\begin{tabular}{c|c|c|c|c|c|c|c|c|c|c|c|c}
 \hline
 & \multicolumn{4}{c}{$N_1$} & \multicolumn{4}{c}{$N_2$} & \multicolumn{4}{c}{$N_3$}    \\
 \cmidrule(lr){2-5} \cmidrule(lr){6-9} \cmidrule(lr){10-13} 
 Game  & Single & PA & PM & Combo & Single & PA & PM & Combo & Single & PA & PM & Combo  \\ [0.5ex] 
 \midrule
 \hline
 Alien & 367.5 & 910.5 & 1.0e3 & 241.2 & 500.1 & 800.9 & 1.0e3 & 249.6 & 553.9 & 773.5 & 1251.3 & 256.7 \\
 \hline
 Assault & 625.7 & 450.4 & 398.4 & 324.9 & 1.1e3 & 598.8 & 417.6 & 346.3 & 1.3e3 & 662.0 & 447.0 & 368.5 \\
 \hline
 Asterix & 481.3 & 2.1e3 & 2.7e3 & 230.8 & 848.4 & 5.5e3 & 3.1e3 & 235.0 & 951.1 & 1.0e4 & 7.7e3 & 258.2 \\
 \hline
 Asteroids & 1.3e3 & 1.2e3 & 1.3e3 & 1.1e3 & 1.3e3 & 1.4e3 & 1.4e3 & 1.3e3 & 1.3e3 & 1.4e3 & 1.6e3 & 1.3e3 \\
 \hline
 Atlantis & 3.5e4 & 4.1e4 & 1.7e4 & 2.1e4 & 6.7e4 & 4.4e4 & 3.3e4 & 2.3e4 & 7.8e4 & 4.6e4 & 5.4e4 & 2.7e4 \\
 \hline
 BankHeist & 16.9 & 69.8 & 35.1 & 502.2 & 23.8 & 138.1 & 51.2 & 896.3 & 28.0 & 194.3 & 87.5 & 1.1e3 \\
 \hline
 BattleZone & 4.9e3 & 4.3e3 & 4.9e3 & 4.2e3 & 7.1e3 & 7.3e3 & 7.0e3 & 4.2e3 & 7.8e3 & 1.5e4 & 1.4e4 & 5.7e3  \\
 \hline
 BeamRider & 473.3 & 1.1e3 & 949.2 & 415.0 & 516.2 & 1.3e3 & 1.1e3 & 457.9 & 543.5 & 1.6e3 & 1.4e3 & 511.4 \\
 \hline
 Bowling & 17.4 & 21.3 & 11.4 & 23.6 & 20.4 & 22.6 & 16.7 & 23.3 & 24.5 & 23.0 & 29.3 & 24.5 \\
 \hline
 Boxing & 1.0 & 0.1 & 31.0 & 95.7 & 4.7 & -6.8 & 60.2 & 98.5 & 6.9 & -6.8 & 60.2 & 98.5 \\
 \hline
 Breakout & 9.4 & 11.3 & 2.4 & 129.7 & 15.1 & 5.4 & 2.8 & 253.8 & 17.2 & 8.2 & 10.3 & 394.7 \\
 \hline
 Centipede & 2.7e3 & 4.2e3 & 4.3e3 & 3.2e3 & 3.0e3 & 4.4e3 & 4.2e3 & 4.2e3 & 2.9e3 & 4.9e3 & 4.8e3 & 4.9e3 \\
 \hline
 CrazyClimber & 3.1e4 & 1.4e4 & 1.7e4 & 1.0e4 & 5.0e4 & 1.2e4 & 1.6e4 & 1.3e4 & 5.7e4 & 1.2e4 & 1.5e4 & 2.8e4 \\
 \hline
 DemonAttack & 264.6 & 1.9e3 & 1.3e3 & 168.8 & 454.3 & 1.8e3 & 1.2e3 & 167.8 & 538.2 & 1.7e3 & 1.4e3 & 202.5 \\
 \hline
 DoubleDunk & -18.7 & -23.4 & -9.3 & -17.3 & -18.2 & -23.3 & -9.3 & -16.6 & -18.0 & -23.3 & -9.3 & -16.6 \\
 \hline
 Enduro & 4.6 & 88.6 & 107.8 & 0.0 & 27.9 & 88.6 & 163.4 & 0.0 & 45.9 & 88.6 & 163.4 & 0.0 \\
 \hline
 FishingDerby & -91.5 & -85.4 & -95.3 & -62.2 & -88.5 & -78.9 & -95.8 & -14.2 & -88.1 & -79.6 & -95.6 & 21.1 \\
 \hline
 Freeway & 8.8 & 22.8 & 27.0 & 11.0 & 11.6 & 22.8 & 28.2 & 11.0 & 13.6 & 22.8 & 28.2 & 11.0 \\
 \hline
 Frostbite & 222.7 & 703.6 & 210.2 & 151.3 & 251.6 & 830.0 & 210.2 & 151.3 & 254.1& 830.0 & 210.2 & 151.3 \\
 \hline
 Gopher & 566.6 & 531.6 & 783.7 & 391.4 & 727.5 & 531.6 & 860.6 & 381.0 & 772.1 & 531.6 & 860.6 & 381.0 \\
 \hline
 Gravitar & 214.5 & 54.7 & 37.5 & 231.8 & 204.9 & 221.1 & 263.8 & 244.6 & 197.5 & 361.6 & 491.7 & 242.8 \\
 \hline
 IceHockey & -8.3 & -12.0 & -7.6 & -8.5 & -7.6 & -10.6 & -7.4 & -9.1 & -7.5 & -9.2 & -5.4 & -8.8  \\
 \hline
 Jamesbond & 81.5 & 135.2 & 186.9 & 41.3 & 221.0 & 126.3 & 326.6 & 47.5 & 268.5 & 318.7 & 446.4 & 63.7 \\
 \hline
 Kangaroo & 239.0 & 425.2 & 817.4 & 49.0 & 510.9 & 1.3e3 & 773.3 & 60.0 & 703.5 & 1.3e3 & 1.3e3 & 81.0 \\
 \hline
 Krull & 3.3e3 & 4.2e3 & 5.4e3 & 2.6e3 & 4.9e3 & 4.9e3 & 6.6e3 & 2.8e3 & 5.6e3 & 4.8e3 & 6.6e3 & 3.0e3 \\
 \hline
 KungFuMaster & 6.8e3 & 1.9e3 & 2.3e3 & 1.5e3 & 1.2e4 & 3.9e3 & 2.3e3 & 1.5e3 & 1.3e4 & 3.9e3 & 2.3e3 & 1.5e3 \\
 \hline
 MsPacman & 680.8 & 1.1e3 & 1.5e3 & 354.5 & 721.9 & 1.1e3 & 1.6e3 & 386.9 & 812.0 & 932.4 & 1.4e3 & 430.3 \\
 \hline
 NameThisGame & 2.5e3 & 3.4e3 & 2.7e3 & 2.4e3 & 3.4e3 & 4.1e3 & 3.0e3 & 2.4e3 & 4.0e3 & 4.6e3 & 3.0e3 & 2.4e3 \\
 \hline
 Pitfall & -51.6 & -45.9 & -69.2 & -93.8 & -58.9 & -45.9 & -61.4 & -65.8 & -36.5 & -45.9 & -61.4 & -65.8 \\
 \hline
 Pong & -20.2 & -11.4 & -7.9 & -0.9 & -17.3 & -1.9 & -8.0 & 6.8 & -14.2 & -2.1 & -0.6 & 10.4 \\
 \hline
 PrivateEye & 40.5 & -42.1 & -38.8 & 44.4 & 46.9 & 0.3 & 9.7 & 47.6 & 37.8 & 50.7 & 100.0 & 56.8 \\
 \hline
 Qbert & 411.6 & 219.8 & 160.3 & 290.6 & 618.8 & 764.7 & 143.8 & 278.2 & 728.0 & 1.2e3 & 153.3 & 292.5  \\
 \hline
 Riverraid & 2.2e3 & 906.2 & 2.3e3 & 1.3e3 & 2.4e3 & 2.3e3 & 2.3e3 & 1.6e3 & 2.4e3 & 2.3e3 & 2.3e3 & 1.6e3 \\
 \hline
 RoadRunner & 3.0e3 & 0.005 & 7.2e3 & 8.6e3 & 9.9e3 & 406.8 & 1.3e4 & 1.5e4 & 1.2e4 & 4.1e3 & 1.6e4 & 1.6e4 \\
 \hline
 Robotank & 2.7 & 4.9 & 5.5 & 2.3 & 3.0 & 6.9 & 5.5 & 2.3 & 3.2 & 6.9 & 5.5 & 2.3 \\
 \hline
 Seaquest & 495.6 & 439.4 & 677.6 & 279.4 & 639.6 & 521.8 & 992.7 & 398.4 & 677.1 & 549.0 & 2.2e3 & 462.8 \\
 \hline
 SpaceInvaders & 245.8 & 603.3 & 487.2 & 161.5 & 299.4 & 753.5 & 610.0 & 181.5 & 325.4 & 853.2 & 637.1 & 206.7 \\
 \hline
 StarGunner & 1.2e3 & 1.4e3 & 1.8e3 & 670.4 & 1.5e3 & 1.4e3 & 1.8e3 & 790.3 & 1.7e3 & 1.4e3 & 1.8e3 & 790.3 \\
 \hline
 TimePilot & 3.5e3 & 2.9e3 & 1.8e3 & 3.6e3 & 3.3e3 & 4.0e3 & 1.7e3 & 3.5e3 & 3.3e3 & 4.1e3 & 2.6e3 & 3.6e3 \\
 \hline
 UpNDown & 3.2e3 & 865.1 & 1.7e3 & 1.5e3 & 8.4e3 & 758.9 & 1.7e3 & 1.4e3 & 1.2e4 & 890.4 & 5.1e3 & 1.6e3 \\
 \hline
 VideoPinball & 2.2e4 & 1.4e4 & 1.3e4 & 2.3e4 & 2.4e4 & 2.8e4 & 2.3e4 & 2.4e4 & 2.4e4 & 5.9e4 & 4.7e4 & 2.4e4 \\
 \hline
 WizardOfWor & 901.2 & 887.7 & 910.2 & 757.2 & 983.6 & 1.5e3 & 1.9e3 & 752.5 & 1.1e3 & 1.5e3 & 2.2e3 & 871.4 \\
 \hline
 Zaxxon & 10.1 & 24.8 & 4.3 & 33.1 & 45.8 & 24.5 & 468.0 & 134.5 & 24.6 & 24.5 & 468.0 & 134.5 \\ [1ex]  
 \hline
\end{tabular}
\caption[htbp]{The accumulated reward scores for the PPO agent at time steps $N_1$ = 2e5, $N_2$ = 5e5, $N_3$ = 7e5.}
\label{table5}
\end{table*}

\begin{table*}[htbp]
\centering
\begin{tabular}{c|c|c|c|c|c|c|c|c|c}
 \multirow{3}{*}{Game} 
 & \multicolumn{9}{c}{Speed-up} \\
 \hline
 & \multicolumn{3}{c}{PA} & \multicolumn{3}{c}{PM} & \multicolumn{3}{c}{Combo}    \\
 \cmidrule(lr){2-4} \cmidrule(lr){5-7} \cmidrule(lr){8-10} 
 & A2C & ACER & PPO & A2C & ACER & PPO & A2C & ACER & PPO \\ [0.5ex] 
 \midrule
 \hline
 Alien & 5502.05 & 0.54 & 882.56 & 3018.21 & 0.897 & 1950.36 & 8.61 & 0.33 & 25.86 \\
 \hline
 Assault & 237.31 & 4.79 & 47.29 & 35.46 & 0.66 & 5.78 & 1.47 & inf & 0.88 \\
 \hline
 Asterix & inf & inf & inf & inf & inf & inf & 230 & inf & 0.95 \\
 \hline
 Asteroids & 8.25 & 0.73 & 97.43 & inf & 5.21 & 1502.12 & 112.69 & 0.88 & 689.88 \\
 \hline
 Atlantis & 160.35 & 0.92 & 6.898 & 173.08 & inf & 31.70 & 80.07 & inf & 1.51 \\
 \hline
 BankHeist & 35.796 & 0.04 & 16.97 & 129.05 & 0.06 & 88.33 & 47.97 & 0.31 & 233.3 \\
 \hline
 BattleZone & inf & inf & 592.48 & inf & inf & 169.84 & 251.06 & 0.26 & 9.48  \\
 \hline
 BeamRider & 553.71 & inf & 998.69 & 88.71 & inf & 147.98 & 212.98 & 1.22 & 17.37 \\
 \hline
 Bowling & 754.85 & 111.45 & 31.93 & 171.12 & 0.43 & 3.90 & inf & 0.98 & 6.42 \\
 \hline
 Boxing & 0.19 & 0.35 & 19.74 & 219.49 & 0.38 & inf & 331.91 & inf & inf \\
 \hline
 Breakout & 109.38 & 0.06 & 612.34 & 20.51 & 0.18 & 15.01 & 76.77 & 1.49 & 1784.22 \\
 \hline
 Centipede & inf & 1.88 & inf & inf & 0.52 & inf & inf & 0.42 & inf \\
 \hline
 CrazyClimber & 4.02 & 0.31 & 1.23 & 5.40 & 1.01 & 3.71 & 4.49 & 1.65 & 2.198 \\
 \hline
 DemonAttack & 54.56 & inf & 68.97 & 34.397 & 3.98 & 58.2 & 3.18 & inf & 1.97 \\
 \hline
 DoubleDunk & 0.17 & nan & 0.06 & 174.65 & inf & inf & 534.85 & nan & 54.15 \\
 \hline
 Enduro & inf & inf & 13.61 & inf & inf & 20.43 & 1.04 & nan & 0.81 \\
 \hline
 FishingDerby & 320.05 & inf & 40.76 & 13.10 & nan & 616.14 & 358.23 & 0.77 & 2077.55 \\
 \hline
 Freeway & inf & inf & 41.99 & inf & inf & 18.69 & 1.22 & inf & 11.09 \\
 \hline
 Frostbite & inf & 1.95 & 199.18 & inf & 0.52 & 126.87 & 2.85 & 2.63 & 0.64 \\
 \hline
 Gopher & 12.94 & 0.32 & 16.82 & 18.62 & 0.28 & 22.34 & 9.08 & 1.48 & 0.55 \\
 \hline
 Gravitar & inf & inf & 573.63 & inf & inf & 2011.69 & 6184.82 & inf & 1.40 \\
 \hline
 IceHockey & inf & inf & 228.42 & inf & inf & inf & inf & nan & 6.63 \\
 \hline
 Jamesbond & inf & inf & 3625.66 & inf & inf & 2808.36 & 3032.05 & inf & 1.37 \\
 \hline
 Kangaroo & 243.60 & inf & 17.37 & inf & inf & 21.26 & 2780.83 & inf & 2.81 \\
 \hline
 Krull & 382.94 & inf & 876.03 & 1228.53 & inf & 4251.09 & 160.64 & 0.11 & 19.84 \\
 \hline
 KungFuMaster & 7 & inf & 1.33 & 5.17 & inf & 4.94 & 1.14 & 0.82 & 0.76 \\
 \hline
 MsPacman & inf & 0.52 & 95.68 & inf & 0.21 & 61.37 & 2.72 & 29.37 & 0.76 \\
 \hline
 NameThisGame & 634.23 & 0.35 & 861.94 & 133.88 & 0.36 & 3.40 & 36.52 & inf & 29.13 \\
 \hline
 Pitfall & 0.03 & inf & 53.82 & 0.18 & nan & 231.52 & 1.25 & nan & 463.05 \\
 \hline
 Pong & 215.57 & inf & 4428.59 & 2665.23 & inf & 4428.59 & 50.01 & 0.795 & 10.22 \\
 \hline
 PrivateEye & inf & inf & 343.52 & 798.98 & inf & 61.63 & 160.81 & 0.38 & 25.797 \\
 \hline
 Qbert & 314.6 & 0.92 & 255.06 & 80.59 & 0.36 & 7.86 & 391.11 & 1.05 & 0.67  \\
 \hline
 Riverraid & 0.92 & 0.36 & 9.51 & 26.64 & 0.27 & 15.96 & 3.29 & 0.82 & 0.48 \\
 \hline
 RoadRunner & 34.43 & 0.53 & 29.44 & 78.61 & 0.53 & 49.88 & 97.86 & 0.93 & 727.94 \\
 \hline
 Robotank & inf & inf & 179.31 & inf & inf & 164.08 & 1996.31 & 0.75 & 1.68 \\
 \hline
 Seaquest & 15.395 & inf & 9.62 & inf & inf & inf & 1.83 & 0.30 & 39.36 \\
 \hline
 SpaceInvaders & 897.32 & 2.63 & 93.02 & 1127.22 & inf & 81.14 & 2.15 & inf & 2.73 \\
 \hline
 StarGunner & 47.11 & 0.62 & 32.08 & 38.97 & 0.45 & 18.91 & 161.30 & 0.94 & 0.53 \\
 \hline
 TimePilot & 1997.68 & inf & 81.95 & 0.199 & inf & 13.13 & 1683.55 & inf & 359.52 \\
 \hline
 UpNDown & 270.05 & 0.06 & 6.44 & 98.43 & inf & 4.32 & 5.96 & inf & 0.797 \\
 \hline
 VideoPinball & inf & 0.72 & 488.97 & inf & 0.45 & 706.17 & 92.05 & inf & 97.15 \\
 \hline
 WizardOfWor & 1300 & inf & 14.85 & 2379.63 & inf & 3413.66 & 300.42 & 4.02 & 34.56 \\
 \hline
 Zaxxon & 146.37 & 0.4 & 4.81 & inf & inf & 53.398 & 30.72 & 0.84 & 15.67 \\ [1ex]  
 \hline
\end{tabular}
\caption[htbp]{Original statistics: Speed-up = $\frac{T}{T_G}$ where $T_G$ represents the number of time steps that the group agents took to achieve the highest performance and $T$ represents the number of time steps that the corresponding single agents took to reach the same performance. The $inf$ values mean that the single agents were never able to reach the same performance. The $nan$ values mean the group agents never reached a satisfactory level of performance. The data are reported for each rule and agent separately.}
\label{table2}
\end{table*}

\begin{figure*}
         \begin{subfigure}[]{0.49\linewidth}
                \includegraphics[width=0.333\linewidth]{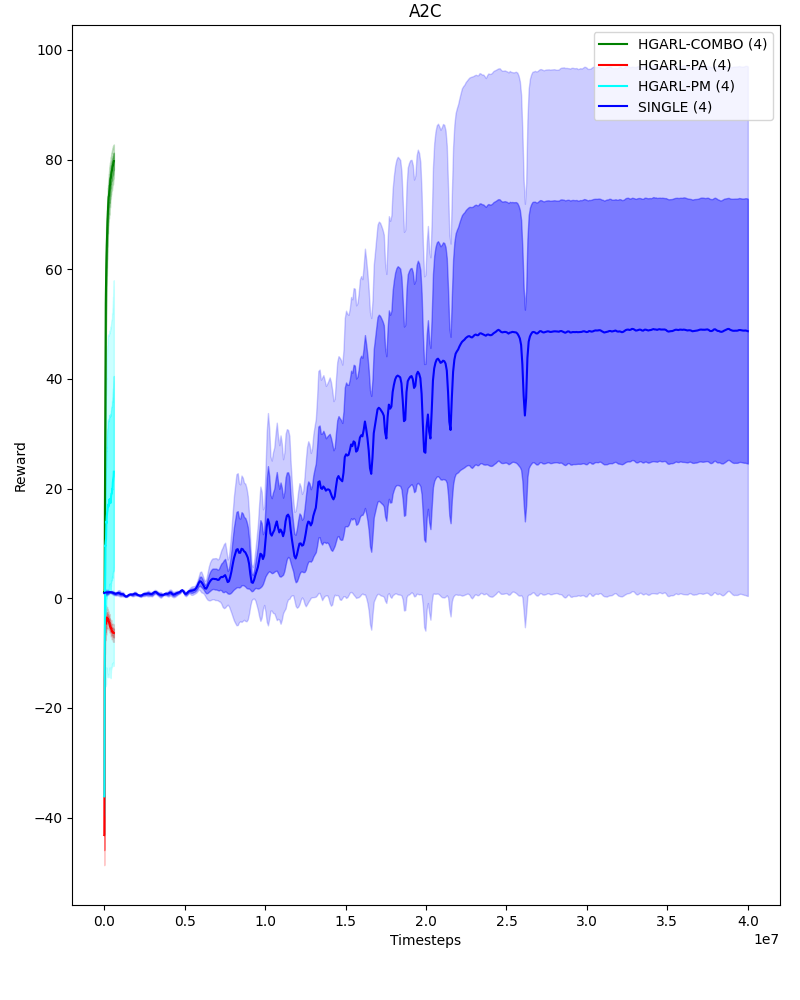}\hfill
                \includegraphics[width=0.333\linewidth]{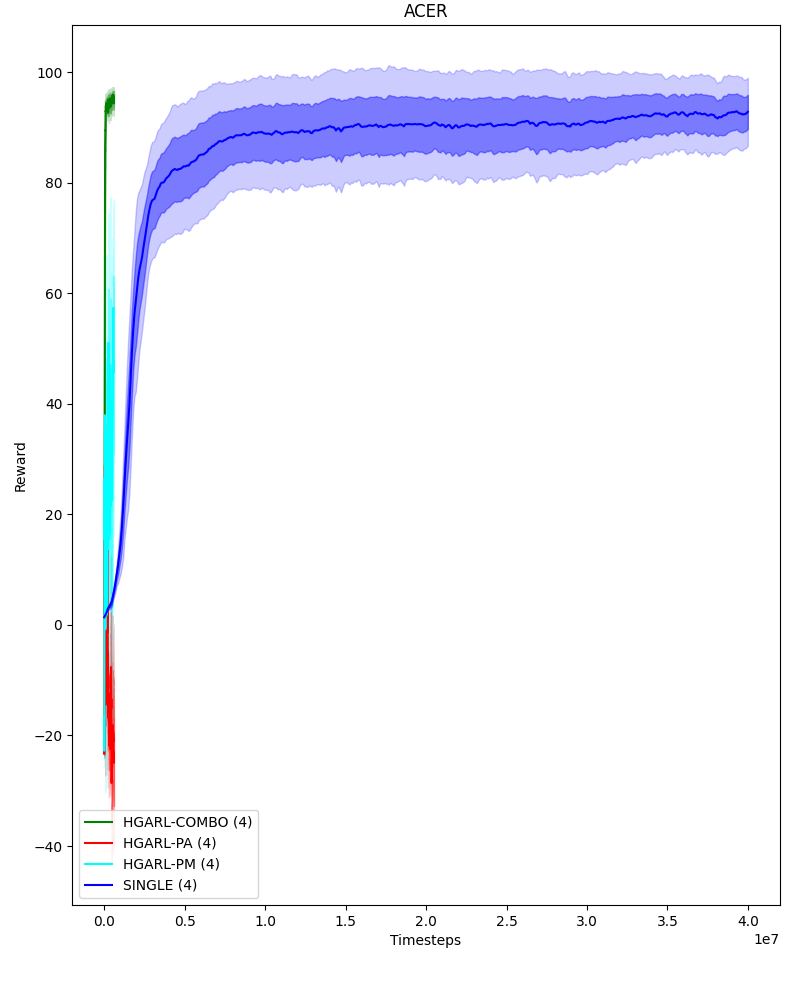}\hfill
                \includegraphics[width=0.333\linewidth]{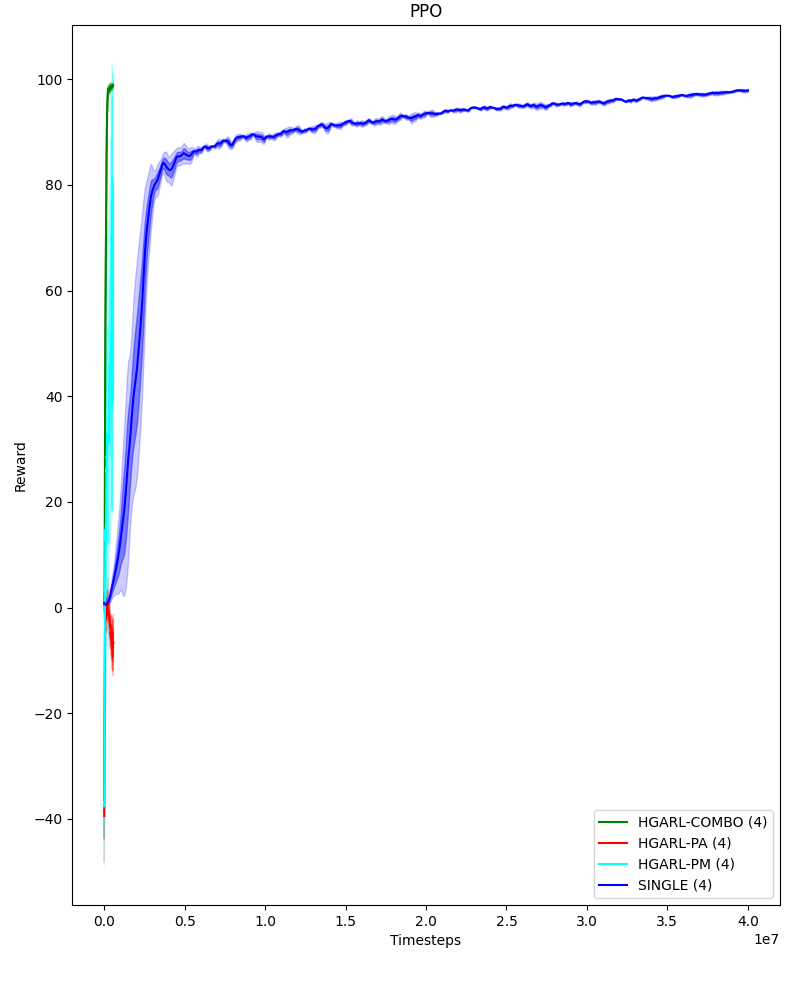}
                \caption{Boxing}
         \end{subfigure}
         \begin{subfigure}[]{0.49\linewidth}
                \includegraphics[width=0.333\linewidth]{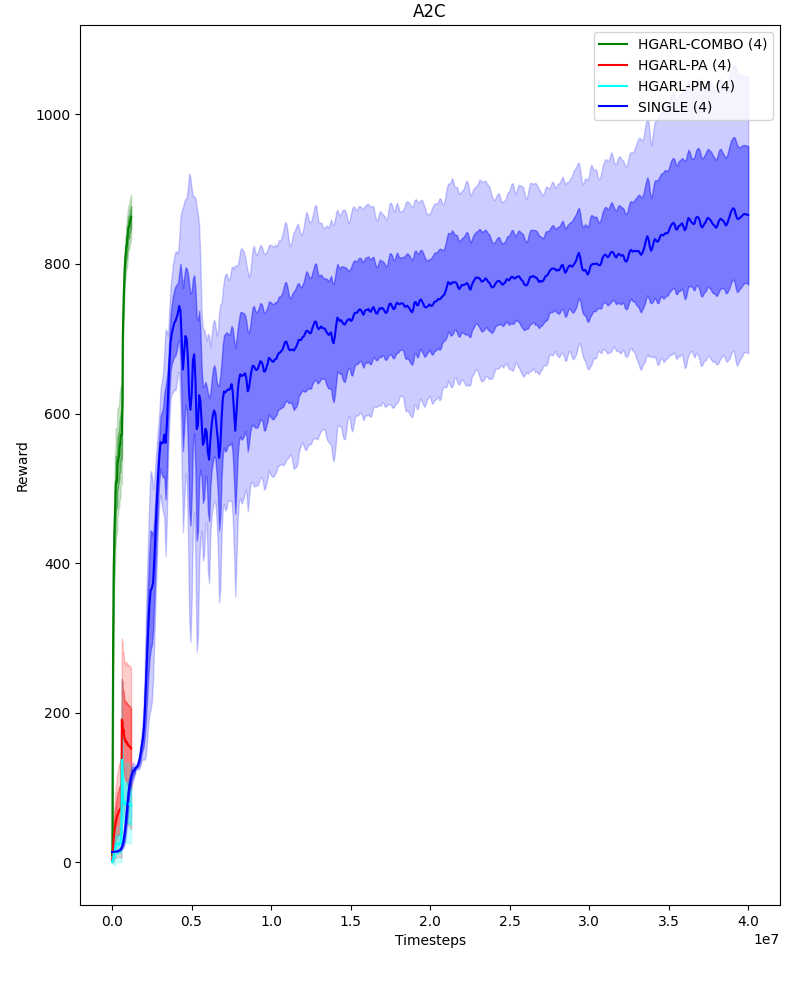}\hfill
                \includegraphics[width=0.333\linewidth]{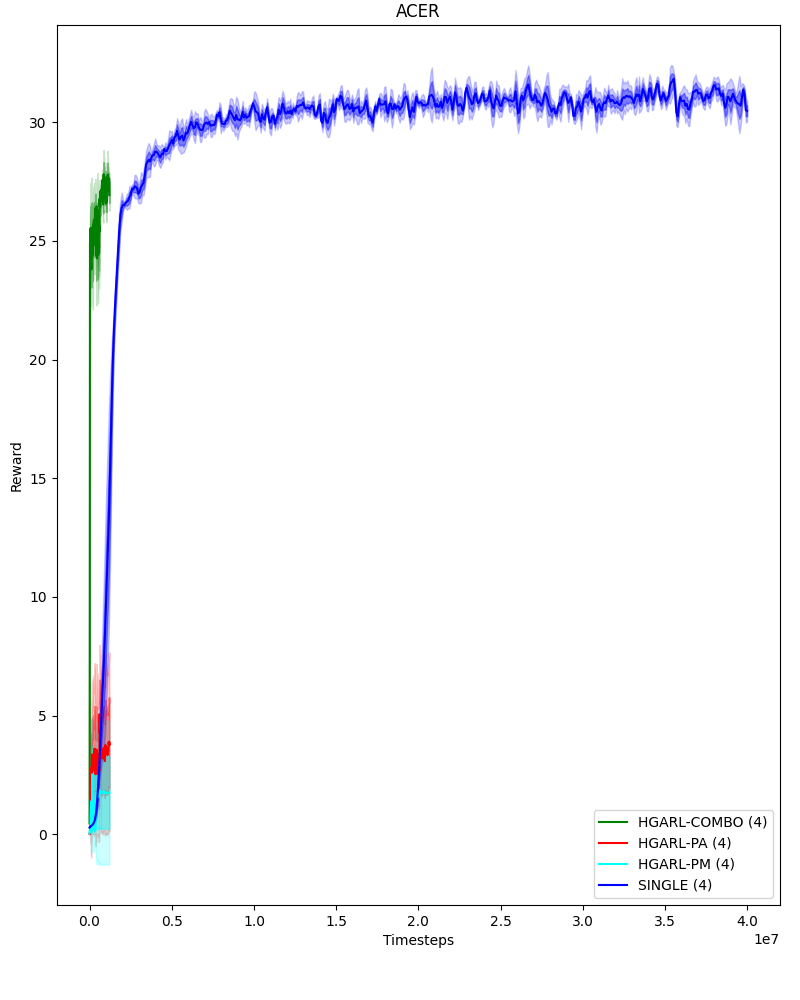}\hfill
                \includegraphics[width=0.333\linewidth]{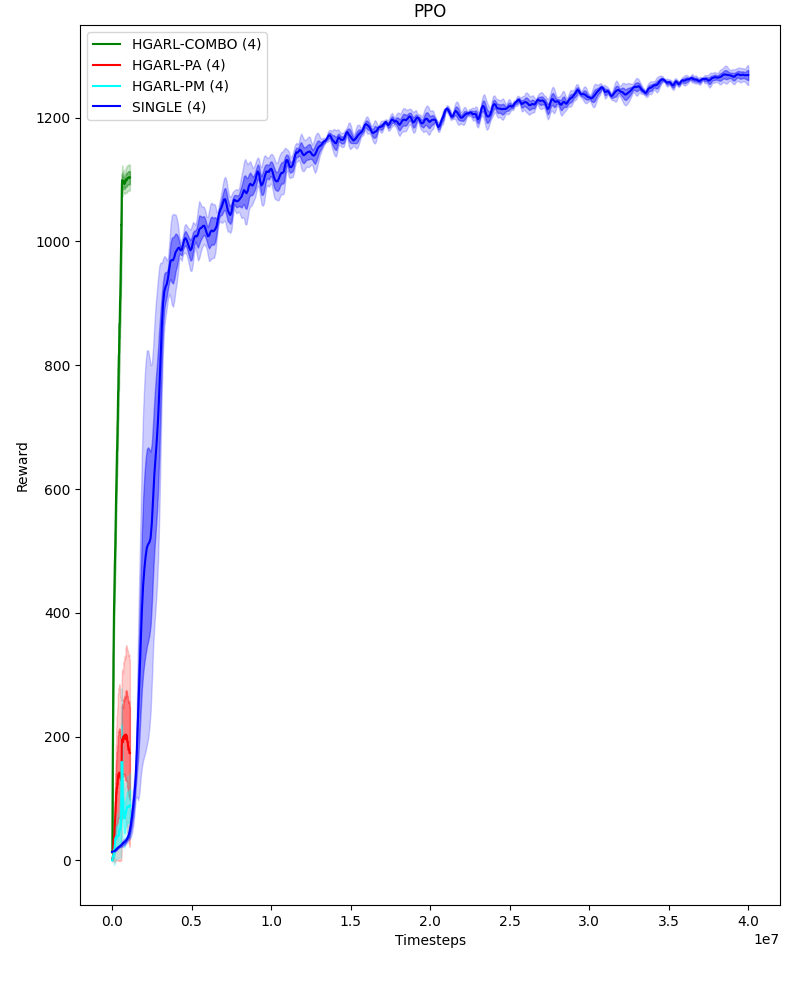}
                \caption{BankHeist}
         \end{subfigure}
         \begin{subfigure}[]{0.49\linewidth}
                \includegraphics[width=0.333\linewidth]{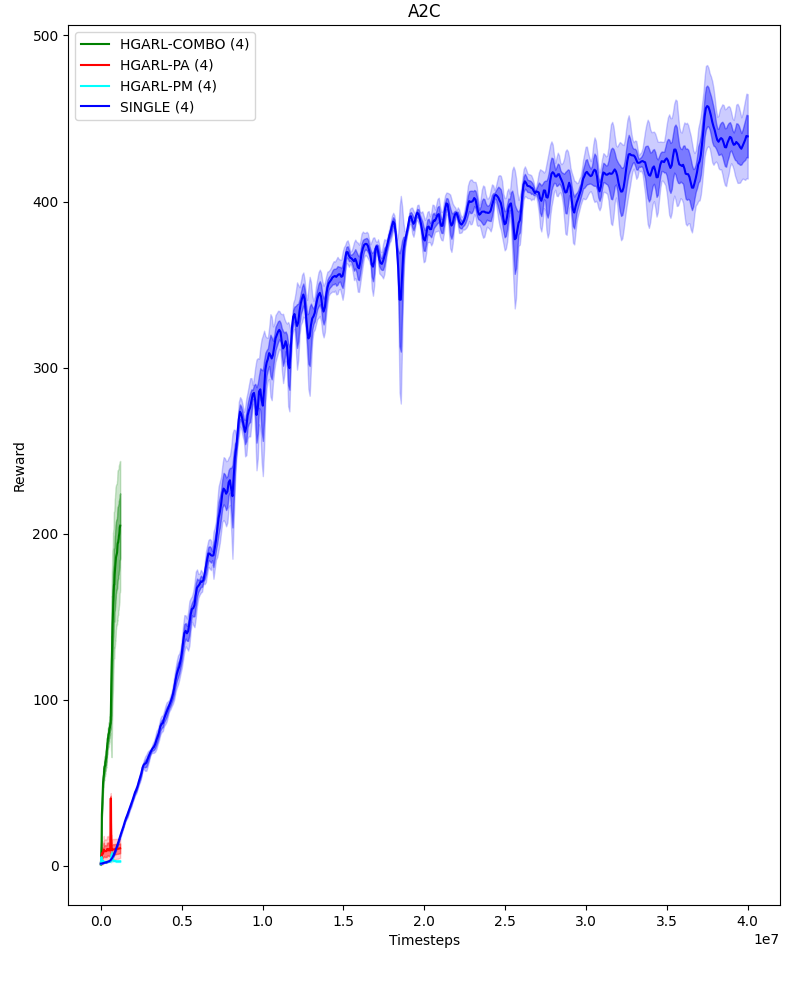}\hfill
                \includegraphics[width=0.333\linewidth]{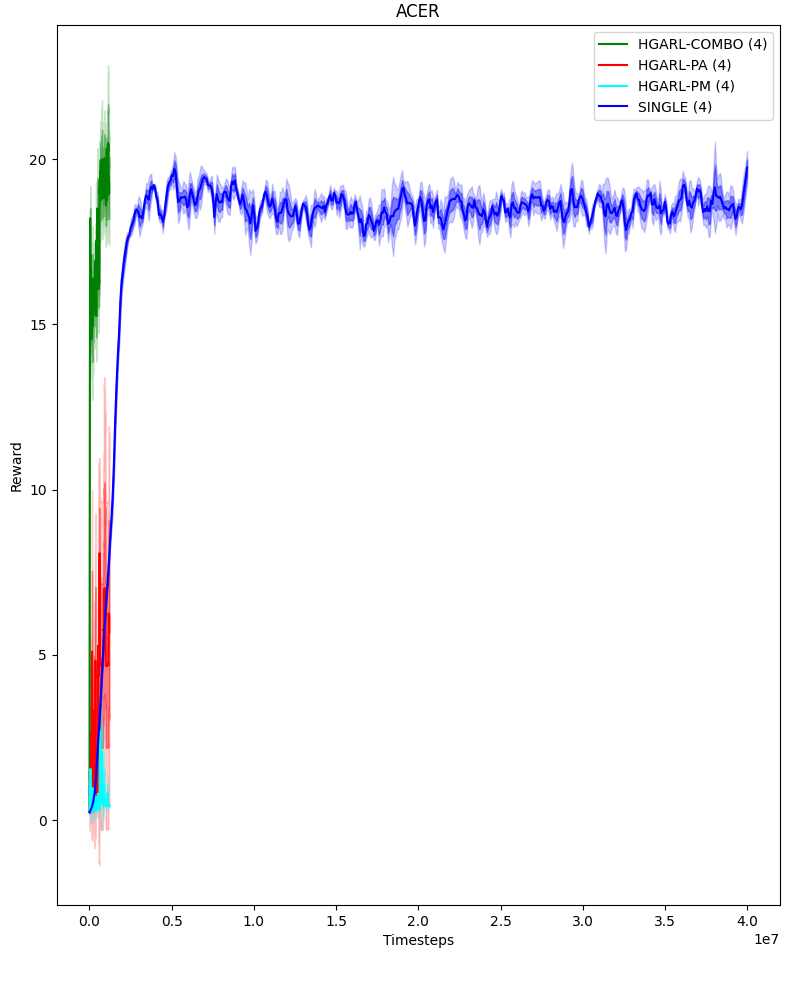}\hfill
                \includegraphics[width=0.333\linewidth]{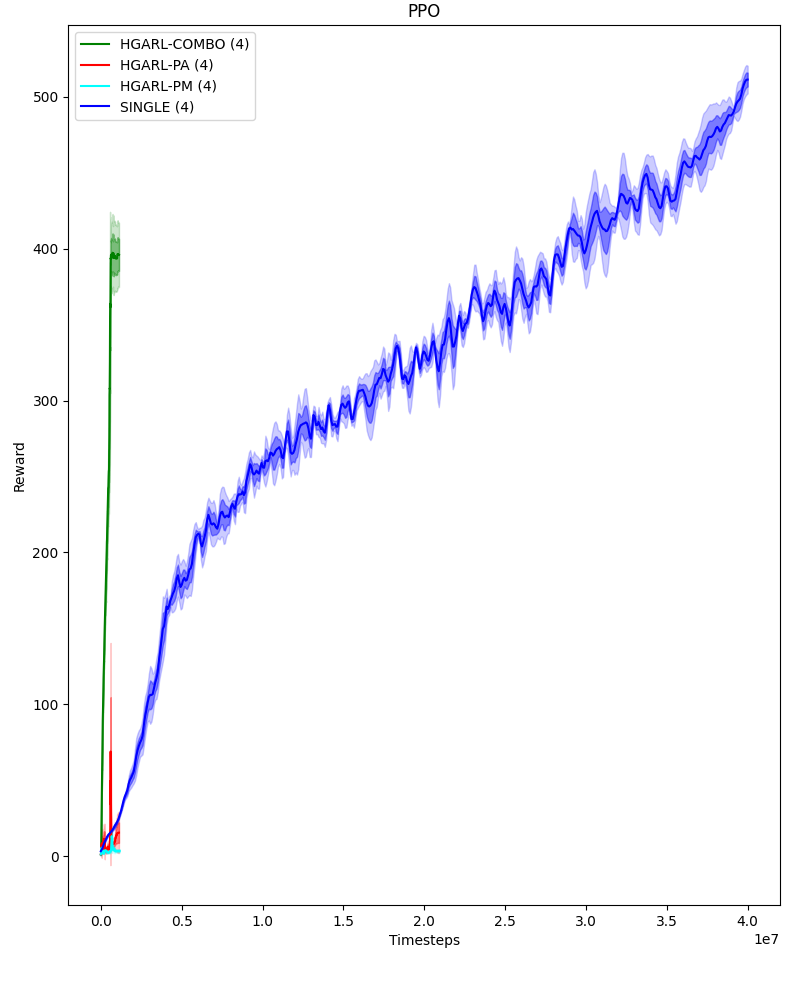}
                \caption{Breakout}
         \end{subfigure}
         \begin{subfigure}[]{0.49\linewidth}
                \includegraphics[width=0.333\linewidth]{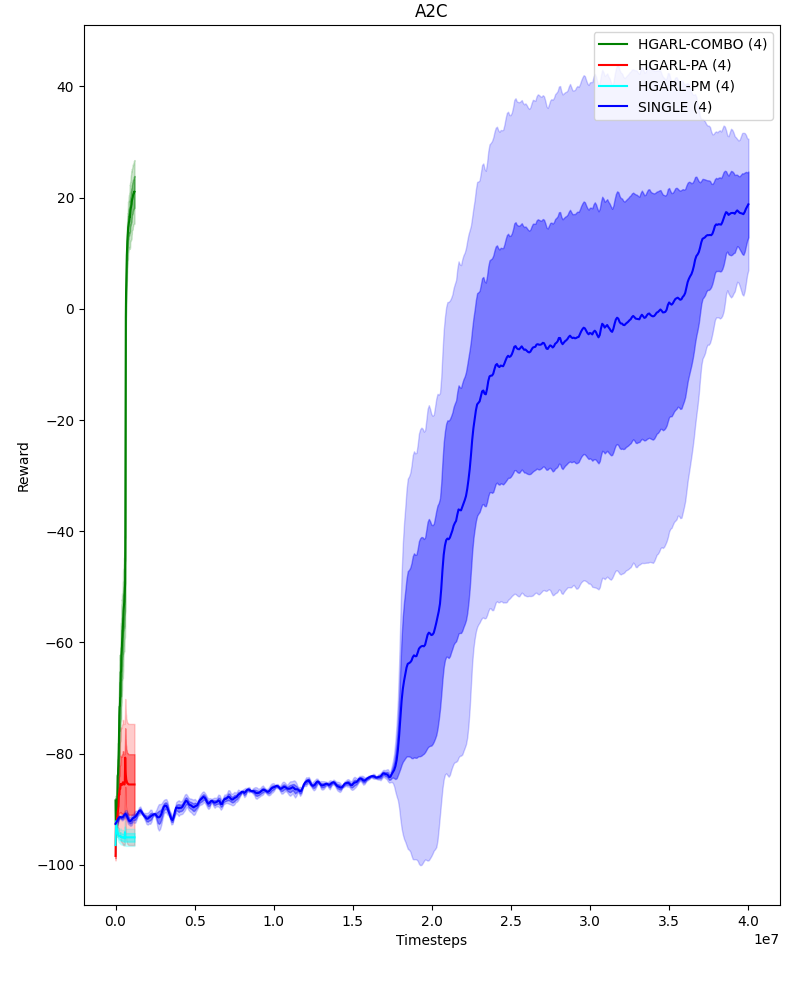}\hfill
                \includegraphics[width=0.333\linewidth]{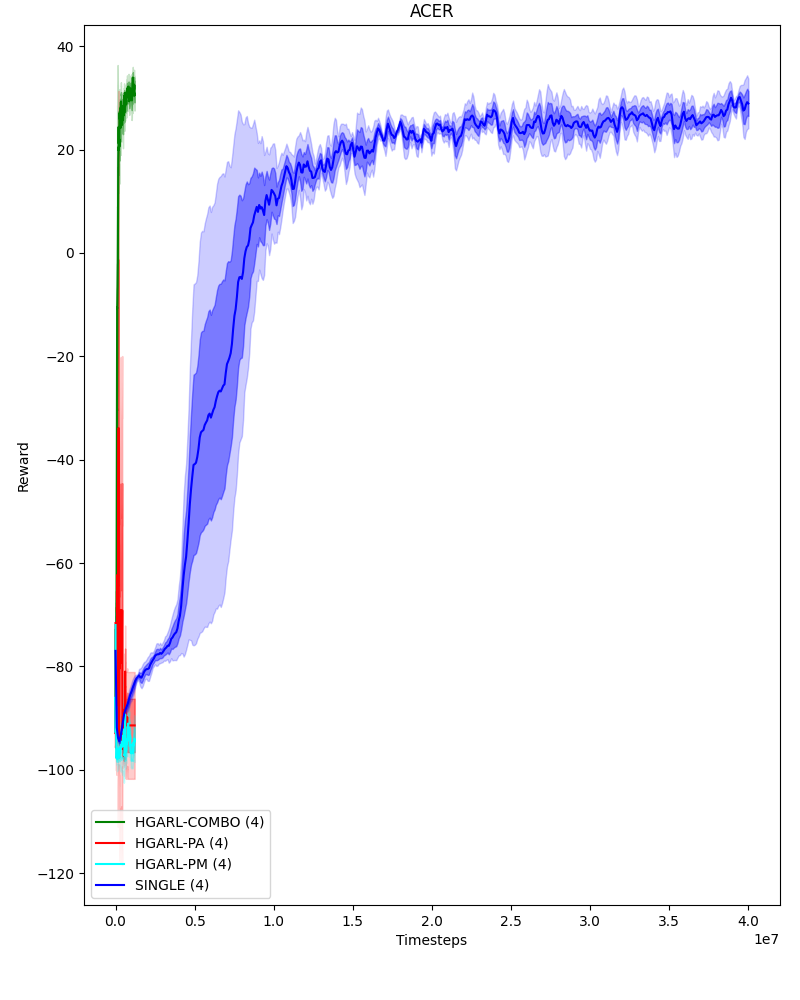}\hfill
                \includegraphics[width=0.333\linewidth]{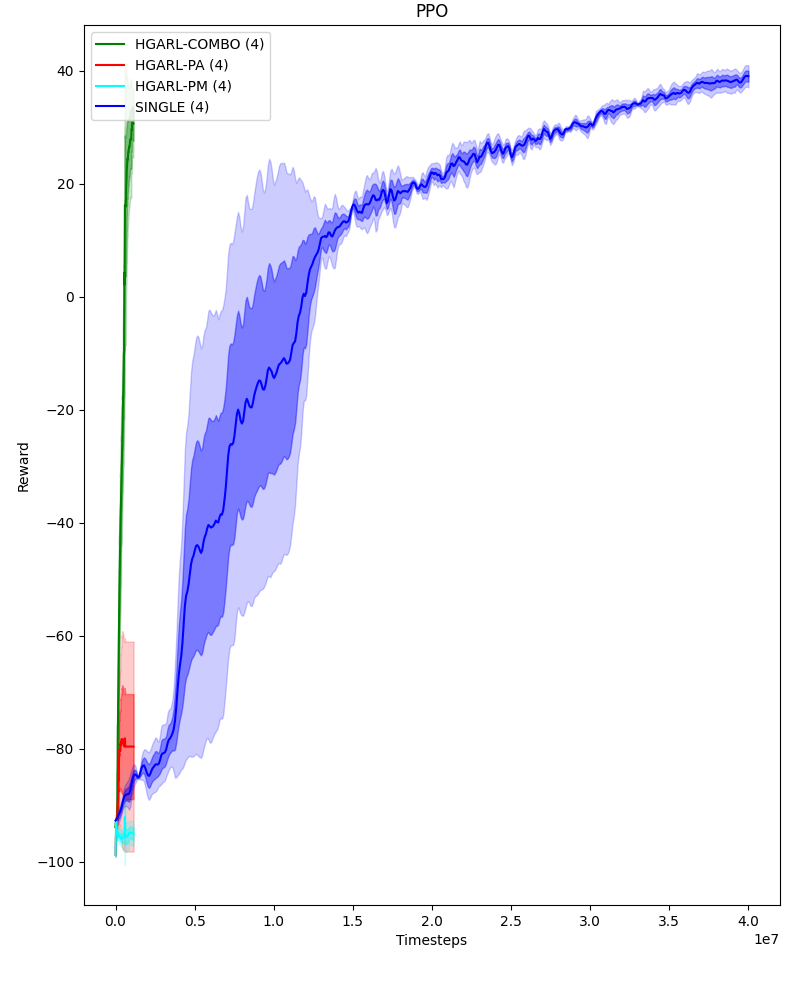}
                \caption{FishingDerby}
         \end{subfigure}
         \begin{subfigure}[]{0.49\linewidth}
                \includegraphics[width=0.333\linewidth]{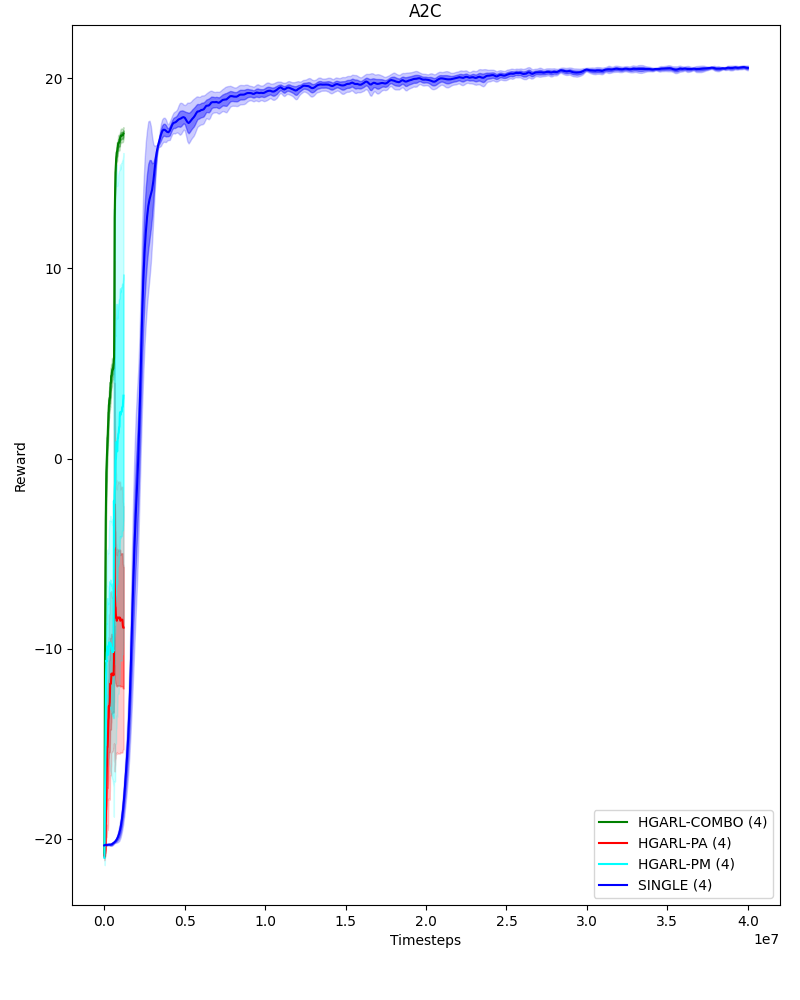}\hfill
                \includegraphics[width=0.333\linewidth]{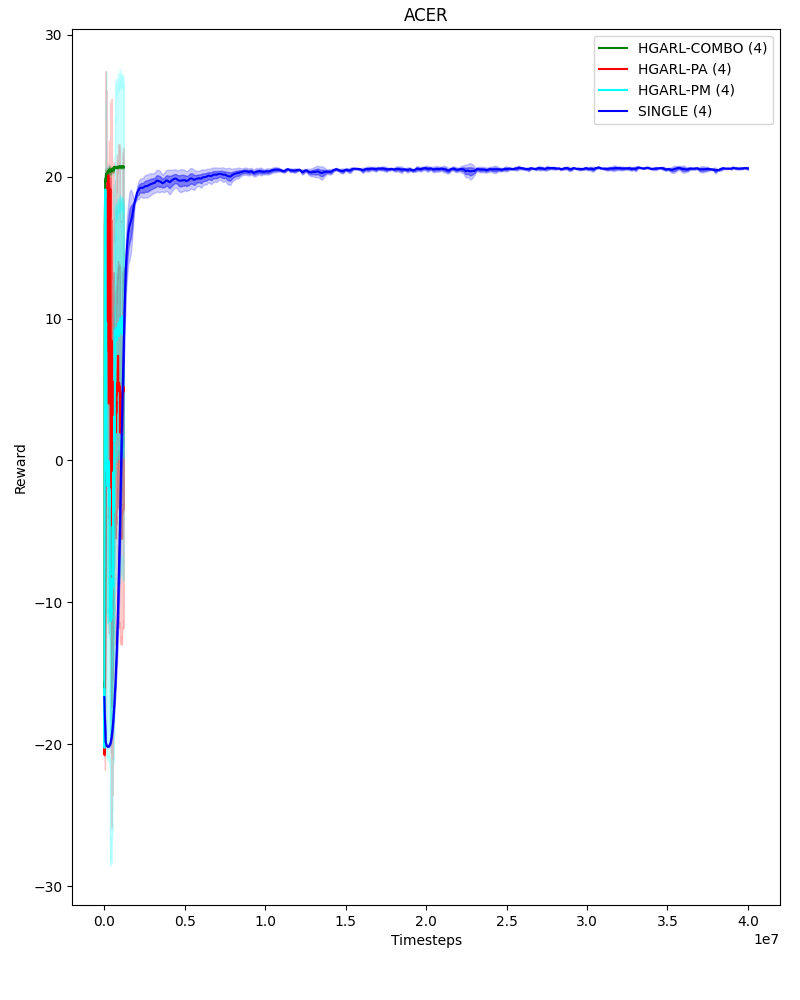}\hfill
                \includegraphics[width=0.333\linewidth]{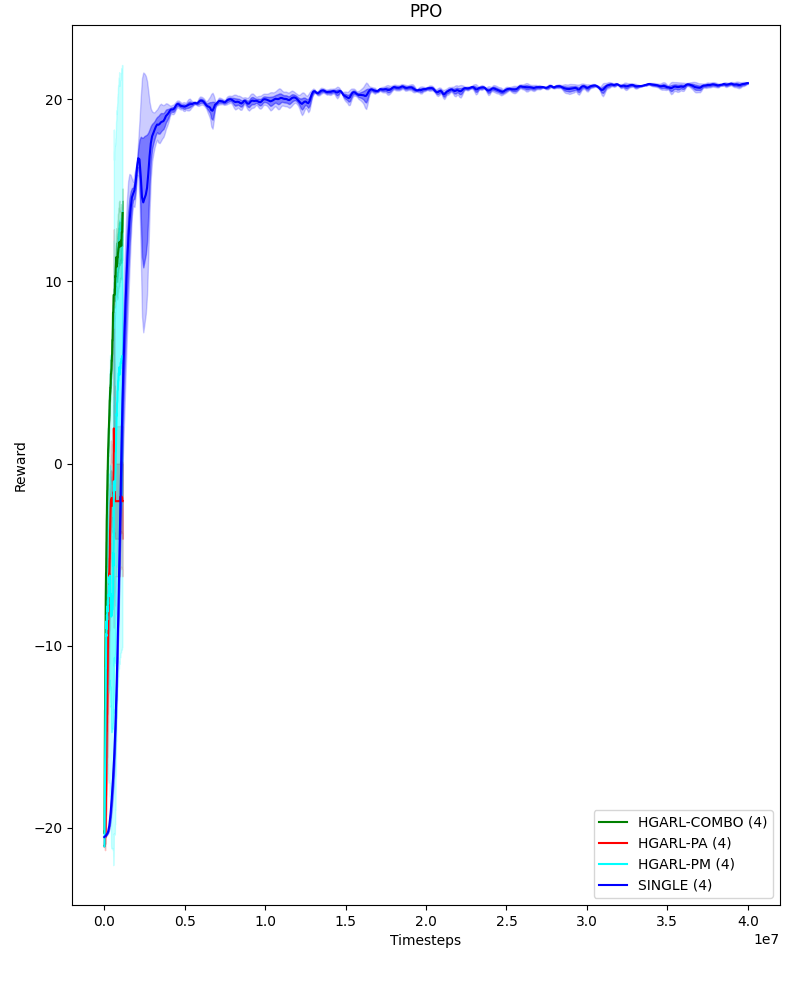}
                \caption{Pong}
         \end{subfigure}
         \begin{subfigure}[]{0.49\linewidth}
                \includegraphics[width=0.333\linewidth]{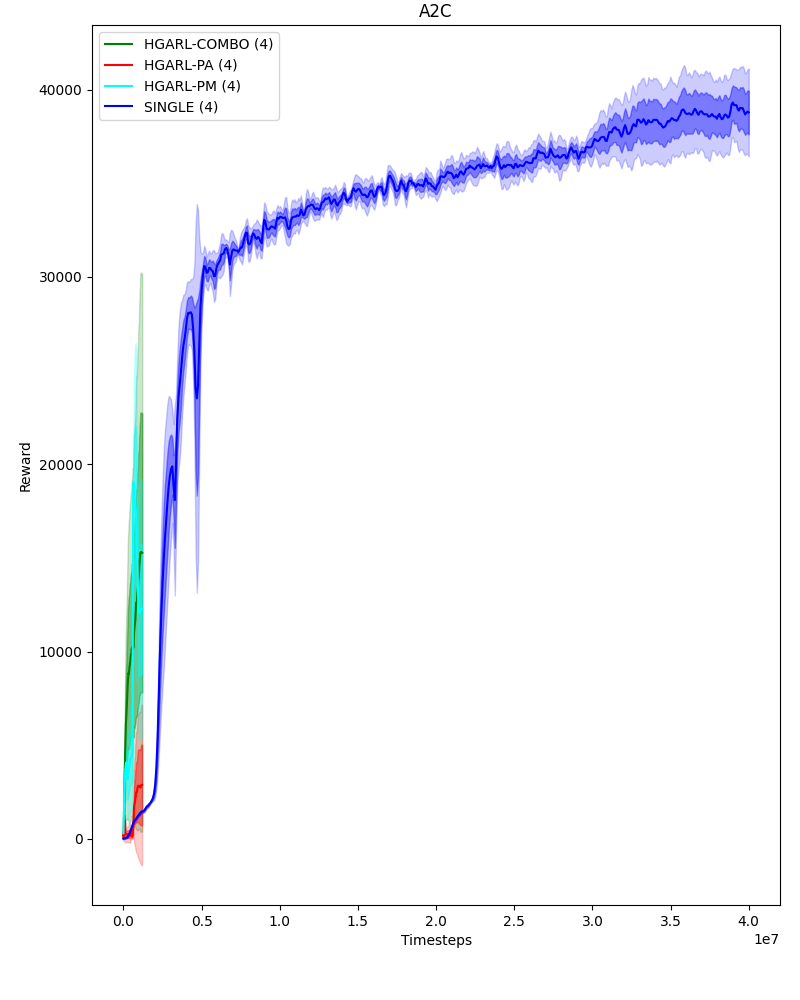}\hfill
                \includegraphics[width=0.333\linewidth]{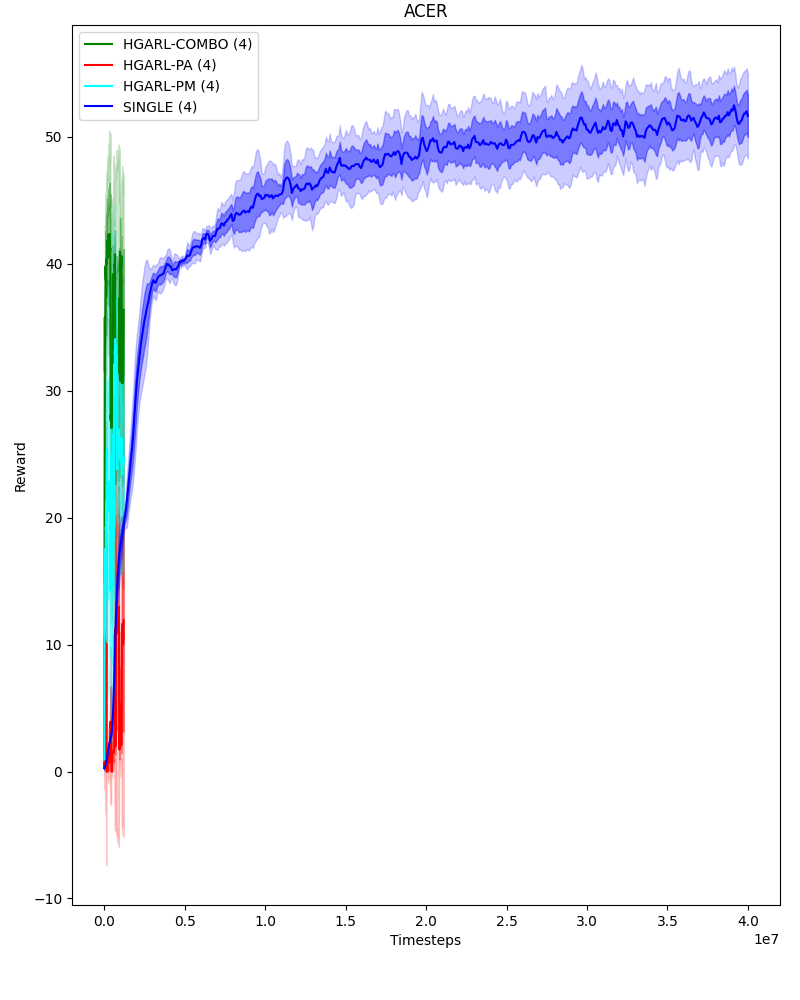}\hfill
                \includegraphics[width=0.333\linewidth]{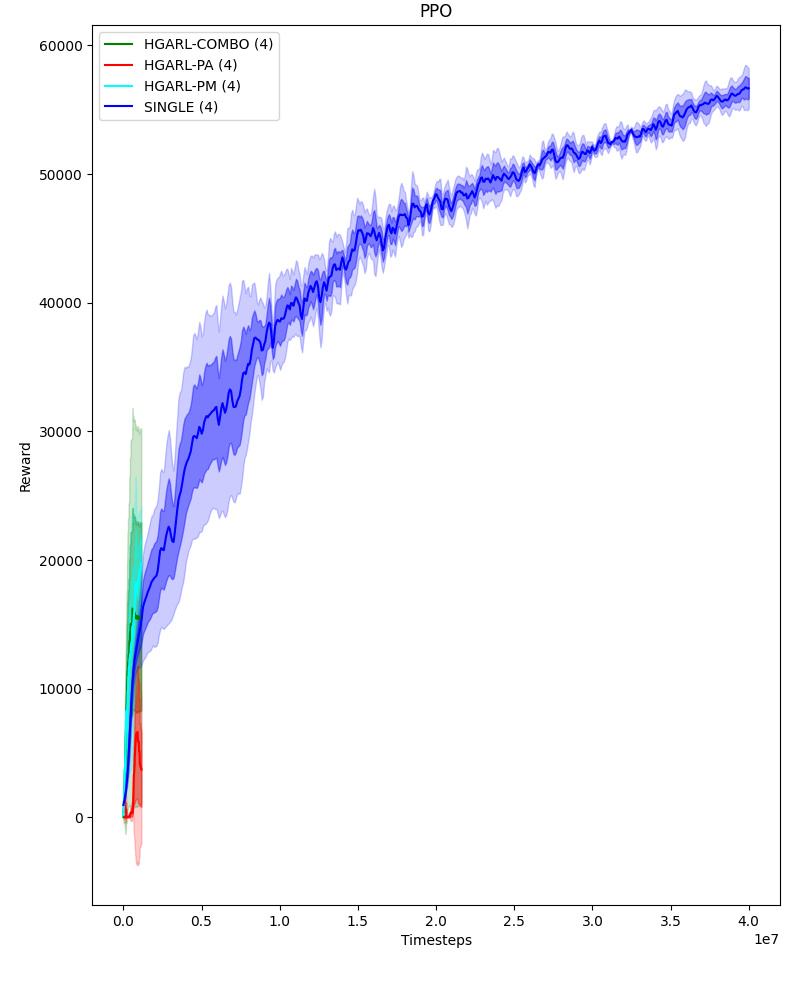}
                \caption{RoadRunner}
         \end{subfigure}
         \caption{Atari 2600 Games: Part 1. The Combo rule shows superb performance for all three agents of A2C, ACER and PPO.}
        \label{Atari1}
\end{figure*}

\begin{figure*}
        \begin{subfigure}[b]{0.49\linewidth}
                \includegraphics[width=0.333\linewidth]{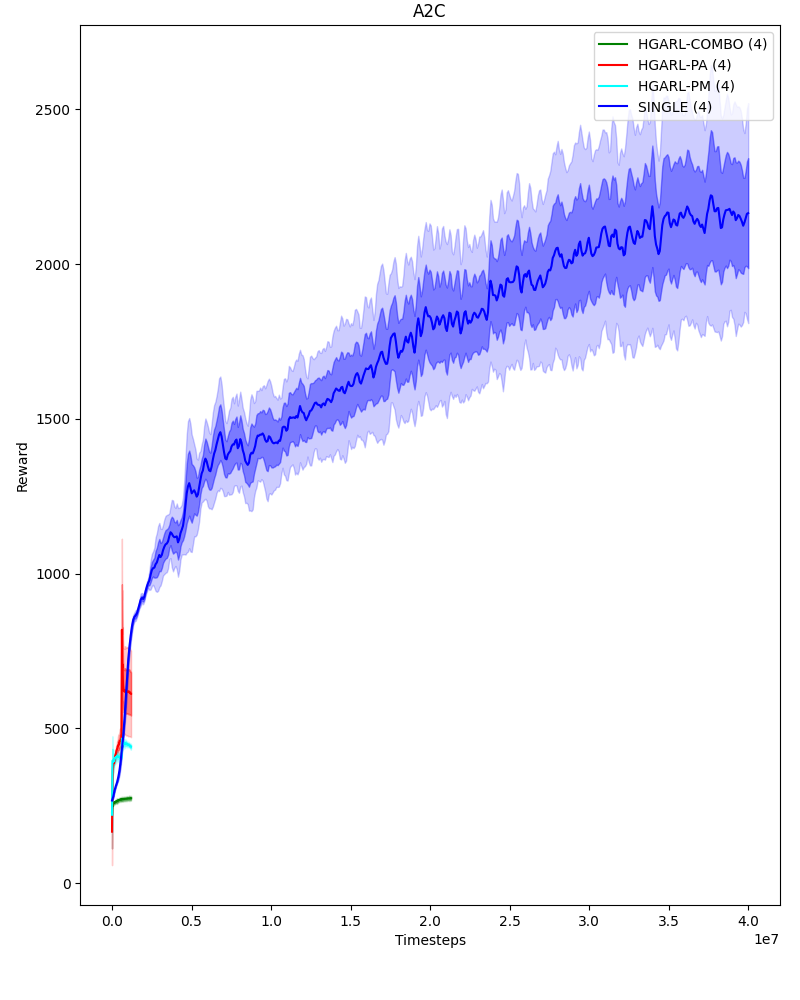}\hfill
                \includegraphics[width=0.333\linewidth]{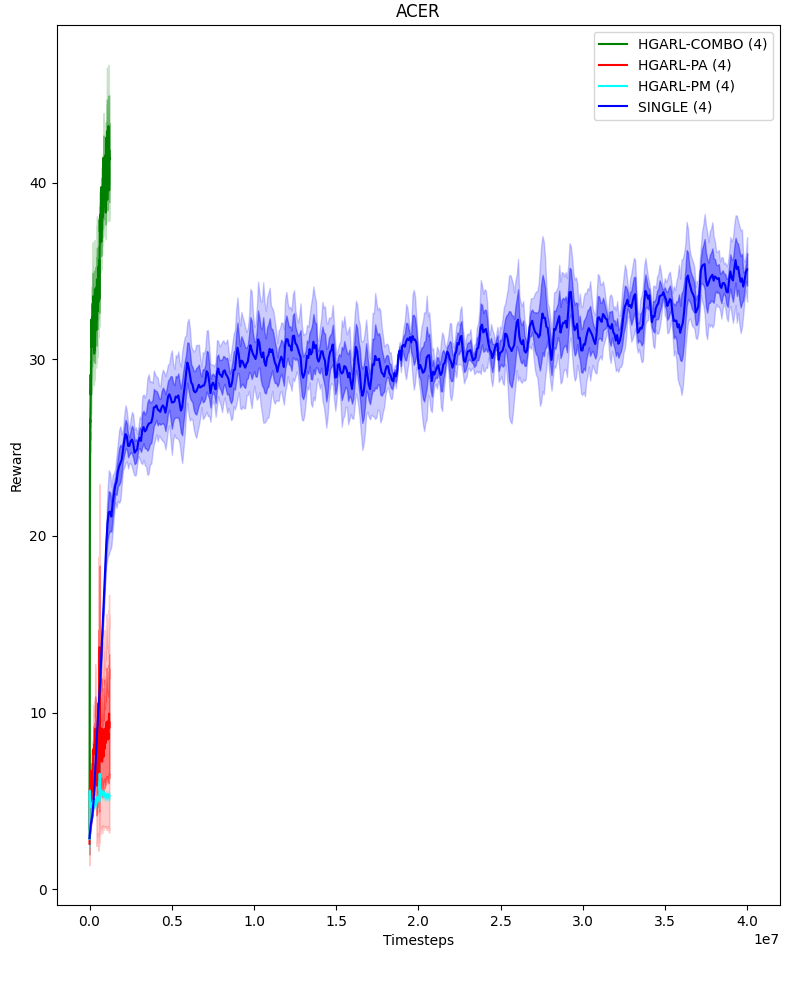}\hfill
                \includegraphics[width=0.333\linewidth]{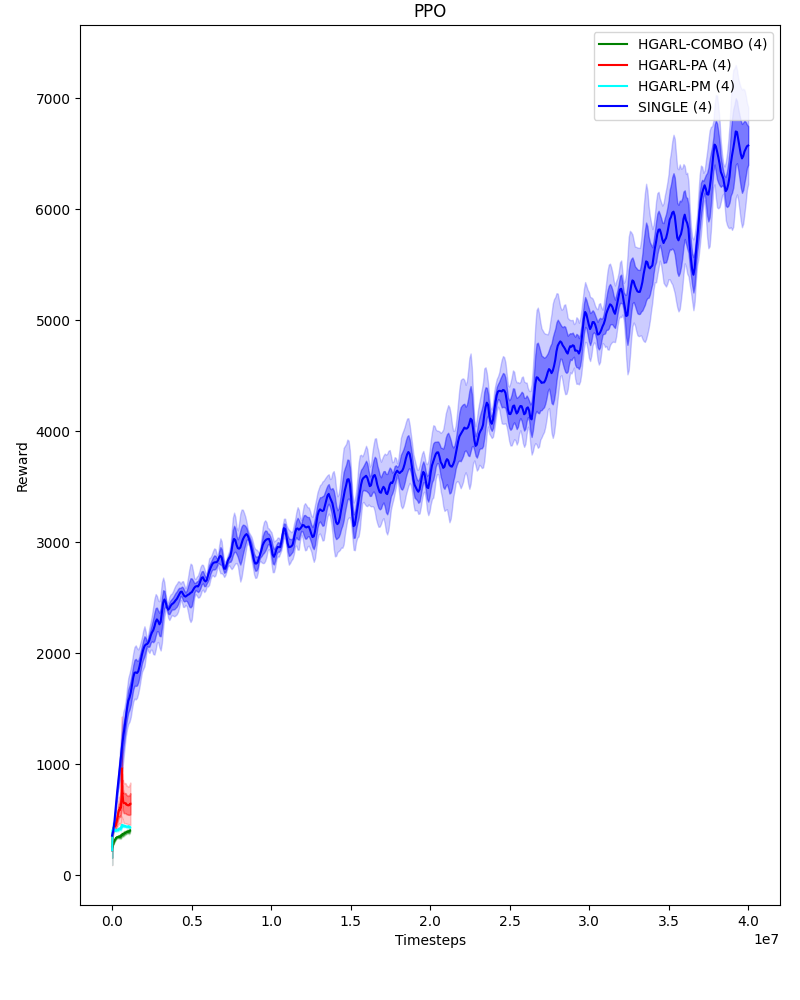}
                \caption{Assault}
         \end{subfigure}
         \begin{subfigure}[b]{0.49\linewidth}
                \includegraphics[width=0.333\linewidth]{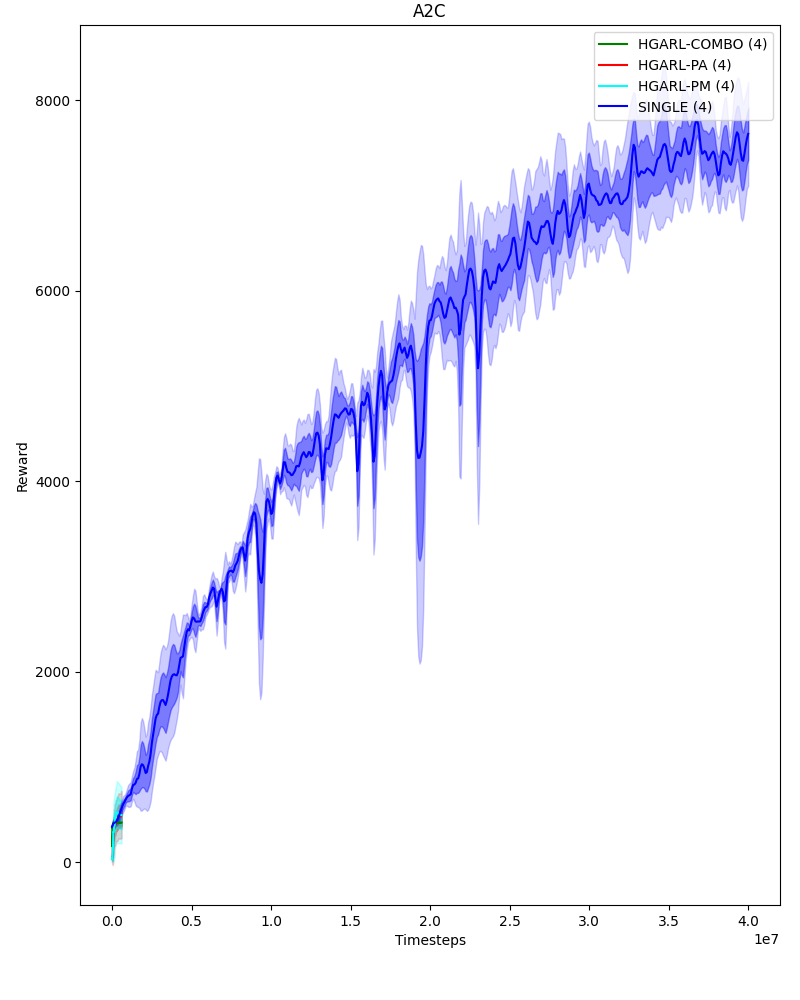}\hfill
                \includegraphics[width=0.333\linewidth]{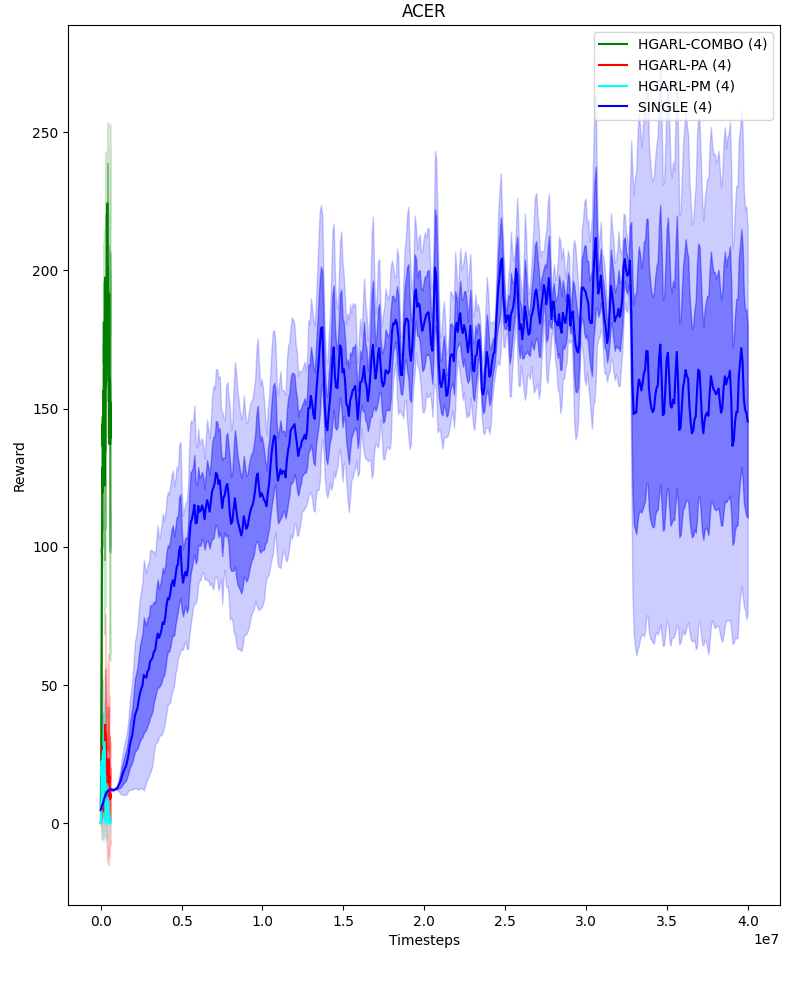}\hfill
                \includegraphics[width=0.333\linewidth]{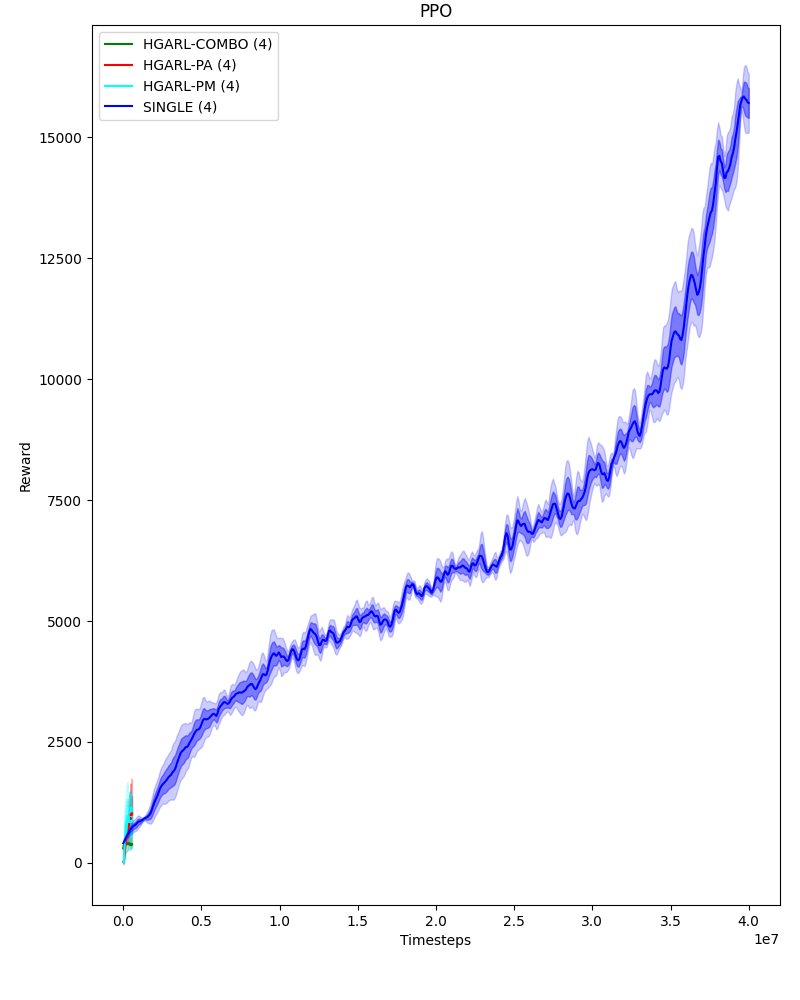}
                \caption{Gopher}
        \end{subfigure}
        \begin{subfigure}[b]{0.49\linewidth}
                \includegraphics[width=0.333\linewidth]{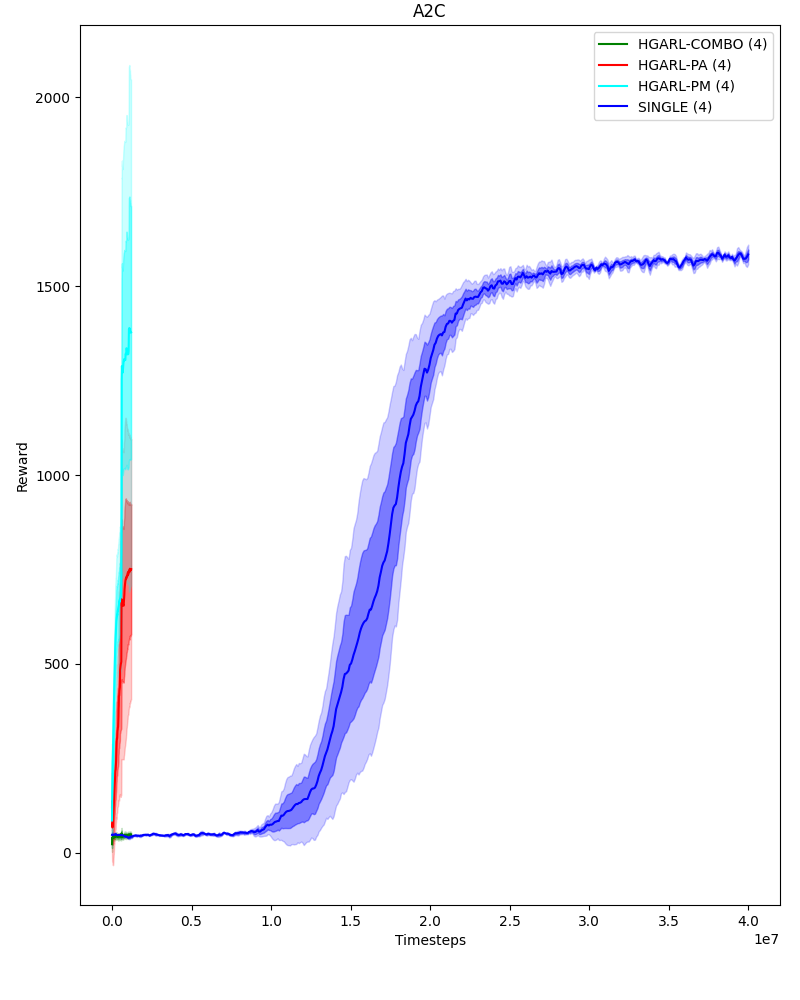}\hfill
                \includegraphics[width=0.333\linewidth]{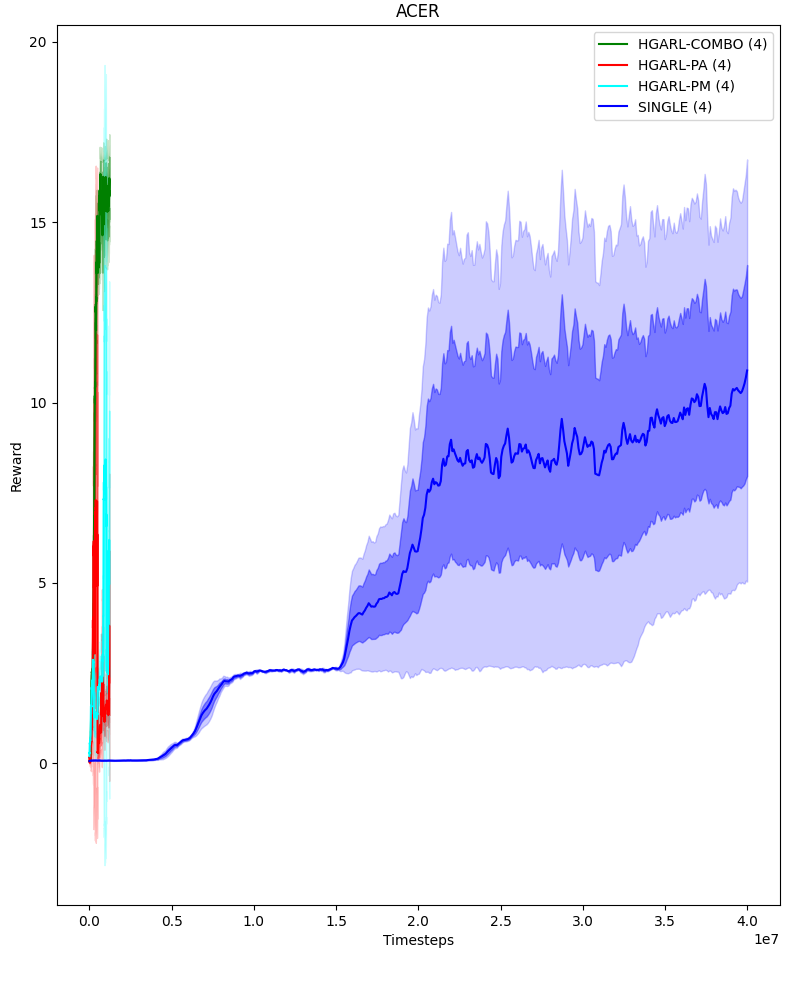}\hfill
                \includegraphics[width=0.333\linewidth]{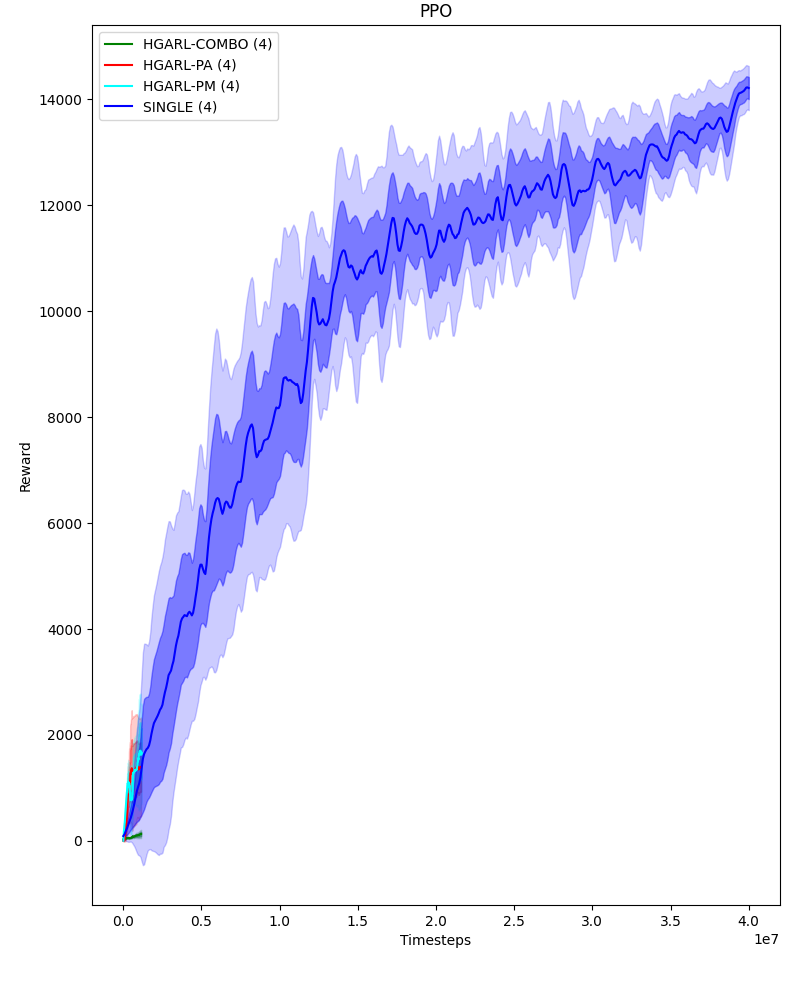}
                \caption{Kangaroo}
        \end{subfigure}
        \begin{subfigure}[b]{0.49\linewidth}
                \includegraphics[width=0.333\linewidth]{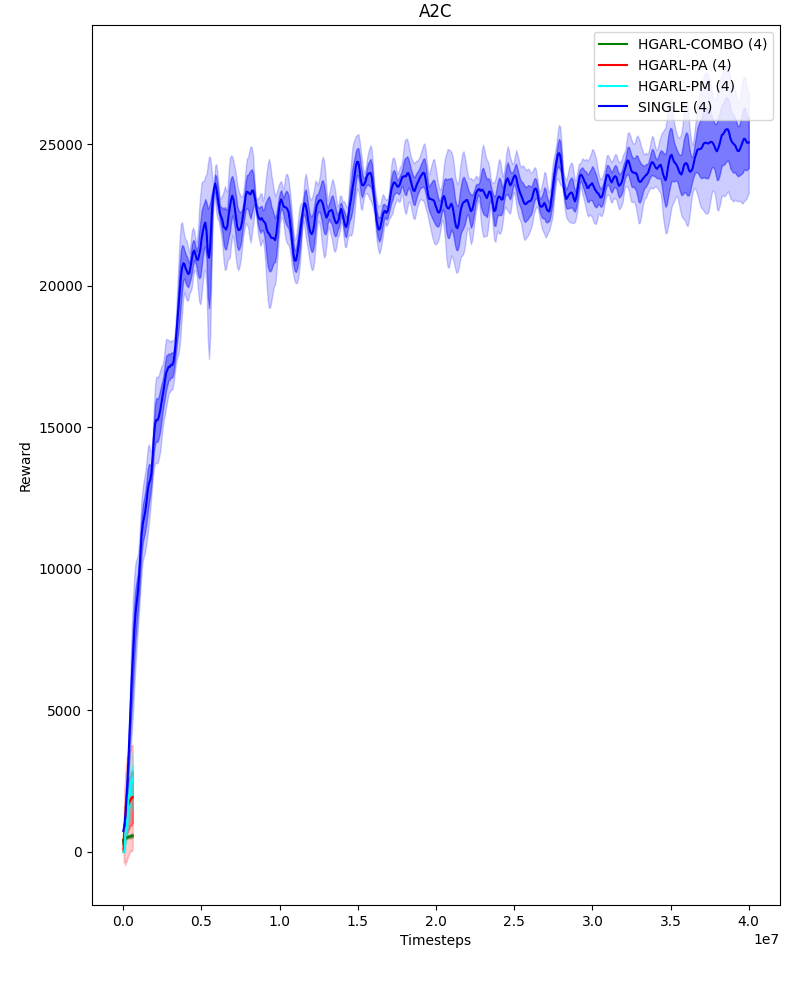}\hfill
                \includegraphics[width=0.333\linewidth]{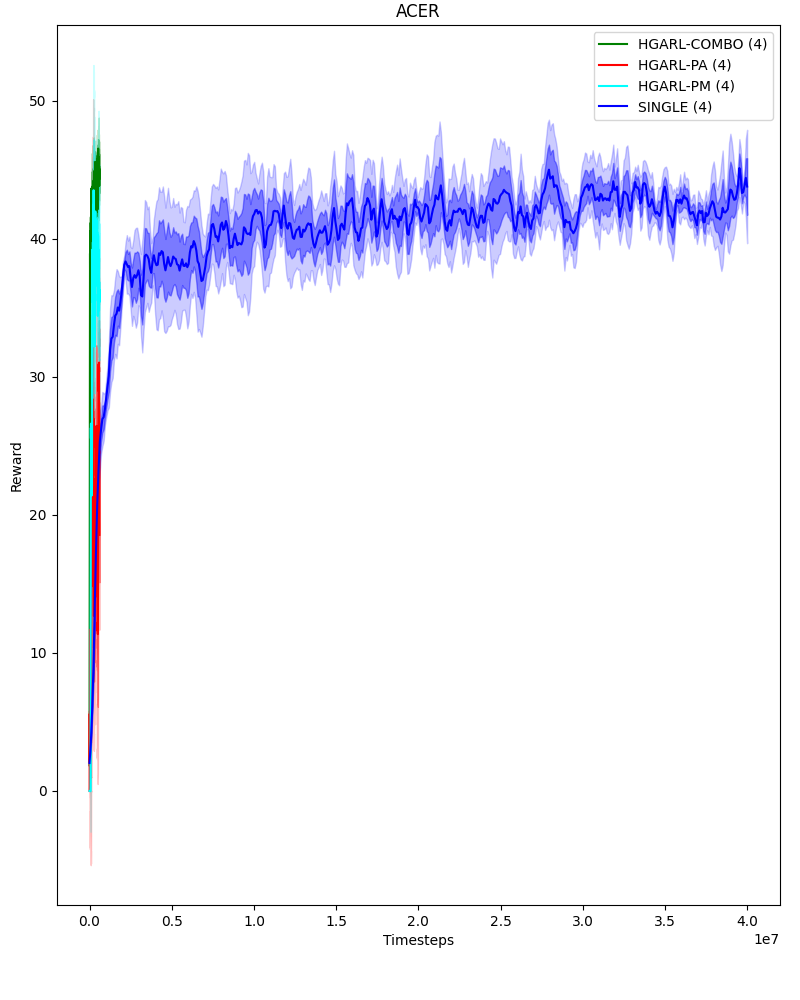}\hfill
                \includegraphics[width=0.333\linewidth]{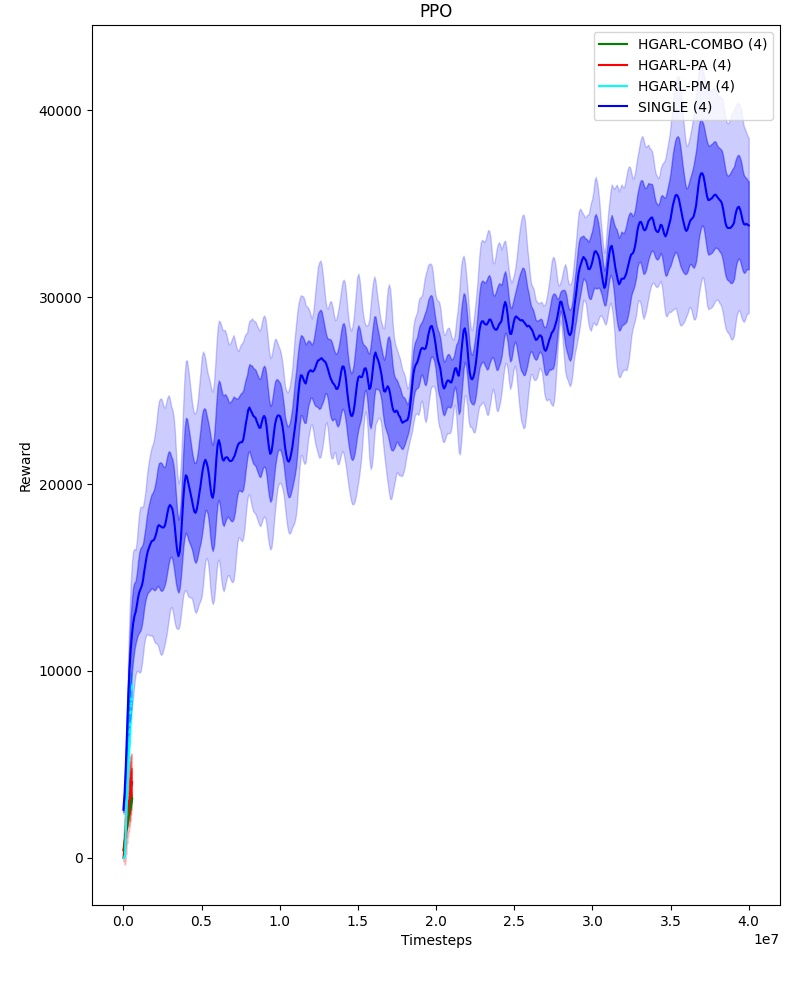}
                \caption{KungFuMaster}
        \end{subfigure}
        \begin{subfigure}[b]{0.49\linewidth}
                \includegraphics[width=0.333\linewidth]{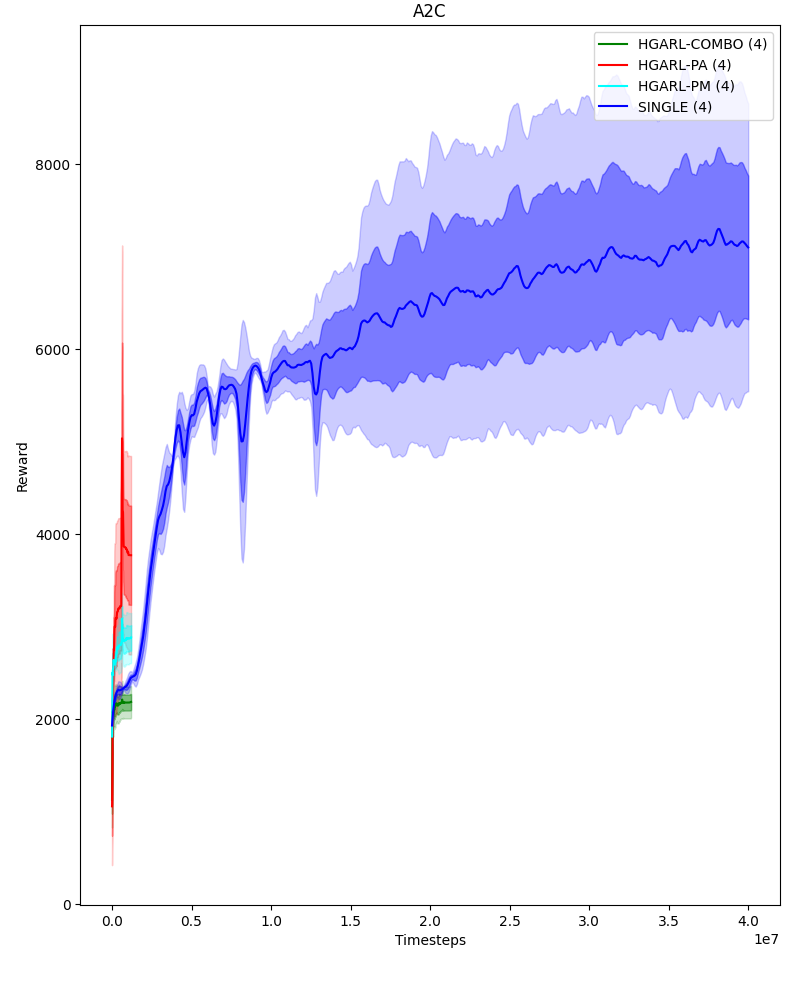}\hfill
                \includegraphics[width=0.333\linewidth]{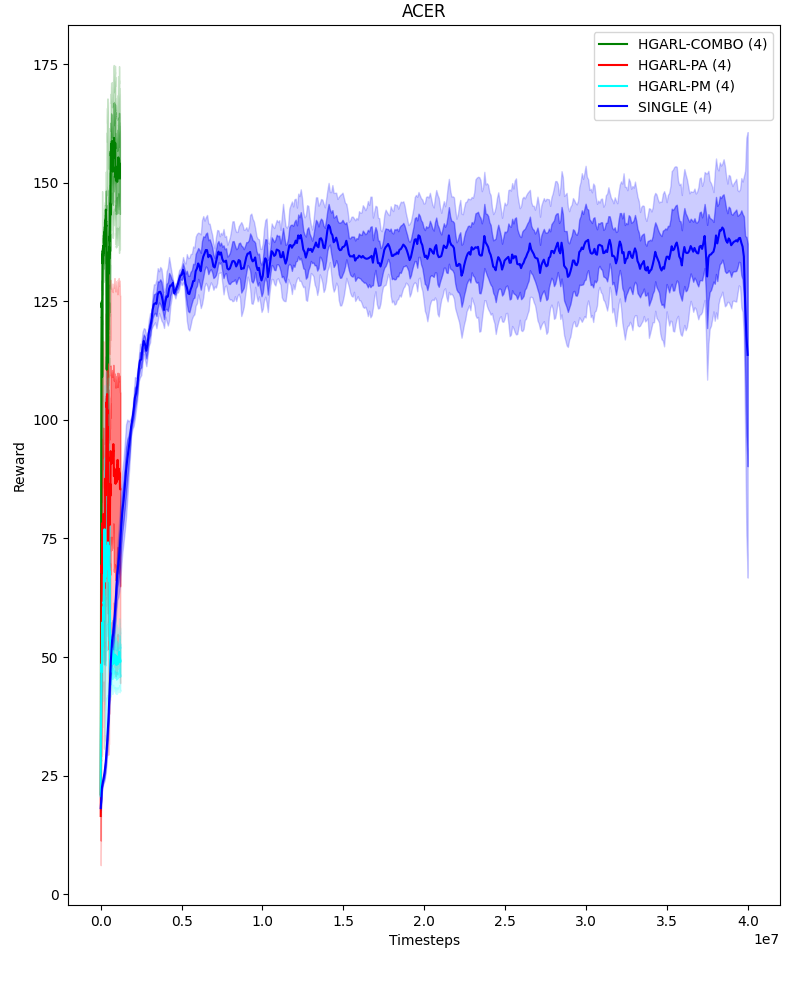}\hfill
                \includegraphics[width=0.333\linewidth]{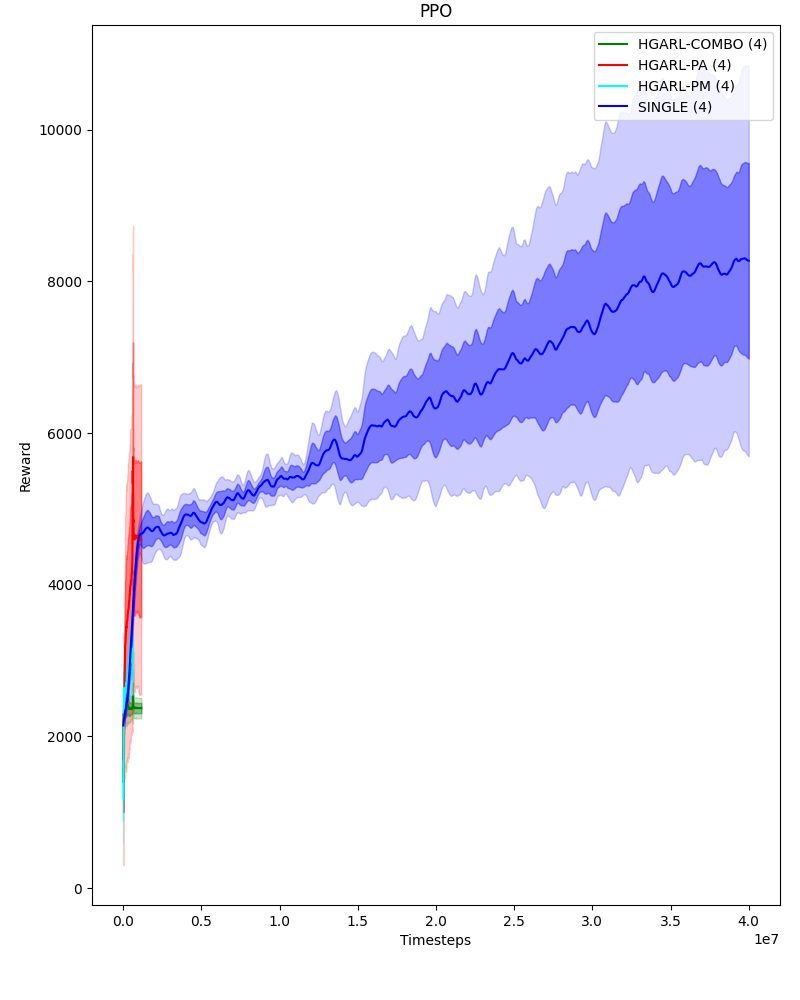}
                \caption{NameThisGame}
        \end{subfigure}
        \begin{subfigure}[b]{0.49\linewidth}
                \includegraphics[width=0.333\linewidth]{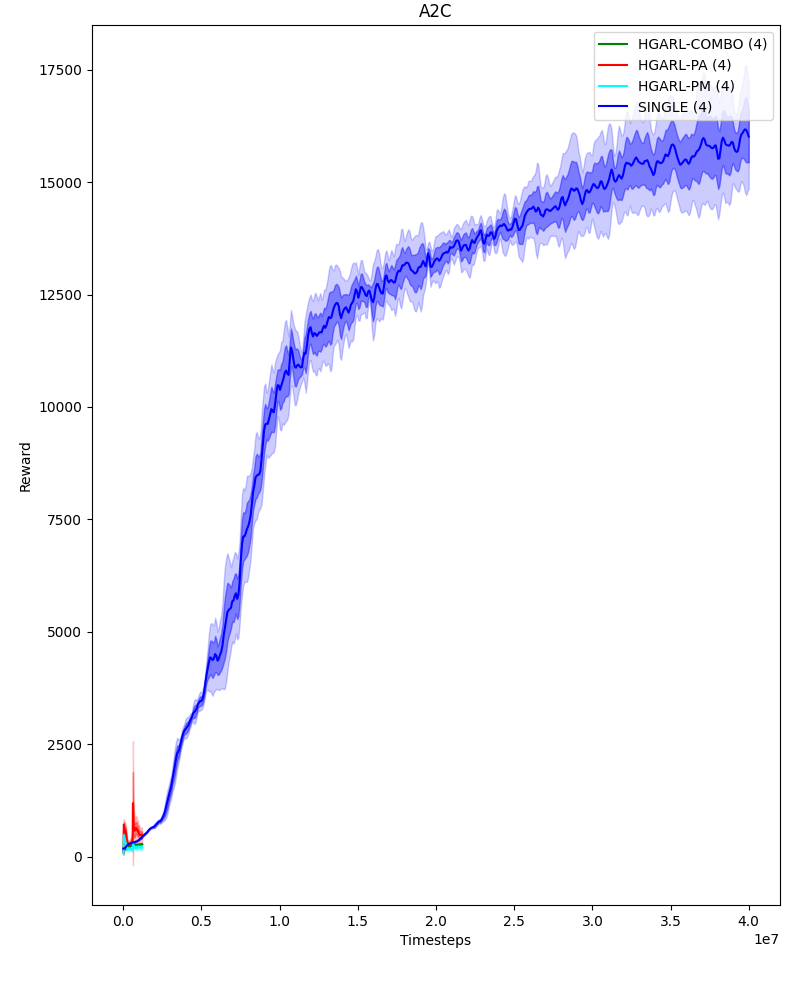}\hfill
                \includegraphics[width=0.333\linewidth]{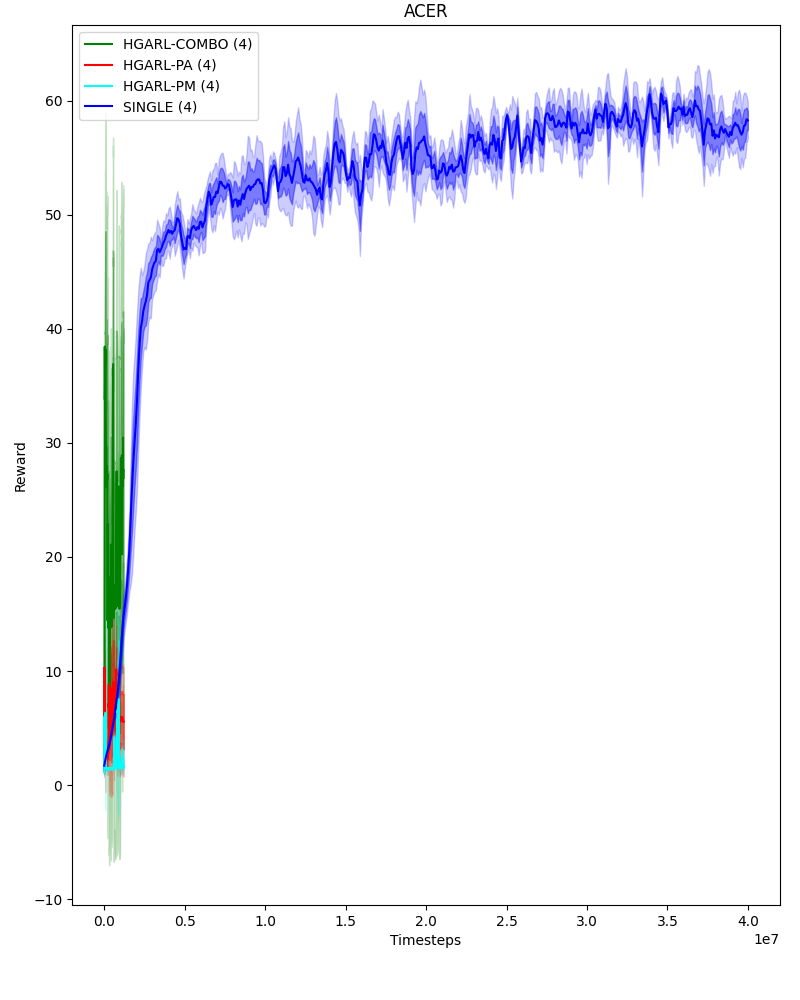}\hfill
                \includegraphics[width=0.333\linewidth]{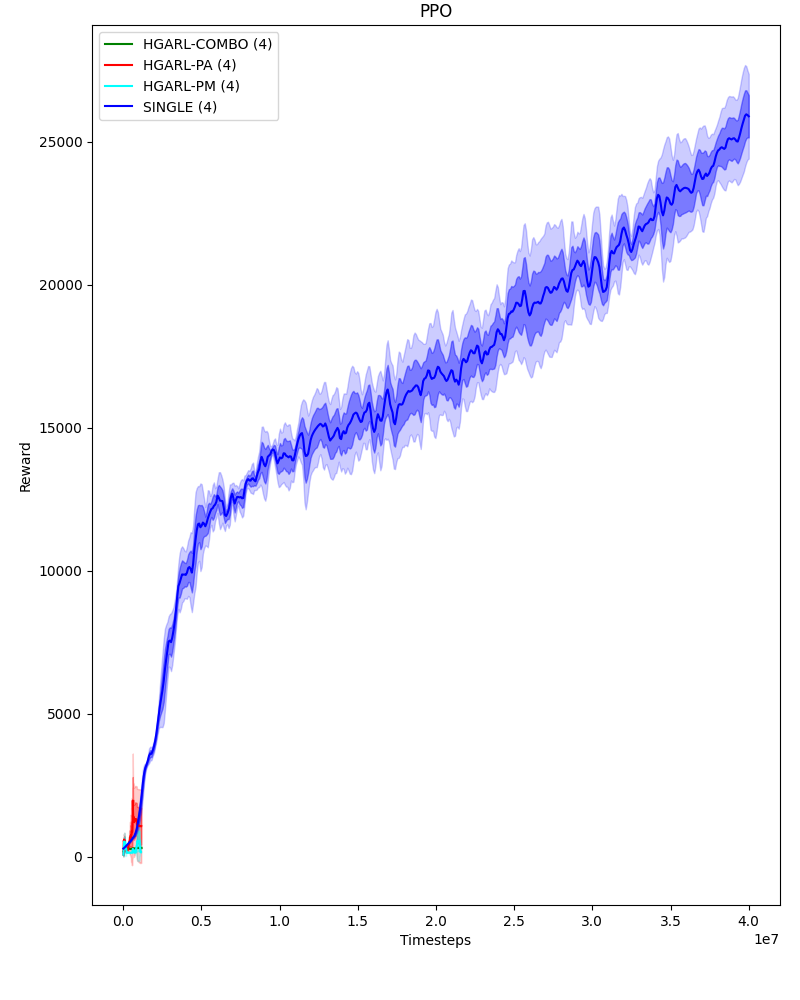}
                \caption{Qbert}
        \end{subfigure}
        \begin{subfigure}[b]{0.49\linewidth}
                \includegraphics[width=0.333\linewidth]{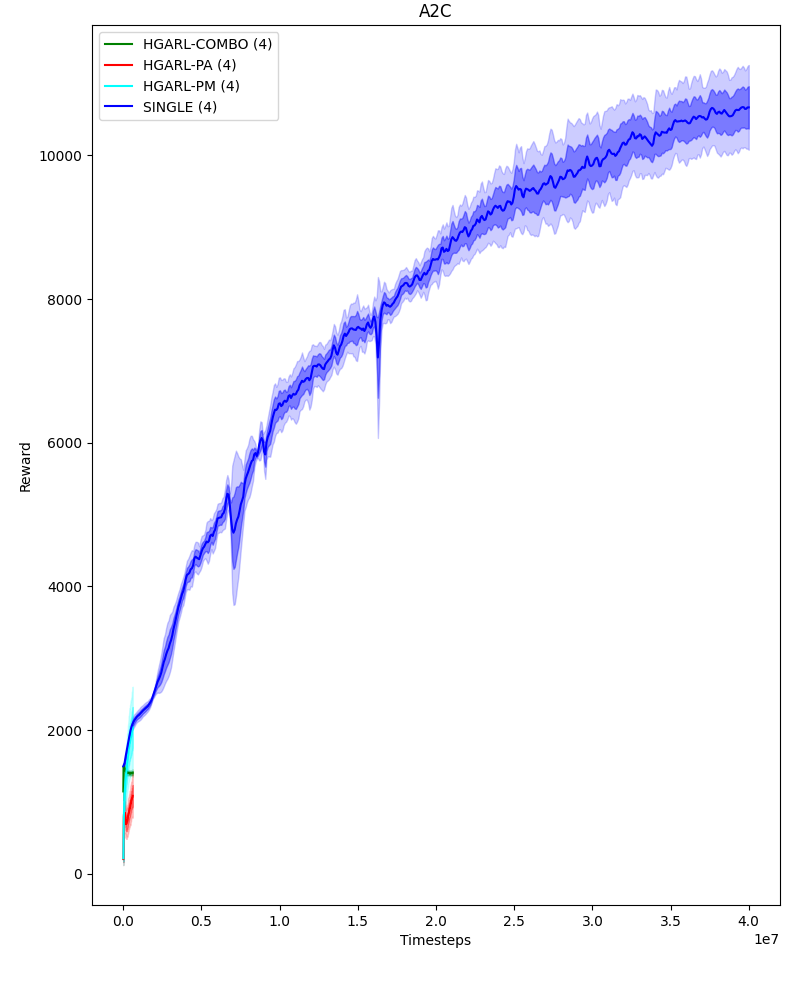}\hfill
                \includegraphics[width=0.333\linewidth]{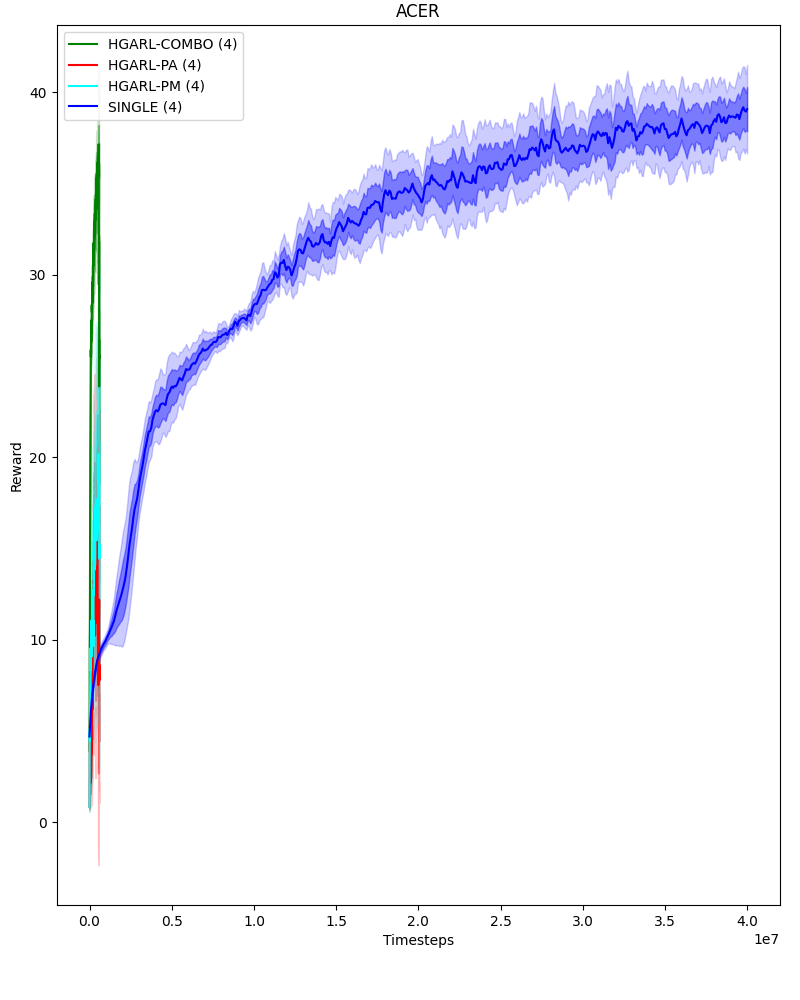}\hfill
                \includegraphics[width=0.333\linewidth]{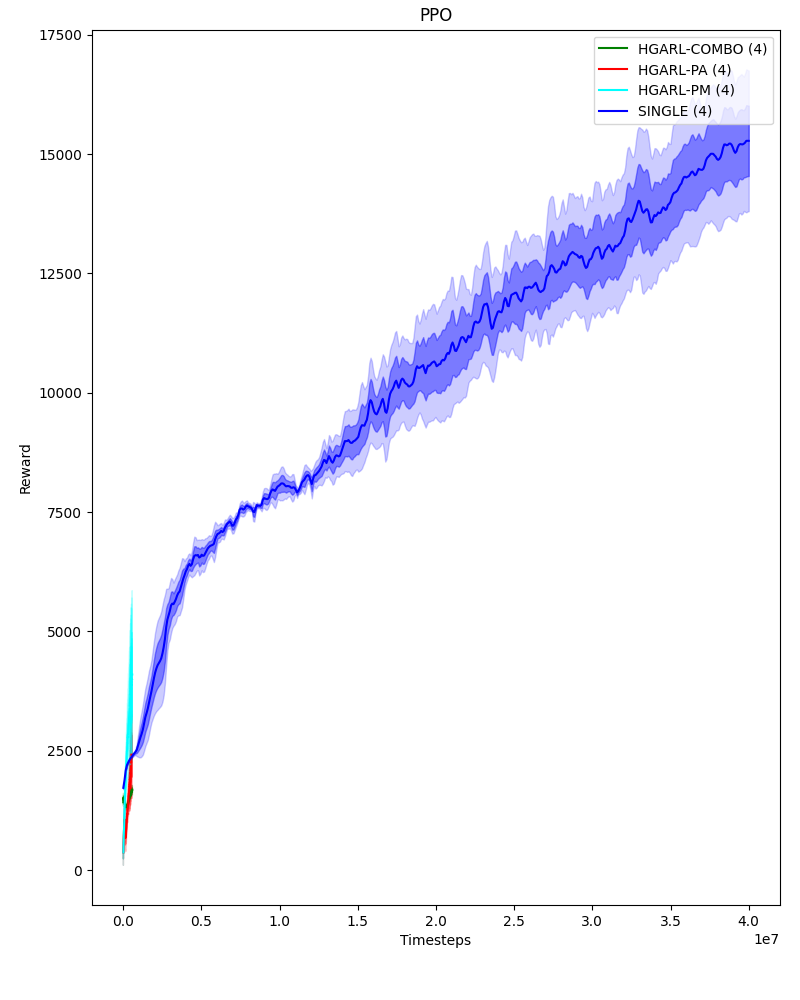}
                \caption{Riverraid}
        \end{subfigure}
        \begin{subfigure}[b]{0.49\linewidth}
                \includegraphics[width=0.333\linewidth]{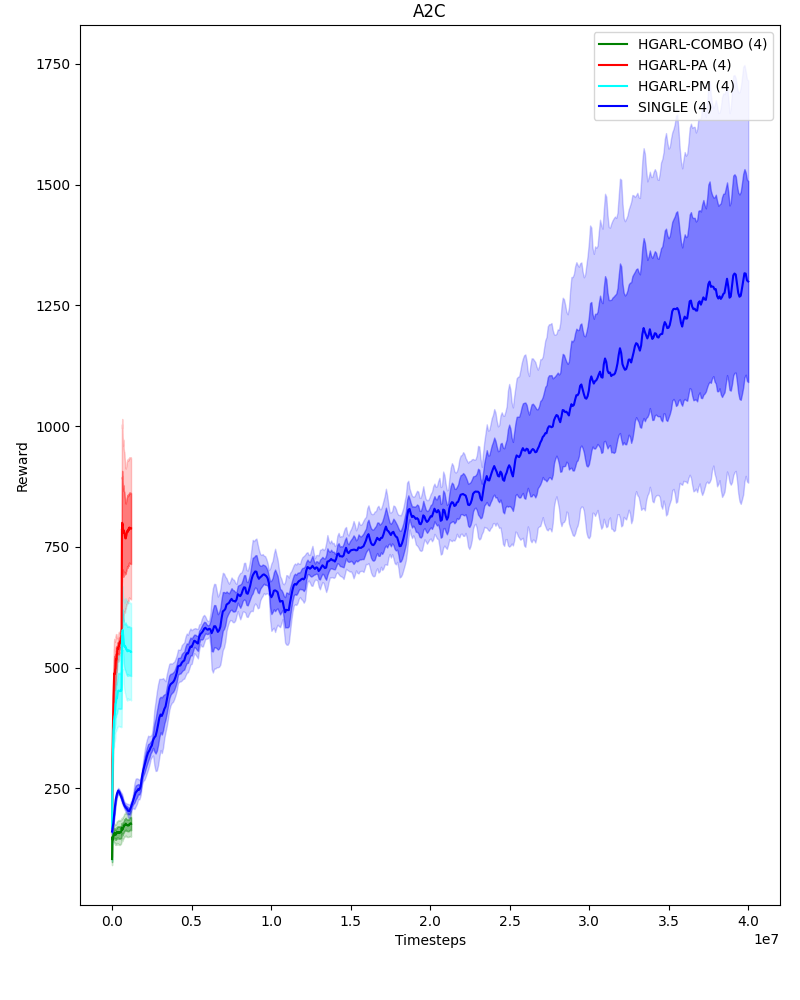}\hfill
                \includegraphics[width=0.333\linewidth]{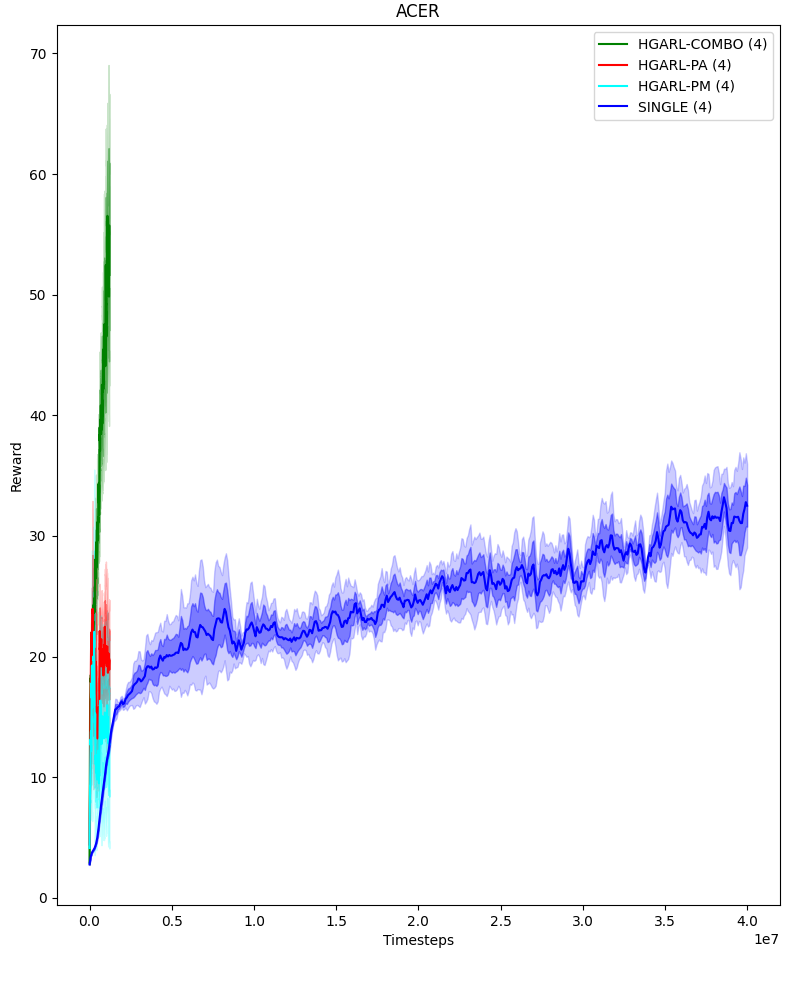}\hfill
                \includegraphics[width=0.333\linewidth]{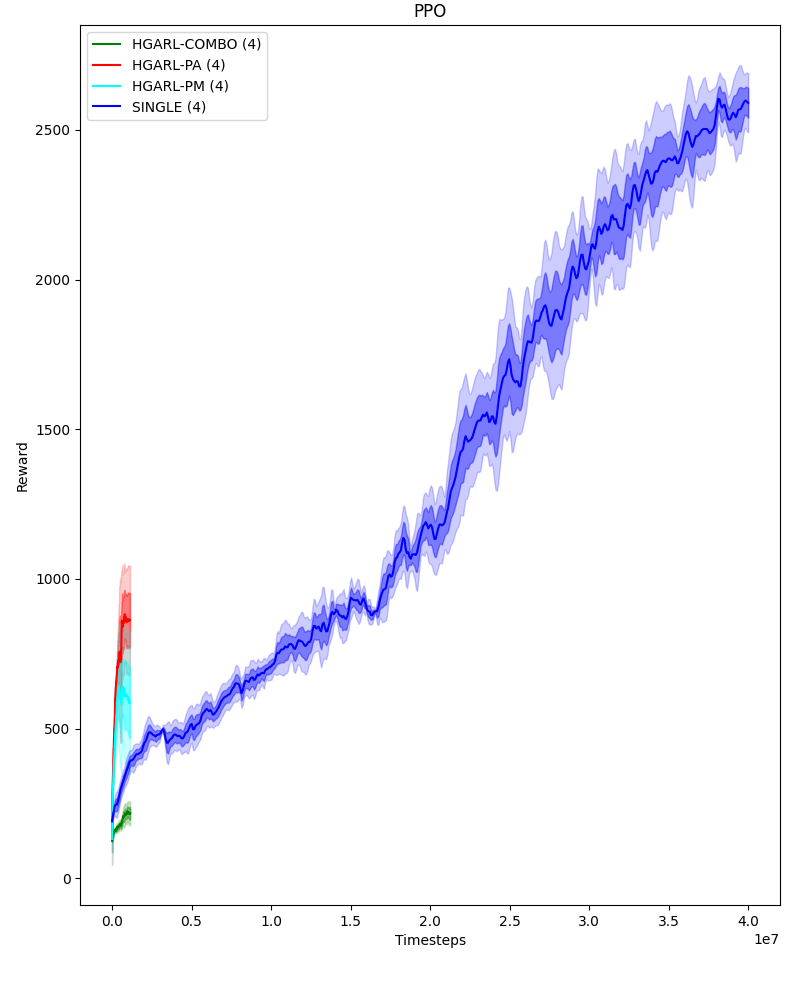}
                \caption{SpaceInvaders}
        \end{subfigure}
        \begin{subfigure}[b]{0.49\linewidth}
                \includegraphics[width=0.333\linewidth]{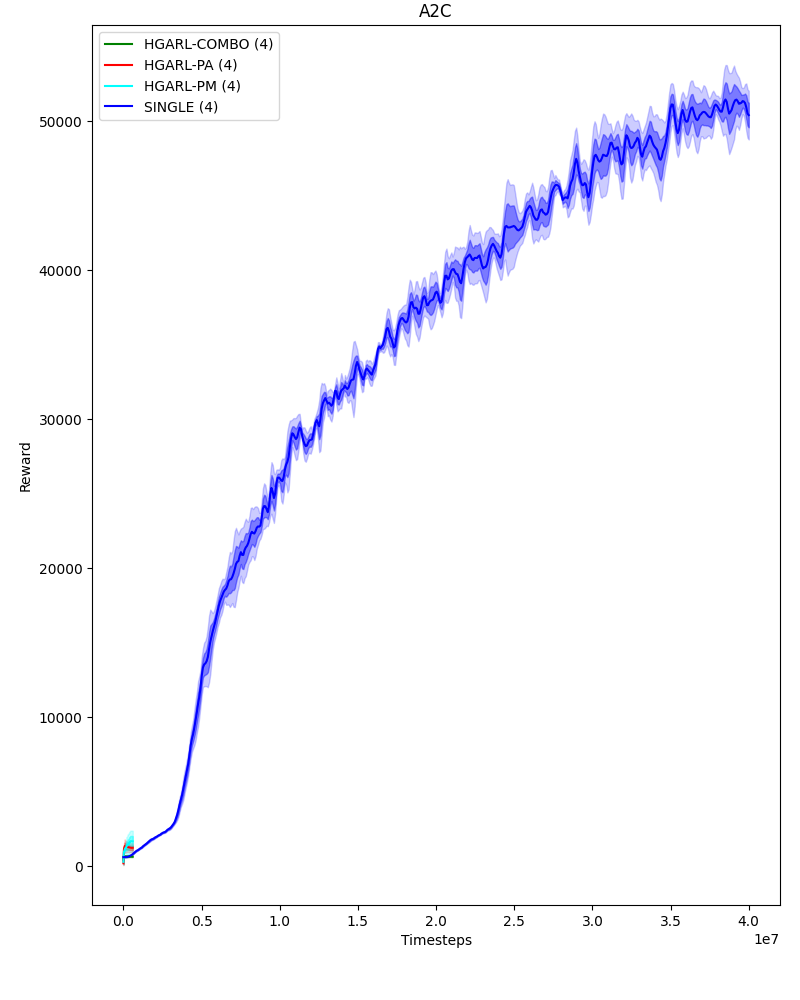}\hfill
                \includegraphics[width=0.333\linewidth]{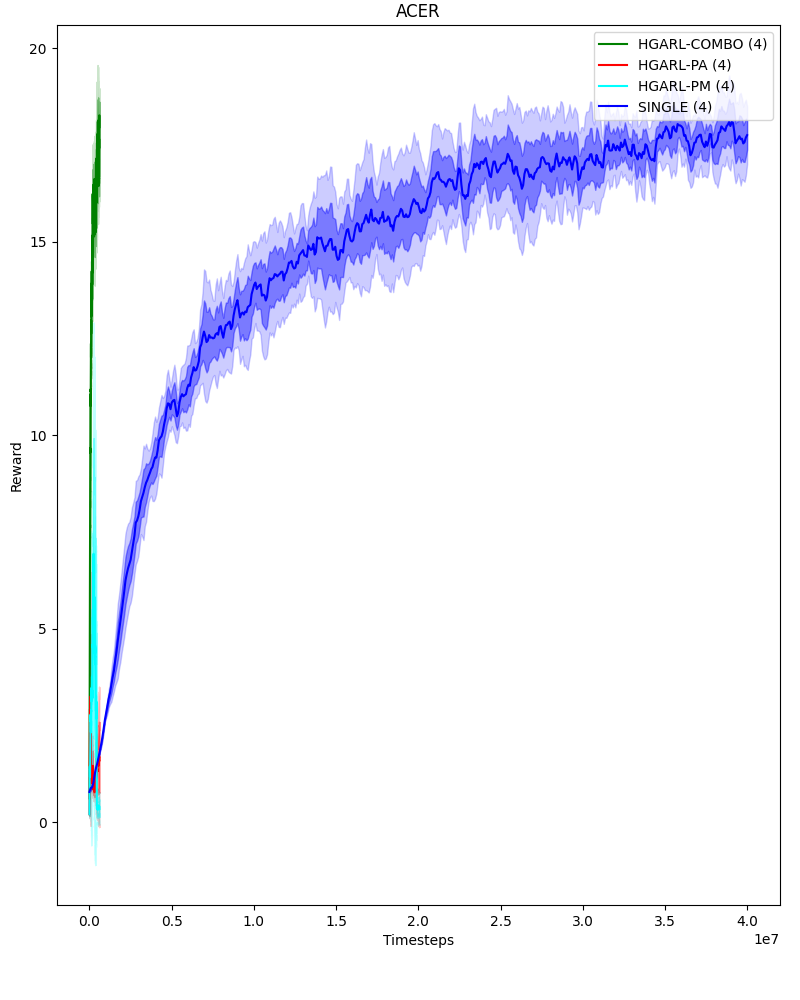}\hfill
                \includegraphics[width=0.333\linewidth]{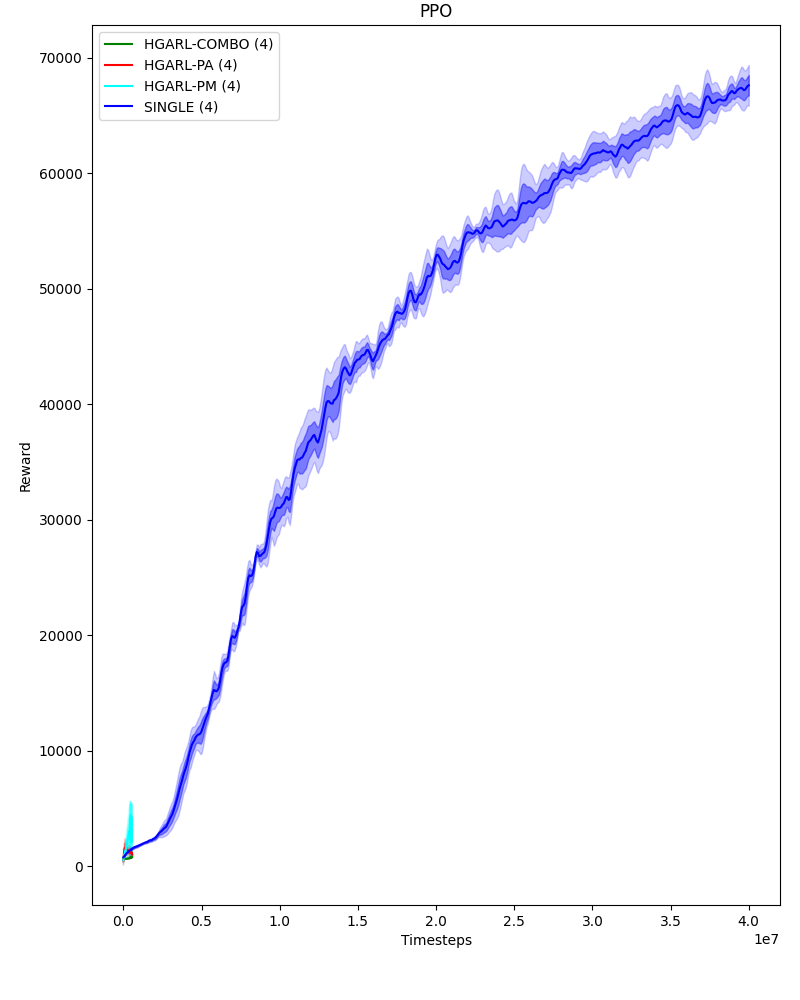}
                \caption{StarGunner}
        \end{subfigure}
        \begin{subfigure}[b]{0.49\linewidth}
                \includegraphics[width=0.333\linewidth]{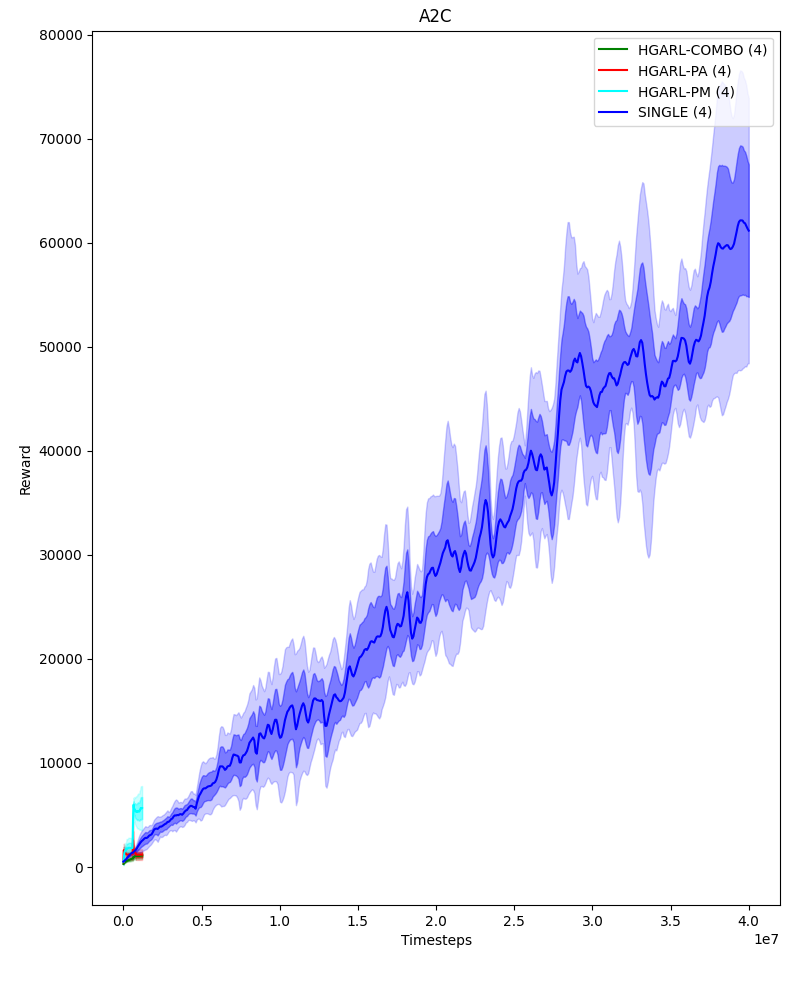}\hfill
                \includegraphics[width=0.333\linewidth]{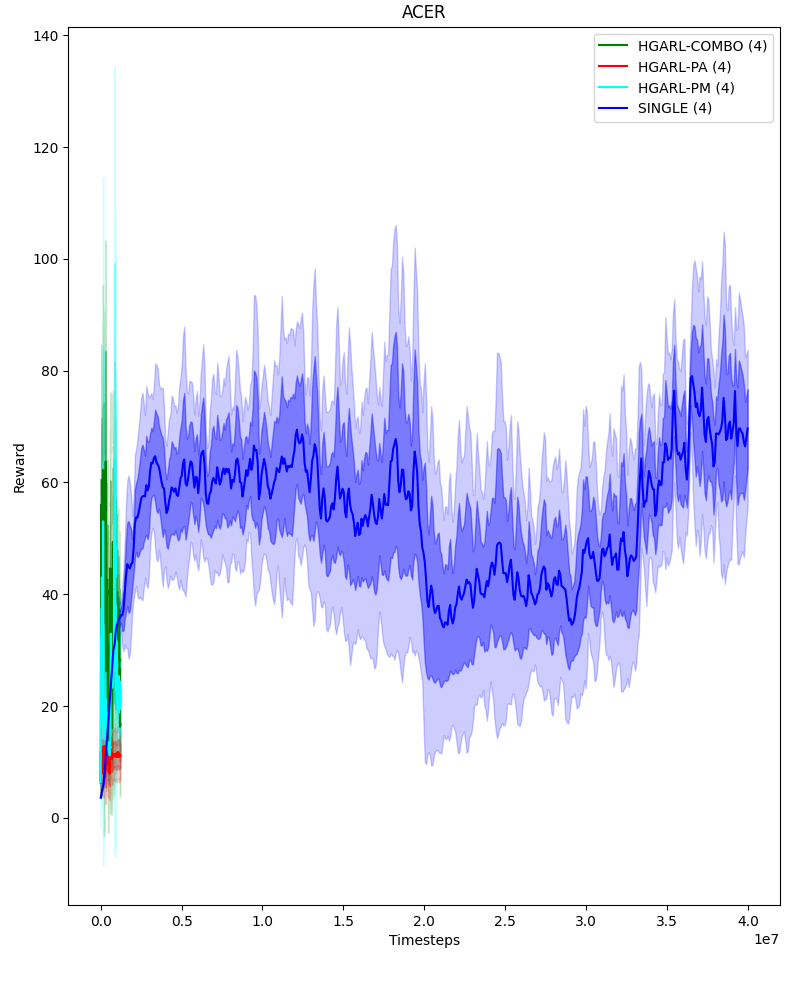}\hfill
                \includegraphics[width=0.333\linewidth]{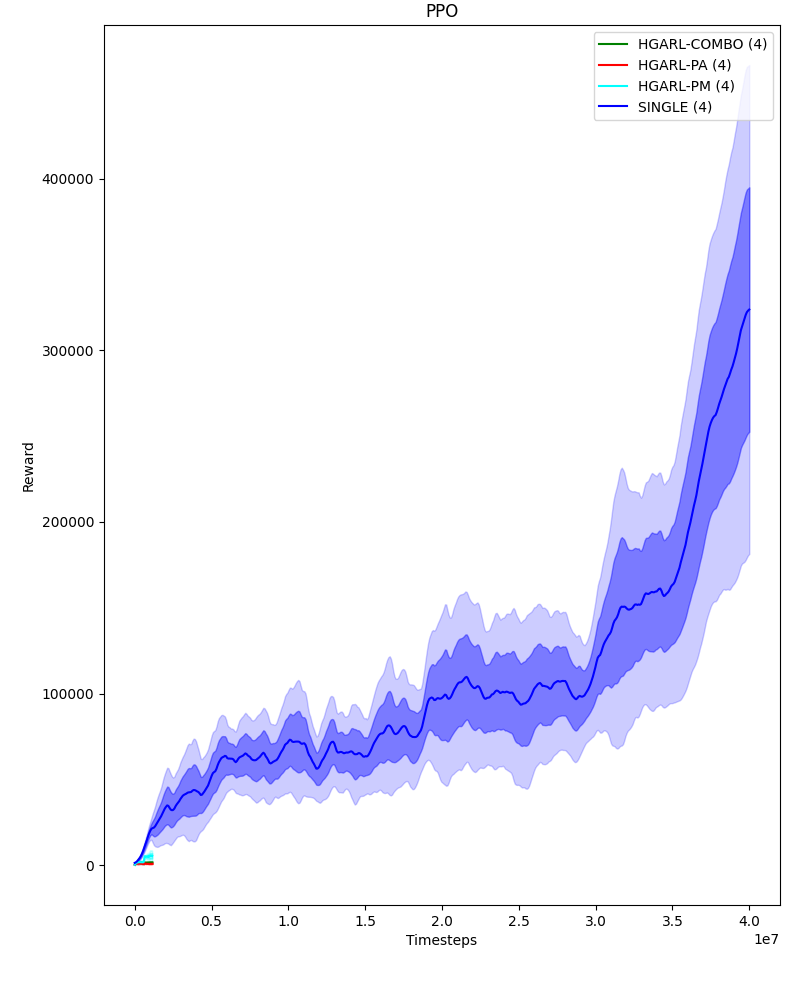}
                \caption{UpNDown}
        \end{subfigure}
        \label{Atari2}
 \end{figure*}

 \begin{figure*}
       \ContinuedFloat 
        \begin{subfigure}[b]{0.49\linewidth}
                \includegraphics[width=0.333\linewidth]{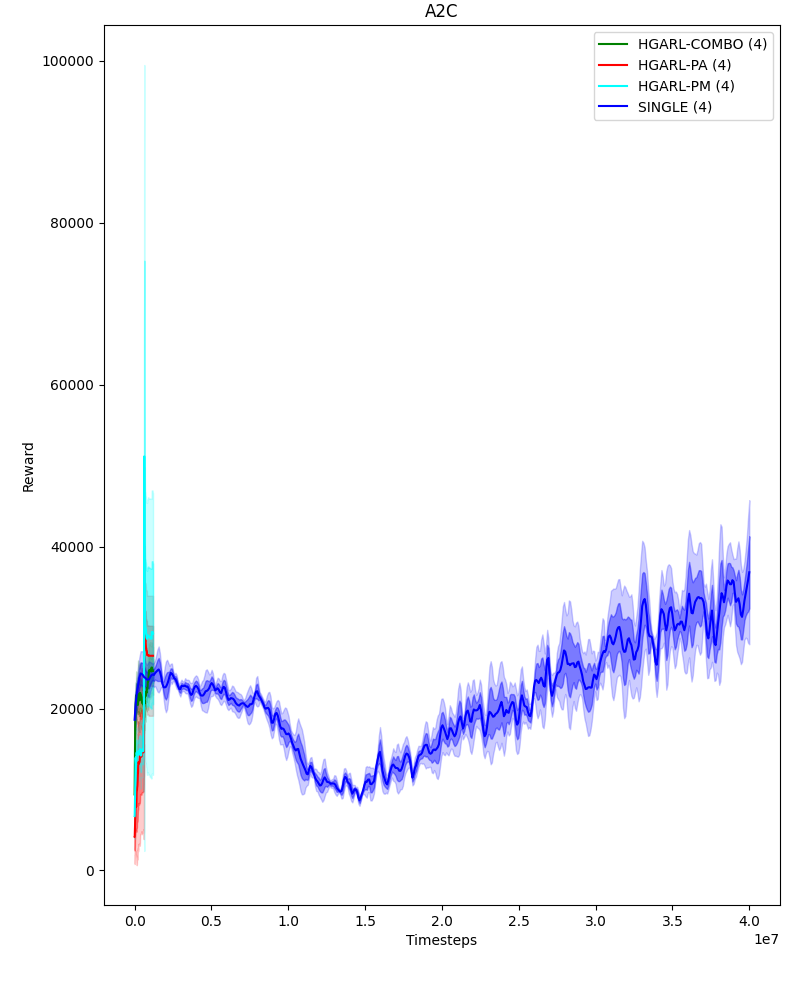}\hfill
                \includegraphics[width=0.333\linewidth]{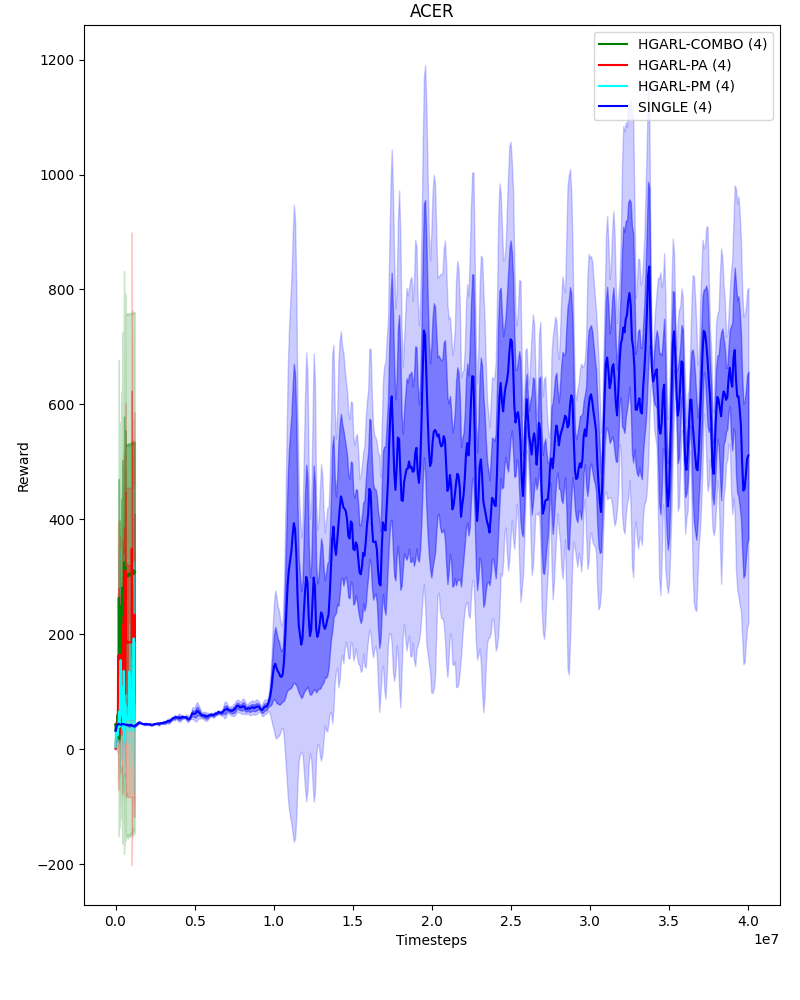}\hfill
                \includegraphics[width=0.333\linewidth]{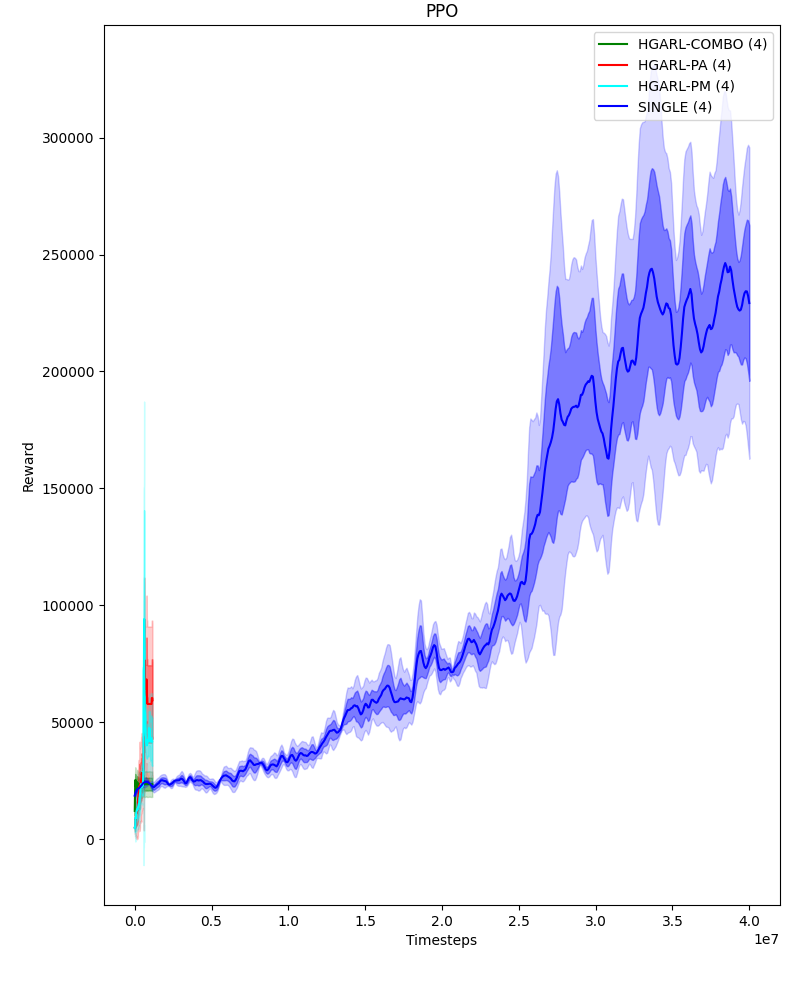}
                \caption{VideoPinball}
        \end{subfigure}
        \begin{subfigure}[b]{0.49\linewidth}
                \includegraphics[width=0.333\linewidth]{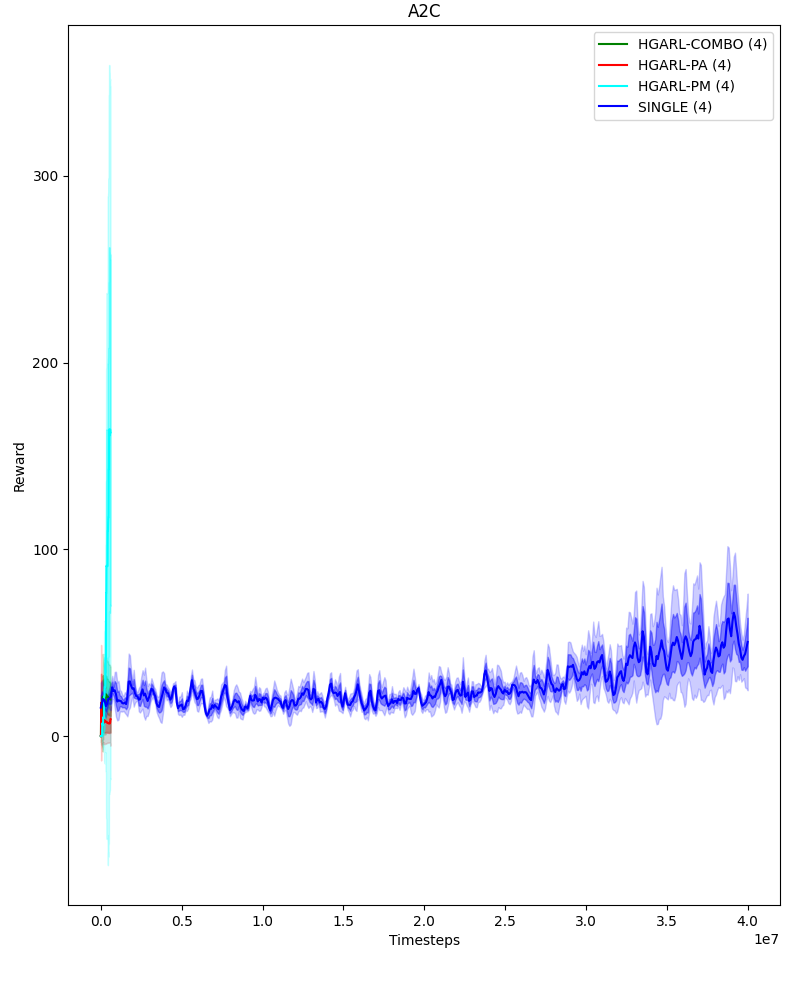}\hfill
                \includegraphics[width=0.333\linewidth]{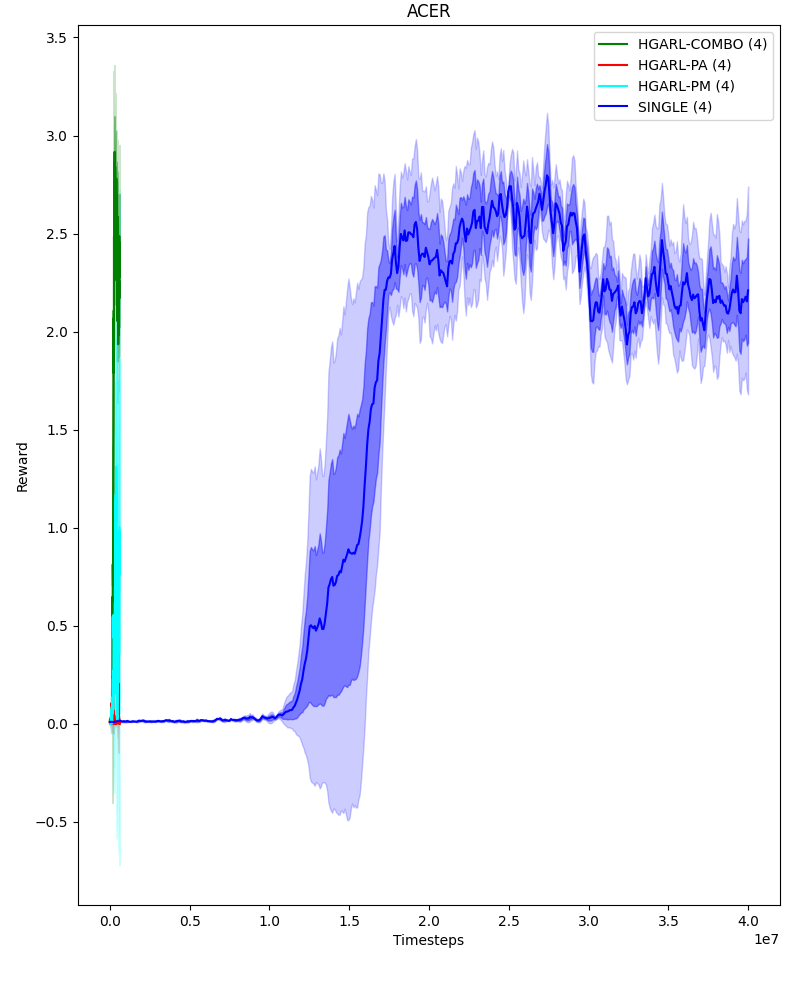}\hfill
                \includegraphics[width=0.333\linewidth]{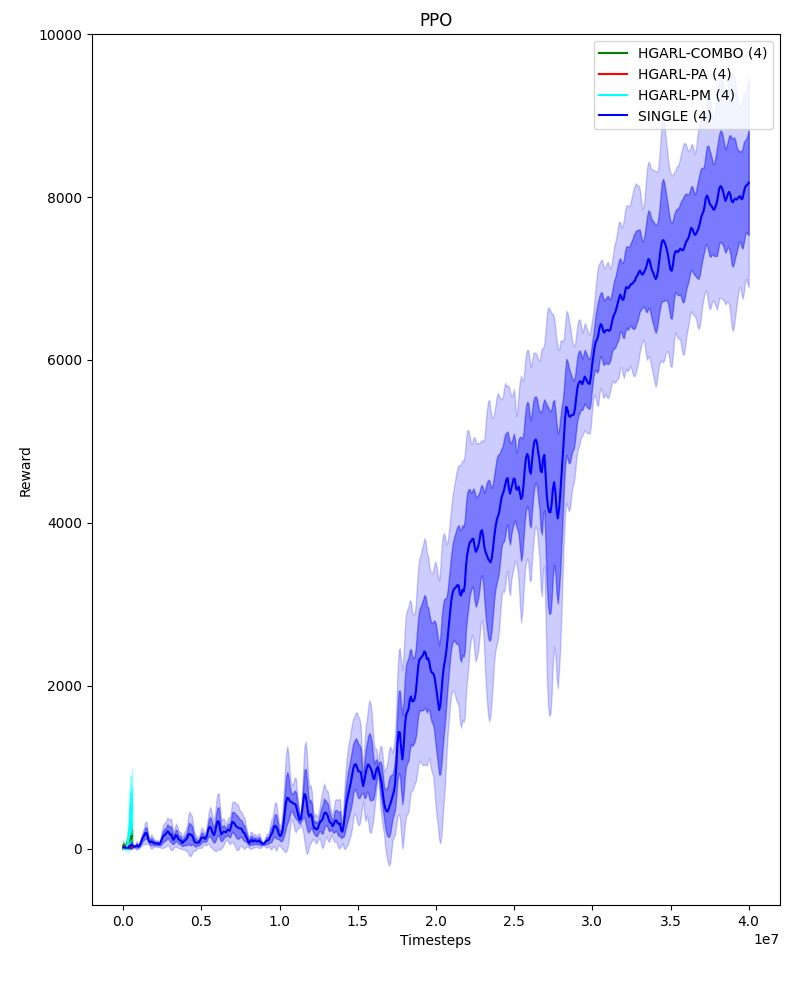}
                \caption{Zaxxon}
        \end{subfigure}
        \caption{Atari 2600 Games: Part 2. The ACER agents are greatly improved under the Combo rule.}
        \label{Atari2}
\end{figure*}

\begin{figure*}
       \begin{subfigure}[b]{0.49\linewidth}
                \includegraphics[width=0.333\linewidth]{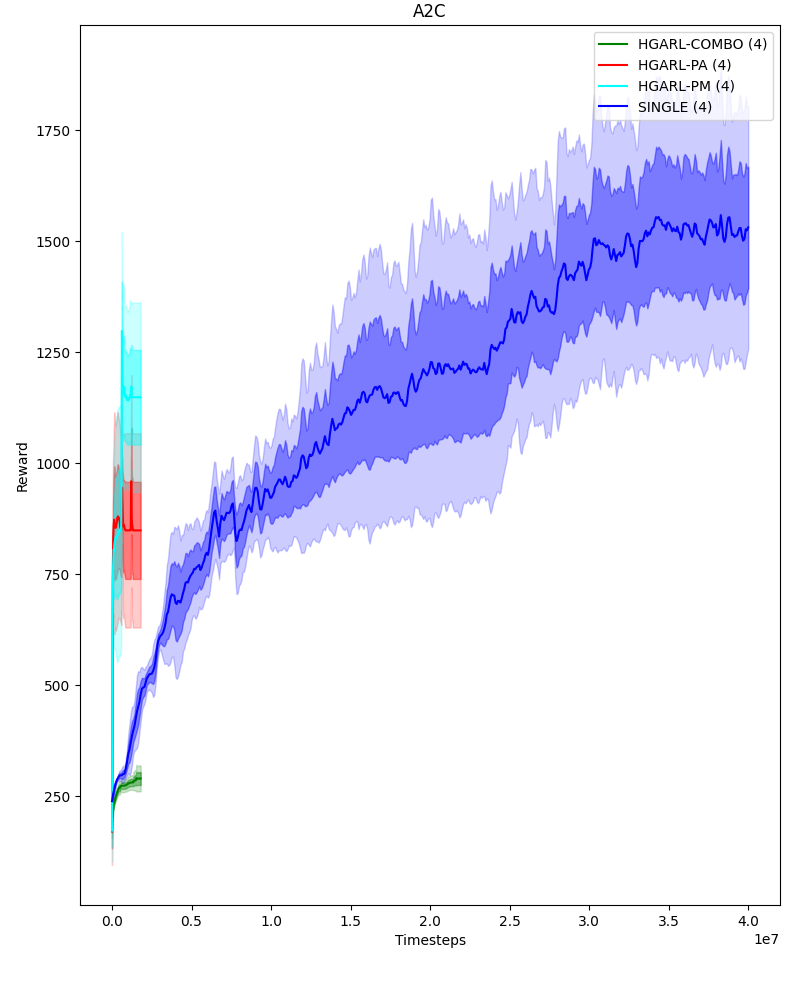}\hfill
                \includegraphics[width=0.333\linewidth]{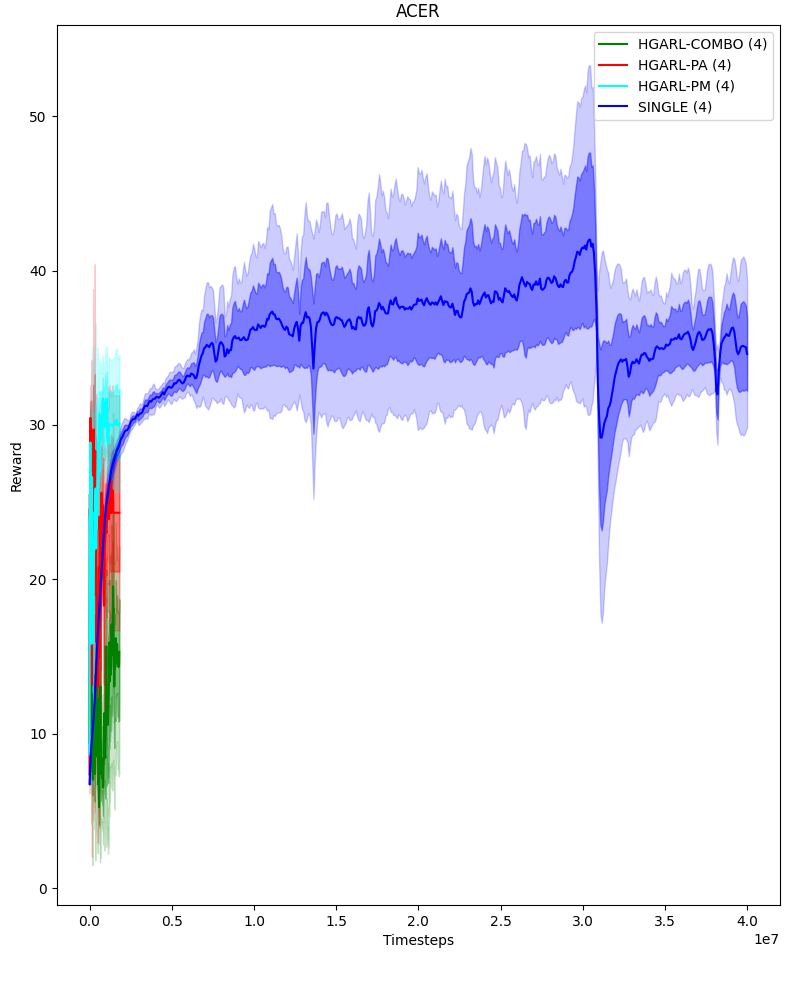}\hfill
                \includegraphics[width=0.333\linewidth]{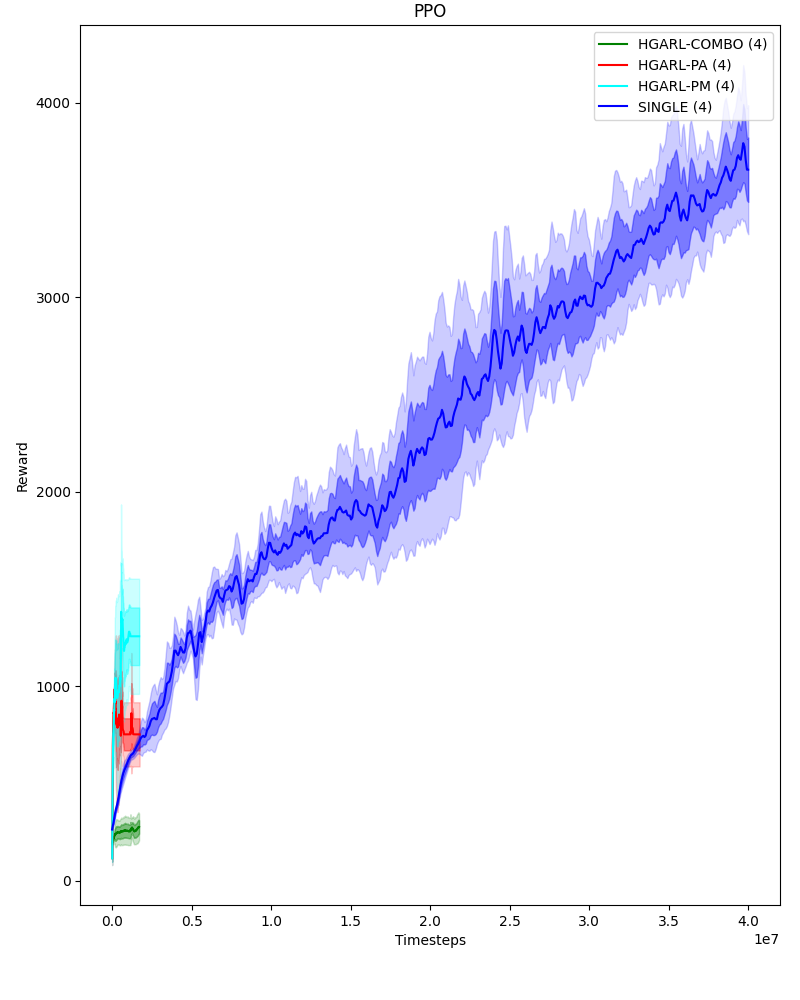}
                \caption{Alien}
        \end{subfigure}
        \begin{subfigure}[b]{0.49\linewidth}
                \includegraphics[width=0.333\linewidth]{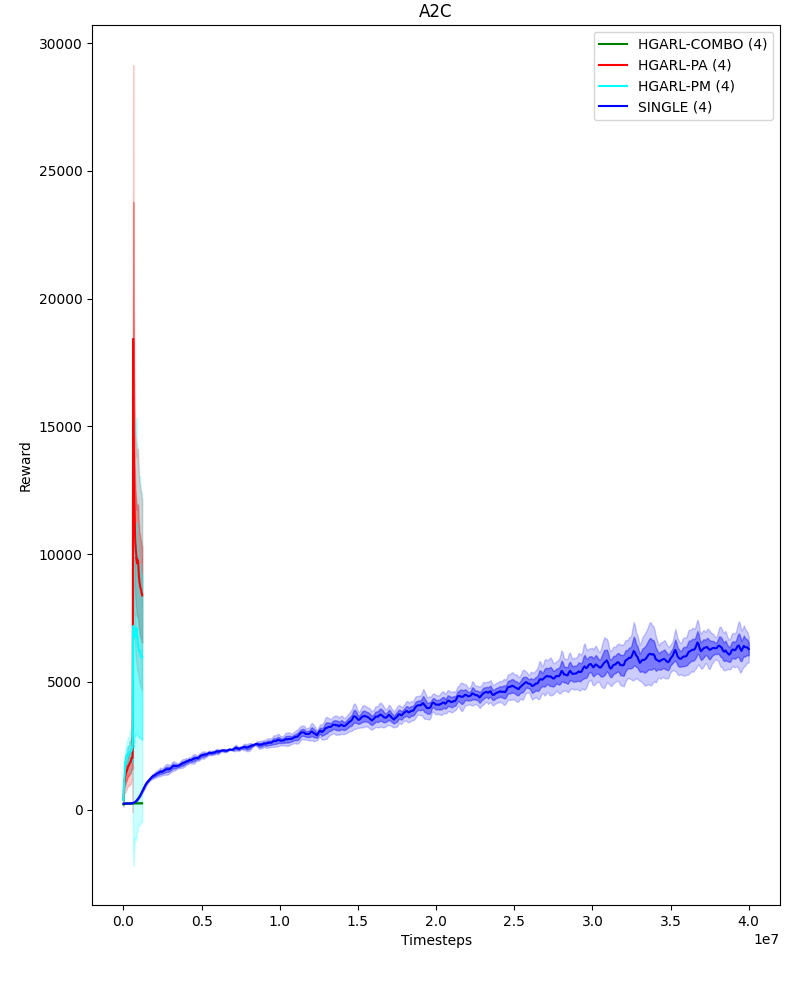}\hfill
                \includegraphics[width=0.333\linewidth]{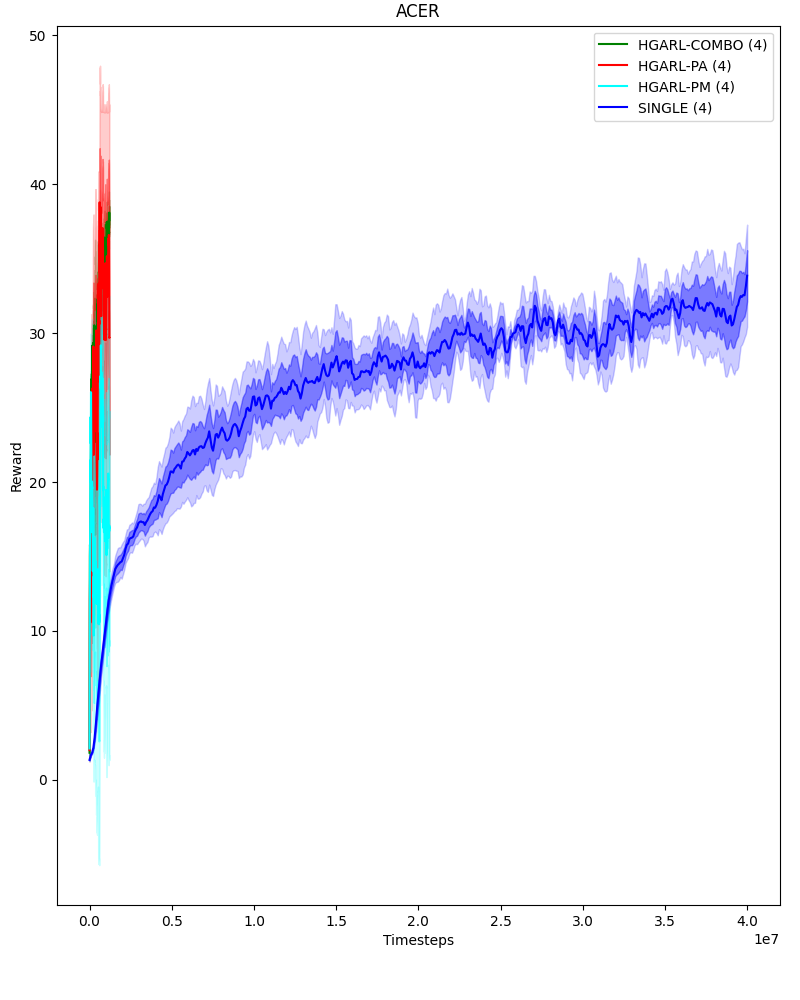}\hfill
                \includegraphics[width=0.333\linewidth]{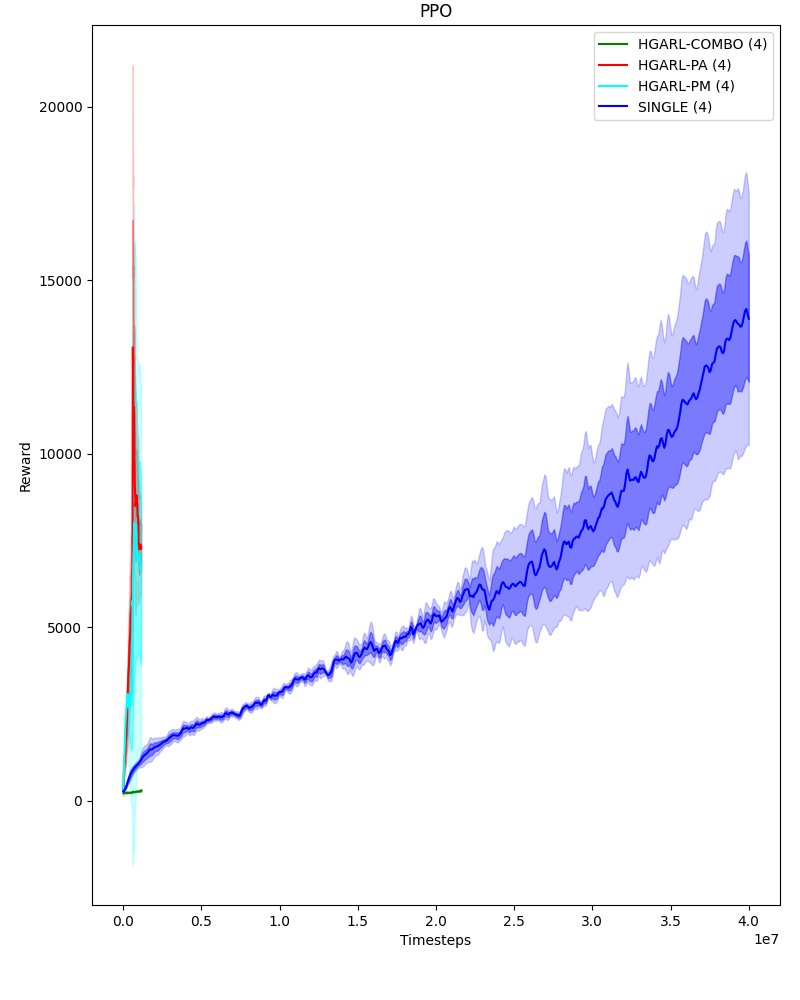}
                \caption{Asterix}
        \end{subfigure}
        \begin{subfigure}[b]{0.49\linewidth}
                \includegraphics[width=0.333\linewidth]{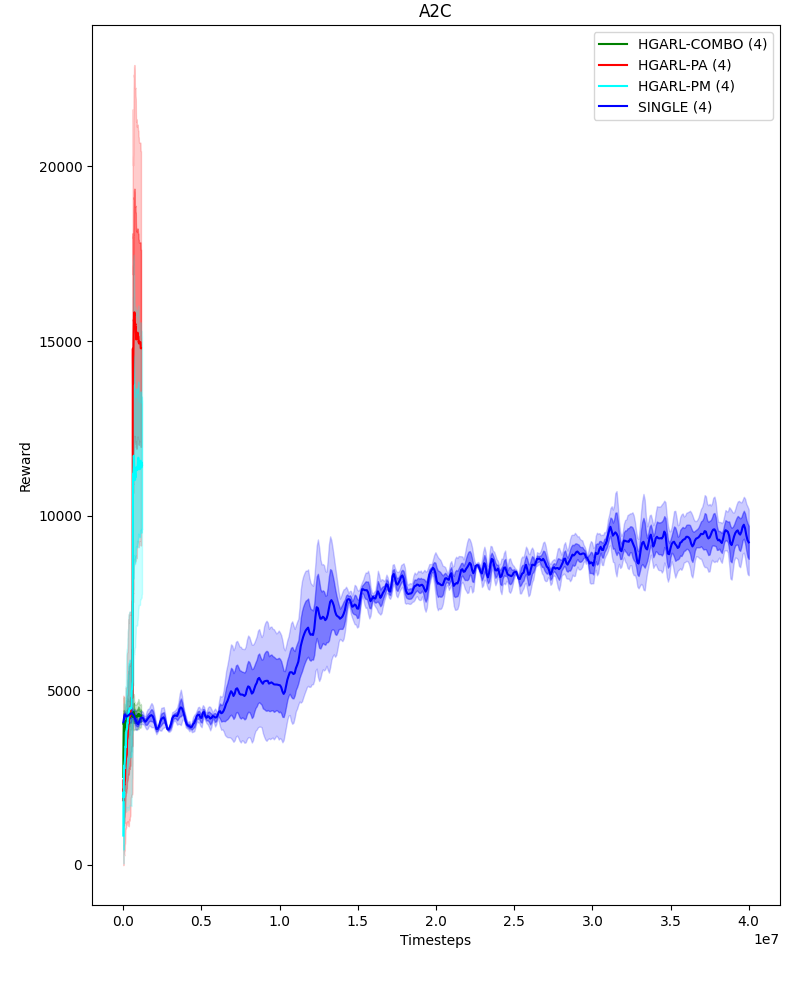}\hfill
                \includegraphics[width=0.333\linewidth]{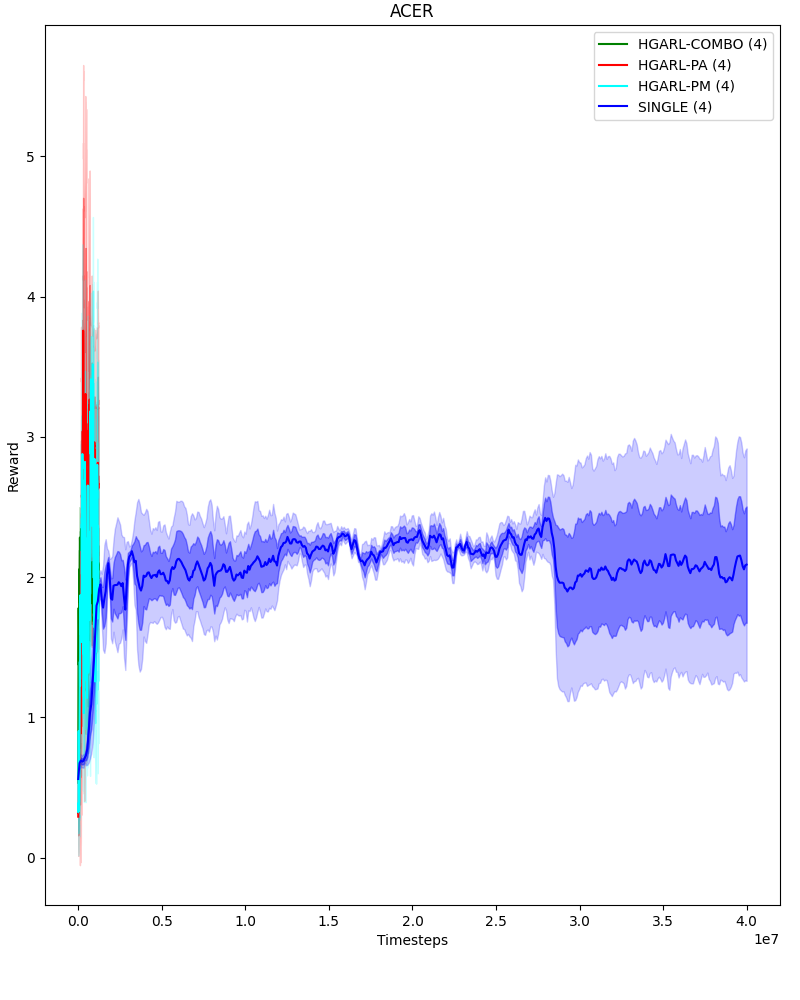}\hfill
                \includegraphics[width=0.333\linewidth]{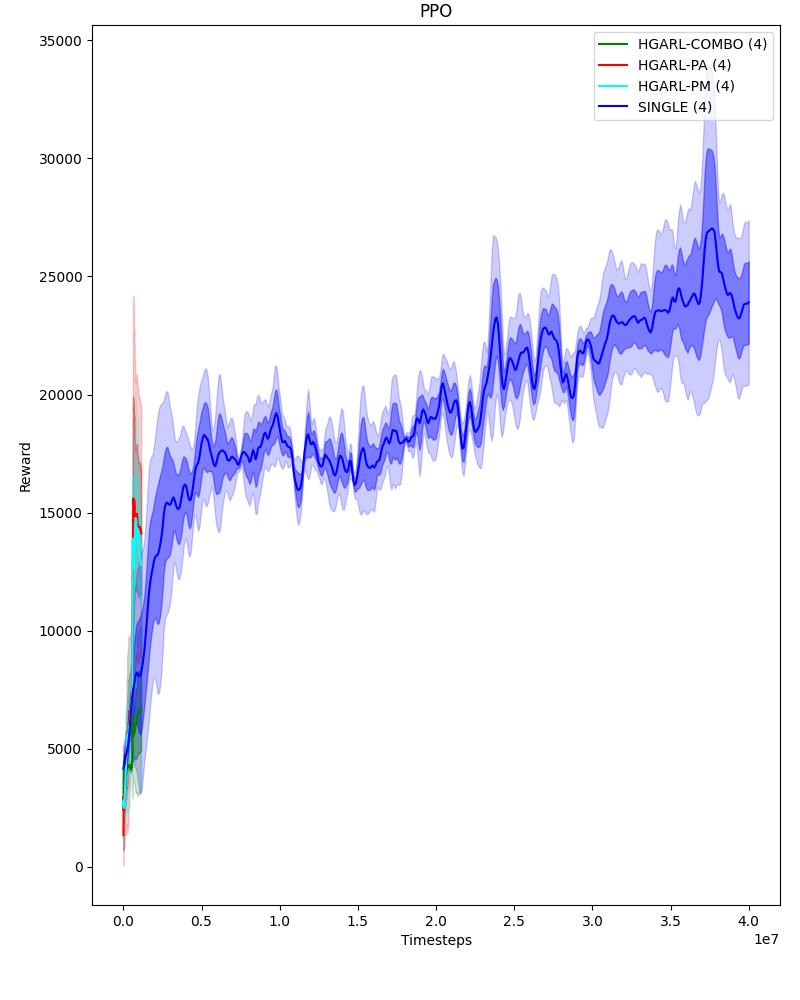}
                \caption{BattleZone}
        \end{subfigure}
        \begin{subfigure}[b]{0.49\linewidth}
                \includegraphics[width=0.333\linewidth]{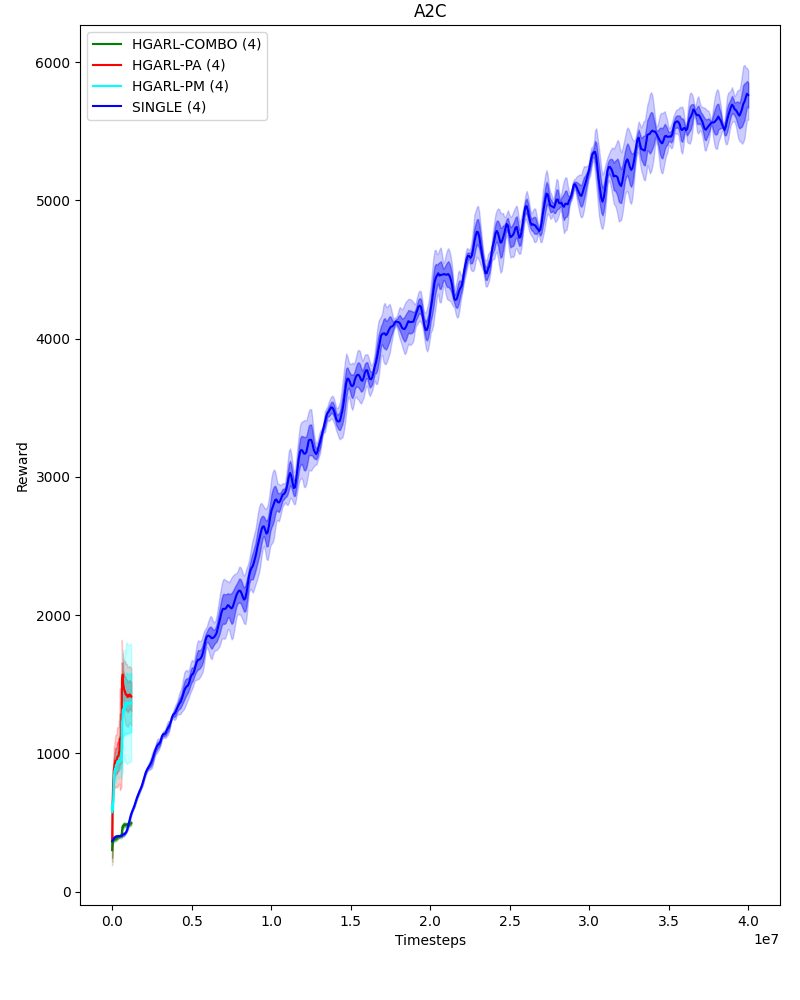}\hfill
                \includegraphics[width=0.333\linewidth]{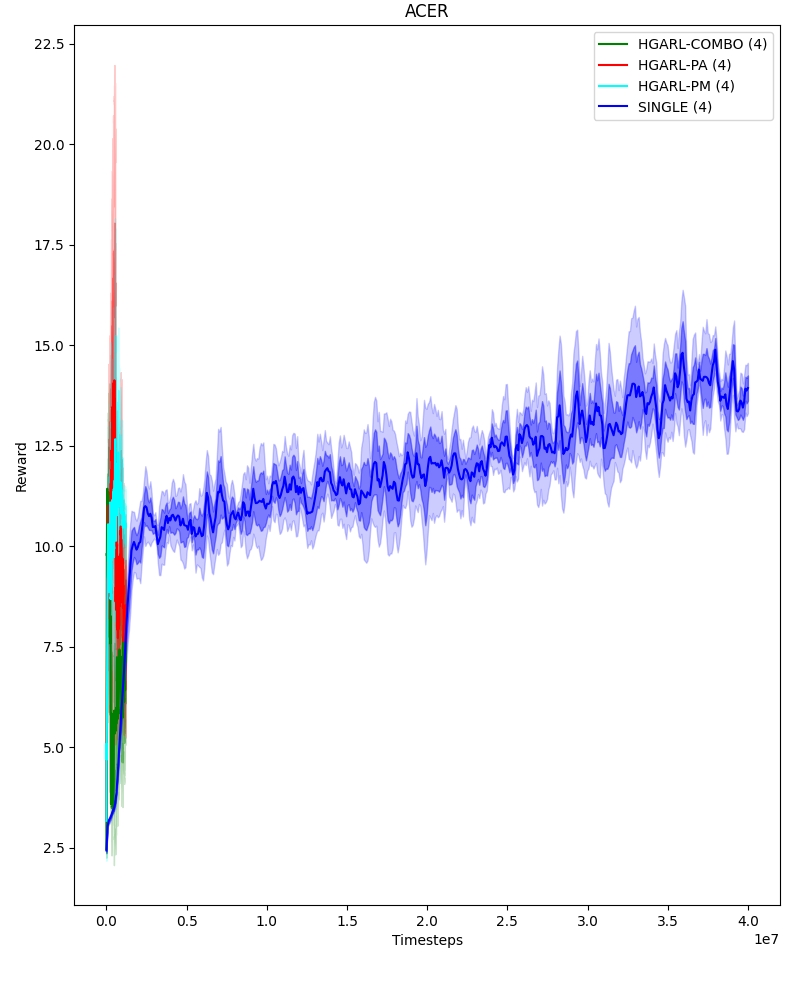}\hfill
                \includegraphics[width=0.333\linewidth]{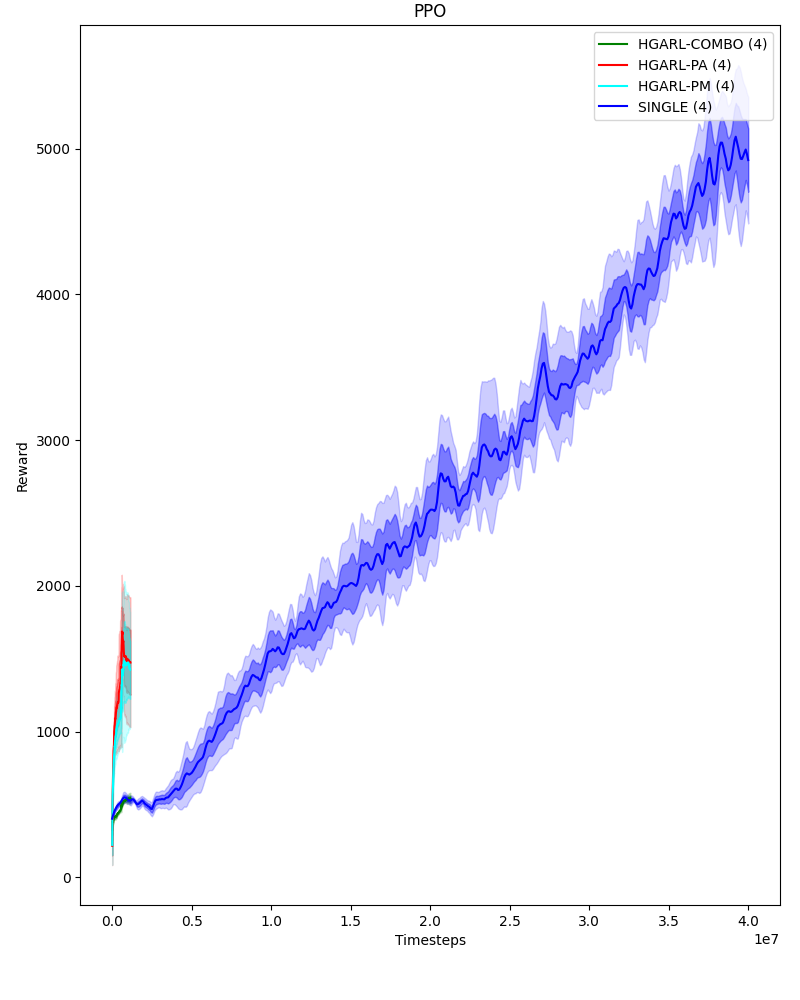}
                \caption{BeamRider}
        \end{subfigure}
        \begin{subfigure}[b]{0.49\linewidth}
                \includegraphics[width=0.333\linewidth]{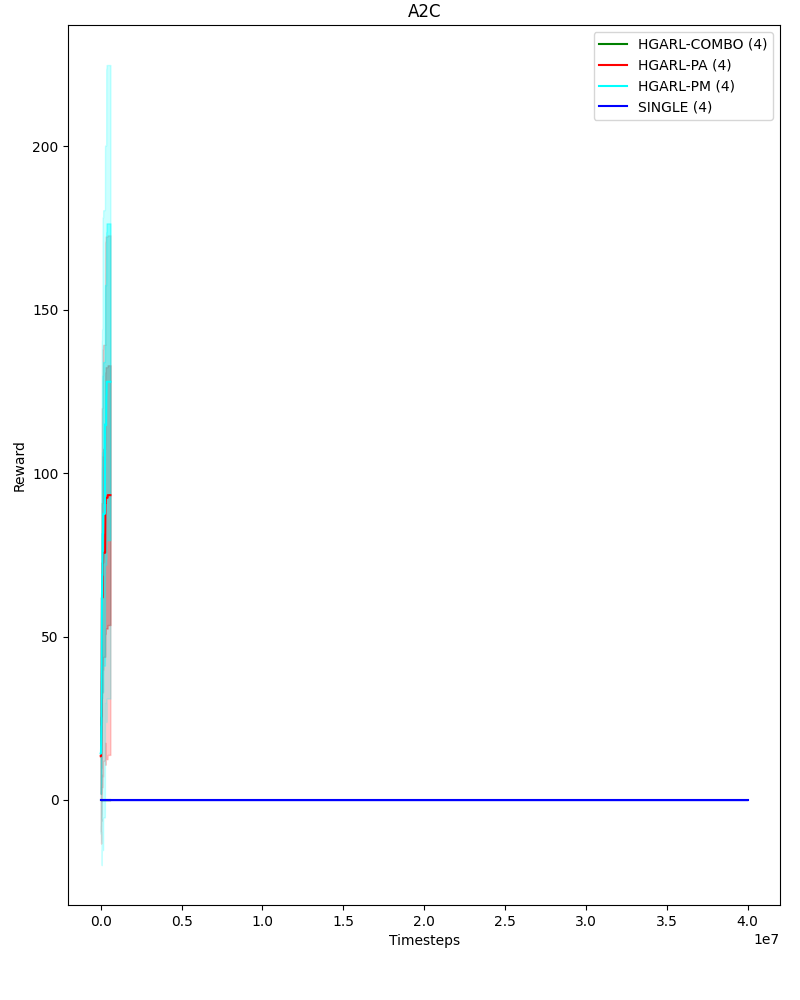}\hfill
                \includegraphics[width=0.333\linewidth]{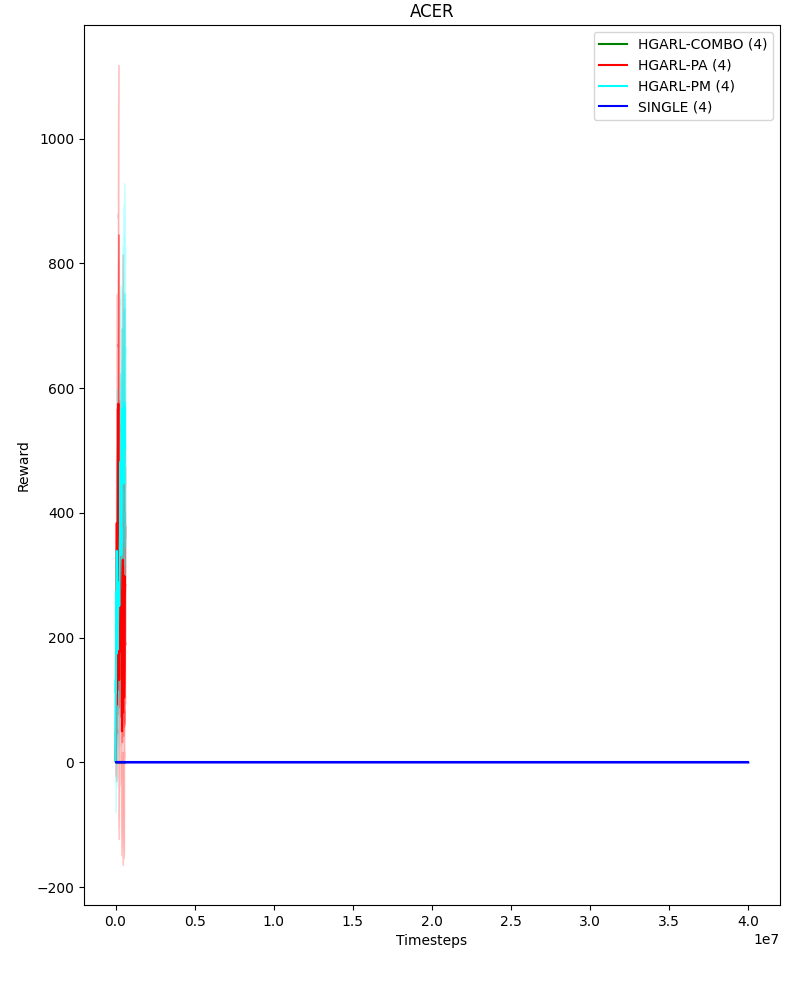}\hfill
                \includegraphics[width=0.333\linewidth]{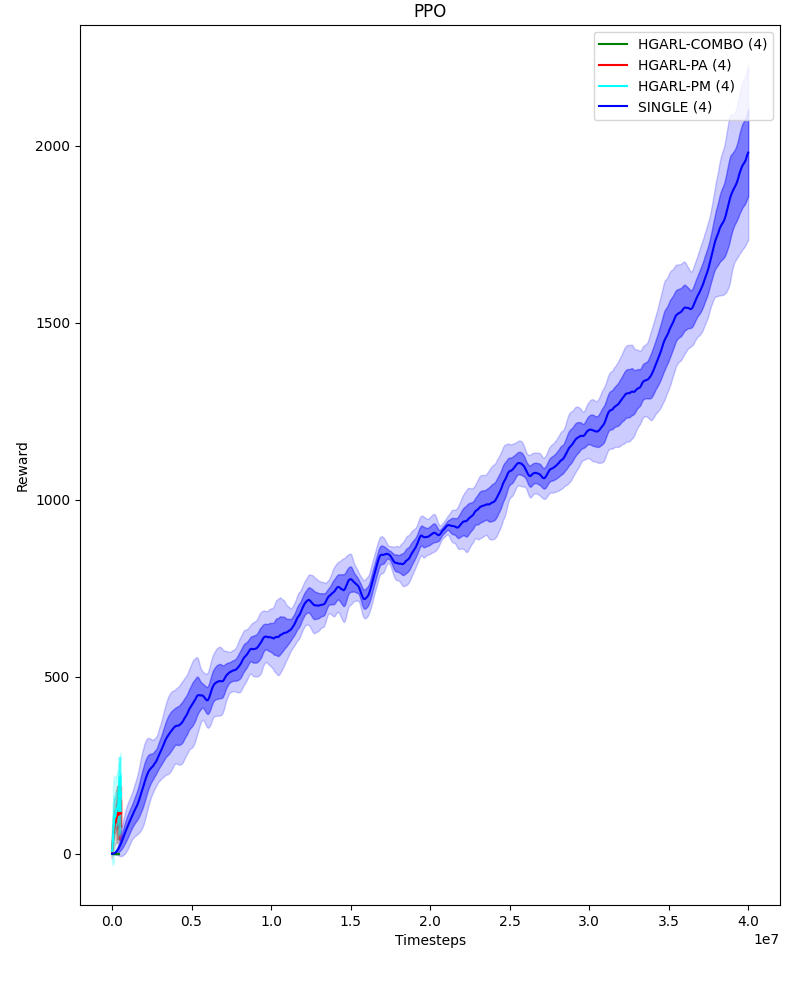}
                \caption{Enduro}
        \end{subfigure}
        \begin{subfigure}[b]{0.49\linewidth}
                \includegraphics[width=0.333\linewidth]{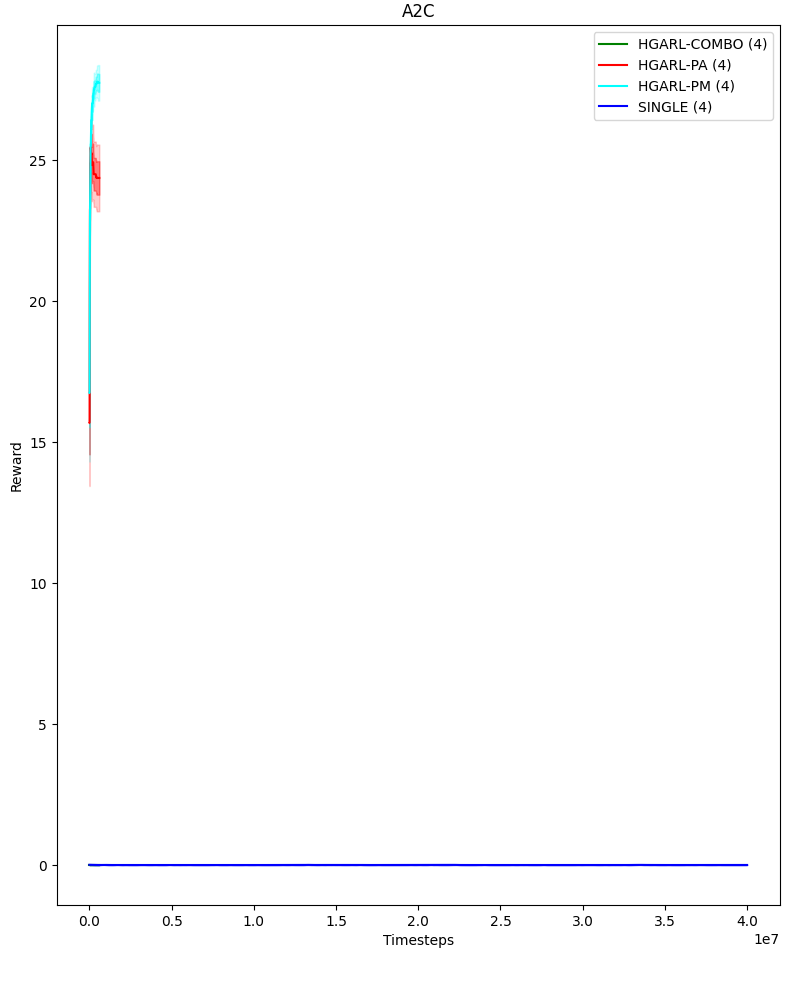}\hfill
                \includegraphics[width=0.333\linewidth]{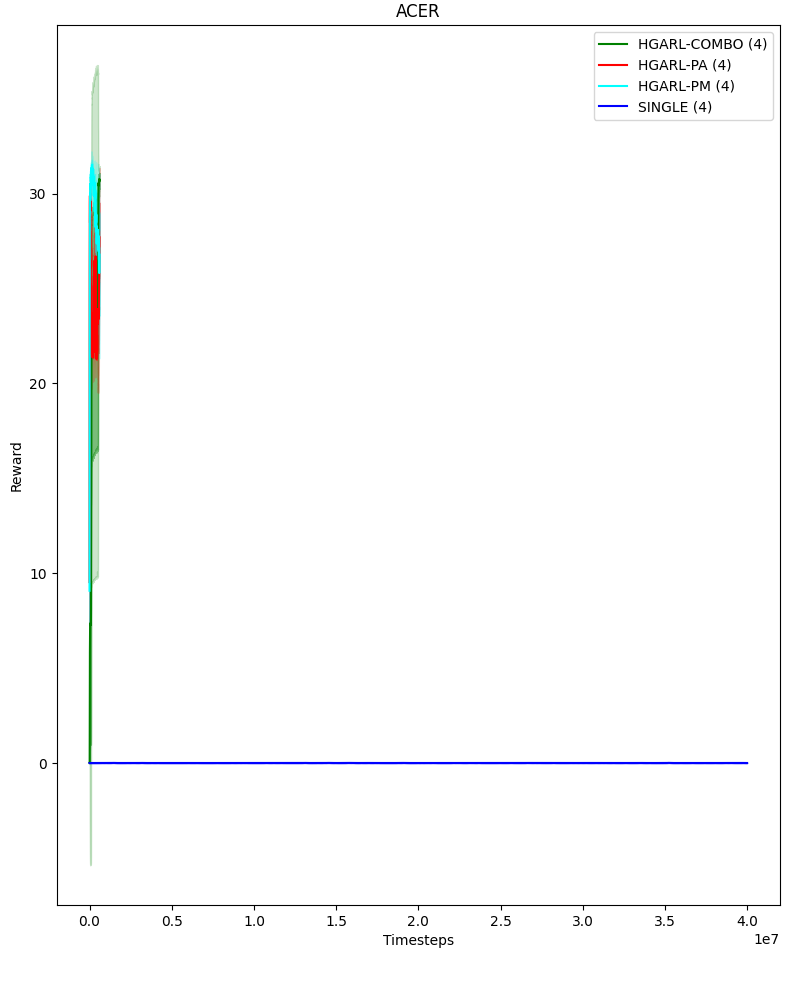}\hfill
                \includegraphics[width=0.333\linewidth]{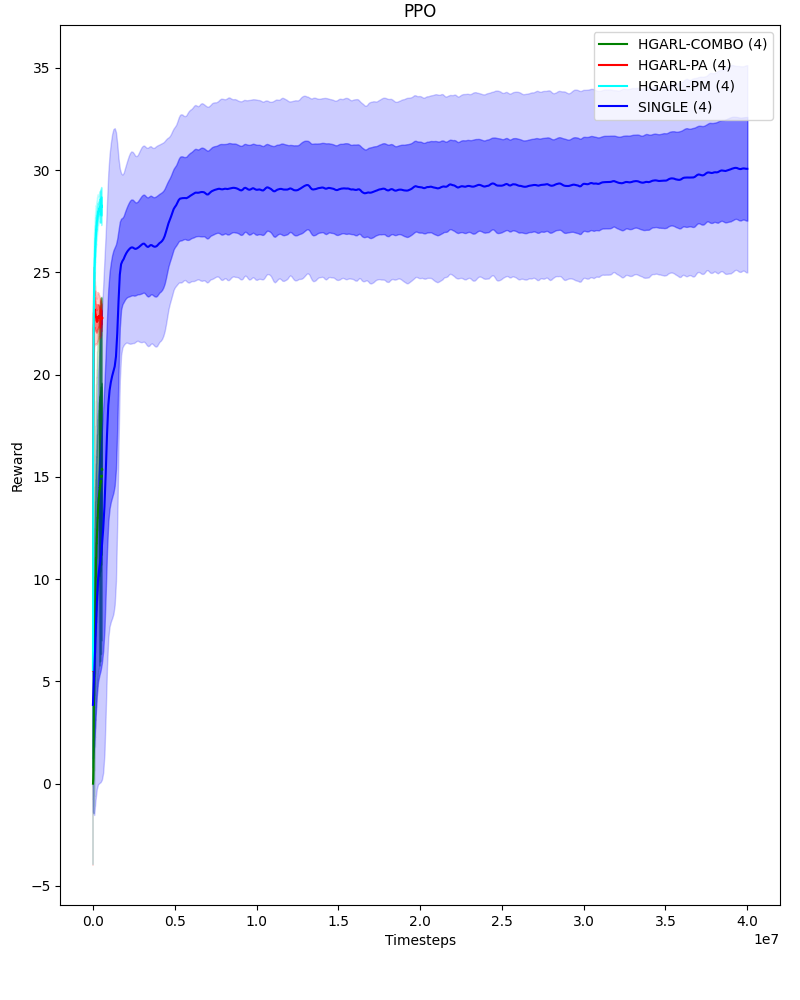}
                \caption{Freeway}
        \end{subfigure}
        \begin{subfigure}[b]{0.49\linewidth}
                \includegraphics[width=0.333\linewidth]{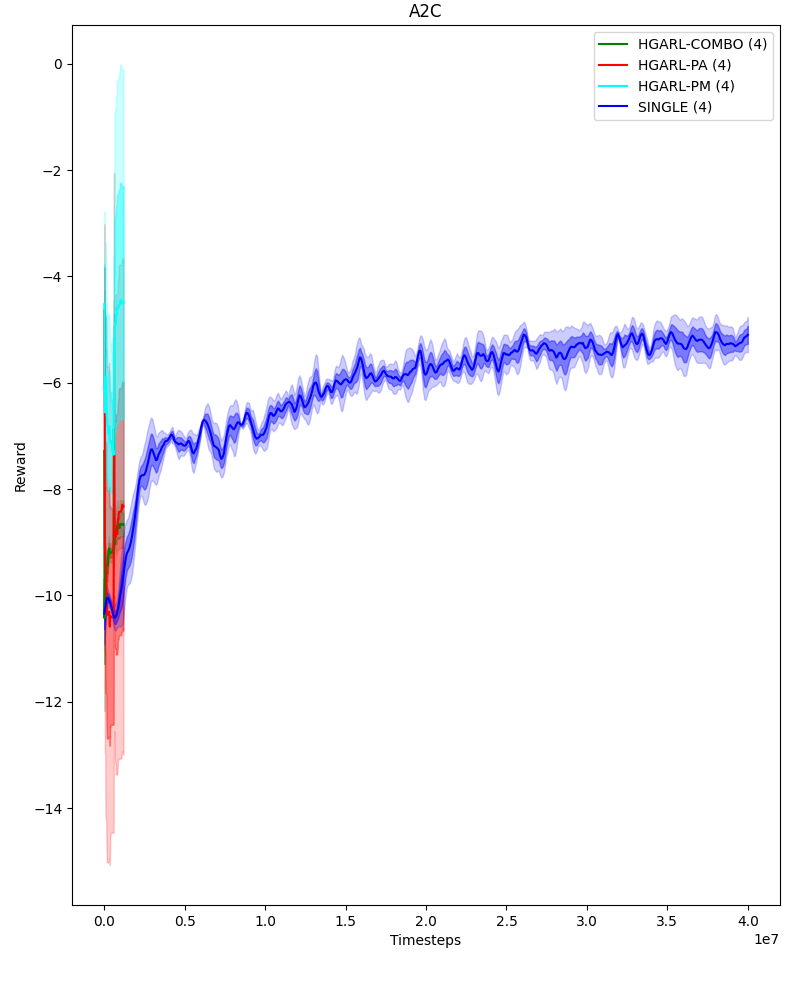}\hfill
                \includegraphics[width=0.333\linewidth]{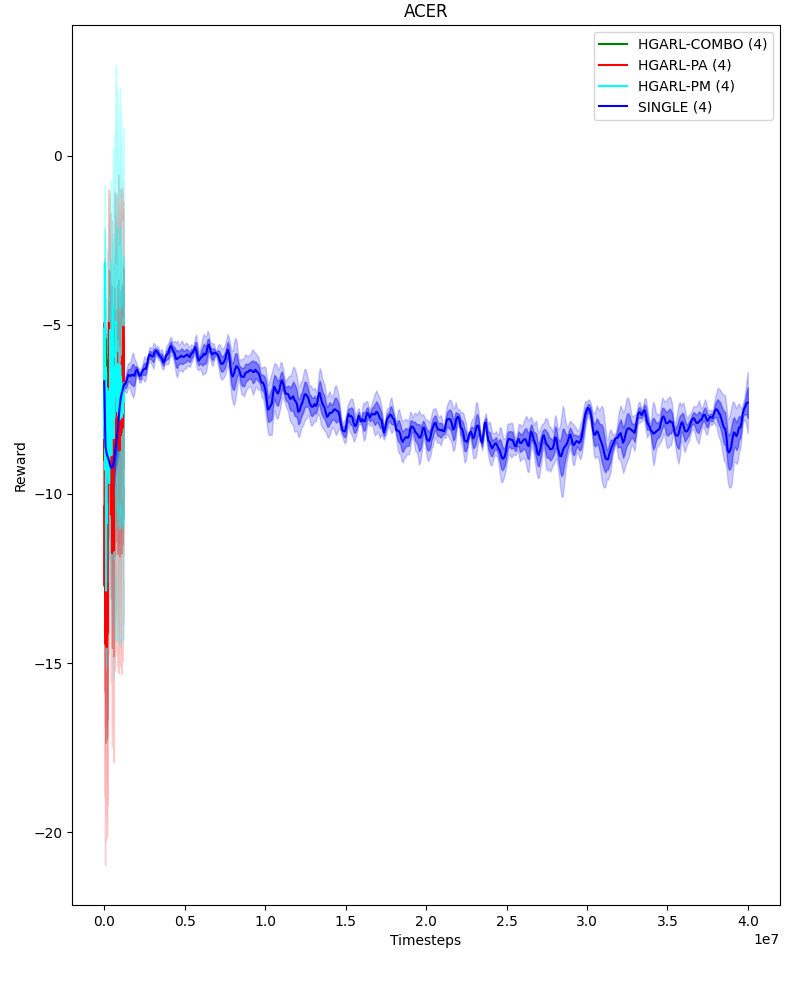}\hfill
                \includegraphics[width=0.333\linewidth]{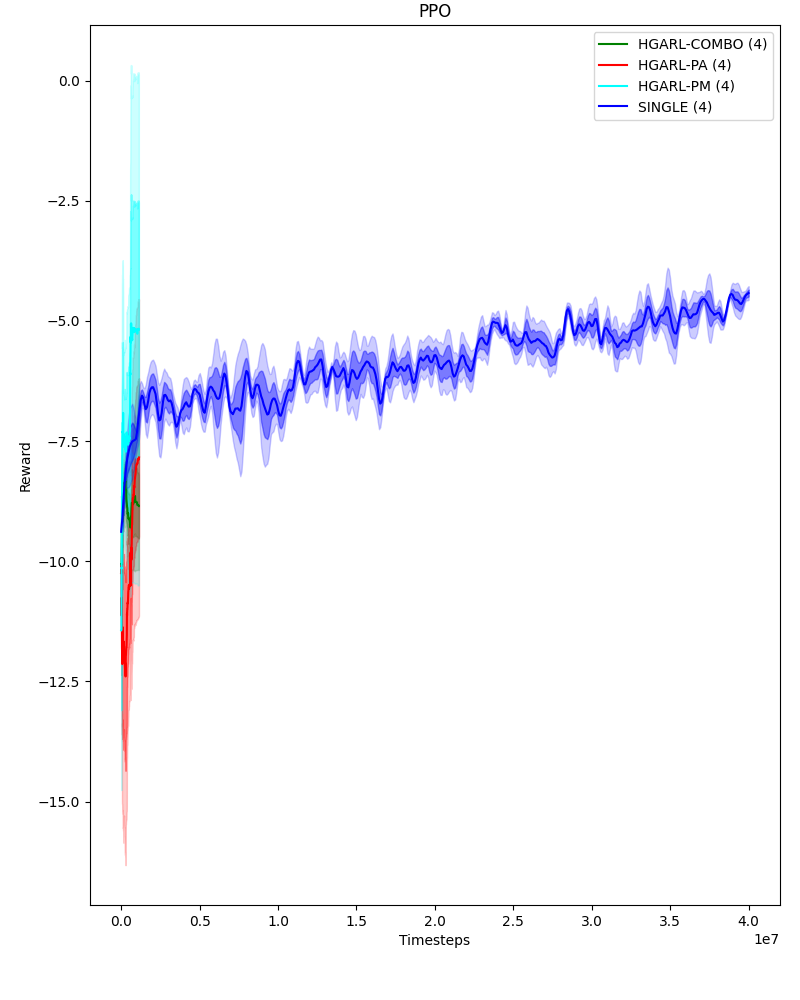}
                \caption{IceHockey}
        \end{subfigure}
         \begin{subfigure}[b]{0.49\linewidth}
                \includegraphics[width=0.333\linewidth]{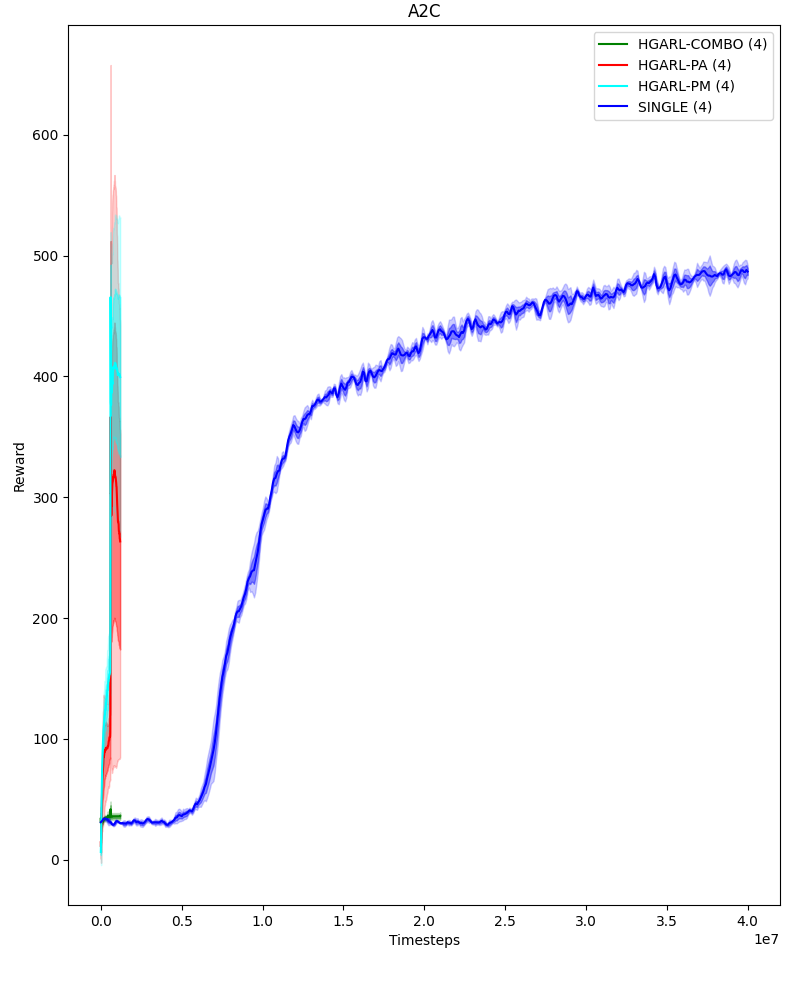}\hfill
                \includegraphics[width=0.333\linewidth]{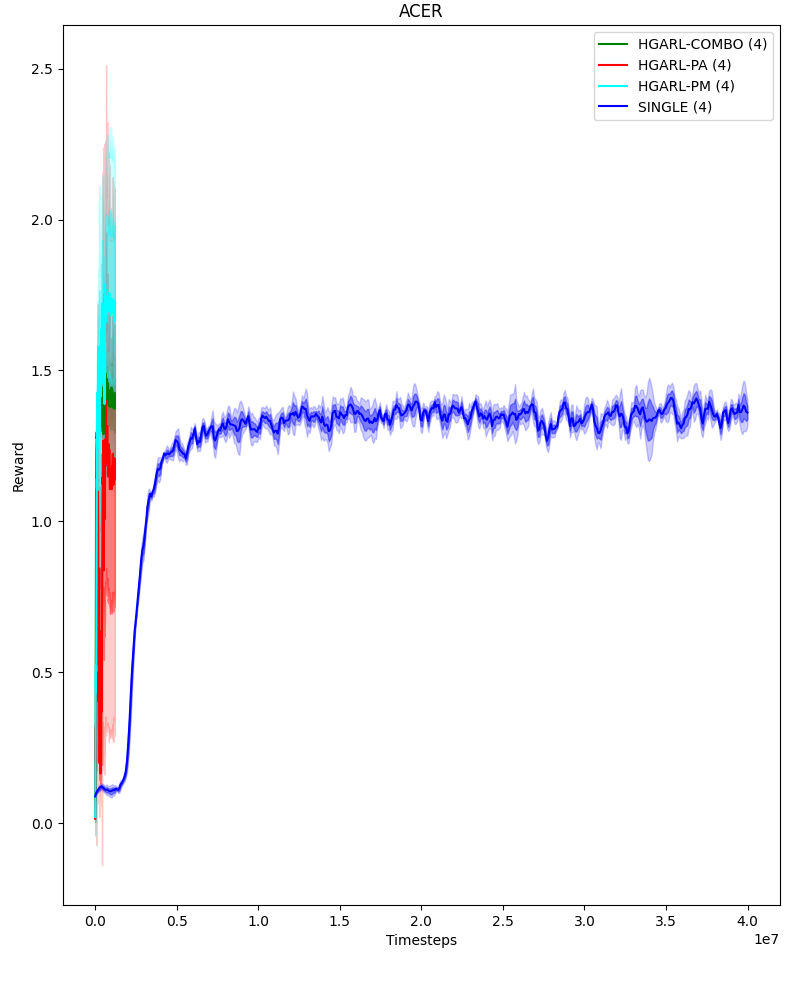}\hfill
                \includegraphics[width=0.333\linewidth]{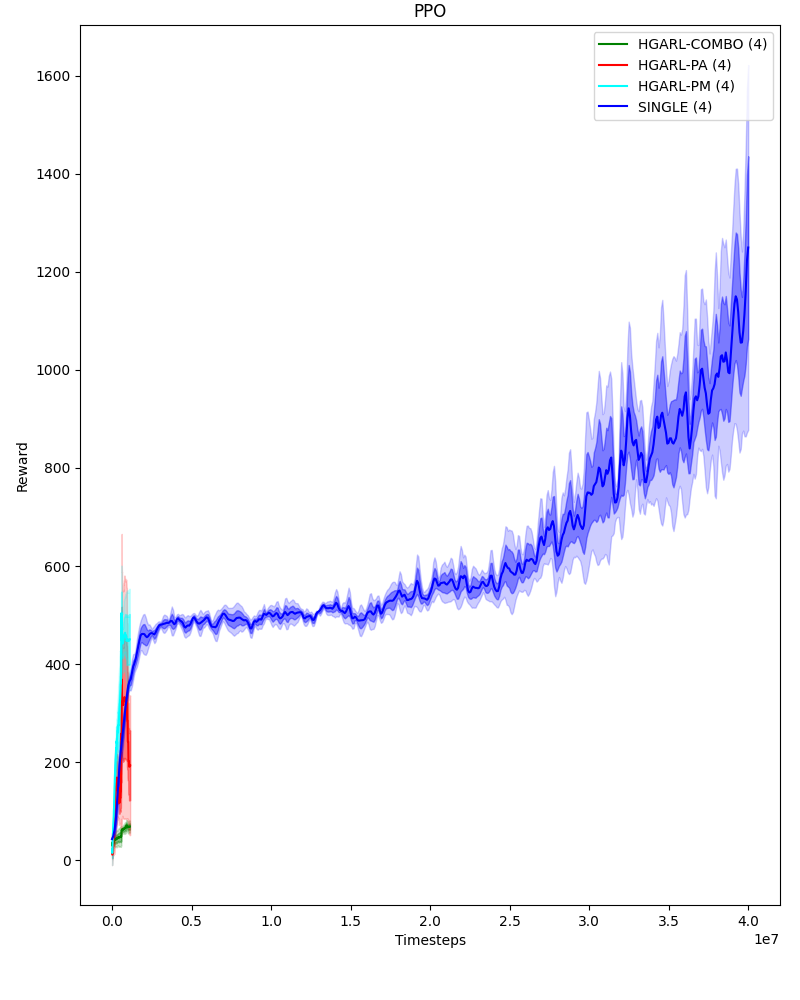}
                \caption{Jamesbond}
        \end{subfigure}
        \begin{subfigure}[b]{0.49\linewidth}
                \includegraphics[width=0.333\linewidth]{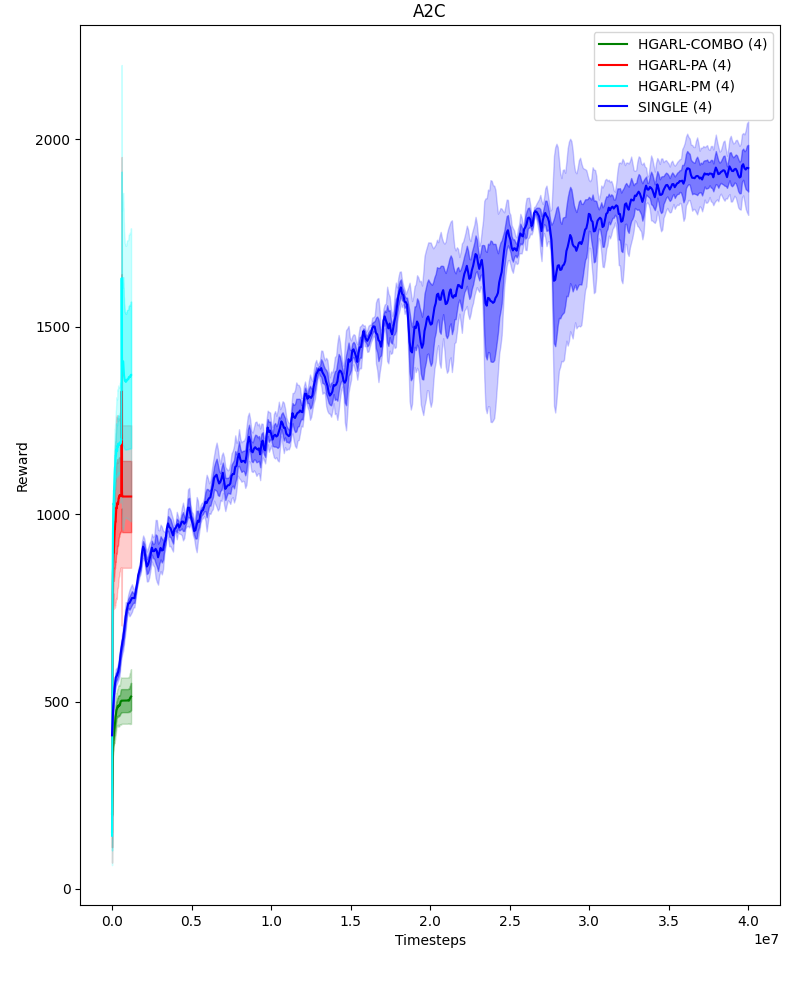}\hfill
                \includegraphics[width=0.333\linewidth]{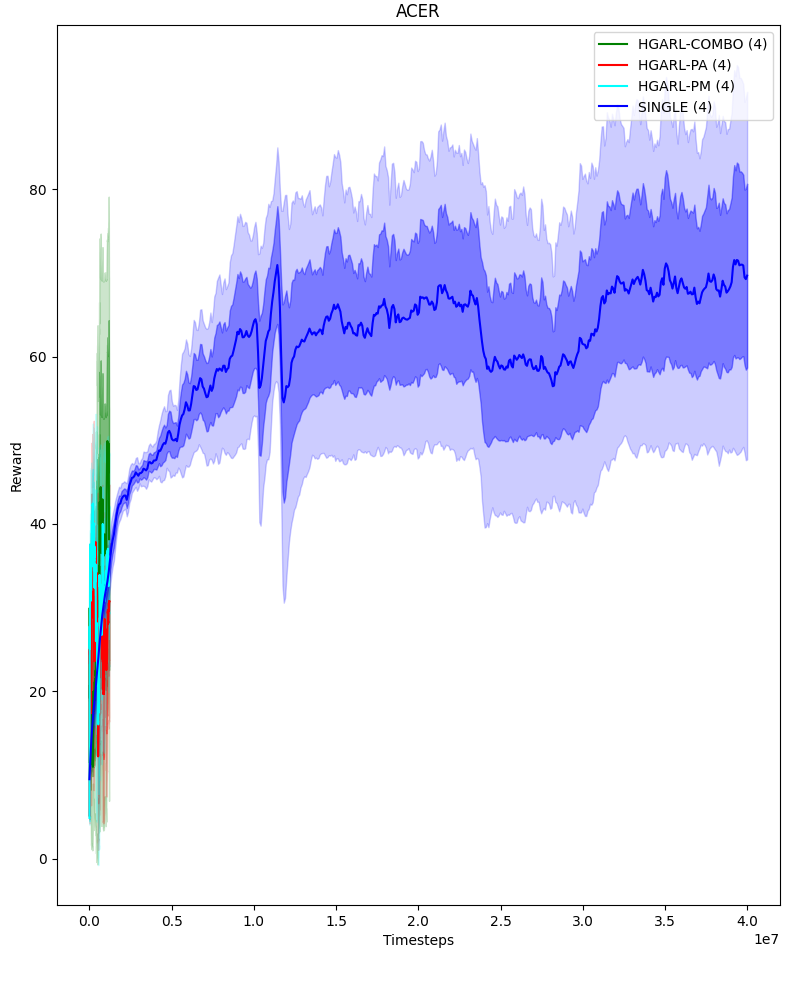}\hfill
                \includegraphics[width=0.333\linewidth]{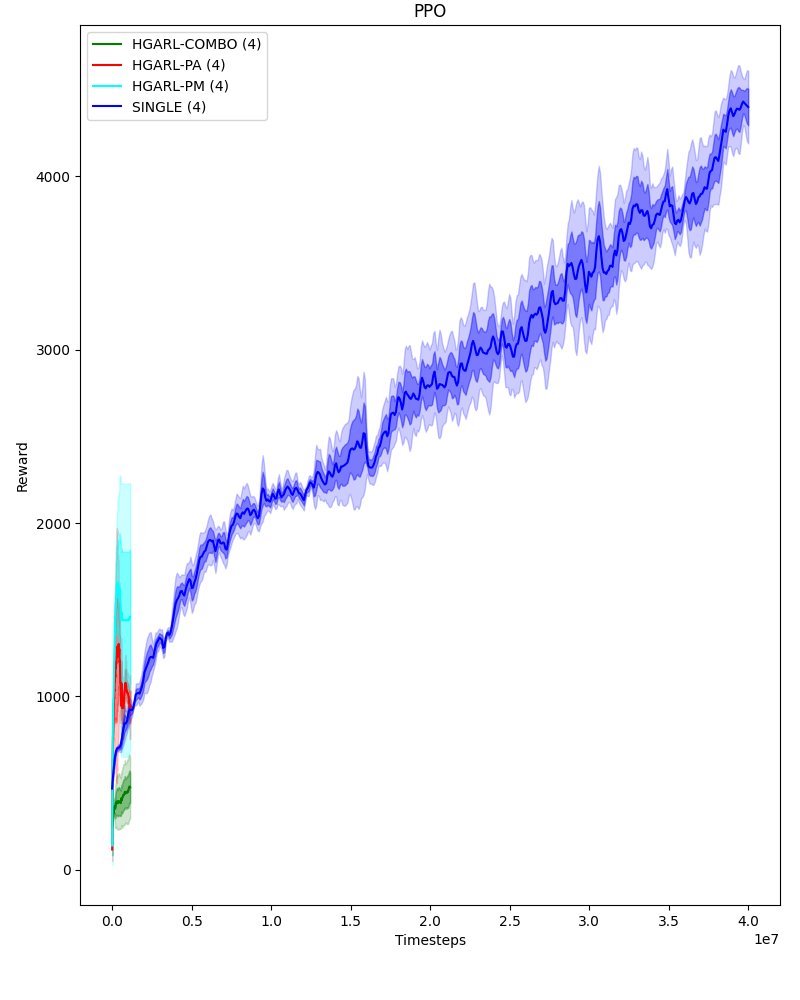}
                \caption{MsPacman}
        \end{subfigure}
        \begin{subfigure}[b]{0.49\linewidth}
                \includegraphics[width=0.333\linewidth]{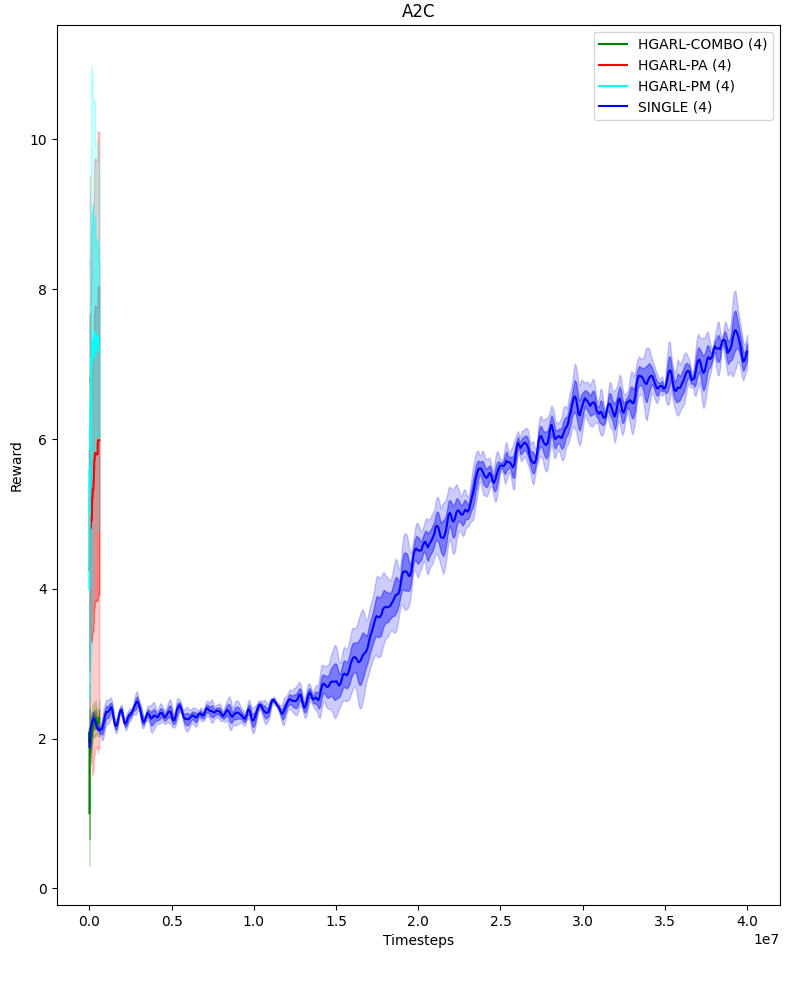}\hfill
                \includegraphics[width=0.333\linewidth]{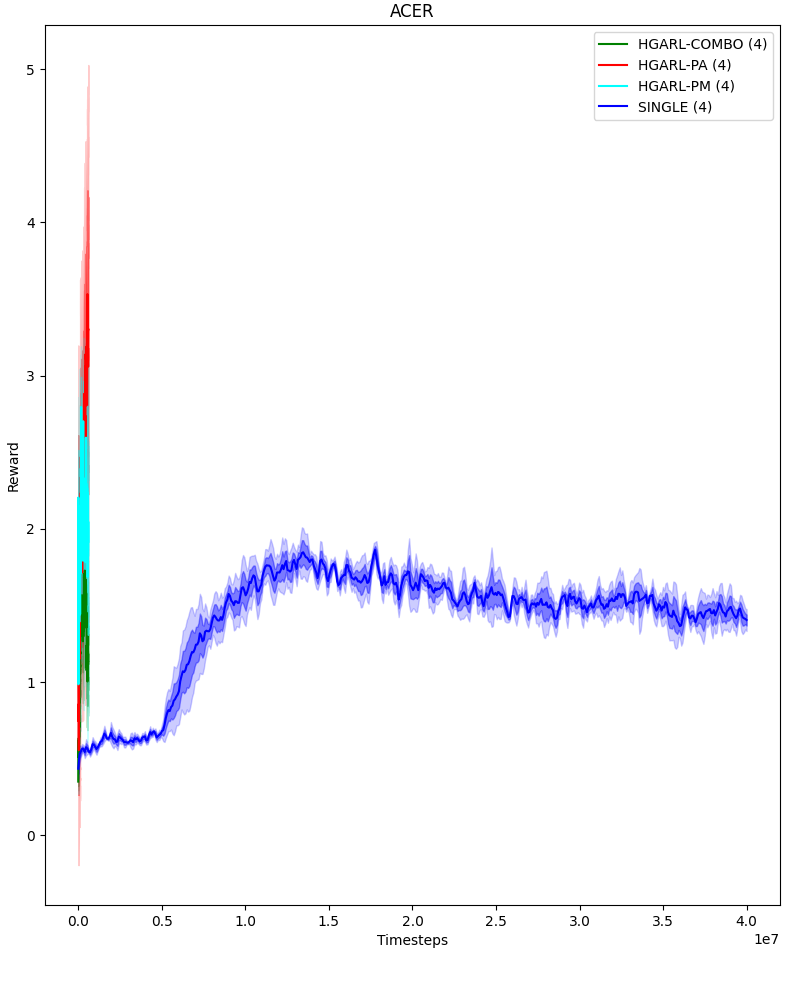}\hfill
                \includegraphics[width=0.333\linewidth]{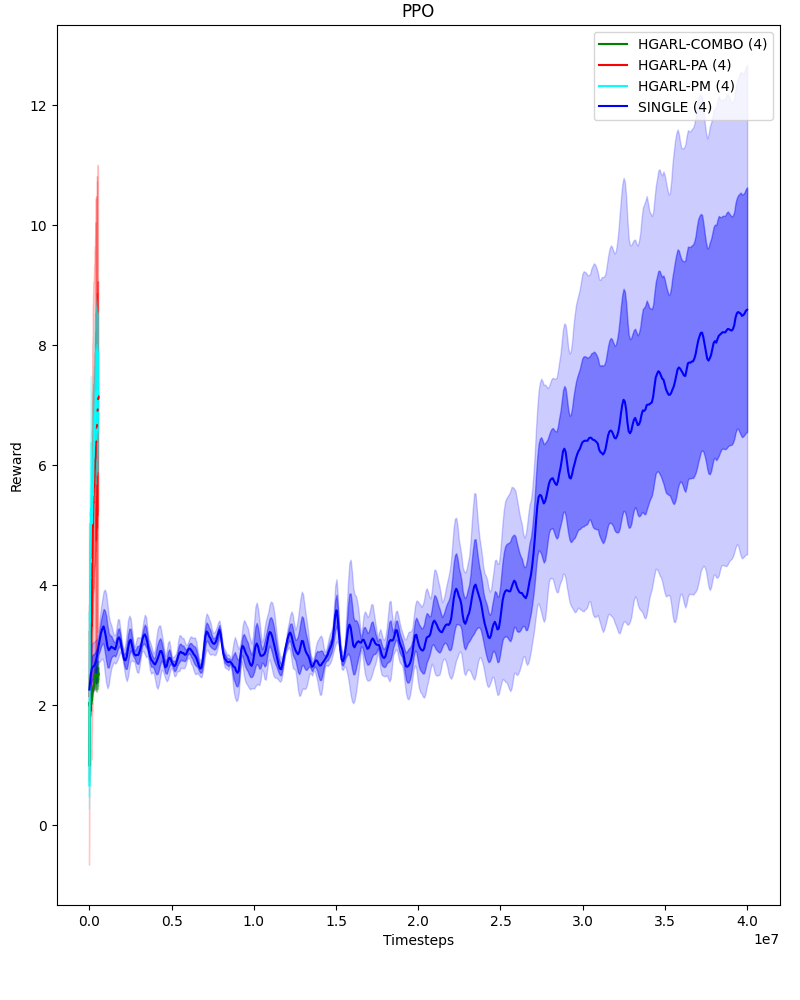}
                \caption{Robotank}
        \end{subfigure}
        \label{Atari3}
\end{figure*}

\begin{figure*}
       \ContinuedFloat
        \begin{subfigure}[b]{0.49\linewidth}
                \includegraphics[width=0.333\linewidth]{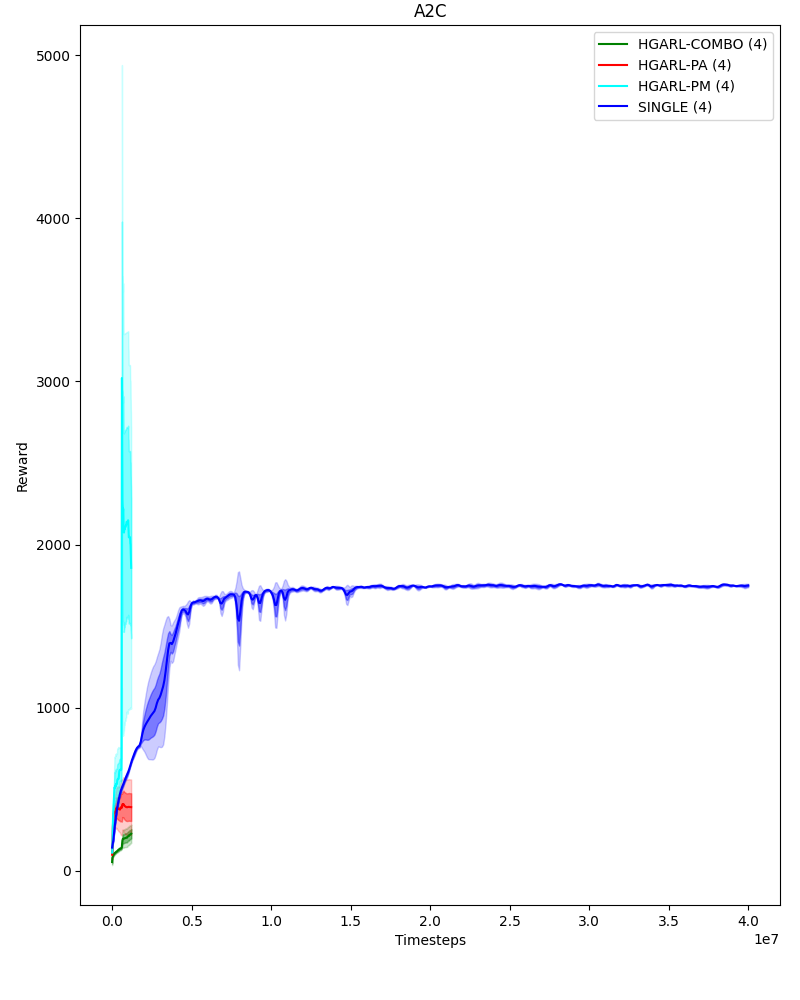}\hfill
                \includegraphics[width=0.333\linewidth]{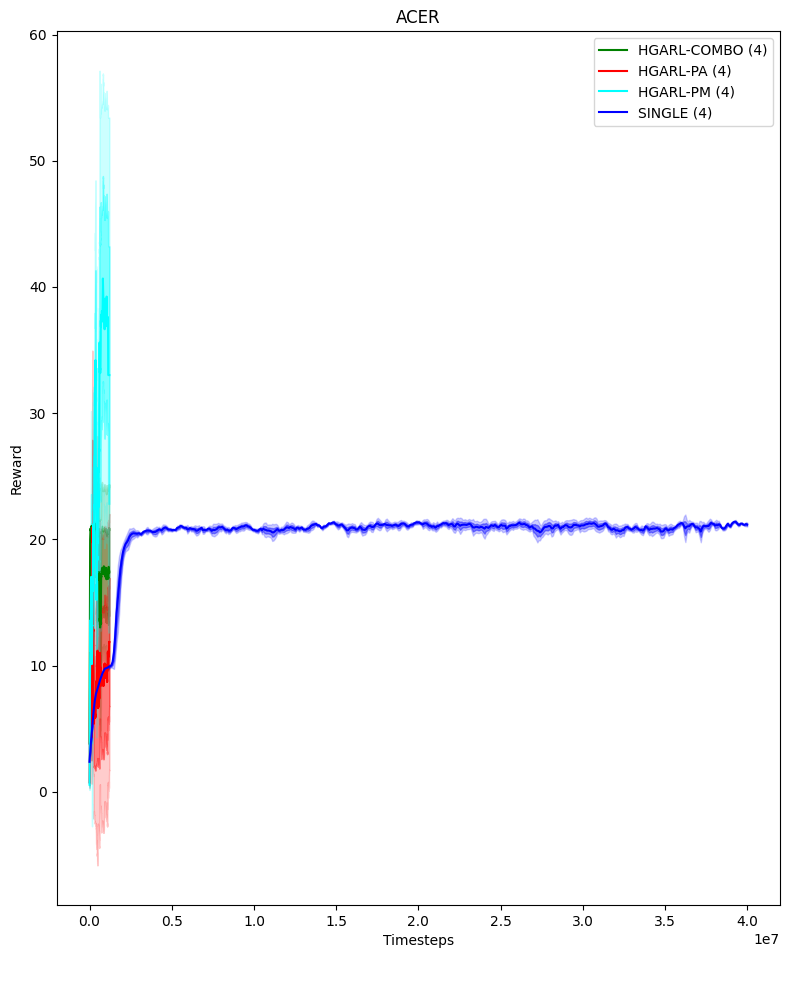}\hfill
                \includegraphics[width=0.333\linewidth]{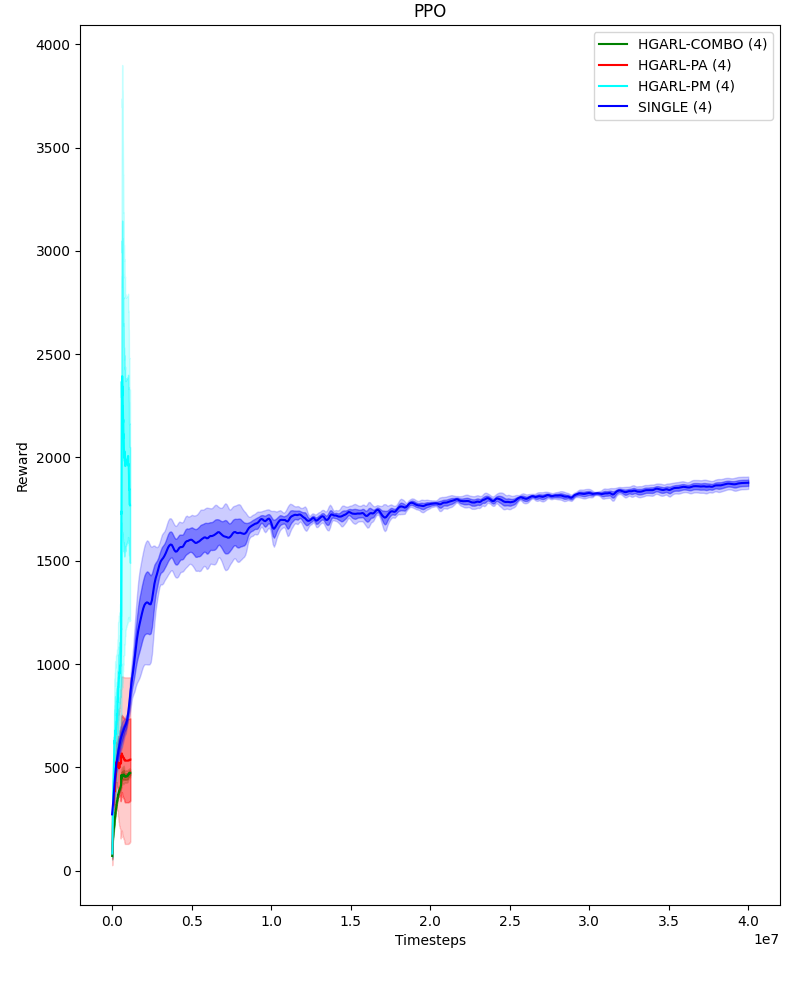}
                \caption{Seaquest}
        \end{subfigure}
        \begin{subfigure}[b]{0.49\linewidth}
                \includegraphics[width=0.333\linewidth]{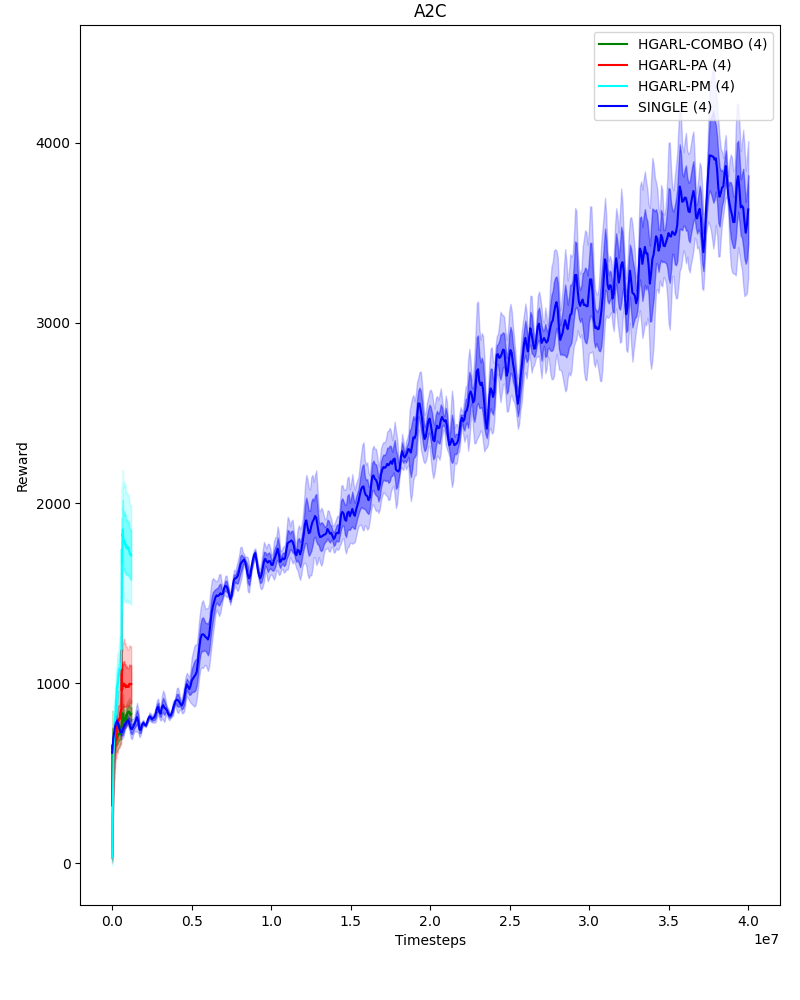}\hfill
                \includegraphics[width=0.333\linewidth]{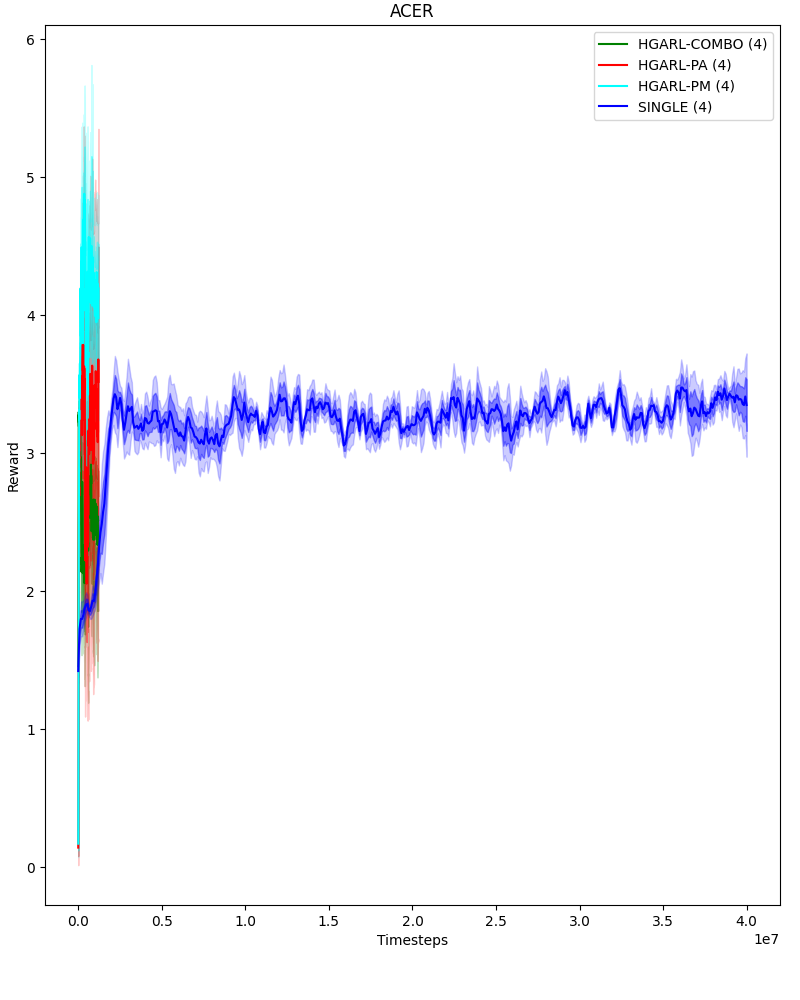}\hfill
                \includegraphics[width=0.333\linewidth]{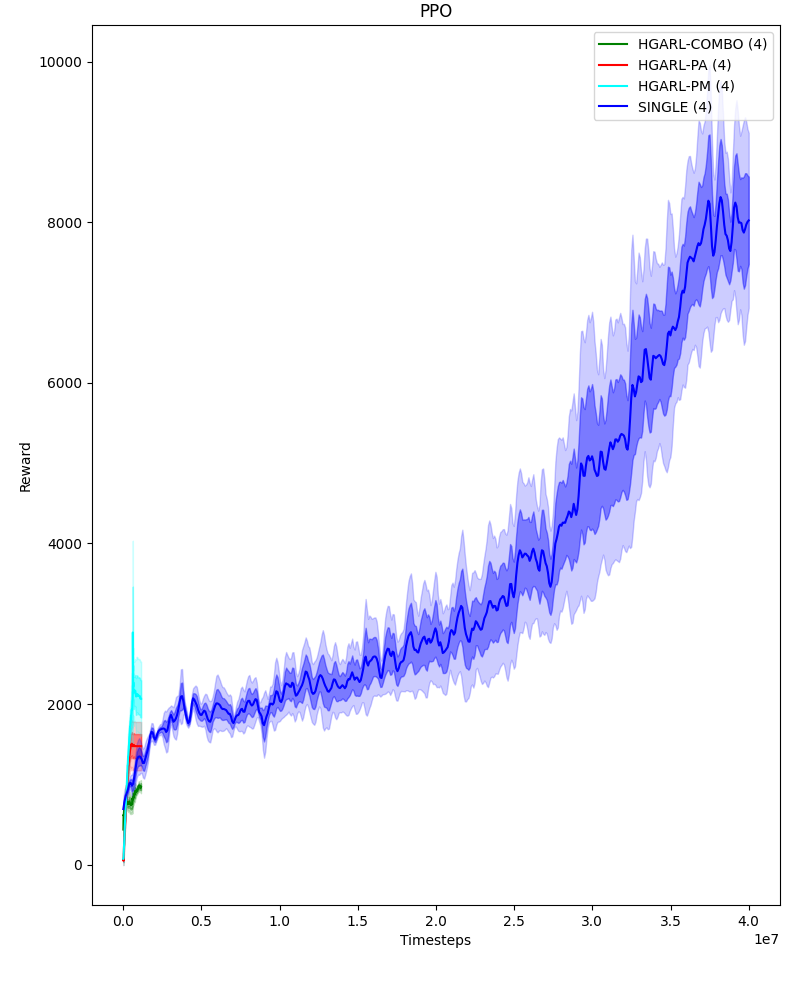}
                \caption{WizardOfWor}
        \end{subfigure}
        \begin{subfigure}[b]{0.49\linewidth}
                \includegraphics[width=0.333\linewidth]{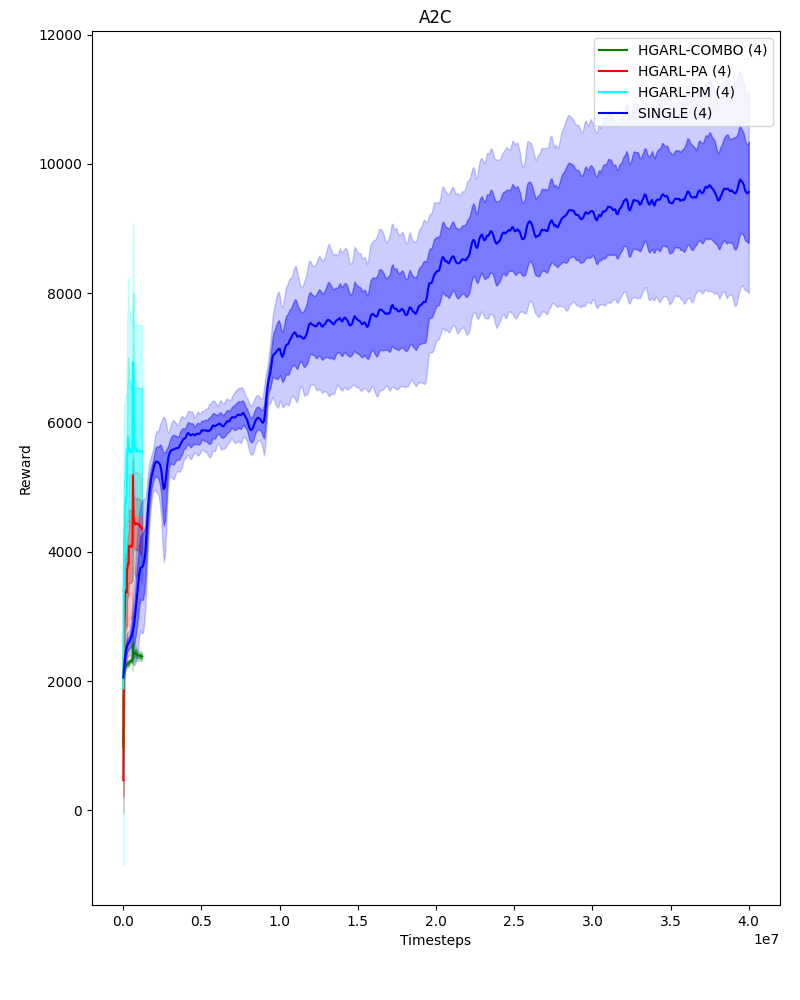}\hfill
                \includegraphics[width=0.333\linewidth]{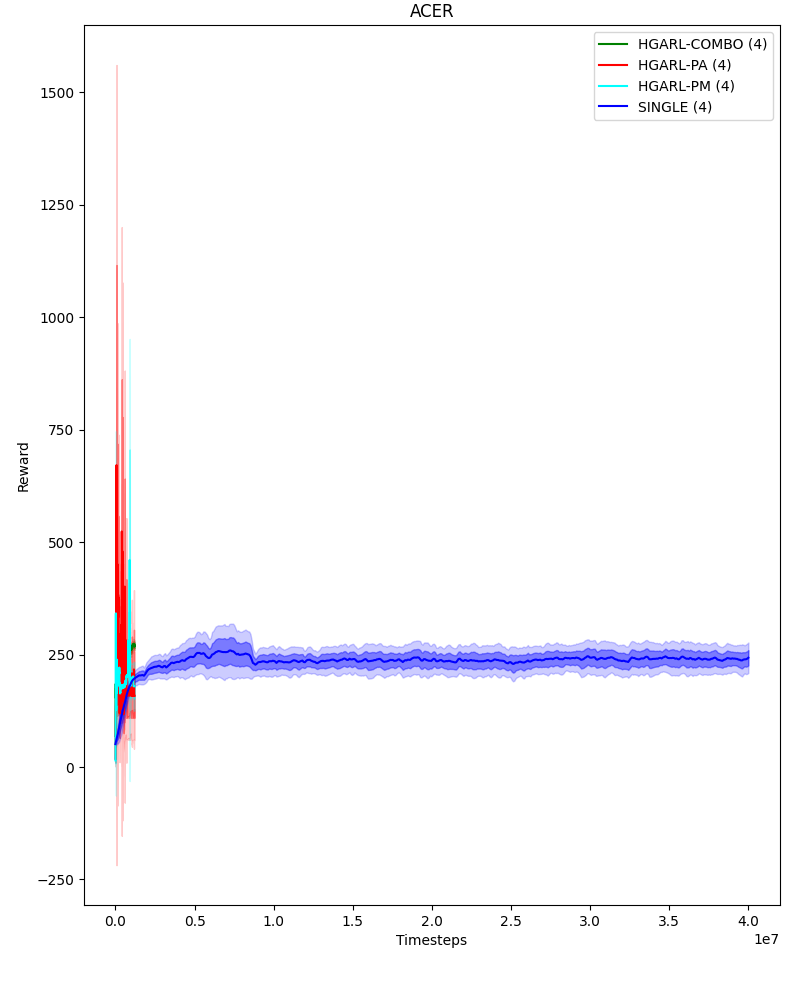}\hfill
                \includegraphics[width=0.333\linewidth]{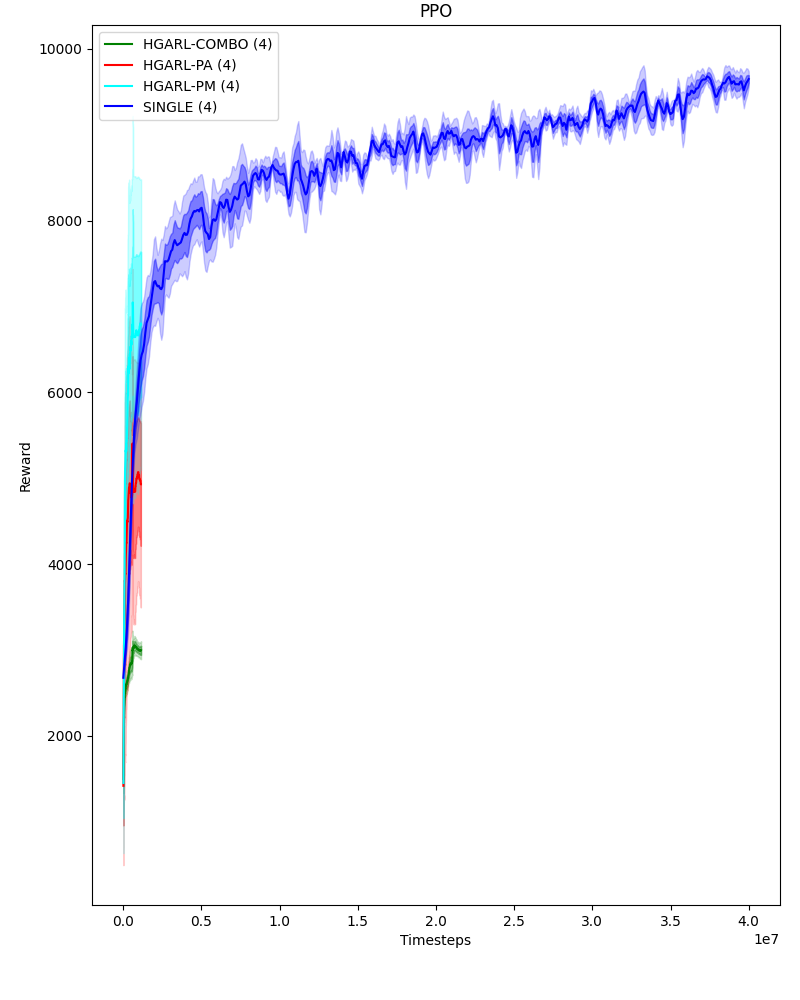}
                \caption{Krull}
        \end{subfigure}
        \begin{subfigure}[b]{0.49\linewidth}
                \includegraphics[width=0.333\linewidth]{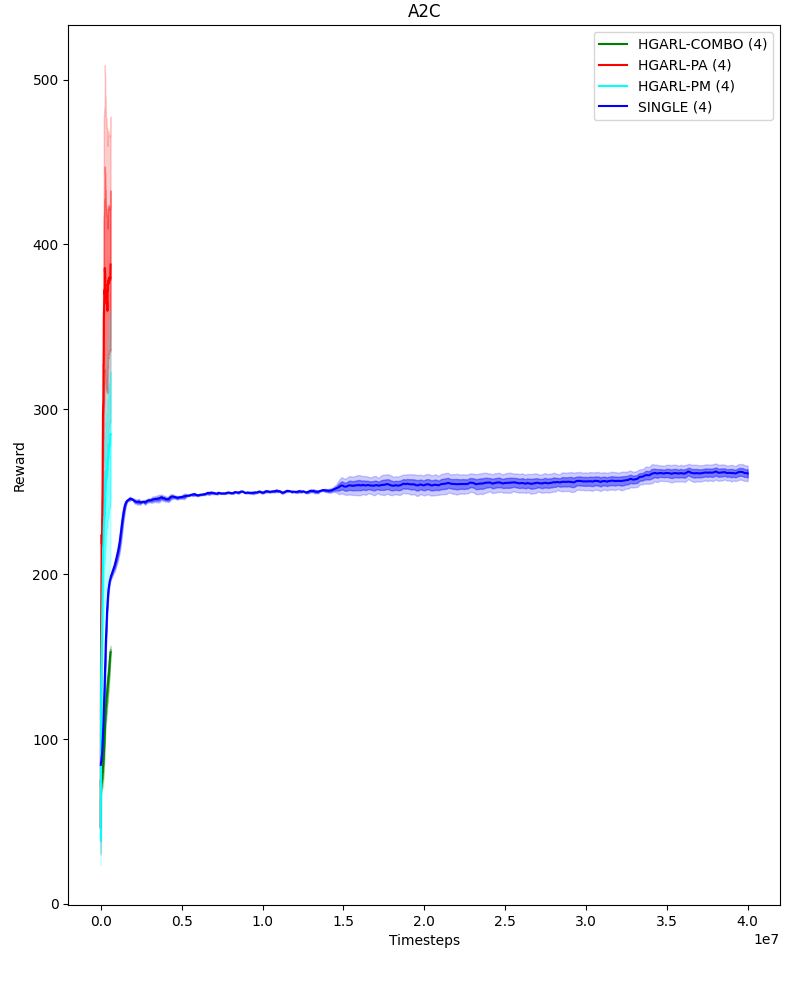}\hfill
                \includegraphics[width=0.333\linewidth]{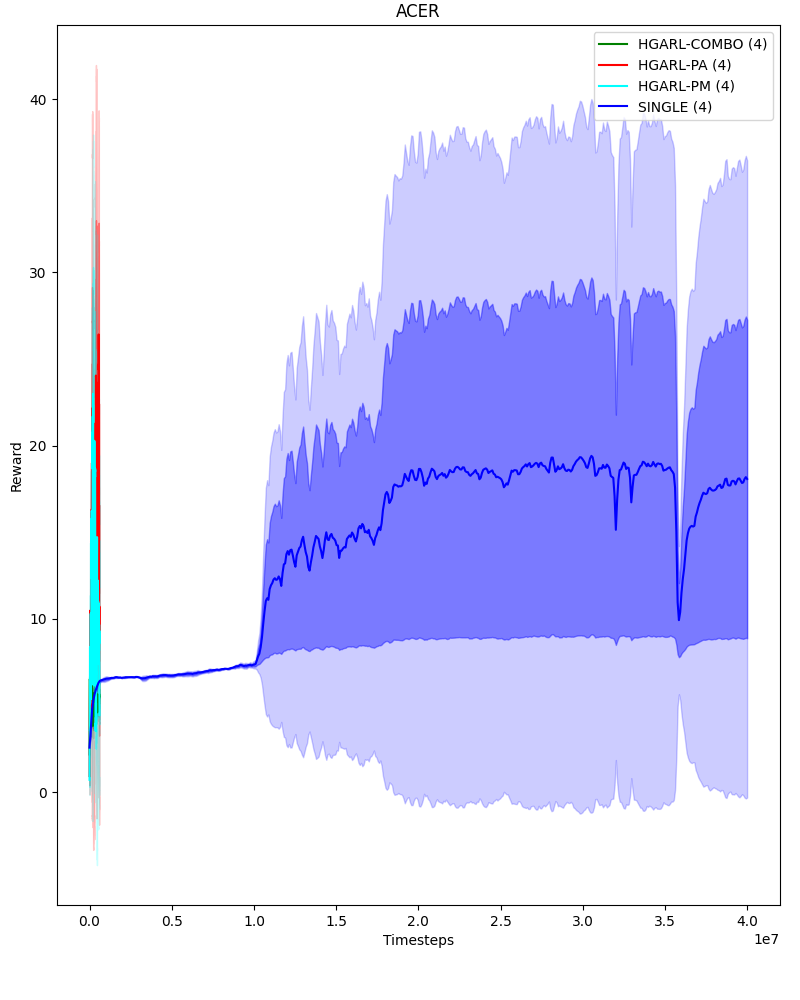}\hfill
                \includegraphics[width=0.333\linewidth]{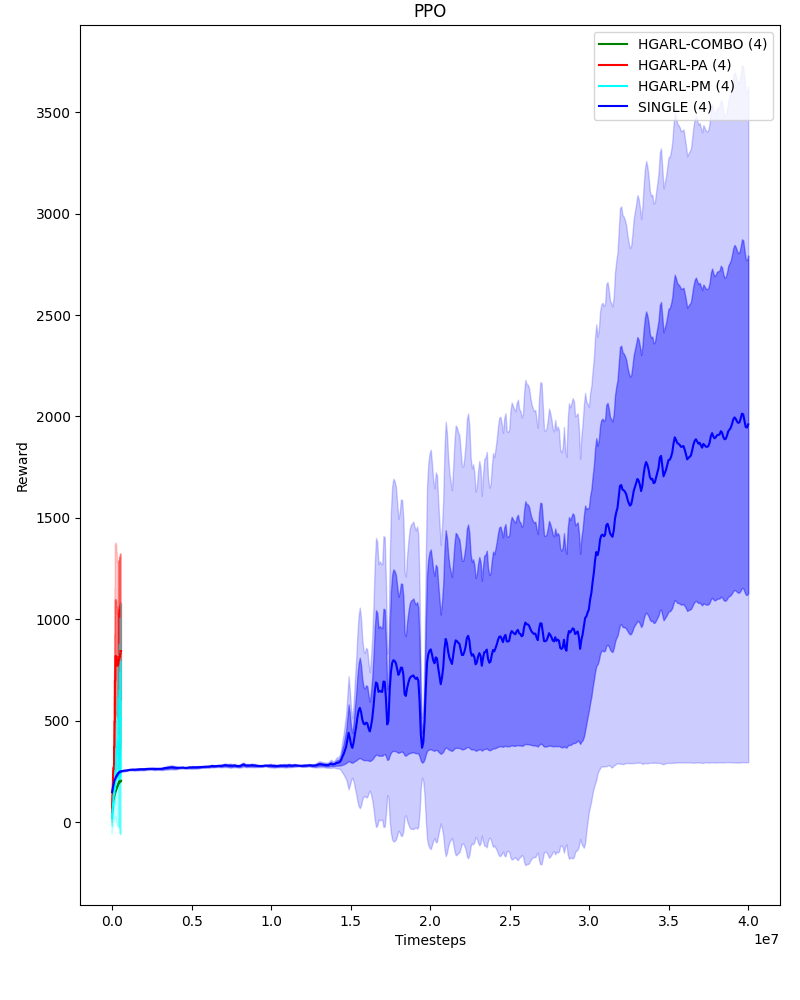}
                \caption{Frostbite}
        \end{subfigure}
        \begin{subfigure}[b]{0.49\linewidth}
                \includegraphics[width=0.333\linewidth]{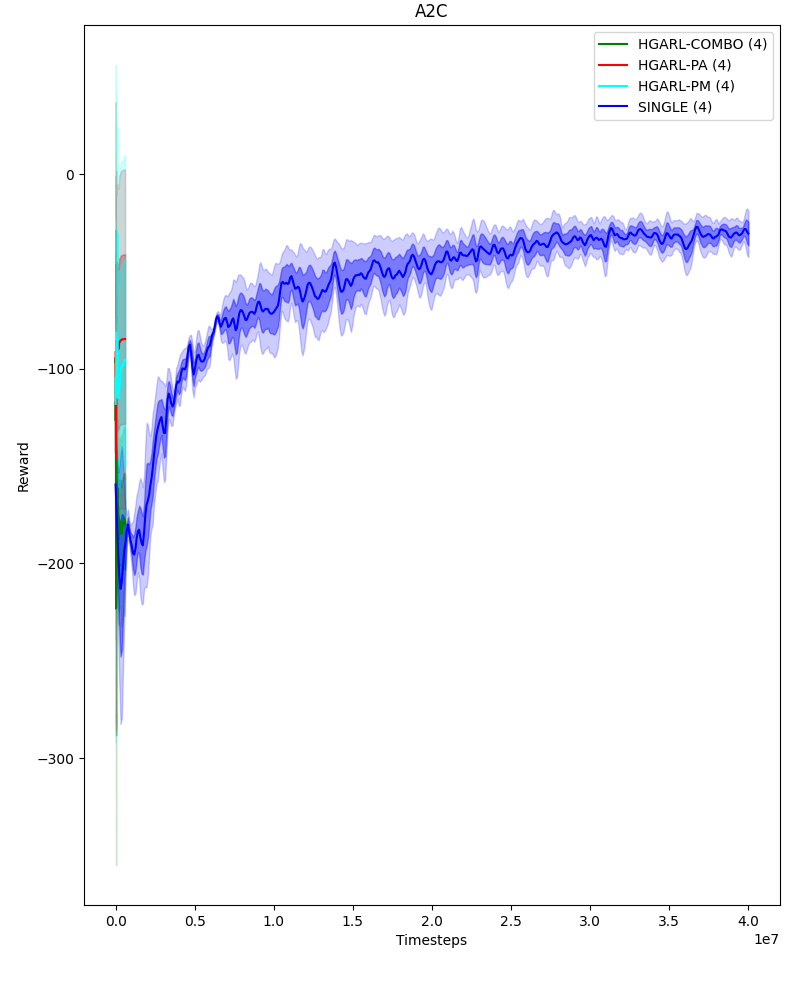}\hfill
                \includegraphics[width=0.333\linewidth]{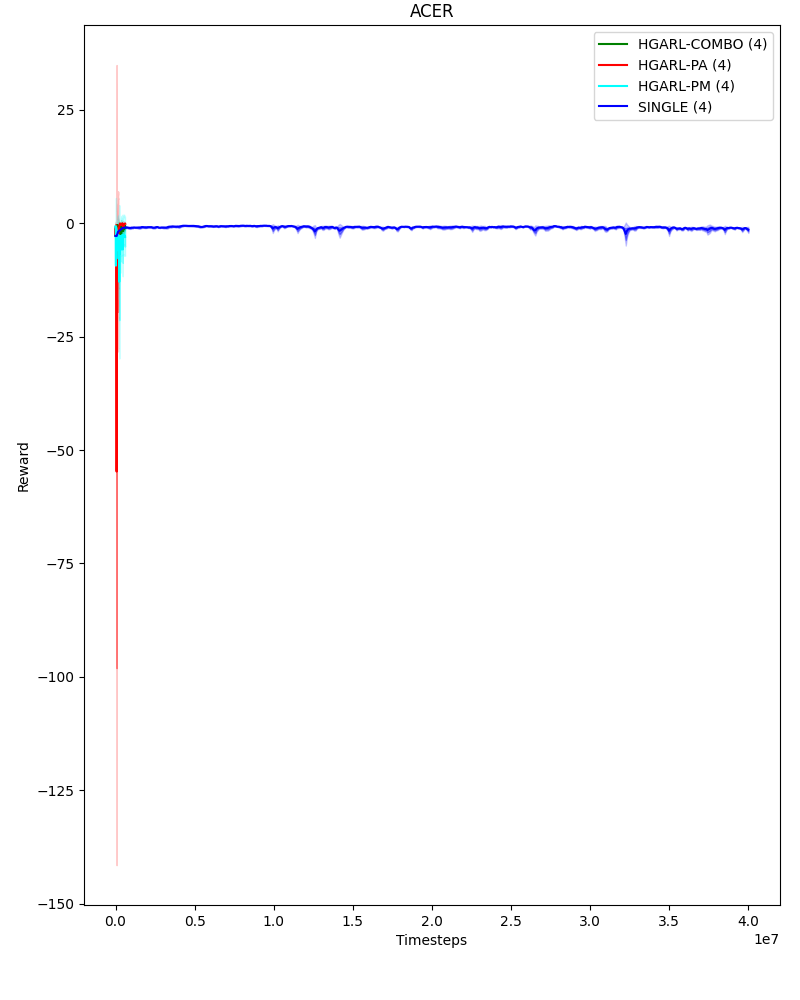}\hfill
                \includegraphics[width=0.333\linewidth]{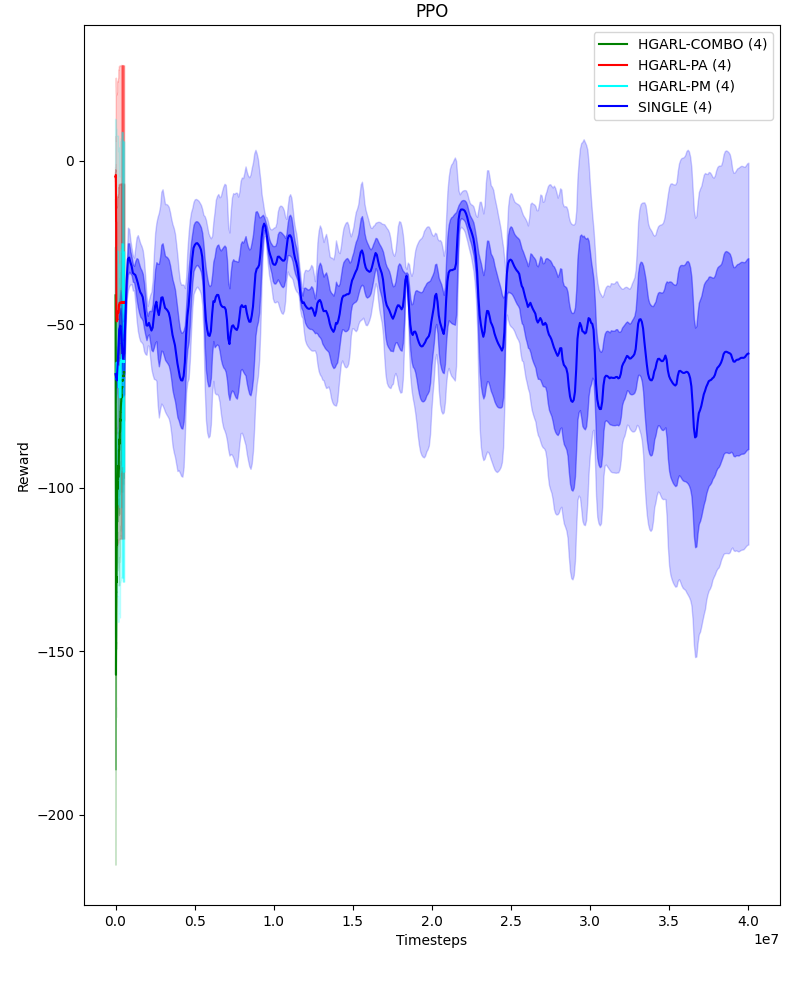}
                \caption{Pitfall}
        \end{subfigure}
        \begin{subfigure}[b]{0.49\linewidth}
                \includegraphics[width=0.333\linewidth]{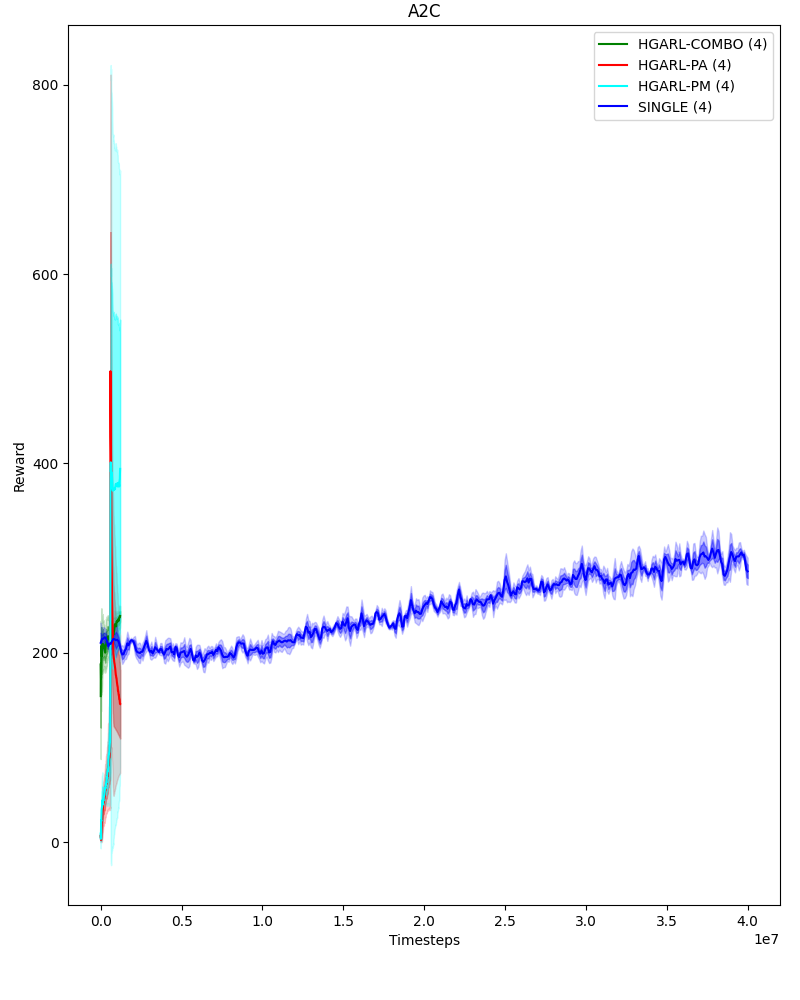}\hfill
                \includegraphics[width=0.333\linewidth]{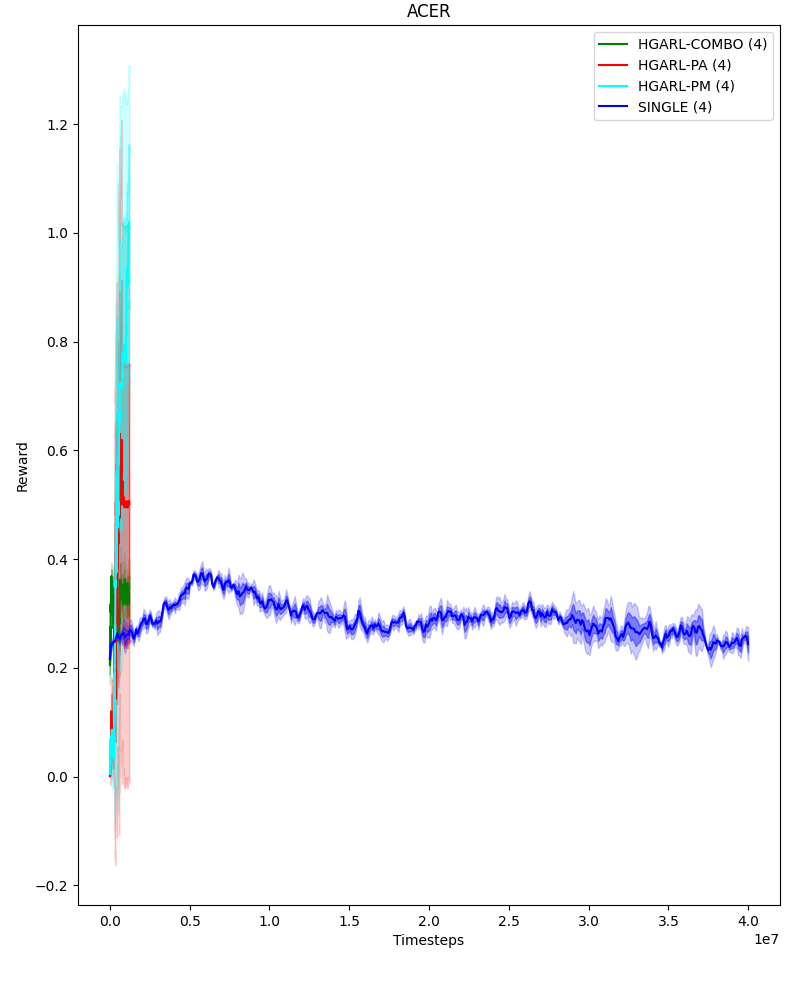}\hfill
                \includegraphics[width=0.333\linewidth]{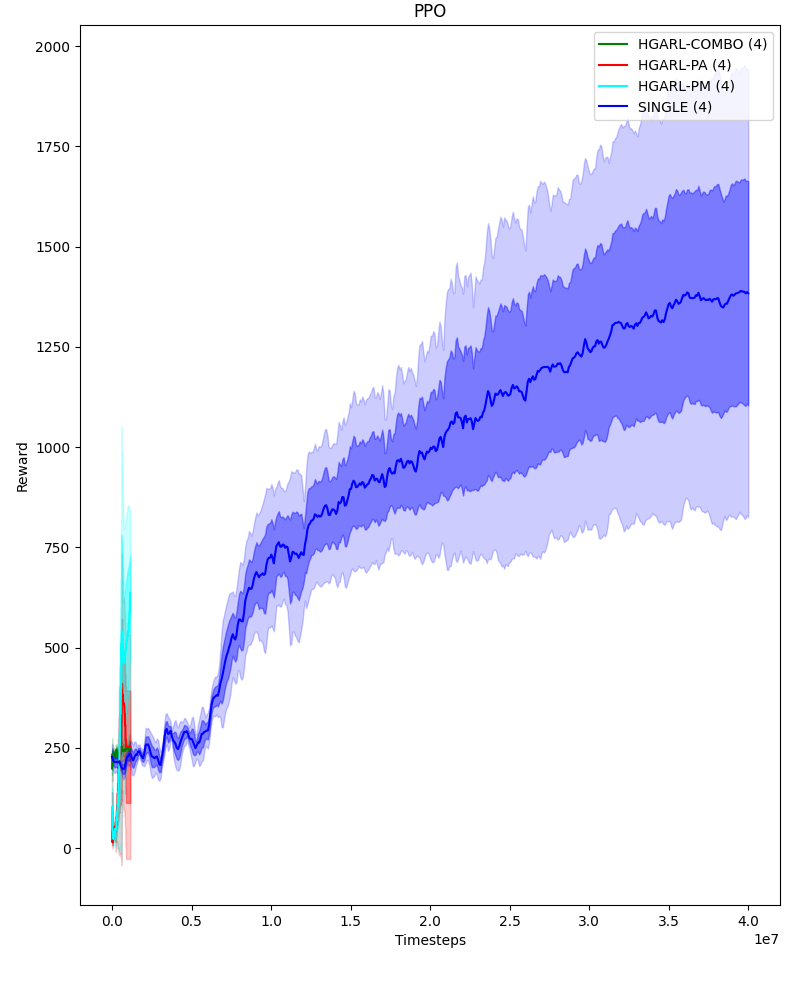}
                \caption{Gravitar}
        \end{subfigure}
        \caption{Atari 2600 Games: Part 3. The PA or PM rule shows great performance.}
        \label{Atari3}
\end{figure*}

\begin{figure*}
        \begin{subfigure}[b]{0.49\linewidth}
                \includegraphics[width=0.333\linewidth]{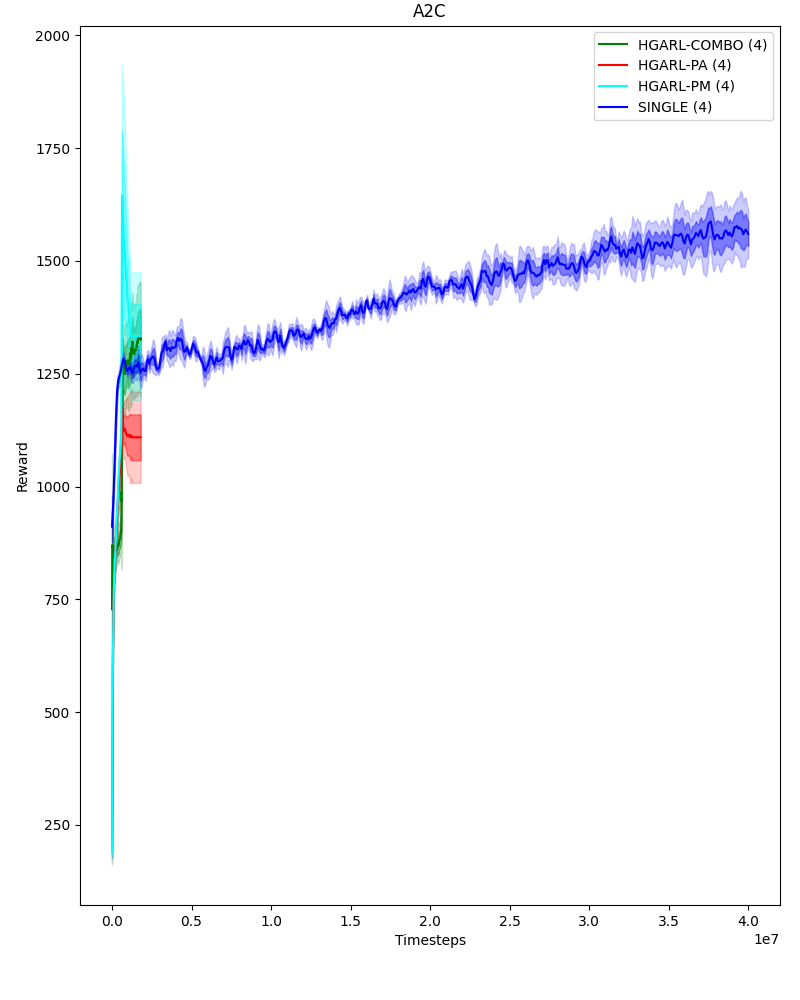}\hfill
                \includegraphics[width=0.333\linewidth]{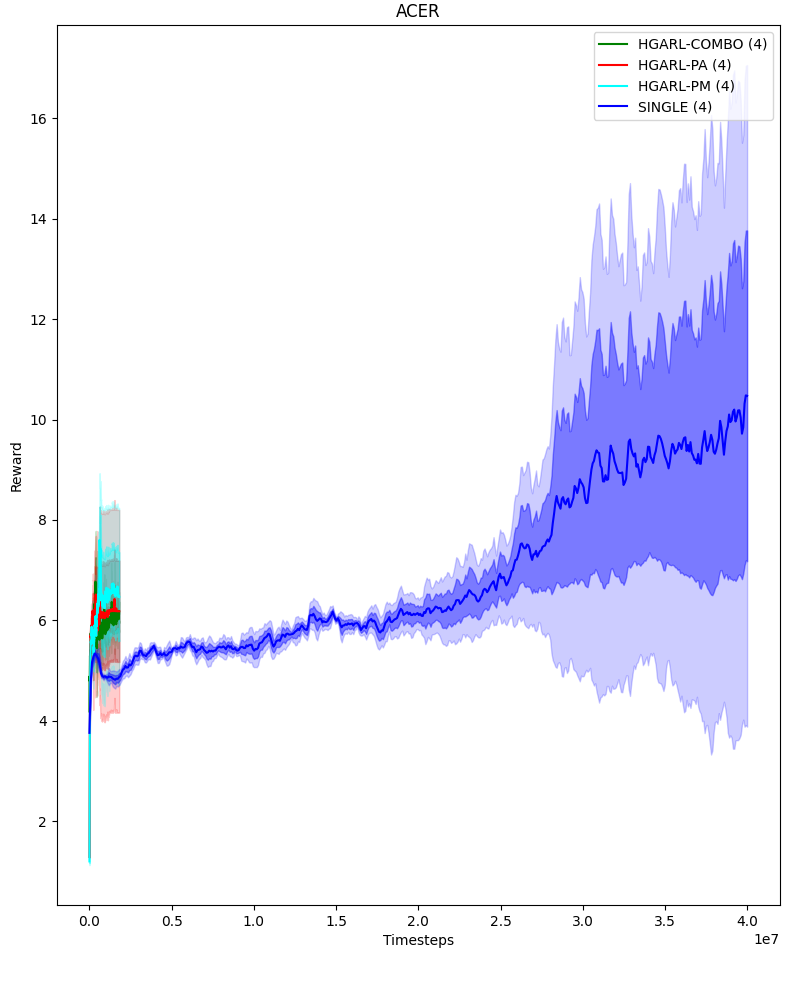}\hfill
                \includegraphics[width=0.333\linewidth]{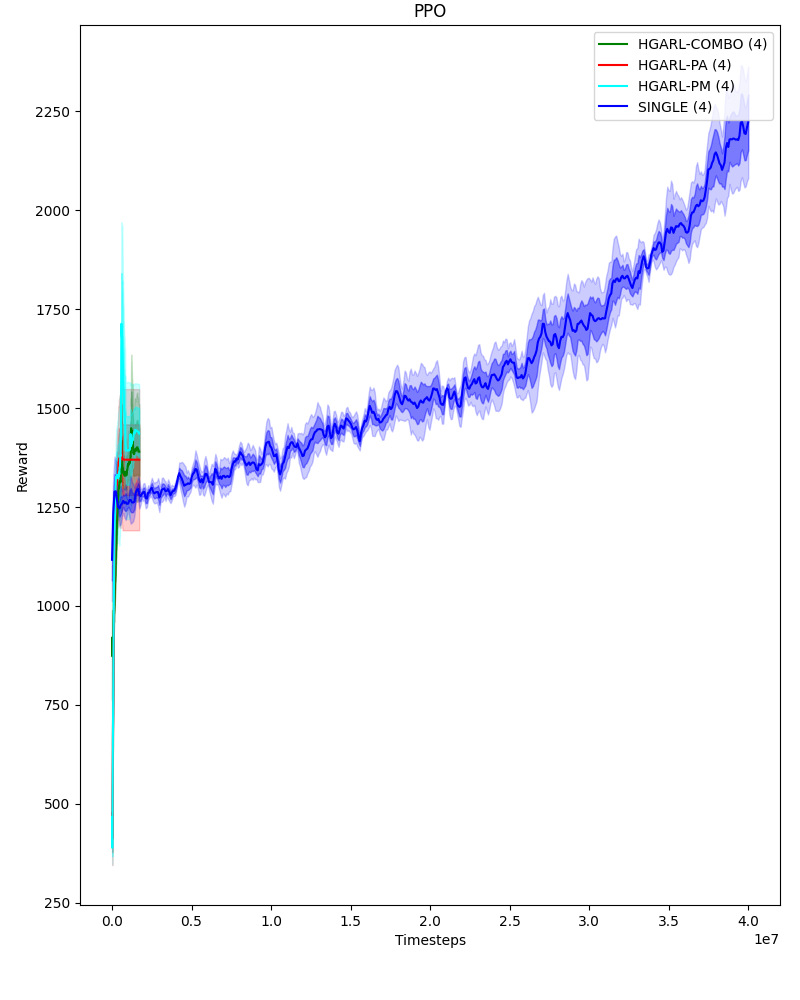}
                \caption{Asteroids}
        \end{subfigure}
        \begin{subfigure}[b]{0.49\linewidth}
                \includegraphics[width=0.333\linewidth]{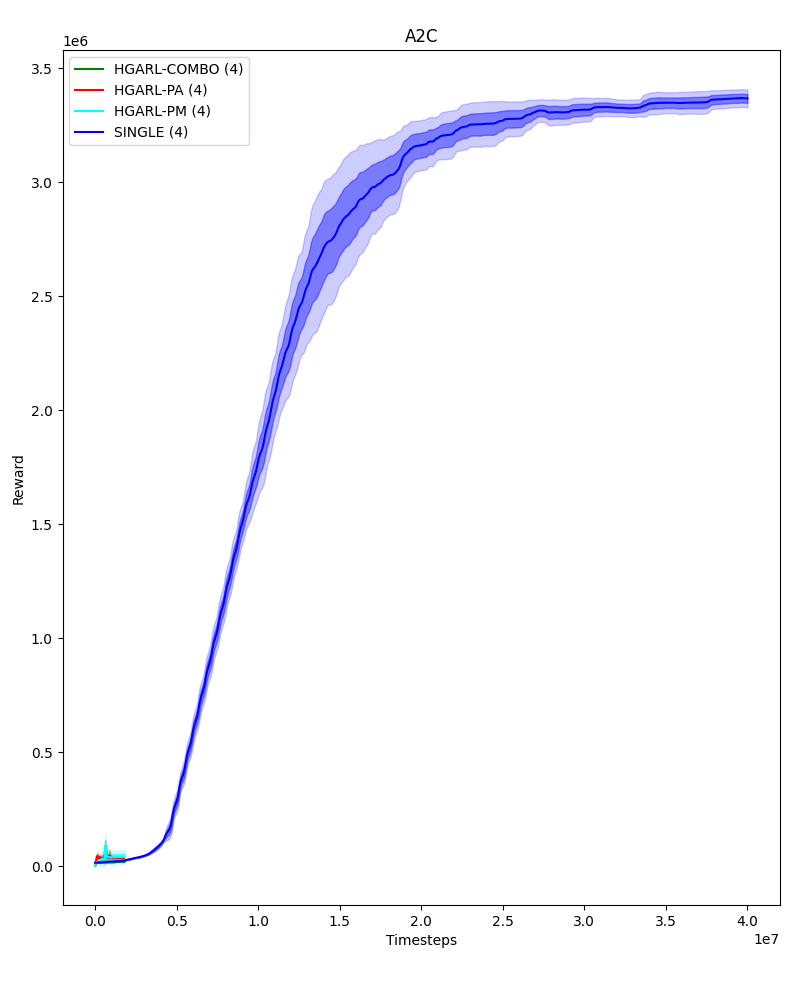}\hfill
                \includegraphics[width=0.333\linewidth]{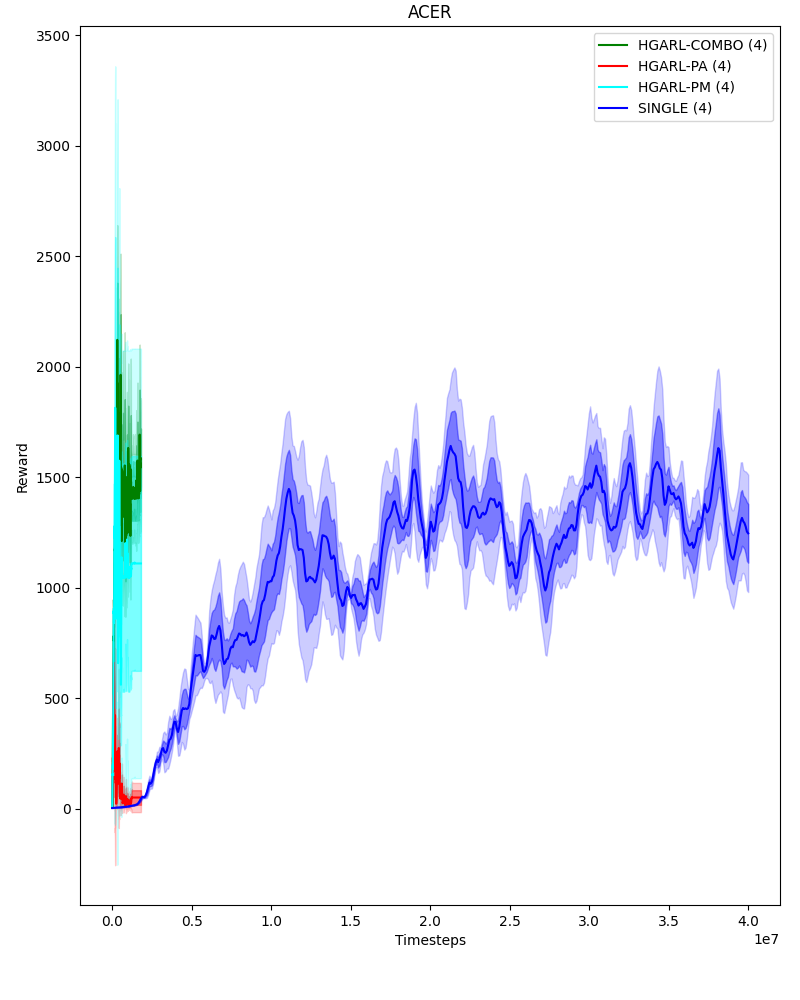}\hfill
                \includegraphics[width=0.333\linewidth]{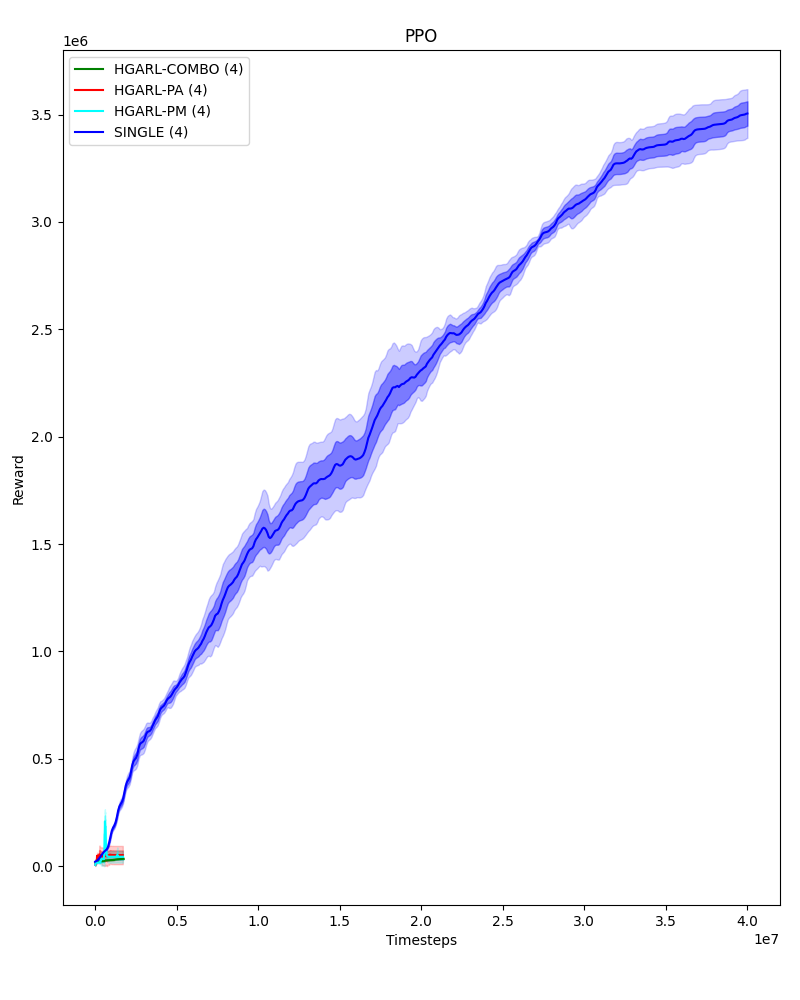}
                \caption{Atlantis}
        \end{subfigure}
        \begin{subfigure}[b]{0.49\linewidth}
                \includegraphics[width=0.333\linewidth]{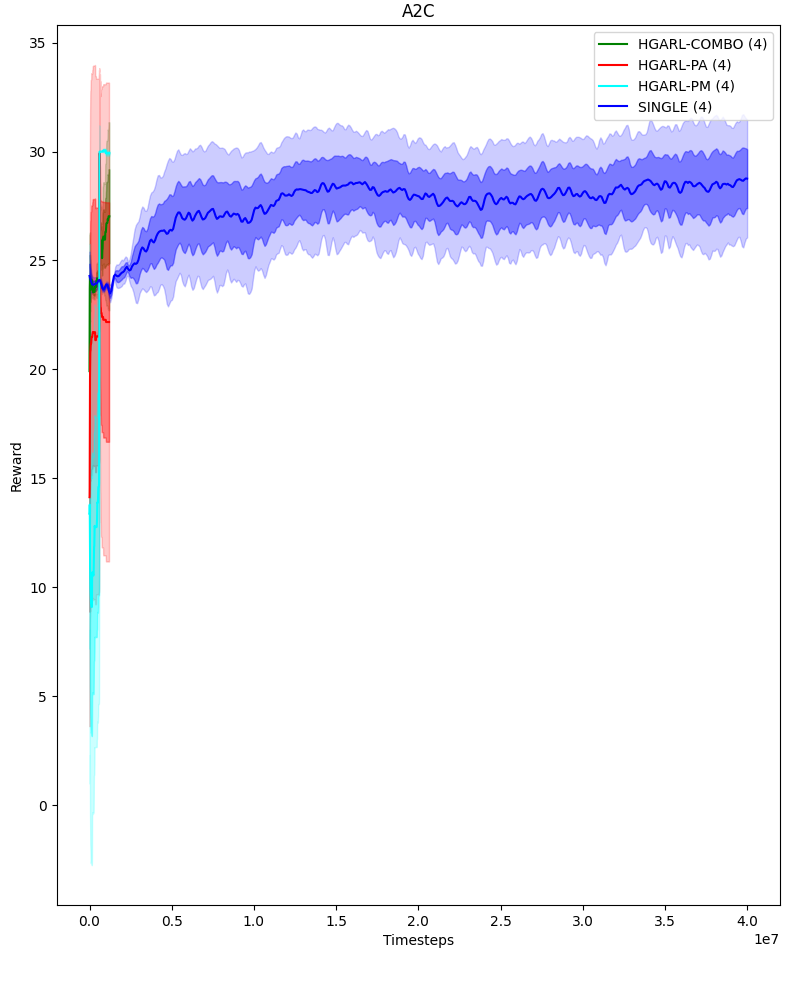}\hfill
                \includegraphics[width=0.333\linewidth]{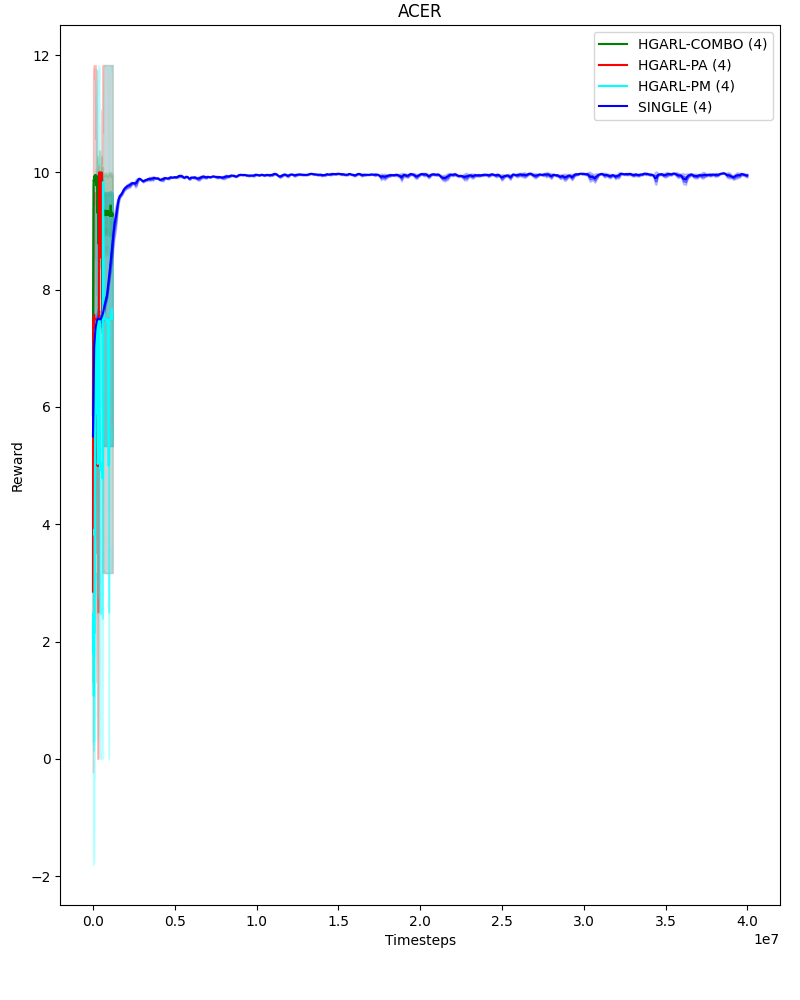}\hfill
                \includegraphics[width=0.333\linewidth]{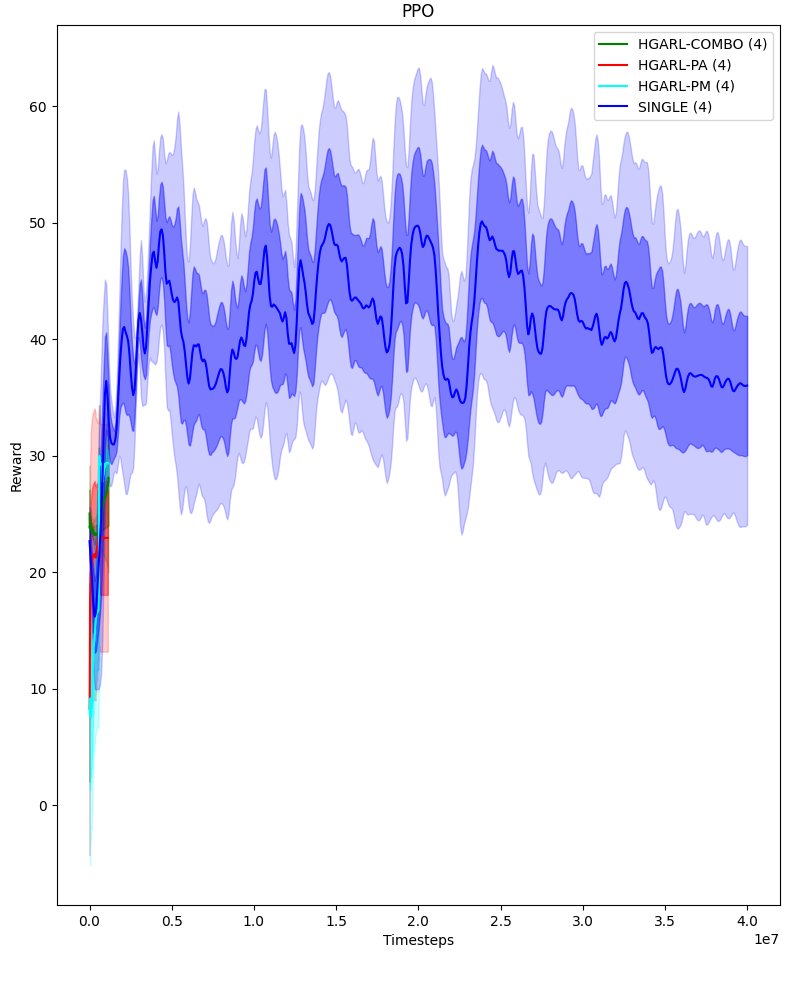}
                \caption{Bowling}
        \end{subfigure}
        \begin{subfigure}[b]{0.49\linewidth}
                \includegraphics[width=0.333\linewidth]{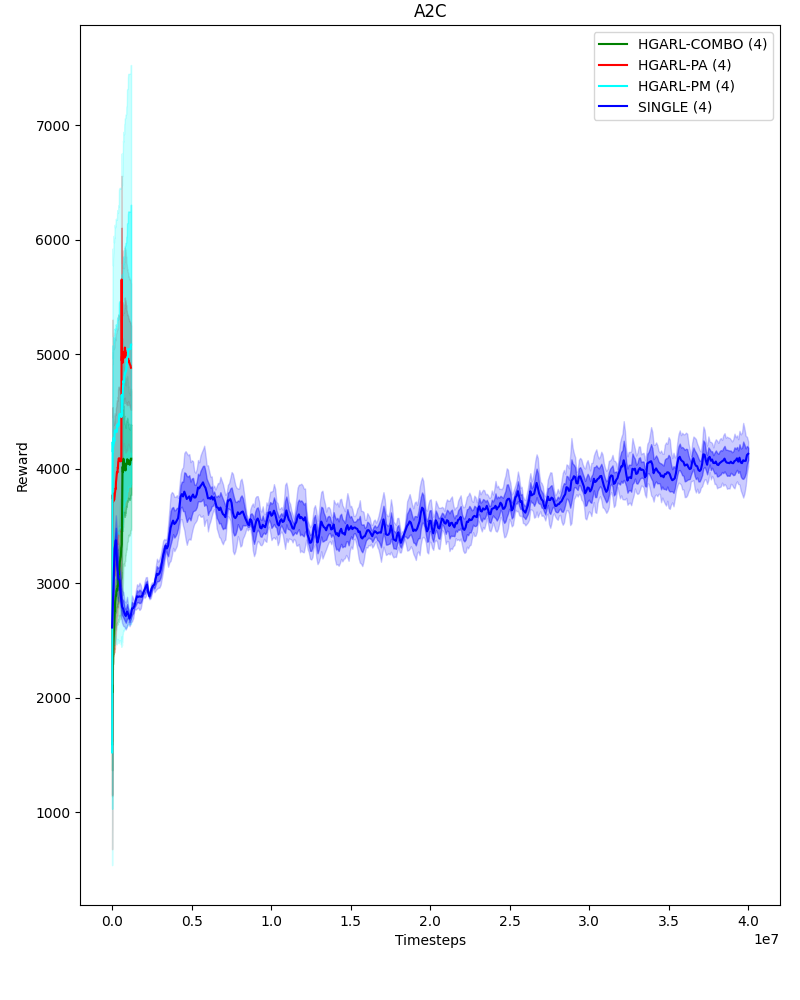}\hfill
                \includegraphics[width=0.333\linewidth]{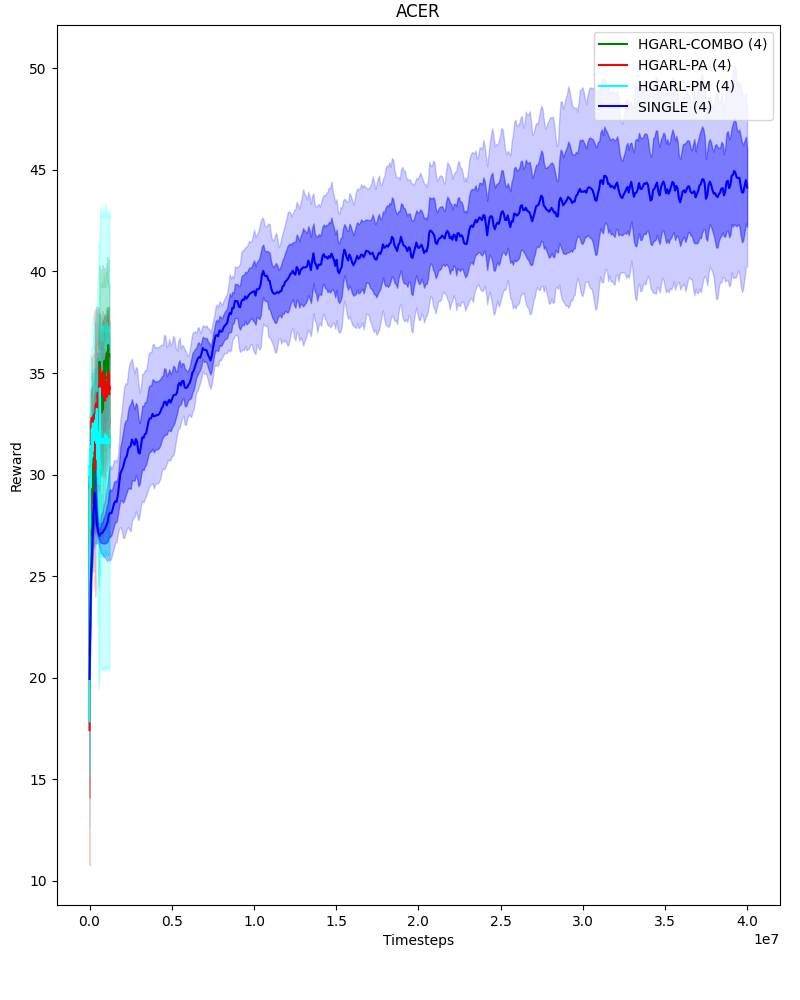}\hfill
                \includegraphics[width=0.333\linewidth]{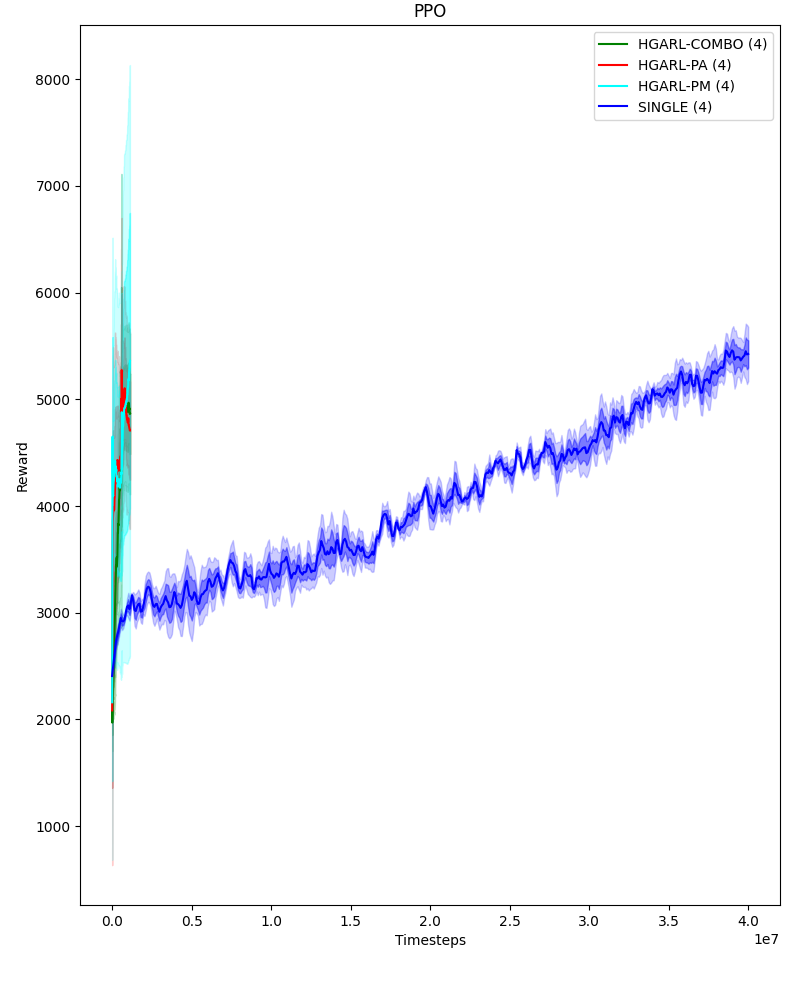}
                \caption{Centipede}
        \end{subfigure}
        \begin{subfigure}[b]{0.49\linewidth}
                \includegraphics[width=0.333\linewidth]{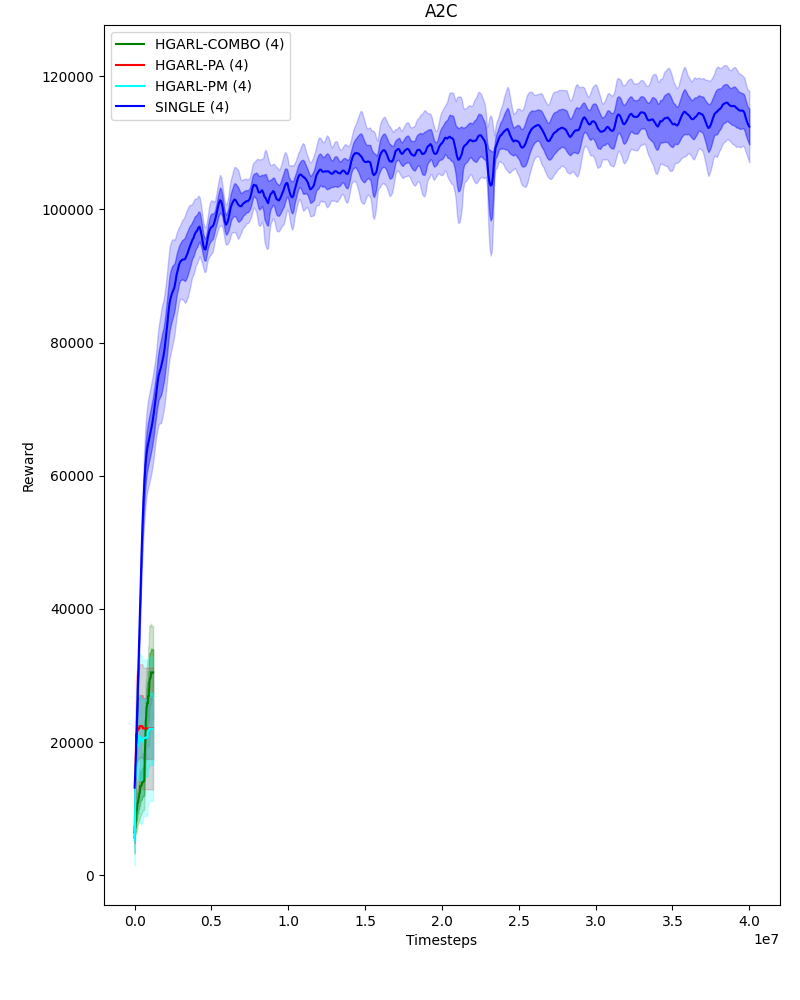}\hfill
                \includegraphics[width=0.333\linewidth]{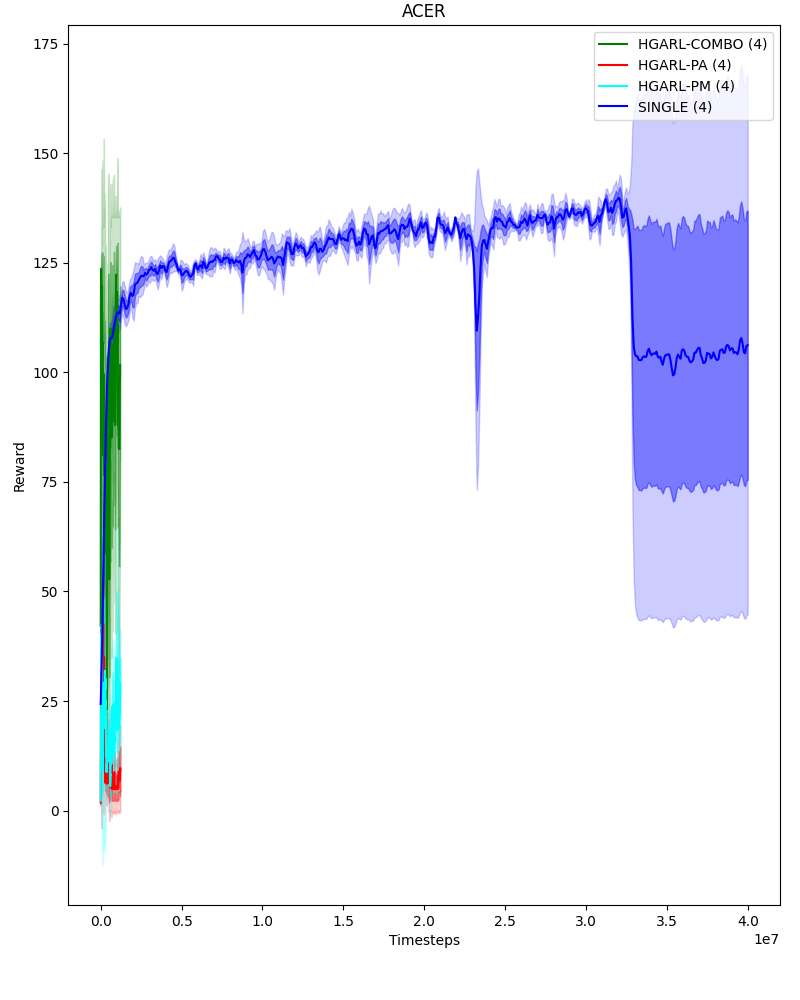}\hfill
                \includegraphics[width=0.333\linewidth]{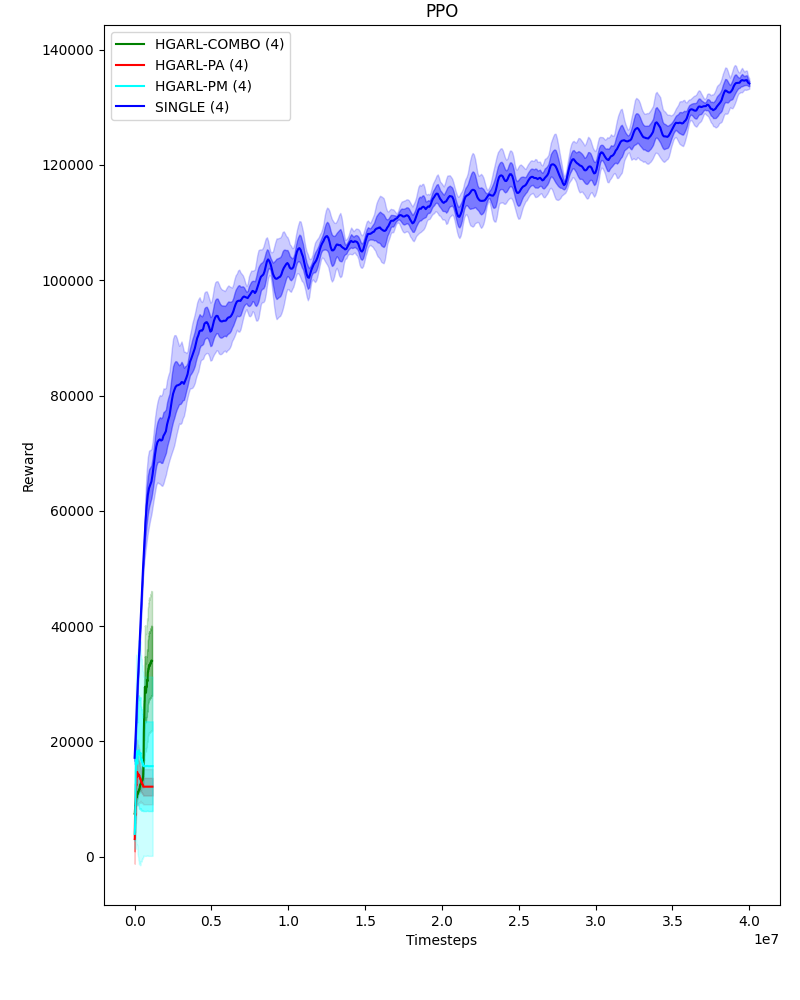}
                \caption{CrazyClimber}
        \end{subfigure}
        \begin{subfigure}[b]{0.49\linewidth}
                \includegraphics[width=0.333\linewidth]{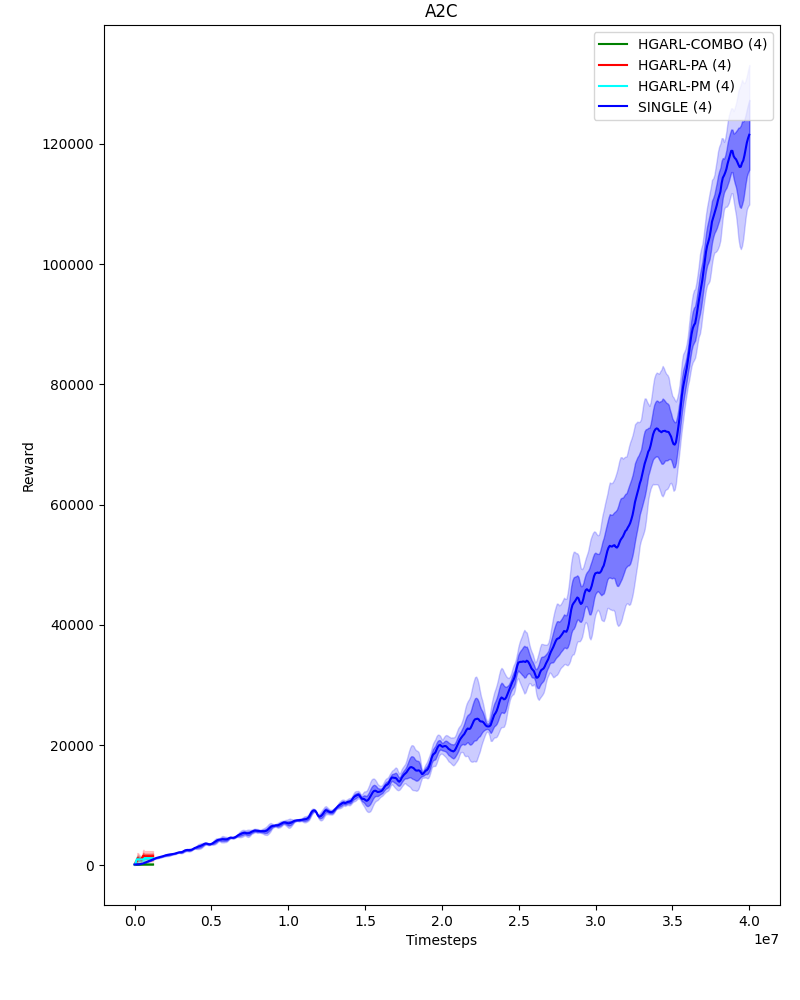}\hfill
                \includegraphics[width=0.333\linewidth]{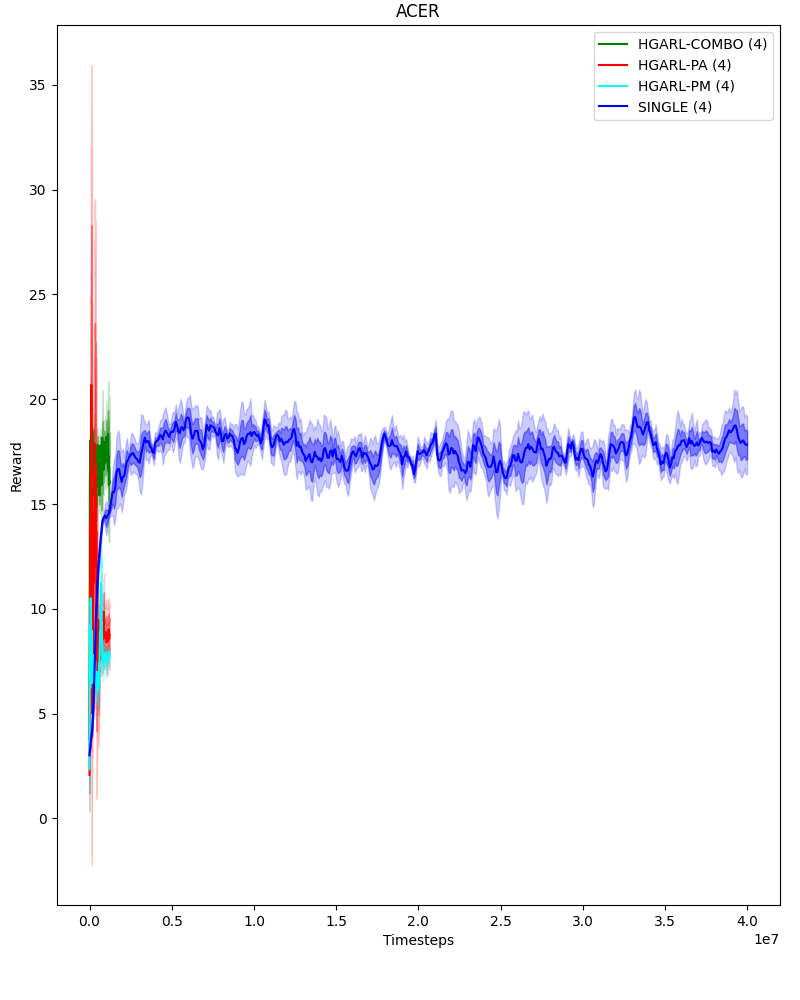}\hfill
                \includegraphics[width=0.333\linewidth]{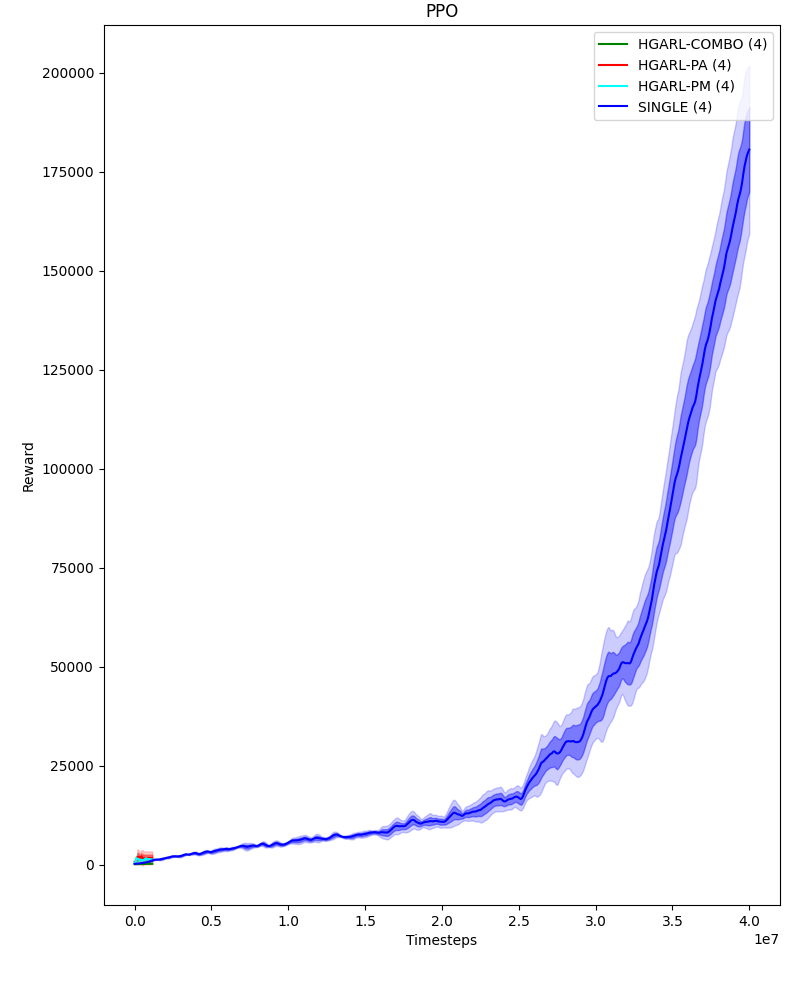}
                \caption{DemonAttack}
        \end{subfigure}
        \begin{subfigure}[b]{0.49\linewidth}
                \includegraphics[width=0.333\linewidth]{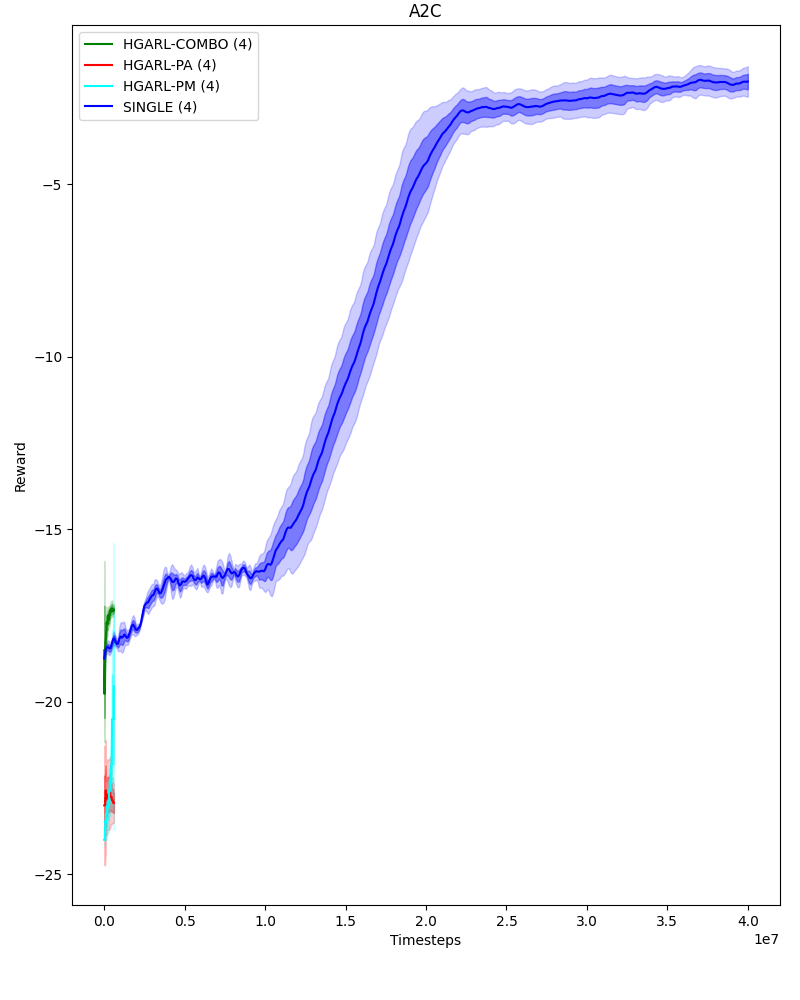}\hfill
                \includegraphics[width=0.333\linewidth]{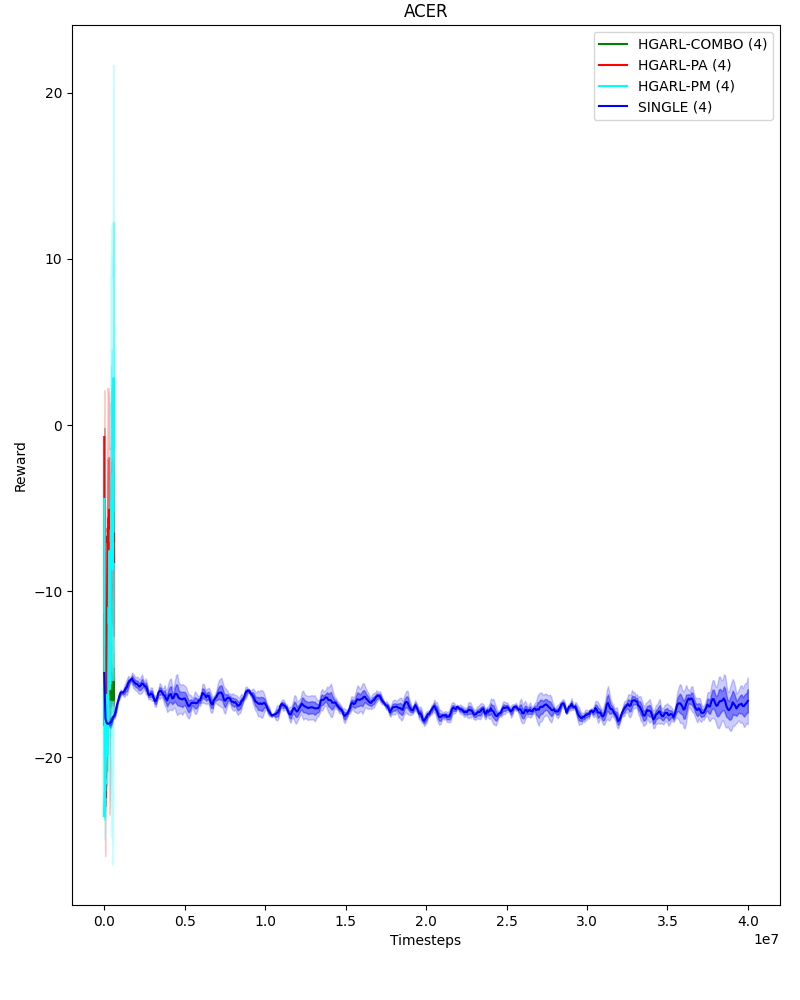}\hfill
                \includegraphics[width=0.333\linewidth]{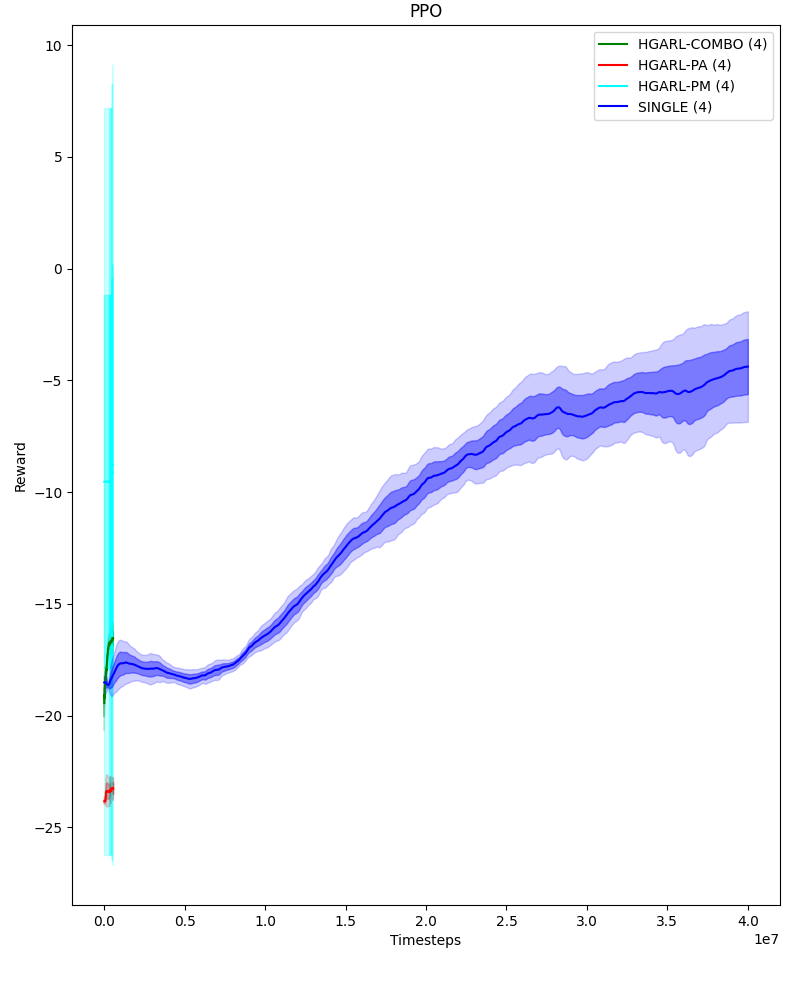}
                \caption{DoubleDunk}
        \end{subfigure}
         \begin{subfigure}[b]{0.49\linewidth}
                \includegraphics[width=0.333\linewidth]{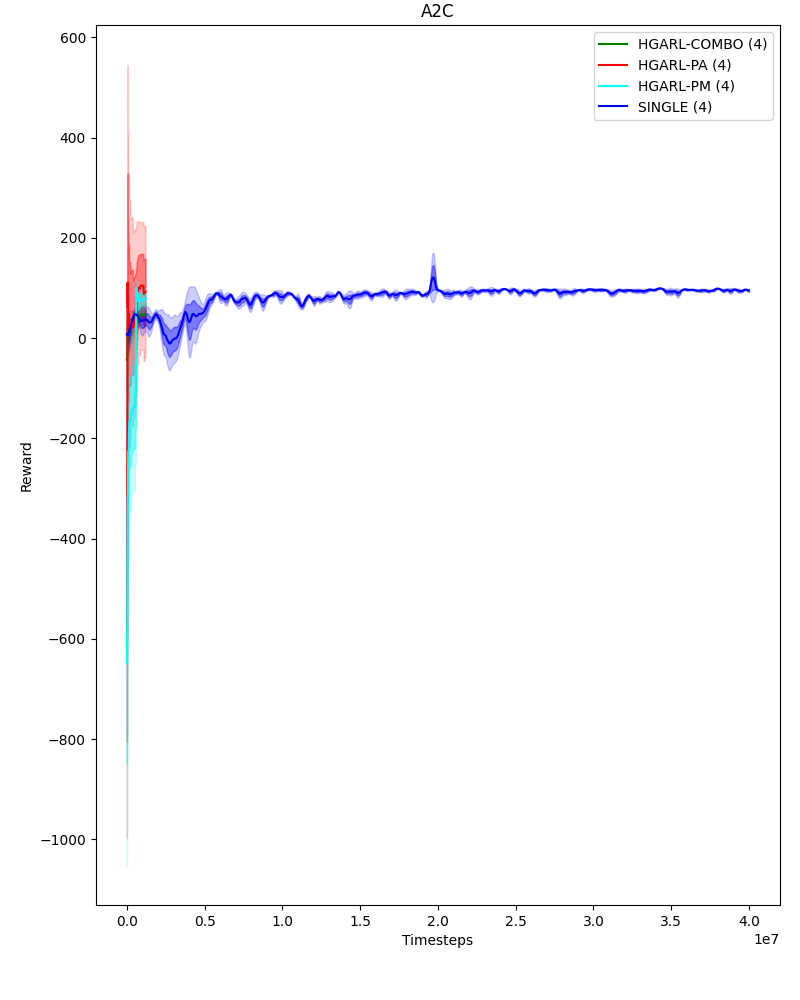}\hfill
                \includegraphics[width=0.333\linewidth]{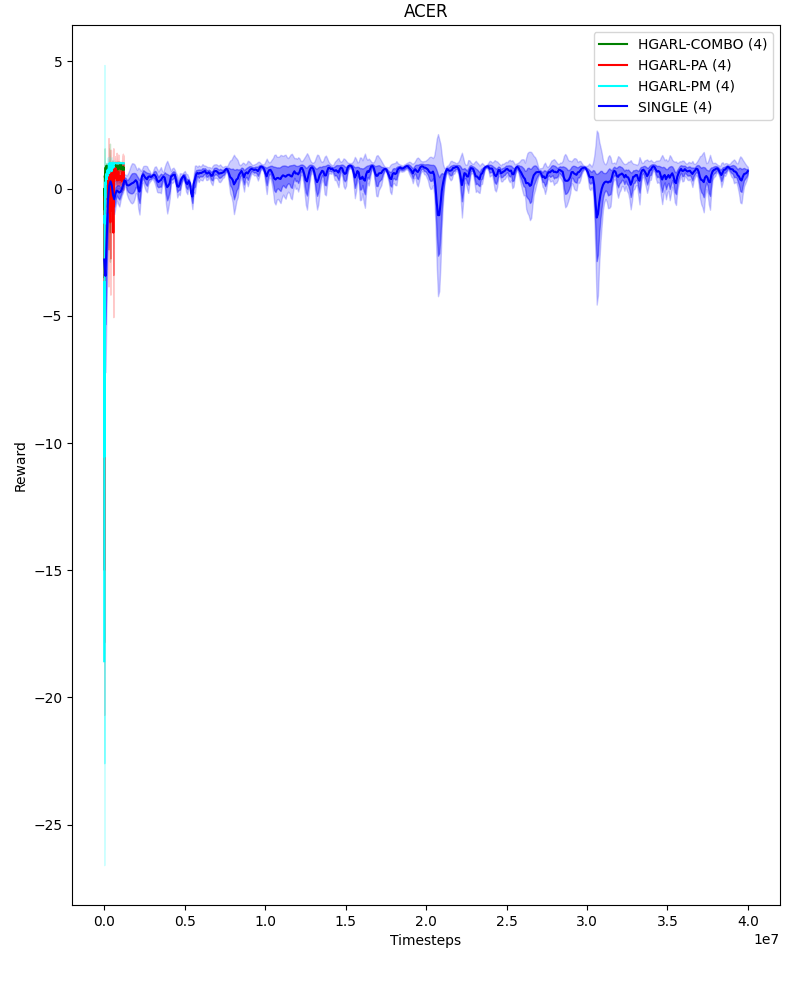}\hfill
                \includegraphics[width=0.333\linewidth]{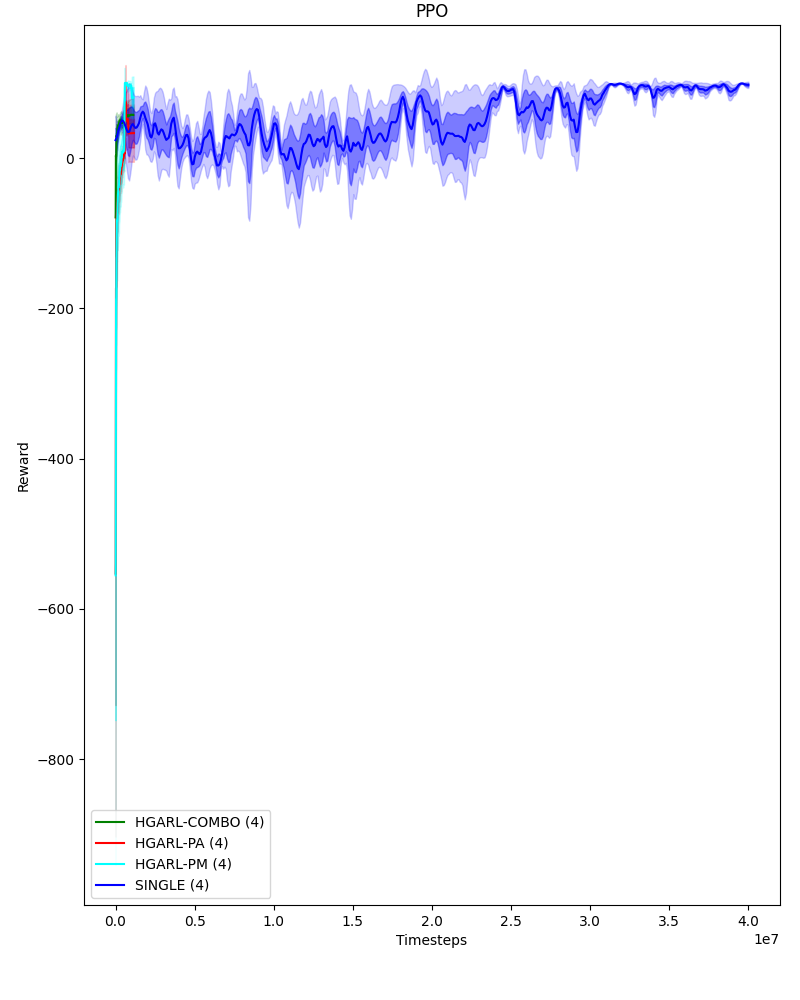}
                \caption{PrivateEye}
        \end{subfigure}
        \begin{subfigure}[b]{0.49\linewidth}
                \includegraphics[width=0.333\linewidth]{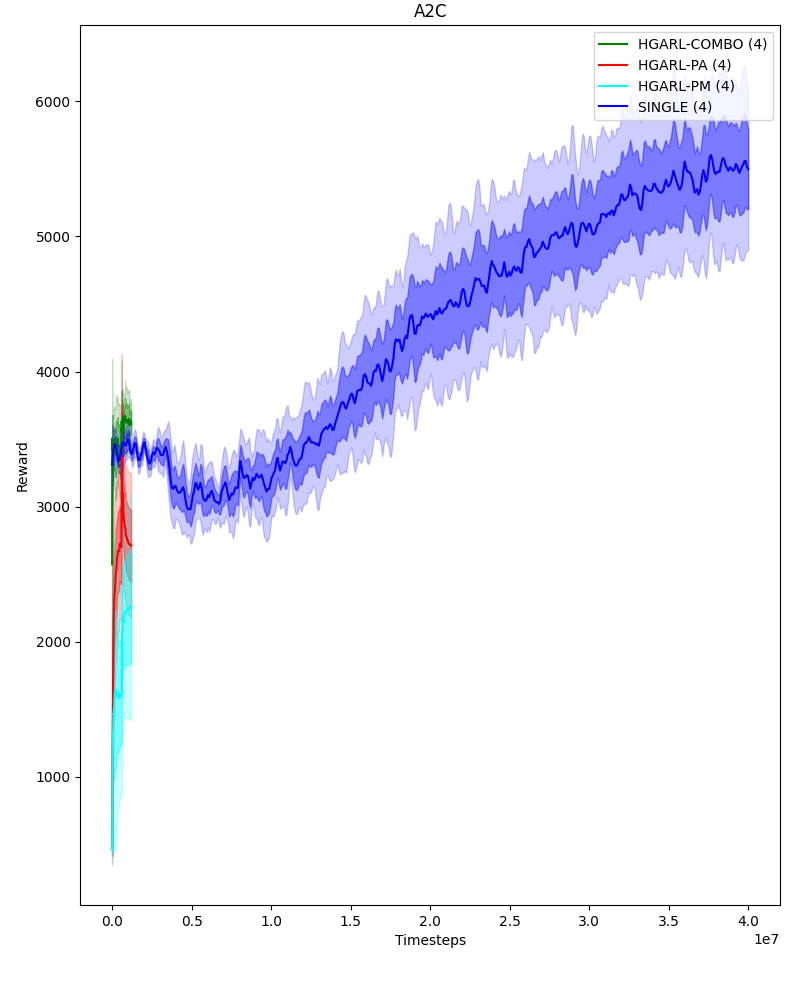}\hfill
                \includegraphics[width=0.333\linewidth]{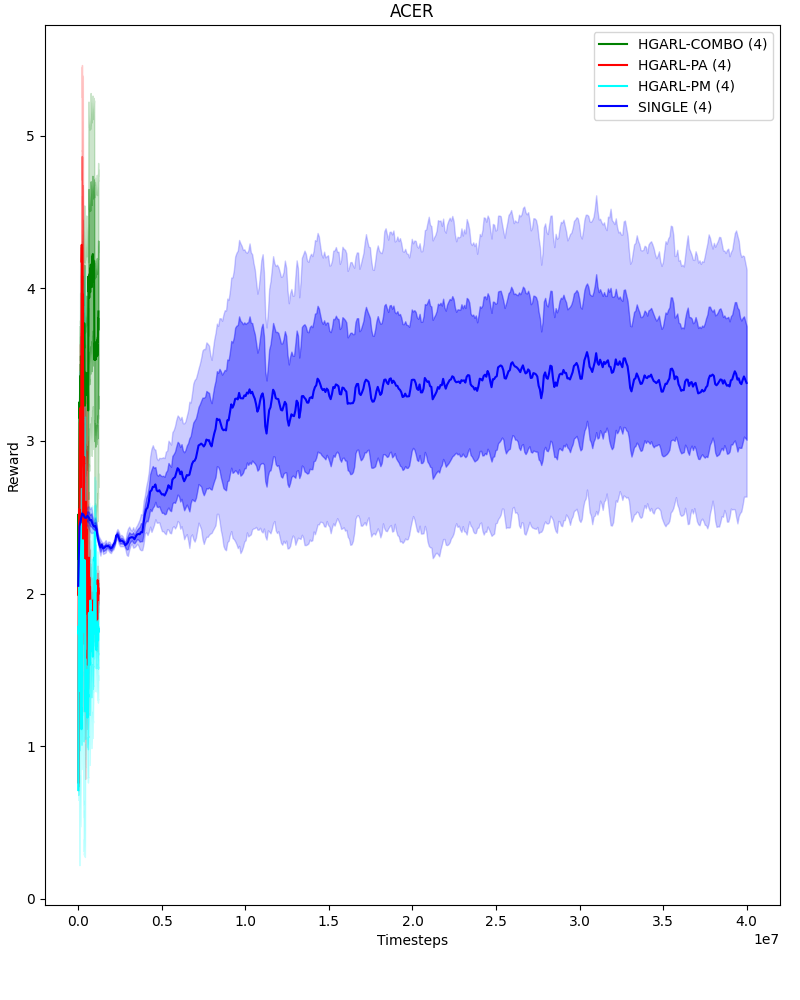}\hfill
                \includegraphics[width=0.333\linewidth]{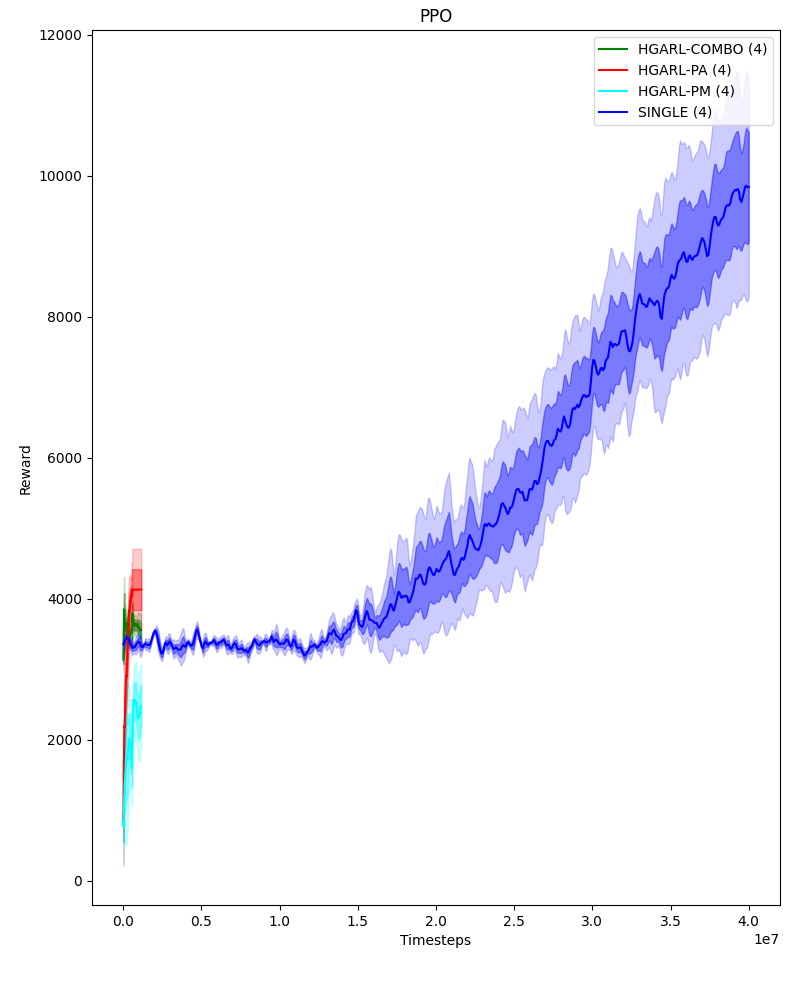}
                \caption{TimePilot}
        \end{subfigure}
        \caption{Atari 2600 Games: Part 4. The three rules produce similar performance that are still much better than single agents.}
        \label{Atari4}
\end{figure*}

\end{document}